\documentclass[10pt]{article}
\usepackage[utf8]{inputenc}
\usepackage[section]{placeins}  
\newif\ifmechgrids \mechgridsfalse
\newif\ifflipfig \flipfigfalse
\newif\ifresfigs \resfigsfalse
\usepackage{amsmath,amssymb,amsthm}
\usepackage[preprint]{tmlr}

\usepackage{array}
\usepackage{booktabs}
\usepackage{multirow}
\usepackage{tabularx}
\usepackage{longtable}
\usepackage{graphicx}
\usepackage{tikz}
\usetikzlibrary{positioning,arrows.meta,calc,fit,backgrounds}
\usepackage[edges]{forest}
\usepackage{microtype}
\usepackage{newunicodechar}
\usepackage[hidelinks]{hyperref}
\usepackage{caption}
\usepackage[nameinlink,capitalise]{cleveref}  
\newunicodechar{≈}{\ensuremath{\approx}}\newunicodechar{≤}{\ensuremath{\le}}\newunicodechar{≥}{\ensuremath{\ge}}
\newunicodechar{∝}{\ensuremath{\propto}}\newunicodechar{∑}{\ensuremath{\sum}}
\newunicodechar{↔}{\ensuremath{\leftrightarrow}}\newunicodechar{∞}{\ensuremath{\infty}}
\newunicodechar{μ}{\ensuremath{\mu}}\newunicodechar{Δ}{\ensuremath{\Delta}}\newunicodechar{τ}{\ensuremath{\tau}}
\newunicodechar{σ}{\ensuremath{\sigma}}\newunicodechar{ω}{\ensuremath{\omega}}\newunicodechar{κ}{\ensuremath{\kappa}}
\newunicodechar{ℤ}{\ensuremath{\mathbb{Z}}}\newunicodechar{∈}{\ensuremath{\in}}
\newunicodechar{⟦}{[}\newunicodechar{⟧}{]}\newunicodechar{√}{\ensuremath{\surd}}\newunicodechar{∘}{\ensuremath{\circ}}
\newunicodechar{≠}{\ensuremath{\neq}}\newunicodechar{⊤}{\ensuremath{^\top}}
\newunicodechar{≡}{\ensuremath{\equiv}}\newunicodechar{≺}{\ensuremath{\prec}}
\newunicodechar{⪯}{\ensuremath{\preceq}}\newunicodechar{⊗}{\ensuremath{\otimes}}\newunicodechar{ℓ}{\ensuremath{\ell}}

\theoremstyle{plain}
\newtheorem{theorem}{Theorem}[section]
\newtheorem{lemma}[theorem]{Lemma}
\newtheorem{proposition}[theorem]{Proposition}
\newtheorem{corollary}[theorem]{Corollary}
\theoremstyle{definition}
\newtheorem{definition}[theorem]{Definition}

\newtheorem{assumption}[theorem]{Assumption}
\theoremstyle{remark}
\newtheorem{remark}[theorem]{Remark}

\newcommand{\HX}{H_{X}}
\newcommand{\HW}{H_{W}}
\newcommand{\R}{\mathbb{R}}
\newcommand{\E}{\mathbb{E}}
\newcommand{\tr}{\operatorname{tr}}
\newcommand{\diag}{\operatorname{diag}}

\DeclareMathOperator*{\argmin}{arg\,min}

\title{Transforms for LLM Quantization:\\ The Great Inversion and Format Co-Design}
\author{\name Ehsan Jokar \email e.jokar.elec@gmail.com}

\begin{document}
\maketitle

\begin{abstract}

Most competitive 4-bit LLM research pipelines now open the same way: apply a linear, function-preserving transform (a rotation, scaling, permutation, or non-orthogonal affine map) so the outlier mass sits more favorably against the group scales, and only then round. Yet we are aware of no survey dedicated to this \emph{transform stage}, and its literature is quietly re-deriving an older theory. We identify and formalize the principle that organizes it, the \emph{Great Inversion}: allocation-flexible coding rewards energy \emph{concentration}, whereas the grouped shared-scale quantization a deployed matrix instruction performs rewards \emph{within-group flattening}. We prove that opposition under within-group majorization and show why, for a generic spectrum, no optimality guarantee transfers across the divide. Through it we survey 200 works to a June~2026 cutoff, classifying 43 transform methods by structure, data-awareness, searched-versus-constructed, and runtime cost, and recording, where the sources report it, how the families compose with GPTQ-class rounding. Tracing that recipe to 1963 transform coding (decorrelate, allocate bits, quantize) organizes them around the allocation-and-scale constraint those classical optima assume away. Classical coders are \emph{allocation-flexible}, spending different bits per coordinate at a fixed total rate, and for a Gaussian source at high rate it is the Karhunen–Loève transform's \emph{concentration} that minimizes distortion. The operand tile a deployed matrix instruction consumes instead carries one absolute-maximum scale per group and equal bits everywhere, with no allocation at all, and on a uniform grid that objective rewards \emph{flattening}, approached calibration-free by Hadamard incoherence; the optimum of a distinct stochastic AbsMax model goes further, adding a data-aware, non-orthogonal whitening no rotation can supply. The two surrogates are extremized at opposite ends of that within-group order, so the prescriptions point in opposite directions, each backed by a proof against its own objective. A second axis, the number format, refines the picture. The non-uniform FP4 element grid makes flattening buy less, and on an \emph{ideal} floating-point grid a rotation can even cost, by destroying a favorable weight--activation anti-alignment; the deployed \texttt{MXFP4} format pulls the other way, its power-of-two block scale still rewarding a rotation confined to that block, while \texttt{NVFP4}'s mantissa-carrying scale largely removes that pull. Which pole a transform should aim for is thus a joint function of allocation regime and format; the majorization surrogates fix the pole, not the transform. One contrast we record without explaining: on \texttt{MXFP4} a \emph{fixed} global rotation craters where a block-confined one helps. We ground the account in published head-to-head numbers, transcribed verbatim under a one-protocol-per-comparison discipline with cross-source displays flagged as such; distill a first-choice guide indexed by deployment regime; and close with the open problems the inversion exposes.

\end{abstract}

\section{Introduction}
\label{sec:intro}

Serving a large language model is dominated by the cost of moving and multiplying
its weights. At the batch sizes and sequence lengths of interest, autoregressive
decoding is memory-bound. Each generated token must stream the entire weight
matrix through the arithmetic units, so both latency and energy scale with the
number of bytes read. Post-training quantization (PTQ) attacks this directly by
storing weights in four or fewer bits, cutting the bytes read during memory-bound
decode. Quantizing the activations too adds a distinct benefit, since 4-bit
operands feed the low-precision tensor cores whose throughput dominates the
compute-bound prefill and large-batch regimes. \texttt{W4A4} is thus driven by
both axes at once, bandwidth and compute. Because PTQ is
training-free and needs only a small calibration set, it has become the default
route to efficient deployment. The frontier has moved steadily downward: from
\texttt{W8A8}, to weight-only \texttt{W4A16}, to the aggressive \texttt{W4A4}
regime in which both operands of every matrix multiply are 4-bit.

The obstacle at these bit-widths is not the average weight or activation but a
handful of extreme ones. Transformer activations, in particular, carry
sparse, large-magnitude \emph{outliers} (a phenomenon first isolated at scale
by \citet{dettmers2022llmint8}). Weight matrices are milder but still not
flat. Either way, a low-bit quantizer shares a single scale across a whole group of values. A distinct full-precision scale per value would be no compression at all. That shared scale is pinned to the group's largest magnitude. One outlier therefore stretches the dynamic range and forces every
ordinary value in its group onto a coarse grid. The quantization error is then
governed not by the typical magnitude but by the ratio of the extreme to the
typical (the \emph{crest factor}, made precise in \Cref{sec:zoo}). That ratio is
the wall that separates comfortable 8-bit quantization from the difficult 4-bit
regime. Simply spending the calibration budget on better rounding
\citep{frantar2022gptq} does not remove it, because the dynamic range is a property of the
representation, not of the rounding.

The idea that unlocked \texttt{W4A4} is not to handle the outliers directly but to
change the \emph{coordinates} in which quantization happens before rounding at all.
A simpler route suggests itself: keep the outlier channels in higher precision,
as in the mixed-precision scheme of \citet{dettmers2022llmint8}. It is telling that the field did not make it the mainline route. Exempting a few values
from the low-bit grid forfeits the uniform dense GEMM, since the outlier columns
then need a separate high-precision path or bespoke sparse kernels. A persistent
minority line, met throughout \Cref{sec:taxonomy,sec:composition,sec:beyond}, does
pay that price. The winning move
keeps the dense low-bit kernel and instead redistributes the outlier energy.
Because a linear layer $Y=XW^\top$ is
unchanged by inserting an invertible map and its inverse, $Y=(XT^\top)(WT^{-1})^\top$, one
may apply a \emph{function-preserving transform} $T$ that redistributes the
outliers. Ideally it spreads each group's mass evenly across its coordinates, so
that no single entry dominates the shared scale. Such a $T$, when it is an orthogonal map of the residual stream or a diagonal one, absorbs into the surrounding weights at no inference cost, as does any map whose activation side lands on a static weight; the rest stay online. A random rotation, realized
cheaply by a randomized Hadamard transform, provably turns a spiky, outlier-ridden vector
into one whose coordinates are sub-Gaussian and whose crest factor is far smaller, from $O(\sqrt d)$ to $O(\sqrt{\log d})$
(\Cref{sec:incoherence}). The layer computes the same function, but on a
distribution the 4-bit grid can represent.

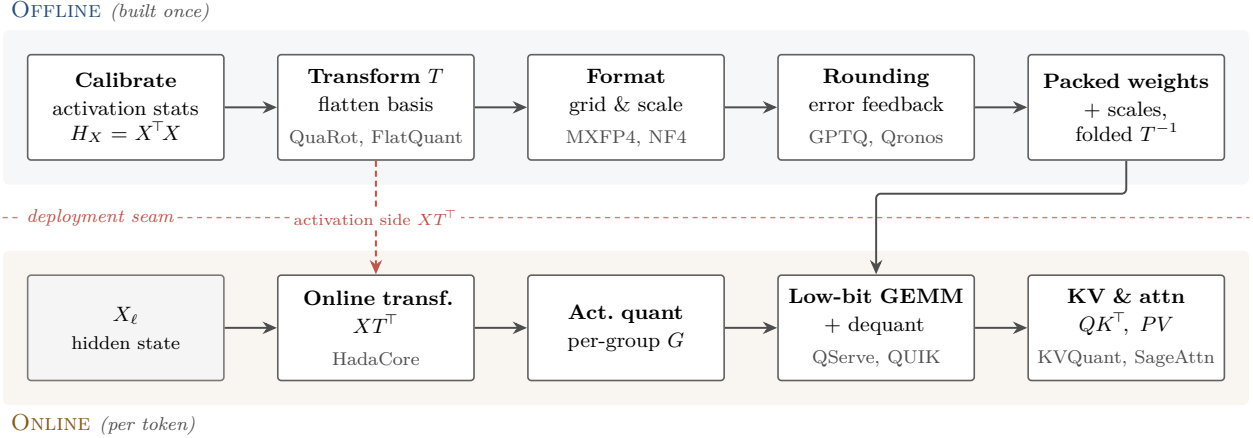
\begin{figure}[t]
\centering
\definecolor{cOff}{HTML}{2F5C8F}   
\definecolor{cOn}{HTML}{9A6A1C}    
\definecolor{cSeam}{HTML}{B23A2E}  
\def\ttl{\bfseries}
\def\met#1{{\footnotesize\color{black!70}#1}}
\resizebox{\linewidth}{!}{%
\begin{tikzpicture}[
  font=\small,
  box/.style ={rounded corners=1.5pt, draw=black!62, line width=0.7pt, fill=white,
               align=center, inner xsep=4pt, inner ysep=4pt, text width=27mm, minimum height=16mm},
  io/.style  ={box, fill=black!4, draw=black!52},
  arr/.style ={-{Stealth[length=2.2mm,width=1.9mm]}, line width=0.85pt, black!70},
  seam/.style={-{Stealth[length=2.2mm,width=1.9mm]}, line width=0.9pt, cSeam!85,
               dash pattern=on 2.6pt off 1.7pt},
  lane/.style={font=\scshape},
]
\def\cA{0} \def\cB{3.8} \def\cC{7.6} \def\cD{11.4} \def\cE{15.2}
\def\rOff{0} \def\rOn{-3.35}
\node[box] (calib) at (\cA,\rOff) {{\ttl Calibrate}\\[1pt]activation stats\\$H_X\!=\!X^{\!\top}\!X$};
\node[box] (tf)    at (\cB,\rOff) {{\ttl Transform $T$}\\[1pt]flatten basis\\[3pt]\met{QuaRot, FlatQuant}};
\node[box] (fmt)   at (\cC,\rOff) {{\ttl Format}\\[1pt]grid \& scale\\[3pt]\met{MXFP4, NF4}};
\node[box] (rnd)   at (\cD,\rOff) {{\ttl Rounding}\\[1pt]error feedback\\[3pt]\met{GPTQ, Qronos}};
\node[box] (pack)  at (\cE,\rOff) {{\ttl Packed weights}\\[1pt]$+$ scales,\\folded $T^{-1}$};
\node[io]  (ain)   at (\cA,\rOn)  {$X_\ell$\\[1pt]{\footnotesize hidden state}};
\node[box] (ker)   at (\cB,\rOn)  {{\ttl Online transf.}\\[1pt]$X T^{\!\top}$\\[3pt]\met{HadaCore}};
\node[box] (aq)    at (\cC,\rOn)  {{\ttl Act.\ quant}\\[1pt]per-group $G$};
\node[box] (gemm)  at (\cD,\rOn)  {{\ttl Low-bit GEMM}\\[1pt]$+$ dequant\\[3pt]\met{QServe, QUIK}};
\node[box] (kv)    at (\cE,\rOn)  {{\ttl KV \& attn}\\[1pt]$QK^{\!\top}\!,\ PV$\\[3pt]\met{KVQuant, SageAttn}};
\foreach \a/\b in {calib/tf,tf/fmt,fmt/rnd,rnd/pack,ain/ker,ker/aq,aq/gemm,gemm/kv}{\draw[arr](\a)--(\b);}
\begin{scope}[on background layer]
  \node[fill=cOff!5, rounded corners=2.5pt, inner sep=3.6mm, fit=(calib)(tf)(fmt)(rnd)(pack)] (offb){};
  \node[fill=cOn!6,  rounded corners=2.5pt, inner sep=3.6mm, fit=(ain)(ker)(aq)(gemm)(kv)]     (onb){};
\end{scope}
\coordinate (sL) at ($(offb.south west)!0.5!(onb.north west)$);
\coordinate (sR) at ($(offb.south east)!0.5!(onb.north east)$);
\draw[cSeam!65, dash pattern=on 2.2pt off 2pt, line width=0.6pt] (sL)--(sR);
\node[fill=white, inner sep=1.5pt, font=\footnotesize\itshape, text=cSeam!88, anchor=west]
      at ([xshift=3mm]sL) {deployment seam};
\draw[seam] (tf.south) -- (ker.north)
      node[midway, fill=white, inner sep=1.2pt, font=\scriptsize, text=cSeam!90]
      {activation side $X T^{\!\top}$};
\draw[arr, rounded corners=2pt] (pack.south) -- ++(0,-0.5) -| (gemm.north);
\node[lane, text=cOff!85!black, anchor=west] at ([yshift=2.8mm]offb.north west)
      {Offline\ {\normalfont\itshape\footnotesize\color{black!70}(built once)}};
\node[lane, text=cOn!85!black, anchor=west]  at ([yshift=-2.8mm]onb.south west)
      {Online\ {\normalfont\itshape\footnotesize\color{black!70}(per token)}};
\end{tikzpicture}%
}
\caption{\textbf{The low-bit post-training-quantization pipeline the surveyed methods assemble, as a two-lane swimlane.} \emph{Top (offline, built once):} calibrate activation statistics $\to$ construct a function-preserving \emph{transform} $T$ (\Cref{sec:taxonomy}) $\to$ fix the target \emph{grid/format} (\Cref{sec:format}) and error-feedback-\emph{round} the transformed weights onto it (\Cref{sec:composition}) $\to$ pack static low-bit weights. \emph{Bottom (online, per token):} the deployed datapath (\Cref{sec:systems}). The transform straddles the \emph{deployment seam} (dashed line, \Cref{sec:live}): its weight side $WT^{-1}$ folds into the packed weights offline (\Cref{def:fpt}), while of its two sides only the activation side $XT^{\!\top}$ crosses it (dashed arrow), to an online kernel unless it too can be absorbed (\Cref{thm:invariance,prop:gamma}). Representative methods are named per stage.}
\label{fig:pipeline}
\end{figure}

This ``transform-then-round'' recipe is
now the opening move of most competitive low-bit pipelines, and of nearly every one that quantizes activations. QuaRot
\citep{ashkboos2024quarot} rotates the residual stream by Hadamard matrices that
fold into adjacent weights, removing activation outliers without changing the output; SpinQuant \citep{liu2024spinquant} learns the rotation instead of fixing
it; QuIP\# \citep{tseng2024quipsharp} pairs incoherence-inducing rotations with
lattice codebooks; FlatQuant \citep{sun2024flatquant} uses learned affine maps;
and WUSH \citep{chen2025wush} derives a closed-form, data-aware transform with
optimality guarantees under a stochastic AbsMax model. These methods differ in almost everything downstream
(the codebook, the rounding algorithm, whether the transform is fixed or learned),
yet they share this first stage. In the aggressive low-bit regime that stage is often the decisive one, as consequential as the choice of rounder.  \Cref{fig:pipeline} lays out that pipeline end to end (the offline construction the transform opens, and the online datapath it must survive) and previews where each part is treated in what follows.

That recipe is what the research literature runs on, and the shipped models draw a sharper line through it than the phrase suggests. On the 4-bit \emph{weight} path the frontier mixture-of-experts releases of 2025--2026 use no transform at all, relying instead on group-wise scales, a protected set of sensitive tensors, and training-time control, quantization-aware training where the card reports it; first-party \texttt{AWQ} and SpinQuant-rotated releases at the same bit-width do carry one, so the shipped weight path is split rather than uniform. The 4-bit \emph{activation} paths that ship divide along the block scale: the one that rotates does so exactly where \Cref{thm:invariance} and \Cref{prop:gamma} offer nothing to fold into, while the \texttt{NVFP4} releases quantize activations at group size~$16$ with no rotation named. The transform stage is thus not yet universal in production, though the shared-scale theory of \Cref{sec:inversion} covers the weight path too; \Cref{sec:sys-deployed} sets out the evidence.

Despite this centrality, we are aware of no survey dedicated to the transform stage. Broad low-bit-LLM surveys treat it as one technique among many.
\citet{gong2024lowbitsurvey}, for instance, catalog systems and algorithms across the whole quantization stack, giving equivalent transformation one subsection, subdivided into shifting, scaling and rotation. The closest adjacent works also stop short of it. The PRISMA systematic review of \citet{czako2025review} folds ``equivalent transformations'' into a four-way catalog of activation-outlier mitigations, frozen in early 2025. \citet{liu2025compeval} decouple published PTQ pipelines into exactly two stages (pre-quantization transformation and error mitigation) and benchmark them under one protocol across \texttt{INT4}, \texttt{MXFP4}, and \texttt{NVFP4}, a controlled empirical comparison of the transform stage that ranks methods rather than organizing the family under a principle. None of the three supplies an optimality theory, and none draws the link to the classical lineage we develop here.

What is missing is a treatment that
organizes the fast-growing family of transforms (diagonal scalings,
permutations, fixed and learned rotations, non-orthogonal maps) under a single
principle, collects the scattered optimality results that explain when and
why a transform helps, and traces the idea to its origin. For the
transform-then-round recipe is not new: it is the sixty-year-old pipeline of
\emph{transform coding} (decorrelate, allocate bits, quantize) re-derived for
neural-network weights. Both halves of the contrast we build on are, individually,
established: for a Gaussian source at high rate, energy \emph{concentration} is the optimum of classical transform coding \citep{goyal2001transform,huang1963block} (with the caveats of \Cref{sec:classical-caveats}), and energy \emph{flattening} (incoherence) bounds the proxy quantization error \citep{chee2023quip} and is the optimal direction under an AbsMax shared-scale model \citep{chen2025wush}, the direction the deployed rotations of \citet{ashkboos2024quarot} take.
What reading the modern literature against the classical theory reveals (and what,
to our knowledge, has not been drawn explicitly) is that the two are the \emph{same} optimization run in opposite regimes, the deployed shared-scale optimum being the reversal of the classical one. 

That reversal is our organizing principle, and its first-order driver is a pair of
freedoms the operand tile removes together: whether the downstream quantizer may allocate
bits per coordinate, which is what makes concentration \emph{strictly} optimal, and whether
each coordinate carries its own scale, which is what makes flattening strictly optimal, on a uniform
grid, once allocation is gone. At a common per-coordinate rate with per-coordinate scales,
and one quantizer shape constant across coordinates as for a Gaussian source, every
orthogonal transform ties (\Cref{sec:inv-fixedrate}); it is the shared scale that breaks the
tie. An \emph{allocation-flexible} quantizer (one that spends a different
number of bits on different coordinates under a fixed total rate, by unequal
deterministic assignment or by entropy coding) rewards transforms that
\textbf{concentrate} energy onto a few coordinates. This is classical transform coding, whose optimum, under those conditions, is the Karhunen--Lo\`eve transform (\Cref{sec:classical}). A \emph{shared-scale} quantizer (one absolute-maximum scale per group, equal bits everywhere, no allocation) instead rewards transforms that \textbf{flatten} energy across coordinates on a uniform element grid. The dense GEMM of every deployed low-bit LLM kernel is shared-scale in this sense, though among the deployed 4-bit formats only the per-group \texttt{INT4} grid is uniform (\Cref{sec:inversion}). That flattening is reached near-optimally by the Hadamard transform, while the optimum of a distinct stochastic AbsMax model goes further, adding a data-aware whitening no rotation can supply (\Cref{thm:wush,rem:inv-scope}).

The two prescriptions are
opposed, extremized at opposite ends of the same
concentrate-versus-flatten order: a geometric-mean-versus-group-peak opposition,
Schur-concave against max-driven, built from the standard majorization toolbox
\citep[Ch.~3]{marshall2011majorization} but paired here along the within-group order. Each is backed by a proof against
its own functional (\citealp{bhadane2021pba}; \Cref{thm:inversion}), and
for a generic spectrum neither transfers to the other regime (\Cref{thm:inversion,cor:notransfer}). We call
this the \textbf{Great Inversion}. The LLM field re-ran the classical
transform-coding playbook in the regime its \emph{optimality theory} never addressed, and
arrived at the opposite prescription: the surrogate it optimizes is extremized at the other pole. The classical literature was not silent on the
shared-scale format itself: \Cref{sec:classical} records a block-floating-point line
that analyzed that format directly from 1970 onward, separately, a 1998 rotation applied expressly to improve a quantizer's overload behavior, and, in 2017, a distributed-mean-estimation result that proved the rotation mechanism itself. What none of them supplied is an optimality theory for that regime; assembling one, from the results the modern literature proves against
its own surrogates, is what this survey does. The second axis, the number format, refines that
reversal, since on a floating-point grid the flattening prescription weakens and can
reverse (\Cref{sec:format}). Every section that follows is a consequence of, or a
documented exception to, these two axes.

We survey function-preserving linear transforms $T$: maps that leave a
layer's output unchanged, up to absorption into adjacent weights or activations,
while changing the coordinates in which quantization occurs. We cover diagonal
scalings, permutations, fixed and learned orthogonal rotations, non-orthogonal
affine maps, and sequence-axis or frame transforms, applied to weights, activations, the
KV-cache, and training-time gradients. We treat the rounding stage (GPTQ and its
relatives) only where it interacts with the transform, and codebooks or lattices
only where a transform enables them. We include the classical transform-coding
lineage as essential background and the modern variable-rate line as the other half
of the inversion. We exclude quantization-aware retraining of the base weights,
mixed-precision bit allocation that uses no transform, and architectural changes.
Where later sections nonetheless touch such methods (a normalized-architecture
pretraining recipe, a low-rank additive branch, a precision-routing scheme) we mark
them as \emph{adjacent} rather than as instances of the definition. The corpus was seeded from the transform-stage methods named above, extended by backward and forward citation tracing from those seeds, and checked against the arXiv \texttt{cs.LG} and \texttt{cs.CL} listings and the main machine-learning and systems venues, with a June~2026 freeze on the literature; the classical lineage was traced back from the modern works' own citations to the 1948--2021 sources that ground it. The freeze is on the search, not on the bibliography: a few entries record a venue that postdates it. The deployment repositories and model cards of \Cref{sec:sys-deployed} were re-checked after that freeze, and the bibliography records a revision for each code repository it quotes. Inclusion follows \Cref{def:fpt}: every work we cite is indexed in \Cref{tab:index}, and the transform methods proper are classified in \Cref{tab:taxonomy}. We aim at complete coverage of the transform-stage \emph{families} to that cutoff, not at an exhaustive census of every instance, still less of low-bit quantization at large.

This survey makes seven contributions.
\emph{(i)}~It connects six decades of transform coding (1963--2021) to the 2022--2026 LLM literature through the allocation/shared-scale inversion. It organizes the
scattered optimality results on both sides into one arc, and it makes precise, as a
within-group Schur opposition built from the standard majorization toolbox, that they optimize
different objectives and do not transfer. The component results belong to the works we
cite; the synthesis and the inversion thesis are ours. The crest factor itself is
classical; what is ours is its use as the axis that organizes the whole family of transform methods.
\emph{(ii)}~It surveys 200 works, indexed completely in \Cref{tab:index} and, for the method families, organized in five per-domain tables. It classifies the 43 transform methods proper along
several axes (structure, data-awareness, searched-versus-constructed, runtime
cost, and, where reported, granularity and composability with rounding). It also names
the composition patterns (\textsc{composes}, with \textsc{enabling} as its limiting case,
\textsc{substitutes}, and \textsc{co-optimizes}) the field has been using without a vocabulary.
\emph{(iii)}~It systematically analyzes three under-examined interactions: how much of a transform's benefit per-coordinate allocation buys instead, and how much of it error-feedback rounding empirically overlaps with; how the shift from integer to hardware-native FP4/microscaling formats changes, and can reverse, which transform helps; and how the transform doubles as a \emph{cost} device on the fixed-rate datapath. There the per-group dequantization tax and the
hardware scale-multiplier make scale granularity a systems variable
(\Cref{sec:sys-dequant,sec:sys-groupsize}).
\emph{(iv)}~It provides a which-transform-when guide indexed by deployment
scenario and an evaluation-pitfalls checklist.
\emph{(v)}~It grounds the analysis in an empirical record (\Cref{sec:empirical}):
a compilation of published head-to-head results, including the one contrast it does not explain, the \texttt{MXFP4} global-versus-block-confined rotation gap (\Cref{sec:fmt-flip}), read under a one-protocol-per-comparison
discipline. That compilation assembles measurements consistent with the thesis: the six-family \texttt{W4A4} comparison, on which some sufficient transform is essential and pure diagonal scaling is not it, the weight-only bit-width sweep, the transform--rounder overlap, the integer-to-\texttt{FP4} flip, the KV cache, and the deployment speed and memory payoff. The concentrate-versus-flatten opposition itself is not on the published record: it is proved against the two surrogate objectives (\Cref{thm:inversion}) and illustrated, at single-layer scale, in \Cref{fig:inversion}. Every measurement reported in \Cref{sec:empirical} is transcribed verbatim from a cited table or figure and re-verified against its source, the few differences and margins we quote being arithmetic on those transcribed values; the mechanism figures (\Cref{fig:crest,fig:inversion,fig:reshape}) and \Cref{tab:flip} are instead our own computations, labeled as such and offered as illustrations of the theory rather than as evaluations.
\emph{(vi)}~It checks the premise against the shipped 4-bit releases it examines (\Cref{sec:sys-deployed}), finding the weight path split between transform-free, \texttt{AWQ} and rotated recipes and the activation path split by block scale, so the survey's subject is delimited by deployment and not only by the literature.

\emph{(vii)}~It argues that the inversion exposes four open problems: joint
transform-and-rounding optimality, a guarantee for the truly deployed
(extreme-value, non-surrogate) objective, the return of allocation-flexible coding to the inference
path, and transform--format co-design.

The rest of the survey follows the arc these contributions set out. \Cref{sec:foundations} fixes the notation and the
quantizer/transform machinery a reader needs; \Cref{sec:classical} traces the
classical inheritance and its concentration optimum; and \Cref{sec:inversion}
states and derives the inversion, the survey's core. \Cref{sec:t4} then assembles the master optimality ledger, cataloging what is proven on each side of the inversion and what is not.

\Cref{sec:taxonomy} sorts the transforms into six families, and \Cref{sec:composition} works out how a transform composes with the rounder and the codebook, substitutes at high rate for per-coordinate allocation, and co-optimizes with the grid. \Cref{sec:format} turns to the number format and the integer-to-\texttt{FP4} co-design, and \Cref{sec:beyond} carries the story beyond the weight matrix to the KV-cache, the attention matmuls, diffusion, and training-time gradients. \Cref{sec:systems} covers the systems layer: the kernels that make transforms cheap, the dequantization tax and the hardware scale-multiplier, the surviving variable-length lane, and what the shipped 4-bit releases actually do (\Cref{sec:sys-deployed}). \Cref{sec:empirical} grounds the theory in published head-to-head numbers; \Cref{sec:practice} distills the evaluation pitfalls and the which-transform-when guide; and \Cref{sec:open} collects the open problems the inversion exposes and concludes.

\section{Foundations and Background}\label{sec:foundations}

Every method in this survey inserts a linear map into a network (reshaping a weight matrix, an activation, or both) before those numbers are quantized, and the literature can be read as a decades-long argument over \emph{which} map. Classical transform coding, the theory behind JPEG and the discrete cosine transform, answered it with the Karhunen--Lo\`eve transform, which \emph{concentrates} energy onto a few coordinates. The low-bit kernel that actually runs a quantized language model, sharing one scale across a whole group, wants the opposite, and that reversal, the \textbf{Great Inversion} (\Cref{sec:inversion}), is the axis this survey turns on.

This section builds, from the ground up, the machinery needed to state that reversal sharply: the decoder-only layer and the three tensors it spends bits on (\Cref{sec:anatomy}), the layer-output proxy that prices a quantizer's error where it is actually felt (\Cref{sec:notation}), the \emph{crest factor} a shared scale is exposed to (\Cref{sec:zoo}) and its price in bits (\Cref{sec:bennett}), the majorization geometry that makes ``concentrate versus flatten'' an exact opposition (\Cref{sec:rate}), the random rotation that reaches the flat pole almost for free (\Cref{sec:incoherence}), the error-feedback rounder that contests part of that gain (\Cref{sec:rounding}), and the cost model that decides where a transform may live at all (\Cref{sec:live}).

\subsection{The decoder-only layer, and where the bits go}
\label{sec:anatomy}A decoder-only layer, in the Transformer lineage of \citet{vaswani2017attention} and in the now-standard pre-norm placement analyzed by \citet{xiong2020layernorm}, maps a hidden-state matrix $X\in\R^{N\times d}$,
holding $N$ token vectors as rows, to an output of the same shape through two
residually-connected sub-blocks. Each opens with an RMSNorm \citep{zhang2019rmsnorm}, a
division by the row's root-mean-square followed by a learned per-channel gain; unlike
LayerNorm it subtracts no mean, a fact that becomes load-bearing in \Cref{sec:live}. Attention normalizes the layer input, the MLP the post-attention residual; we write $\bar X$ for the output of whichever norm opens the sub-block under discussion. A dense layer holds seven weight matrices (\Cref{sec:beyond-moe} takes up the mixture-of-experts case). Five of them ($W_q,W_k,W_v$ in attention and $W_{\mathrm{gate}},W_{\mathrm{up}}$ in the MLP) read such a normalized stream directly and are linear maps of the standard form $\bar X\,W^{\!\top}$; the remaining two, $W_o$ and $W_{\mathrm{down}}$, one in each sub-block, apply the same form to the intermediate their own sub-block has just produced.

The first sub-block, self-attention, projects $\bar X$ to queries, keys and values, $Q=\bar X W_q^{\!\top}$, $K=\bar X W_k^{\!\top}$ and $V=\bar X W_v^{\!\top}$, reshapes them into heads, applies a rotary position embedding \citep{su2021roformer} to the queries and keys, forms the causally masked, scaled score product, passes it through a row-wise softmax to the
row-stochastic $P$, takes the context product $PV$, and mixes the per-head outputs through an output projection $W_o$.
Under \emph{grouped-query attention} \citep{ainslie2023gqa}, now standard in deployed models, several query heads share one key/value head, which shrinks the cache. The second sub-block is a
position-wise gated MLP in the SwiGLU form \citep{shazeer2020glu}: $W_{\mathrm{gate}}$ and $W_{\mathrm{up}}$ lift the stream to a wider intermediate of width $d_{\mathrm{ff}}$, an elementwise gate
modulates that intermediate, and $W_{\mathrm{down}}$ contracts it back. The ratio $d_{\mathrm{ff}}/d$ is a design parameter rather than a constant: \citeauthor{shazeer2020glu}'s parameter-matched form sets it to $\tfrac83$, and across the gated-MLP releases we counted it runs from roughly that value to three times it (\texttt{Gemma-2-27B}, where $d_{\mathrm{ff}}=8d$).

The keys and values are \emph{cached} across generation steps, the KV-cache, so decoding
token $t{+}1$ reuses the $K,V$ of tokens $1,\dots,t$ rather than recomputing them, the cached keys feeding the score product and the cached values the context product. That cache grows linearly with both context length and batch size, and at long context or large batch it, not the weights, dominates memory \citep{hooper2024kvquant}.

Those seven matrices carry essentially all the layer's parameters, the three MLP matrices holding the bulk of them: $80.8\%$ on \texttt{Llama-3-8B}, against $66.8\%$ on the older multi-head \texttt{Llama-2-7B}, the shift driven both by the shrunken $W_k,W_v$ of grouped-query attention and by the wider intermediate.\footnote{Counting the seven matrices from the public configurations gives $80.8\%$ for \texttt{Llama-3-8B} ($d=4096$, $d_{\mathrm{ff}}=14336$, $32$ query and $8$ key/value heads) and $82.4\%$ for \texttt{Llama-3-70B}, against $66.8\%$ for \texttt{Llama-2-7B}, which is multi-head with $d_{\mathrm{ff}}=2.69d$. The share is no constant of the architecture: it tracks $d_{\mathrm{ff}}/d$ and the head geometry together, and across the releases we counted, it runs from $60.0\%$ on \texttt{Qwen3-0.6B}, whose fixed $128$-wide heads leave $W_q,W_o$ large against a narrow $d$, to $91.4\%$ on the original \texttt{gemma-2b}, from the first Gemma release rather than the Gemma-2 family ($d=2048$, $d_{\mathrm{ff}}=8d$, eight $256$-wide query heads over a single key/value head). These are our own counts, not transcribed measurements.} All seven are the matrices weight quantization compresses, and each is a linear map of the same form $Y=XW^{\!\top}$, with $X$ that projection's own input activation, which the next subsection isolates and analyzes.

The distinction that organizes everything downstream is not attention versus MLP but the two kinds of matrix multiply inside them. The seven projections each multiply a data-dependent activation by a \emph{static} weight matrix fixed at training time. The two operations at the heart of attention, the score product $QK^\top$ and the context product $PV$ that follows it, instead multiply \emph{two activations} together, neither operand a stored weight. This is the difference the whole transform story turns on. A function-preserving transform (\Cref{def:fpt}) has a weight side that folds into a static weight offline for free, and when it attaches at a site that can absorb it (an orthogonal map on the residual stream by computational invariance, \Cref{thm:invariance}; a diagonal map by norm-folding, \Cref{prop:gamma}) the whole map costs nothing at inference. Orthogonal maps inserted mid-block stay online unless both their sides land on static weights: the pre-down-projection Hadamard is paid at runtime by an online kernel (the online lane of \Cref{fig:pipeline}), while the head-wise value rotation folds into the value and output projections (\Cref{sec:live}). Having a static operand to absorb the weight side into is what makes the seven projections the natural home for a transform. The two attention matmuls have no static operand to absorb one, so quantizing them calls for different instruments (an axis choice, a scale trick, or an online transform) and is taken up separately in \Cref{sec:beyond}.

With the layer's structure in place, we can say where the bits go. Three tensors are quantized, under three schemes the survey keeps distinct. The weights of the seven projections are the target of weight-only quantization, the \texttt{W4A16} regime of \Cref{sec:taxonomy} (activations high-precision, only the stored parameters shrink). That regime dominates memory-bound single-stream decoding, where the binding cost is the bandwidth to read the weights. The activations entering those projections, the normalized stream $\bar X$ for five of them and the attention context and gated intermediate for $W_o$ and $W_{\mathrm{down}}$, are quantized too in the \texttt{W4A4} regime, which lets the projection GEMMs run on low-bit tensor cores and matters for throughput-bound prefill and large-batch serving. The \emph{KV-cache} is quantized in its own right, for the cache growth above and for an outlier structure in its keys unlike anything in the weight matrices (\Cref{sec:beyond}). The shorthand $\mathrm{W}b\mathrm{A}c\mathrm{KV}d$ records the three widths independently: \texttt{W4A16} touches only the weights, \texttt{W4A4} adds the activations, \texttt{W4A4KV4} the cache as well. With the quantization sites in view, we now zoom in on a single projection, one weight-times-activation linear, and ask what its quantization actually costs.

\subsection{Notation, the linear layer, and the layer-output proxy}
\label{sec:notation}

Take a single one of the projections just inventoried (\Cref{sec:anatomy}): any of the seven, in isolation. Following the batch-first convention of the deep-learning literature, we stack the $N$ calibration tokens as the \emph{rows} of the activation matrix $X\in\R^{N\times d_{\mathrm{in}}}$ and write the weight matrix $W\in\R^{d_{\mathrm{out}}\times d_{\mathrm{in}}}$. A linear layer therefore computes $Y=XW^\top\in\R^{N\times d_{\mathrm{out}}}$ (a single token is a row $x\in\R^{1\times d_{\mathrm{in}}}$ with $y=xW^\top$). The classical transform-coding sections (\Cref{sec:classical}) instead follow that field's universal convention and treat an abstract source sample as a column vector $x$ coded by an orthogonal $U$, so that a transformed second moment is written $U^{\!\top}\Sigma U$; that is the convention of \Cref{sec:rate,sec:classical}. Tensors elsewhere keep the batch-first row layout above, while a single vector is written as a column wherever the algebra is cleaner. We use the Frobenius norm $\|\cdot\|_F$, the spectral norm $\|\cdot\|_2$, the entrywise max $\|\cdot\|_{\max}$, and $\ell_p$ vector norms $\|\cdot\|_p$. Two second-moment matrices recur throughout:
\begin{equation}
\HX \;=\; X^\top X \in\R^{d_{\mathrm{in}}\times d_{\mathrm{in}}},
\qquad
\HW \;=\; W^\top W \in\R^{d_{\mathrm{in}}\times d_{\mathrm{in}}}.
\end{equation}
$\HX$ is the (uncentered) \emph{activation second-moment matrix} (the token second moment $\sum_t x_t^{\!\top}x_t$ over the token rows $x_t$, equivalently the Gram of the $d_{\mathrm{in}}$ channel columns of $X$); up to the standard factor of two it is the layerwise Hessian of the reconstruction loss below, and it drives every data-aware method in this survey. (The OBC/GPTQ Hessian is $2\HX$; our batch-first $X^\top X$ is the matrix GPTQ writes as $XX^\top$ in the feature-first convention, where its activation matrix is our $X^\top$.) $\HW$ is the \emph{weight Gram}.

We write a quantizer as a map $Q(\cdot)$ with error $e=Q(x)-x$, and the quantized-minus-original weight as $\Delta W=\widehat W-W$. Through the theory we mostly use the two-width form $\mathrm{W}b\mathrm{A}c$ of the bit-width shorthand fixed in \Cref{sec:anatomy}.

A quantizer is not judged on the weights in isolation; what almost every calibration procedure in this survey optimizes is the change it induces at the layer output. Suppose we quantize the weights to $\widehat W=W+\Delta W$ and the activations to $\widehat X=X+\Delta X$. The output error $\widehat Y-Y=X\,\Delta W^\top+\Delta X\,W^\top+\Delta X\,\Delta W^\top$ then expands, term by term, into
\begin{equation}
\label{eq:output-error-expansion}
\|\widehat Y-Y\|_F^2
=\underbrace{\|X\,\Delta W^\top\|_F^2}_{\text{weight side}}
+\underbrace{\|\Delta X\,W^\top\|_F^2}_{\text{activation side}}
+\underbrace{2\,\langle X\,\Delta W^\top,\ \Delta X\,W^\top\rangle_F}_{\text{cross term}}
+\underbrace{\mathcal R}_{\text{second order}},
\end{equation}
where the remainder
\[
\mathcal R=2\big\langle X\,\Delta W^\top+\Delta X\,W^\top,\ \Delta X\,\Delta W^\top\big\rangle_F+\|\Delta X\,\Delta W^\top\|_F^2
\]
collects every term carrying the product $\Delta X\,\Delta W^\top$.

Two assumptions collapse the expansion to a clean sum of one-sided terms.
\emph{(i)~The cross term has zero mean.} Under conditionally-unbiased rounding (e.g.\ stochastic rounding) the weight- and activation-side errors are each zero-mean and mutually independent, so their cross-product has zero expectation,
\[
\E\langle X\,\Delta W^\top,\ \Delta X\,W^\top\rangle_F=\tr\!\big(\E[\Delta W]\,X^\top\,\E[\Delta X]\,W^\top\big)=0 .
\]
This independent-unbiased model is precisely the one under which the bilateral split's cross-term vanishes in expectation \citep{chen2025wush}.
\emph{(ii)~The second-order term is dropped.} By the same independence $\E\mathcal R=\E\|\Delta X\,\Delta W^\top\|_F^2$, a product of \emph{two} rounding errors and hence of higher order in the quantization step than either one-sided term. Neglecting it is exact only asymptotically and merely a good approximation at the aggressive \texttt{W4A4} widths this survey emphasizes, where we do not quantify it (\Cref{sec:bennett}).
What survives is the \emph{bilateral proxy}
\begin{equation}
\label{eq:bilateral-proxy}
\E\|\widehat Y - Y\|_F^2
\;\approx\;
\underbrace{\E\,\tr\!\big(\Delta W\, \HX\, \Delta W^\top\big)}_{\text{weight-side proxy}}
\;+\;
\underbrace{\E\,\tr\!\big(\Delta X\, \HW\, \Delta X^\top\big)}_{\text{activation-side proxy}},
\end{equation}
having rewritten the two surviving norms as $\|X\,\Delta W^\top\|_F^2=\tr(\Delta W\,\HX\,\Delta W^\top)$ and $\|\Delta X\,W^\top\|_F^2=\tr(\Delta X\,\HW\,\Delta X^\top)$. The weight-side term is the layerwise objective of the OBC/GPTQ rounding family \citep{frantar2022obc,frantar2022gptq}, whose Hessian is $2\HX$. For weight-only quantization ($\Delta X=0$) the cross and second-order terms vanish identically, so the weight-side term equals $\|X\,\Delta W^\top\|_F^2$ \emph{exactly} on the calibration set, with no approximation. Its data-aware, sensitivity-weighted form has CNN-compression antecedents: \citet{young2021transform} weight instead by the covariance of the end-to-end output gradients.

Every transform in this survey reshapes how error is apportioned between the two proxy terms, and \Cref{sec:inversion} shows that the classical coding surrogate and the shared-scale AbsMax cost the deployed kernel actually charges pull in opposite directions on a single geometric axis. The maps that carry out this redistribution while leaving $Y=XW^\top$ unchanged, \emph{function-preserving transforms}, are the survey's central object, defined next.

\begin{definition}[Function-preserving transform]
\label{def:fpt}
An invertible $T\in GL(d_{\mathrm{in}})$ is \emph{function-preserving} for the layer $Y=XW^\top$ if replacing $(W,X)\mapsto(WT^{-1},\,XT^\top)$ leaves the output unchanged: $(XT^\top)(WT^{-1})^\top=XW^\top$. The weight side $WT^{-1}$ is computed offline; the activation side $XT^\top$ must be produced at inference unless it can be \emph{absorbed} (\Cref{sec:live}).  More generally we call a map function-preserving when its action on one operand is undone exactly by a compensating offline change to the adjacent static weights or bias, by an invariance of the following operation, or by an inverse applied downstream; the linear input-axis case above is the one \Cref{sec:inversion} is stated for, and the other variants are named as such where they appear.
\end{definition}

The definition looks almost too permissive (any invertible $T$ is admissible), but its cost structure is sharply asymmetric, which is what makes the choice of $T$ a real design problem rather than a free lunch. The weight side $WT^{-1}$ is folded in offline and free, while the entire running price of that input-axis case is the activation side $XT^\top$, recomputed for every token. So the space of useful transforms is not the space of \emph{helpful} ones but the space of helpful ones whose activation side is \emph{cheap}, ideally free. That tension, free-on-the-weights against costly-on-the-activations, is the whole economics of \Cref{sec:live}, and the network already ships with one transform that sits at the free end of it.

The learned gain $\gamma$ of the RMSNorm that opens each sub-block is itself a diagonal function-preserving transform, and it folds into each consuming weight at zero cost:
\begin{proposition}[$\gamma$-folding]
\label{prop:gamma}
For a norm with gain $\gamma$ feeding a linear map $W$, $\;\big(x_{\mathrm{norm}}\odot\gamma\big)W^\top=x_{\mathrm{norm}}\big(W\diag(\gamma)\big)^\top$. Hence setting $W\leftarrow W\diag(\gamma)$, $\gamma\leftarrow \mathbf 1$ preserves the function exactly and moves a per-channel scale entirely onto the weights. Where a norm feeds several projections, as both norms of \Cref{sec:anatomy} do, the substitution is applied to each of them before $\gamma$ is reset.
\end{proposition}
\Cref{prop:gamma} is the simplest member of the \emph{absorption algebra} (\Cref{sec:live}); the diagonal-scaling family of \Cref{sec:tax-diagonal} is its data-aware generalization. The same absorb-offline logic will later place transforms at the residual stream, the KV cache, and the attention path, each carrying its own inference cost (\Cref{sec:live}). But absorption is a question of cost, and cost is the section's last concern; the first is more basic: what a quantizer actually does to the object the proxy is built from, one group of values at a time.

\subsection{The scalar quantizer zoo}
\label{sec:zoo}

The bilateral proxy \eqref{eq:bilateral-proxy} weighs a whole layer, but a deployed kernel never quantizes a layer as a single object. Instead, it partitions the coordinates into small \emph{groups} and gives each group its own scale. The proxy is therefore a sum over those groups, and the layer's error is assembled one group at a time; we can understand the entire mechanism on one group and then sum. So fix a group: concretely, picture sixteen weights, fifteen of them ordinary and one an outlier several times larger, the running example we return to at every stage and that \Cref{fig:crest} draws. Three elementary questions decide its fate: what map sends its real values to $b$-bit codes, how is that map's one free parameter, its scale, chosen, and what property of the group then governs the resulting error? Answering them, in that order, isolates the single scalar the rest of the survey turns on.

\begin{definition}[Scalar quantizer]
A $b$-bit scalar quantizer is a map $Q:\R\to\mathcal C=\{c_1,\dots,c_L\}$, $L=2^b$, defined by decision boundaries $-\infty=b_0<b_1<\cdots<b_L=+\infty$ and reproduction points $c_i$, with $Q(x)=c_i$ for $x\in(b_{i-1},b_i]$. It factors into an \emph{encoder} $\alpha:\R\to\{1,\dots,L\}$ storing the index ($b$ bits) and a \emph{decoder} $\beta$ mapping indices to $\mathcal C$. Its distortion is $D=\E[(X-Q(X))^2]$.
\end{definition}

Only the index $\alpha(x)$, $b$ bits, is stored and moved; the reproduction points $\mathcal C$ sit in a tiny table shared by the group. Every design choice from here is where to put those points and their boundaries, and the deployed choice is the plainest imaginable.

The deployed workhorse, the uniform affine quantizer, places the $c_i$ on an evenly spaced grid of step $\Delta$. The asymmetric form $q=\mathrm{clip}(\mathrm{round}(x/s)+z,\,0,\,2^b-1)$ maps a range $[x_{\min},x_{\max}]$ onto all $2^b$ codes with scale $s=\Delta=(x_{\max}-x_{\min})/(2^b-1)$ and integer zero-point $z$, dequantized as $\widehat x=s(q-z)$. The symmetric form ($z=0$) uses a signed grid with $s=\max_i|x_i|/(2^{b-1}-1)$. Inside the represented range the error is bounded by $\Delta/2$; outside it (an \emph{overload}) it is unbounded: a tension the choice of scale must resolve.

That scale need not be shared across the whole tensor, and how widely it is shared is the second deployment knob.
\begin{definition}[Granularity]
A scale may be shared \emph{per-tensor} (one $s$), \emph{per-row} (one $s$ per output row of $W$, the \emph{per-output-channel} weight scale, which the weight-only literature usually just calls \emph{per-channel}, or one $s$ per token row of $X$, the \emph{per-token} activation scale that is the deployed default), \emph{per-column} (one $s$ per input channel, the axis a \emph{per-channel activation} scale shares), or \emph{per-group/block} (contiguous groups of $G$ entries, e.g.\ $G\in\{32,128\}$). Finer granularity lowers distortion at the cost of storing more scales; in the limit a full-precision scale per element would be no compression at all, so a scale must be amortized over a group, which is what makes a single outlier in that group unavoidable. The group size $G$ is the single most important deployment knob and reappears in the block-scaled hardware formats of \Cref{sec:live}.
\end{definition}

Granularity settles how widely a scale is shared, but not the \emph{shape} of the grid it scales; so before choosing a scale, it is worth asking whether the evenly spaced shape is the right one at all. For a fixed density $p$ it is not: the distortion-minimizing $b$-bit grid is in general \emph{non-uniform}, crowding its levels where the probability mass is, and any optimal grid must satisfy the Lloyd--Max conditions.
\begin{proposition}[Lloyd--Max optimality \citep{lloyd1982lsq,max1960quantizing}]
\label{prop:lloydmax}
Assume $\E[X^2]<\infty$ and that the support of $X$ contains at least $L$ points, so that no optimal cell has zero probability. Then every optimal fixed-rate scalar quantizer satisfies, up to sets of probability zero, (i) the \emph{nearest-neighbor} condition $b_i=\tfrac12(c_i+c_{i+1})$ for the interior boundaries $i=1,\dots,L-1$ and (ii) the \emph{centroid} condition $c_i=\E[X\mid X\in(b_{i-1},b_i]]$. Alternating (i) and (ii) (Lloyd's algorithm) never increases $D$, though its stationary points need not be local minima.
\end{proposition}
Both conditions come from optimizing one half at a time: with the boundaries fixed, minimizing $\E[(X-c_i)^2\mathbf 1\{X\in(b_{i-1},b_i]\}]$ over $c_i$ gives the conditional mean, finite by $\E[X^2]<\infty$ and well defined because the cell carries positive probability; with the levels fixed, sending each $x$ to the nearest level puts the boundary at the midpoint of its neighbors. Each half-step weakly decreases $D$, which is why the back-and-forth never increases it. That alternation is the engine behind several of the data-driven codebooks we meet later. The vector-quantization family of \Cref{sec:vq} runs the same loop in many dimensions at once, and its Hessian-weighted version is the CNN-era sibling of GPTQ's objective, in the same second-order tradition (\Cref{sec:classical-nn}). Not every non-uniform codebook is a Lloyd--Max fixed point, however: NF4 \citep{dettmers2023qlora} instead spaces its levels at equal-probability \emph{quantiles} of a Gaussian, an entropy-maximizing rather than MSE-minimizing choice, distinct from the centroid grid above (\Cref{sec:vq}).

Yet deployment overwhelmingly declines the Lloyd--Max optimum, for one hardware reason. A grid whose decode is arithmetic, the uniform integer grid and equally the hardware-native E2M1 float grid of \texttt{MXFP4}/\texttt{NVFP4} (\Cref{sec:live}), needs nothing beyond the shared-scale rescale, whereas a \emph{fitted} codebook must be decoded by a lookup table or a per-element computation, which keeps it off the shared-scale GEMM datapath, leaving it to the weight and KV-cache paths (\Cref{sec:vq,sec:beyond}). The mechanism below is derived for the uniform integer grid, the one the crest factor prices; \Cref{sec:fmt-flip} gives the floating-point exception. The operative question is therefore not whether the uniform grid is optimal, but what governs its error, and that is decided entirely by its one free parameter, the scale.

Two philosophies set the scale $s$ of a uniform group. \emph{AbsMax} (no clipping) pins $s\propto\max_i|x_i|$ so that even the largest value is representable; \emph{MSE-optimal} instead chooses $s$, equivalently a clipping threshold, to minimize $\E[(x-Q(x))^2]$, deliberately sacrificing the rare large values to shrink the step for the bulk. But the field does not clip the outliers \emph{away}, because they are not noise: these extreme values carry output-critical signal, and zeroing the outlier feature dimensions alone degrades a model far more than removing as many random dimensions \citep{dettmers2022llmint8}. Clipping is therefore used as a \emph{complement} to a transform rather than as a substitute for one, as a fitted threshold applied to already-flattened, near-Gaussian values (\Cref{sec:tax-learned-orthogonal,sec:fmt-codesign,sec:guide}). What clipping cannot do on its own is remove the outlier's cost, since it only trades a large error on a few important coordinates for a smaller one on the bulk. A transform, being function-preserving, instead \emph{redistributes} the outlier's energy without discarding any of it, which is why flattening rather than clipping is the mechanism the rest of the survey pursues.

Because the outliers must be represented rather than clipped, the deployed default is AbsMax, and AbsMax is precisely what makes an outlier expensive. Write $M=\max_i|x_i|$ for the group's peak and $\sigma$ for its RMS. The AbsMax step $\Delta=M/(2^{b-1}-1)$ is pinned to the \emph{peak}, yet the signal it must resolve has power $\sigma^2$, set by the \emph{bulk}. The per-element noise power $\Delta^2/12$ (derived in \Cref{sec:bennett}) thus grows with $M^2$ while the signal grows with $\sigma^2$, and everything turns on the ratio between the two,
\begin{equation}
\label{eq:crest}
\mathrm{CF} \;=\; \frac{M}{\sigma}\quad\text{(the \emph{crest factor}, or loading factor)} .
\end{equation}
This ratio is the survey's central object. It is a pure peak-over-typical ($\ell_\infty/\ell_2$) quantity, with a floor of $1$ at a perfectly flat group and a ceiling of $\sqrt G$ for a group of $G$ coordinates (reached when all the energy sits on one). Our sixteen-value running group, with its one outlier, sits at $\mathrm{CF}=3.7$ (\Cref{fig:crest}), already close to the ceiling $\sqrt{16}=4$ that sixteen coordinates allow, though that ceiling grows with the group ($5.7$ at $G=32$, $11.3$ at $G=128$): the same lone outlier in a $128$-coordinate group would give $\mathrm{CF}\approx6.6$, since the bulk it is measured against grows while the peak does not. The outlier channels documented at scale by \citet{dettmers2022llmint8} run an order of magnitude or more above the typical channel \citep{xiao2023smoothquant}, so real groups sit far above our $3.7$. Because the step is pinned to the peak, a single large coordinate lifts $\Delta$ for the whole group and coarsens every ordinary value in it; this is the mechanism, promised in the opener, by which one outlier taxes fifteen innocents. \emph{It is this quantity (not the variance, and not the fourth-moment kurtosis) that the shared-scale grid actually charges for, and that every flattening transform in this survey ultimately targets, whatever surrogate it optimizes} (\Cref{sec:incoherence,sec:inversion}). What we have not yet said is how many bits a given crest factor actually costs. That conversion is the one thing still missing, and the next subsection supplies it, at which point the three-bit penalty of a $\mathrm{CF}=8$ group becomes an $18$\,dB identity.

\subsection{Quantization noise: Bennett's model and SQNR}
\label{sec:bennett}

The crest factor of \Cref{sec:zoo} is so far only a geometric ratio, a property of where a group's energy sits and not yet a statement about error. Two claims were made there and not derived: that the noise power is $\Delta^2/12$ once the grid is fine, and that a group at $\mathrm{CF}=8$ forfeits some three bits. Both presuppose a model of the quantization error: how large it is, and how it shrinks as the grid refines. The high-resolution model of \citet{bennett1948spectra} supplies one. That model is the hinge of the section, the step that turns the \emph{geometry} of a group into a \emph{price}, first in decibels and then in bits, the currency in which every transform we later study is paid.

The model rests on one observation. Fix a single quantizer cell of width $\Delta$. If the grid is fine enough that the input density $p$ barely varies across it, a value landing in the cell is nearly uniform there, so the rounding error $e=Q(x)-x$ is nearly uniform on $[-\Delta/2,\Delta/2]$ and carries power $\int_{-\Delta/2}^{\Delta/2} e^2\,de/\Delta=\Delta^2/12$. That is the origin of the factor of twelve that recurs through the section. Averaging this per-cell power against the density over the granular cells gives the distortion, and because the density charges no mass outside that range no unbounded overload term survives: the content of the lemma.

\begin{lemma}[Bennett's integral \citep{bennett1948spectra}]
\label{lem:bennett}
For a high-resolution quantizer with local step $\Delta(x)$ and input density $p$ whose overload distortion is negligible, as holds in particular when $p$ is supported within the quantizer's granular range, the distortion is $D=\tfrac{1}{12}\int \Delta(x)^2 p(x)\,dx$, the integral taken over that range. For a uniform quantizer ($\Delta$ constant, overload negligible) this reduces to the additive-white-noise model $D=\Delta^2/12$, with quantization error approximately uniform on $[-\Delta/2,\Delta/2]$ and decorrelated from the signal.
\end{lemma}

This additive white-noise picture is what the rest of the section runs on. It is an approximation, exact only as $\Delta\to0$ and merely good at four bits, and it rests on two premises: the same high-resolution premise that justified dropping the second-order term of the proxy in \Cref{sec:notation}, and the negligible-overload premise, which an AbsMax scale satisfies by construction and a fitted clipping threshold does not. Every quantitative claim that follows inherits both caveats, which we flag wherever a result leans on them.

With a noise power in hand, the crest factor's cost is a one-line computation. The signal-to-quantization-noise ratio of a group is its signal power $\sigma^2$ divided by the noise power $\Delta^2/12$; substituting the symmetric AbsMax step $\Delta=M/(2^{b-1}-1)$ from \Cref{sec:zoo} makes the crest factor emerge on its own.

\begin{corollary}[The 6\,dB/bit rule]
\label{cor:6db}
Under the high-resolution noise model of \Cref{lem:bennett}, for a symmetric $b$-bit uniform quantizer with AbsMax scale on a group of RMS $\sigma$ and max $M$, the signal-to-quantization-noise ratio is
\begin{equation}
\mathrm{SQNR}
= \frac{\sigma^2}{\Delta^2/12}
= 12\,(2^{b-1}-1)^2\,\Big(\tfrac{\sigma}{M}\Big)^2 ,
\qquad
\mathrm{SQNR}_{\mathrm{dB}} \approx 6.02\,b + 4.77 - 20\log_{10}\mathrm{CF}.
\end{equation}
\end{corollary}
The decibel form rewards a slow reading, because each of its three terms has a plain meaning. Taking $10\log_{10}$ of the middle expression and writing $2^{b-1}-1\approx2^{b-1}$ splits it into
\begin{equation}
\label{eq:sqnr-split}
\mathrm{SQNR}_{\mathrm{dB}}
\;\approx\;
\underbrace{\big(20\log_{10}2\big)}_{6.02}\,b
\;+\;
\underbrace{\big(10\log_{10}3\big)}_{4.77}
\;-\;
20\log_{10}\mathrm{CF},
\end{equation}
where the constant uses $10\log_{10}12-20\log_{10}2=10\log_{10}3$. The slope $6.02$\,dB is the doubling in the number of levels that each extra bit buys; the constant $4.77$\,dB collects the $\tfrac1{12}$ of the noise model against the signed grid's $2^{b-1}$ levels; and the last term, the one we came for, is the outlier tax. Its consequence is the exchange rate the whole survey runs on: because the crest factor enters only through $-20\log_{10}\mathrm{CF}$, \emph{every doubling of $\mathrm{CF}$ costs the group $6$\,dB, whatever the width $b$}. That $6$\,dB is one full bit in the high-resolution limit, and somewhat less at deployed widths, where the exact corollary's $(2^{b-1}-1)^2$ makes the fourth bit worth $7.4$\,dB rather than $6.02$ (and the fifth $6.6$), so $6$\,dB buys about $0.8$ of a bit at $b=4$. A transform is, in this ledger, a bid to lower $\mathrm{CF}$, its worth quoted in the bits of \Cref{cor:6db} that the reduction buys. The $\mathrm{CF}=8$ group we promised a three-bit loss in \Cref{sec:zoo} is now cashed out: $20\log_{10}8\approx18$\,dB, three times six, which is three bits in the high-resolution limit and about two and a half at $b=4$.

The same exchange rate also fixes where the ``$4$-bit wall'' falls. Because the tax is a fixed number of decibels, its bite is relative to the total budget. A flat $8$-bit group starts near $53$\,dB of SQNR and absorbs an $18$\,dB ($\mathrm{CF}=8$) tax with ample signal to spare, while a flat $4$-bit group starts just under $28$\,dB (the exact corollary; $28.9$\,dB under the $2^{b-1}-1\approx2^{b-1}$ form of \eqref{eq:sqnr-split}), so the same tax consumes most of its budget. The wall is simply where the width-independent tax stops being affordable, which is why $8$-bit quantization is comfortable and $4$-bit is not, and why the transforms that lower $\mathrm{CF}$ matter most at $4$ bits, though they still earn their keep at $8$ wherever activation outliers drive a group's crest to an extreme.

\begin{figure}[t]
\centering
\includegraphics[width=\linewidth]{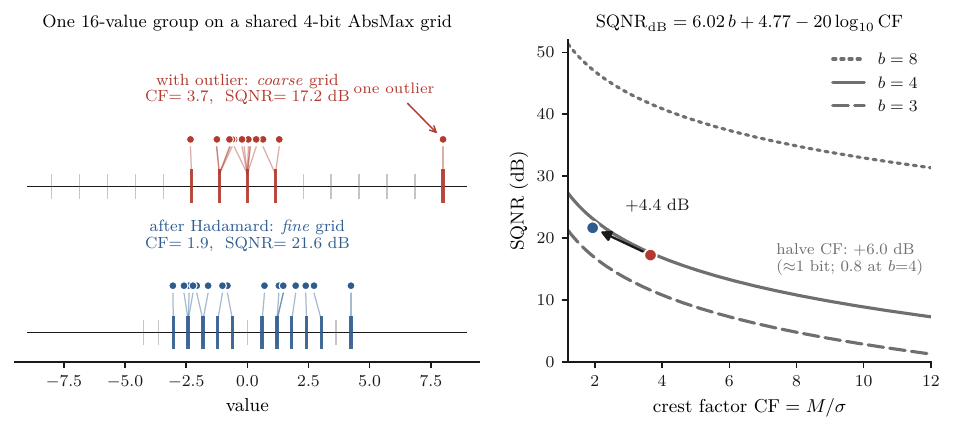}
\caption[The outlier tax and the 6\,dB/bit law]{\textbf{The outlier tax and the
$6$\,dB/bit crest-factor law (\Cref{cor:6db}).} \emph{Left:} one \emph{synthetic} $16$-value
group (fifteen $\mathcal{N}(0,1)$ draws plus one injected outlier) quantized on a shared
$4$-bit AbsMax grid; each true value (dot, above the line) is joined to the grid level it
rounds to (bold tick, on the line), so the slant of the join is its quantization error and
the bold ticks are the levels actually used. The bulk values (near the origin) snap to a
handful of levels ($\mathrm{CF}=3.7$, $\mathrm{SQNR}=17.2$\,dB); after a per-group Hadamard
the grid tightens over the now-flat group ($\mathrm{CF}=1.9$,
$\mathrm{SQNR}=21.6$\,dB). \emph{Right:} the analytic law
$\mathrm{SQNR}_{\mathrm{dB}}=6.02\,b+4.77-20\log_{10}\mathrm{CF}$ for $b\in\{3,4,8\}$. The
two marked points are the \emph{measured} \texttt{INT4} groups from the left panel; they both
sit just below the analytic $b=4$ curve (the flattened group further below): the
$2^{b-1}-1\approx2^{b-1}$ constant simplification shifts the exact curve down by
$\approx1.2$\,dB at $b=4$, and a sub-dB Bennett-model residual of either sign accounts
for the rest.}
\label{fig:crest}
\end{figure}

\Cref{fig:crest} follows the sixteen-value group that recurs through this section. With its lone outlier it sits at $\mathrm{CF}=3.7$ and a measured $\mathrm{SQNR}=17.2$\,dB. A per-group Hadamard (constructed in \Cref{sec:incoherence}) spreads the outlier's energy, drops the crest factor to $1.9$, and lifts the measured SQNR to $21.6$\,dB. The illustration uses the deterministic Hadamard, which happens to flatten this particular group, while the worst-case guarantee of \Cref{prop:rht} needs the random signs. The several decibels the transform recovers are most of the bit that nearly halving $\mathrm{CF}$ is worth by \eqref{eq:sqnr-split}; both measured points sit just below the analytic $b=4$ curve, for the reasons the caption details. Once the quantizer is no longer confined to a uniform grid pinned to the group peak, spending that freedom either on where it places its levels or on how long its codewords are, a different law governs, and it will return to displace this one (\Cref{sec:classical-highrate}).

\subsection{Rate regimes, rate--distortion, and majorization}
\label{sec:rate}

The $6$\,dB law priced an outlier under a shared AbsMax scale, but it left the deeper question hanging: is concentrating a group's energy onto a few coordinates a good idea or a disaster? \Cref{sec:zoo} made it look like a disaster, since concentration is what inflates the crest factor that punishes a shared scale. Yet the entire classical theory of transform coding is built on the opposite conviction, that concentrating energy is the whole point. Both cannot be right in the same setting, and they are not: the answer flips on two properties of whatever quantizer sits downstream: whether it may spend its bits \emph{unequally} across coordinates, and whether each coordinate carries its own scale. This subsection pins both down, recalls just enough classical rate--distortion theory to see why concentration wins once bits may be re-allocated, and then makes the whole opposition exact through one inequality from the theory of majorization.

\begin{definition}[Shared-scale uniform vs.\ allocation-flexible]
\label{def:rate}
We contrast two settings by two freedoms they grant: to allocate bits across coordinates, and to give each coordinate its own scale. A quantizer is \emph{shared-scale uniform} if every coefficient of a group is stored at the same bit-width on a uniform grid with a single shared scale, so no bits can be re-allocated between coordinates: the per-group \texttt{INT4} grid of a deployed low-bit kernel (\texttt{MXFP4} and \texttt{NVFP4} are shared-scale on a \emph{non-uniform} element grid, a second axis taken up in \Cref{sec:format}). It is \emph{allocation-flexible} if the coder may spend a different number of bits on different coordinates under a fixed \emph{total}-rate budget: realized by unequal (but deterministic) bit assignment, by entropy coding of the indices, or both, as in a classical transform coder. \textbf{Every deployed LLM GEMM kernel is shared-scale, fixed-allocation}, for the datapath reasons developed in \Cref{sec:inversion}; the classical transform-coding theory of \Cref{sec:classical} is allocation-flexible.
\end{definition}
\begin{remark}[Terminology]
\label{rem:terminology}
The operative distinction is the conjunction of per-coordinate \emph{allocation} and a
per-coordinate \emph{scale}, the first governing which pole is strictly optimal on the
coding side and the second on the shared-scale side (\Cref{sec:inv-fixedrate}). Allocation
is broader than the textbook fixed- versus variable-\emph{length} (entropy-coding) axis: the KLT optimum below holds under a fixed total rate with optimal deterministic allocation, entropy coding being optional. For brevity we write \emph{fixed-rate} for shared-scale with fixed allocation, whatever the element grid, and \emph{variable-rate} for allocation-flexible, shorthands for that conjunction, not claims about code length; \Cref{prop:lloydmax} above uses \emph{fixed-rate} in the classical fixed-\emph{length} sense, which this shorthand does not disturb.
\end{remark}

The allocation-flexible regime is the native home of classical rate--distortion theory. For a source $X$ with differential entropy $h(X)=-\int p\log_2 p$, the rate--distortion function $R(D)$ is the least rate (bits/sample) achieving expected distortion $\le D$; for independent Gaussians at a fixed total rate the optimum is \emph{reverse water-filling}, which spends more bits on higher-variance coordinates and none at all on those below the water level (\Cref{sec:classical} develops both). Per-coordinate allocation, whether by deterministic unequal bit assignment or by entropy coding, is what a shared-scale kernel forbids.

With allocation on the table we can say why concentration helps. At a \emph{common} per-coordinate rate, with each coordinate quantized by its own scale-invariant quantizer of a common shape, the high-resolution distortion is proportional to $\tr(\Sigma)$, invariant under any orthogonal $U$, so the transform earns nothing (the shared-scale case of \Cref{sec:inversion} is different precisely because one scale must cover the whole group); the gain appears only once bits are re-allocated toward high-variance coordinates, at which point the distortion is governed by the \emph{geometric} mean of the transformed variances rather than their arithmetic mean: minimizing $\sum_k\sigma_k^2\,2^{-2R_k}$ over the per-coordinate rates at fixed total $\sum_k R_k$ gives, by a Lagrange/AM--GM argument and provided every coordinate draws positive rate (\Cref{prop:waterfill}), a distortion $\propto(\prod_k\sigma_k^2)^{1/d}$ with $\sigma_k^2=(U^\top\Sigma U)_{kk}$. That form is what \Cref{prop:schur}(i) consumes and what \Cref{thm:klt} takes as its starting point. The \emph{coding gain}
\begin{equation}
\label{eq:codinggain}
G_{\mathrm{TC}}(U) \;=\; \frac{\tfrac1d\sum_{k}(U^\top \Sigma U)_{kk}}{\big(\prod_{k}(U^\top \Sigma U)_{kk}\big)^{1/d}}
\end{equation}
(arithmetic over geometric mean of the transformed diagonal variances, with $\Sigma$ the source covariance) measures the reduction. By the AM--GM inequality $G_{\mathrm{TC}}\ge1$, with equality iff all transformed variances are equal. $G_{\mathrm{TC}}$ is maximized by making them maximally unequal, i.e.\ by \textbf{concentrating} energy, achieved by the Karhunen--Lo\`eve transform (KLT) that diagonalizes $\Sigma$ (\Cref{sec:classical}). Indeed, by the Schur--Horn theorem the KLT's transformed-variance vector (the eigenvalues of $\Sigma$) majorizes the diagonal of $U^\top\Sigma U$ for every orthogonal $U$. The KLT diagonal is therefore the most concentrated diagonal any orthogonal transform can produce, in the exact majorization sense of \Cref{prop:schur} below. The KLT sits at the allocation-flexible (``variable-rate'') pole.

The opposition between the two poles is a majorization statement, so we fix the vocabulary here.
\begin{definition}[Majorization]
For $a,b\in\R^d_{\ge0}$ with equal sums, $a$ \emph{majorizes} $b$ ($a\succeq b$) if $\sum_{i=1}^k a_{[i]}\ge\sum_{i=1}^k b_{[i]}$ for all $k$, where $x_{[1]}\ge\cdots\ge x_{[d]}$ are the sorted entries. Intuitively $a$ is ``more concentrated'' than $b$.\footnote{A terminology caution held throughout the survey: we use \emph{concentration} in the transform-coding sense: energy packed onto few coordinates. Part of the quantization literature uses the word in the \emph{opposite} sense, for low distributional spread, so that a Hadamard, which we call \emph{flattening}, \emph{increases} their ``concentration'' \citep{federici2026dissecting}. The two senses are inverses; we always mean the former.} A function $f$ is \emph{Schur-convex} if $a\succeq b\Rightarrow f(a)\ge f(b)$, and \emph{Schur-concave} if the inequality reverses. Fixing the partition of the coordinates into scale groups, we write $a\succeq_G b$ ($a$ majorizes $b$ \emph{within groups}) when the two agree on every group's sum and the restriction of $a$ to each group majorizes the restriction of $b$. A block-diagonal doubly stochastic map is doubly stochastic, so $a\succeq_G b$ implies $a\succeq b$, and every Schur-concave function is therefore monotone along $\succeq_G$ as well. The converse fails, and that failure is what matters below: a sum of per-group maxima depends on the grouping, so it is not a functional of $\succeq$ at all, and the opposition of \Cref{prop:schur} is stated along $\succeq_G$.
\end{definition}
Before stating the opposition, we tie the abstract energy vector $v$ back to \Cref{sec:zoo,sec:bennett}. Write $v_k$ for the (transformed) per-coordinate energy; a group's mean energy $\tfrac1G\sum_{k\in g}v_k$ is its $\sigma^2$, while its \emph{peak} energy $\max_{k\in g}v_k$ is the $M^2$ to which an AbsMax scale pins the per-element noise power $\Delta^2/12$ (\Cref{cor:6db}). The variable-rate cost then reads off the whole vector's geometric mean, the shared-scale cost off its per-group peaks, and, as functionals of $v$, the two are extremized at opposite ends of the within-group majorization order.
\begin{proposition}[The two regimes are opposed]
\label{prop:schur}
Fix the total energy $\sum_k v_k$ over the transformed per-coordinate energy vector $v=(v_k)\ge0$. (i)~The variable-rate distortion is governed by the geometric mean $\mathrm{GM}(v)=(\prod_k v_k)^{1/d}$: exactly proportional to it when every coordinate draws positive rate and the per-coordinate quantizers share one shape constant (as under the Gaussian source of \Cref{thm:klt}), and bounded below by the water-filling floor of \Cref{prop:waterfill} when the allocation is not interior; $\mathrm{GM}$ is symmetric and Schur-concave, hence minimized at the most concentrated (majorizing) $v$; equivalently the coding gain \eqref{eq:codinggain}$\,=\mathrm{AM}(v)/\mathrm{GM}(v)$ is Schur-convex. (ii)~The uniform-grid fixed-rate distortion is a sum of per-group maxima, $D_{\mathrm{fr}}=\sum_g\max_{k\in g}v_k$, which by the max-vs-mean bound satisfies $D_{\mathrm{fr}}\ge\frac1G\sum_k v_k$ with equality iff every group is internally \emph{uniform}; it is therefore minimized by within-group flattening and driven to its maximum by concentration. As functionals on the energy vector $v$, then, the two surrogate objectives are extremized at opposite ends of the \emph{within-group} majorization order, at fixed, strictly positive group sums. Because $D_{\mathrm{fr}}$ depends on the grouping it is not permutation-symmetric, hence not a functional of the global order at all, and the across-group placement of energy is a separate lever, the one the permutation family of \Cref{sec:tax-permutation} exploits; which energy vectors a given transform can actually realize, and hence whether these extrema are attainable, is the separate question taken up in \Cref{sec:inversion}.
\end{proposition}
\begin{proof}[Proof sketch]
(i)~At fixed sum, $\prod_k v_k$ is maximal at the flat $v$ and decreases under majorization (equivalently $-\sum_k\log v_k$ is Schur-convex, Hardy--Littlewood--P\'olya), so $\mathrm{GM}$ is Schur-concave and minimized at the concentrated extreme. (ii)~Within each group $\max_{k\in g}v_k\ge\frac1G\sum_{k\in g}v_k$ (max $\ge$ mean), with equality iff the group is uniform; summing over groups gives $D_{\mathrm{fr}}\ge\frac1G\sum_k v_k$, attained at within-group uniformity and maximal ($=\sum_k v_k$) when each group's mass sits on one coordinate. \end{proof}
\Cref{prop:schur} is the mathematical spine of the ``Great Inversion'' (\Cref{sec:inversion}): concentration and flattening are the two extremes of the within-group majorization order, and the rate regime decides which one the deployed objective rewards. It also settles the paradox this subsection opened with. Both intuitions were right, each in its own regime: the classical coder, free to allocate bits, is Schur-concave and minimized at the concentrated pole, so it wants the KLT. The shared-scale kernel is a sum of group maxima, minimized at the flat pole, so it wants the opposite.

Neither half of the opposition is new mathematics. That a Schur-concave rate objective is minimized by the energy-concentrating KLT is a classical result \citep{goyal2001transform}, restated in Schur terms by \citet[whose \emph{fixed-rate} is the fixed-length sense of \Cref{rem:terminology}, its allocation being per-coordinate]{bhadane2021pba}, and that a Schur--Horn majorization argument can be turned into an energy-flattening rotation for shared-scale quantization is by now used in practice \citep{xu2026torq}. The two ingredients that place the two objectives at opposite poles are textbook: a separable sum is Schur-concave for a concave summand (here $\log$) and Schur-convex for a convex one \citep[Prop.~3.C.1]{marshall2011majorization}; and, on a single group's coordinates, the maximum is symmetric and Schur-convex directly from the definition. The sum over groups is therefore monotone under \emph{within-group} majorization, but, depending on the grouping, it is not Schur-convex on the whole vector. What \Cref{prop:schur} contributes is to place both, as opposite ends of one within-group order, at the foundation of the transform taxonomy.

We have now proven what the fixed-rate objective of \Cref{def:rate} rewards: a group as flat as possible, its energy spread until the crest factor sits near its floor of $1$. What we have not shown is that such flatness is \emph{achievable}: cheaply, and without knowing in advance where the outliers lie. That is the next subsection's business, and the answer turns out to be almost suspiciously easy.

\subsection{Incoherence and the flattening law}
\label{sec:incoherence}

Neither a weight matrix nor an activation arrives flat. Both arrive carrying exactly the outliers that motivated the survey, and we do not know in advance which coordinates hold them. Remarkably, we need not know. A single random rotation flattens any vector, whatever its outlier pattern, and does so without inspecting the data at all: a direct consequence of concentration of measure in high dimensions. This subsection makes that precise, and in the process meets the crest factor from the other side: the property a rotation manufactures, \emph{incoherence}, turns out to be nothing but the crest factor read over a whole matrix.

\begin{definition}[$\mu$-incoherence \citep{chee2023quip}]
A matrix $A\in\R^{m\times n}$ is $\mu$-incoherent if $\|A\|_{\max}\le \mu\,\|A\|_F/\sqrt{mn}$; i.e.\ no entry is much larger than the root-mean-square entry. \citet{chee2023quip} pair this with a Hessian form: a symmetric positive-semidefinite $H=Q\Lambda Q^{\!\top}$ is $\mu$-incoherent when its eigenvectors satisfy $|Q_{ij}|\le\mu/\sqrt n$, the convention their proxy-error bound is stated in. Incoherence processing is the application of a transform that makes $W$ incoherent, and, in the activation-quantizing (\texttt{W4A4}) regime, the activations $X$ as well, before quantization.
\end{definition}
Incoherence is a crest factor in disguise: $\mu$ is the ratio of the largest entry to the root-mean-square entry, the same $\ell_\infty/\ell_2$ quantity as $\mathrm{CF}$ in \eqref{eq:crest} (\Cref{fig:crest}), read over a whole matrix rather than one group. Making $A$ incoherent \emph{is} flattening its crest factor, and the $\approx 6$\,dB/bit accounting of \Cref{cor:6db} applies in the same form, at whole-matrix rather than per-group granularity.

\begin{remark}[The second factor: operand alignment]
\label{rem:alignment}
Spread, the crest factor, is the axis a shared scale is most exposed to, and the one this
survey tracks. It is not the only one. \citet{federici2026dissecting} show that, under
negligible clipping and uncorrelated quantization noise, the quantization SNR factors into a
concentration term and an \emph{alignment} term, how far the dominant directions of the weights
and activations agree, and that \emph{no} orthogonal transform can change the second: flattening
is provably powerless on it. Improving alignment requires a non-orthogonal, data-aware map, and
the alignment-optimal transform is a geometric mean of the weight second moment and the
\emph{inverse} activation second moment, the same $\HX$/$\HW$ balancing the whitening family
performs (\Cref{sec:tax-affine}, WUSH). Two consequences run through what follows. Alignment is
part of why the taxonomy must climb past rotation to whitening, and part of why the realized
deployed error remains an open problem (\Cref{sec:open}). It is a distinct quantity from the
joint statistic $\Delta_{\mathrm{FP}}$, a normalized sum over coordinates of the product of the two operands' squared magnitudes, that governs the floating-point sign reversal of \Cref{sec:fmt-flip}: both are properties of the operand \emph{pair} rather than of either
profile alone, but $\Delta_{\mathrm{FP}}$ moves under a common rotation and this factor does not. We organize the survey around spread, because that
is what the shared-scale grid charges for directly and what a transform can move, and we flag
alignment wherever it, and not spread, decides the outcome.
\end{remark}

Why should a rotation flatten anything? Because it turns a spiky vector into one whose coordinates each behave like a small Gaussian sample of the vector's total energy. And the largest of many Gaussians exceeds their typical size only by a $\sqrt{\log}$ factor, not by the arbitrarily large factor a raw outlier can reach. The precise vehicle is the standard maximal bound for sub-Gaussian variables.
\begin{lemma}[Sub-Gaussian max, the $\sqrt{\log G}$ law]
\label{lem:subg}
Let $z_1,\dots,z_G$ be mean-zero sub-Gaussian with proxy variance $\tau^2$. Then $\E\max_k|z_k|\le \tau\sqrt{2\log(2G)}$, and $\Pr[\max_k|z_k|\ge t]\le 2G\exp(-t^2/2\tau^2)$. Consequently, if in addition the group's realized RMS satisfies $\sigma\ge c\,\tau$ for a constant $c>0$, then $\E[\mathrm{CF}]\le c^{-1}\sqrt{2\log(2G)}$, and $\mathrm{CF}\le c^{-1}\sqrt{2\log(2G/\delta)}$ with probability at least $1-\delta$; in either form $\mathrm{CF}=O(\sqrt{\log G})$, the logarithm natural.
\end{lemma}
\begin{proof}[Proof sketch]
Union-bound the $G$ sub-Gaussian tails $\Pr[|z_k|\ge t]\le 2e^{-t^2/2\tau^2}$; integrating the tail (or the standard maximal inequality) gives the expectation bound. Setting $t=\tau\sqrt{2\log(2G/\delta)}$ makes that union bound $\delta$, which is the high-probability form. With $\sigma\ge c\,\tau$, $\mathrm{CF}=O(\sqrt{\log G})$ in either form. \end{proof}

\begin{proposition}[Rotation flattens \citep{suresh2017dme,chee2023quip,tseng2024quipsharp}]
\label{prop:rht}
Let $R$ be either Haar-distributed on the orthogonal group or a randomized Hadamard $R=HD$ with $D$ a diagonal of random signs (a \emph{deterministic} Hadamard does not give the guarantee for an arbitrary $w$; nor does an arbitrary random orthogonal $R$, for instance a random signed permutation, which leaves the crest factor untouched). Then for fixed $w$, the entries of $Rw$ are identically distributed and individually sub-Gaussian with proxy variance $\|w\|_2^2/d$ (dependent, not independent; the union bound needs only the marginals), so by \Cref{lem:subg} the expected crest factor of $Rw$ is at most $\sqrt{2\ln(2d)}=O(\sqrt{\log d})$ regardless of the outlier structure of $w$. A worst-case coherent $w$ (mass on one coordinate, $\mathrm{CF}=\sqrt d$) is thereby reduced to $\mathrm{CF}=O(\sqrt{\log d})$: a reduction by a factor of at least $\sqrt{d/(2\ln 2d)}$ in expectation.
\end{proposition}
\begin{proof}[Proof sketch]
For $R=HD$ with $H$ normalized ($H_{ji}=\pm1/\sqrt d$), $(Rw)_j=\sum_i\varepsilon_i H_{ji}w_i$ is a Rademacher sum, hence sub-Gaussian by Hoeffding's lemma with proxy variance $\sum_i H_{ji}^2w_i^2=\|w\|_2^2/d$; since $\varepsilon_i H_{ji}$ is distributed as $\varepsilon_i/\sqrt d$, the marginals are identical across $j$. For Haar $R$, $(Rw)_j=\|w\|_2u_j$ with $u$ uniform on the sphere, and the standard spherical tail bound gives the same proxy variance. In both cases orthogonality fixes the realized RMS at $\tau=\|w\|_2/\sqrt d$ exactly, so \Cref{lem:subg} applies with $\sigma=\tau$. The two exclusions are one line each: for a \emph{deterministic} $H$ the flat vector $w=H^{\!\top}e_1$ is mapped to $Hw=e_1$, crest factor $\sqrt d$, so no bound can hold for every $w$; and a signed permutation leaves both $\max_i|w_i|$ and $\|w\|_2$ unchanged, hence the crest factor with them. \end{proof}
The part of the guarantee a deployed scale actually needs survives the grouping it imposes. Since the entries of $Rw$ are identically distributed and marginally sub-Gaussian with proxy variance $\|w\|_2^2/d$, \emph{any} $G$-coordinate subset, any group the shared scale actually acts on, is itself sub-Gaussian with that same proxy variance. That already bounds what a shared scale is charged for, since the scale \emph{is} the group
maximum: by \Cref{lem:subg}, $\E\max_{k\in g}|(Rw)_k|\le\tau\sqrt{2\ln(2G)}$ with
$\tau=\|w\|_2/\sqrt d$, for every group and every $w$. A per-group \emph{crest factor} bound
needs more, because it divides by that group's own realized RMS, and \Cref{lem:subg}'s extra
premise $\sigma_g\ge c\,\tau$ is not automatic: for a Sylvester $H$ at $d=2048$, with $w$
placing equal mass on coordinates $1$ and $1025$, the two active Hadamard columns agree
across the whole of the first group, so that group's RMS is $0$ or $\sqrt2\,\tau$ with
probability $\tfrac12$ each and its crest factor is undefined half the time. The mechanism is
cancellation rather than support. For a $G$-aligned group and a Sylvester $H$, two columns
agree across the group up to a common sign exactly when their indices are congruent modulo
$G$, and the group's realized mean energy is
$\sigma_g^2=\tau^2\sum_c (A^{(g)}_c)^2/\|w\|_2^2$, where $A^{(g)}_c=\sum_{i\equiv c\,(G)}s^{(g)}_i w_i$ and
$s^{(g)}_i=\pm1$ is the sign column $i$ of $R$ carries across group $g$, its own sign draw
times its block sign; balance therefore fails downward exactly where
coordinates sharing a class interfere destructively on net, and, because those block signs
change with the group, where they instead reinforce that group sits above $\tau$, the group
average of $\sigma_g^2$ being exactly $\tau^2$. It therefore suffices that $w$ carry at most one
coordinate per class, and then $\sigma_g=\tau$ \emph{exactly}, for every group and every sign
draw, the worst-case coherent $w$ being the extreme instance. For such a \emph{balanced}
group each group's crest factor is $O(\sqrt{\log G})$ by the same lemma, in expectation and
with probability $1-\delta$; otherwise $\sigma_g\ge c\,\tau$ has to be carried as the premise
it is. By \Cref{cor:6db} a crest-factor reduction by a factor $F$ is worth $\log_2 F$ bits of AbsMax precision, in that corollary's asymptotic currency. At whole-vector granularity against a worst-case coherent $w$ that is the $\approx\tfrac12\log_2\!\big(d/(2\ln 2d)\big)$ bits above; at the deployed group granularity it is strictly less, because a group of $G$ coordinates cannot start above the ceiling $\mathrm{CF}\le\sqrt G$ of \eqref{eq:crest}, which caps any gain at $\tfrac12\log_2 G$. For a balanced group that does start at that ceiling, \Cref{lem:subg} bounds the post-rotation crest factor in expectation by $\sqrt{2\ln 2G}$, so the expected gain is at least $\tfrac12\log_2\!\big(G/(2\ln 2G)\big)$: $1.8$ bits at $G=128$ and $1.0$ at $G=32$. On a real operand the rotation delivers what the bound promises; what differs is the starting point. \Cref{fig:reshape} measures $\mathrm{CF}\,28\!\to\!5$ over a whole real $3072\times2048$ activation, about $2.5$ bits at that whole-matrix granularity against the ${\approx}3.2$ a maximally coherent operand reaching the same post-rotation crest factor would gain there. The realized crest factor sits well inside the ${\approx}7.8$ that \Cref{lem:subg} allows with probability at least $0.95$ over all $6.3$M entries once the spread of the token norms (a factor $1.25$ here) is charged against the realized RMS; the $\sqrt{2\ln 2d}\approx4.1$ of \Cref{prop:rht} is the per-token figure, not the whole-matrix one. The remaining $0.7$-bit gap is a starting-point effect: a real operand begins at $\mathrm{CF}\,28$ rather than the $\sqrt{d}\approx45$ of a maximally coherent one. Those bits, weighed against the cost of applying $R$, are the quantitative case for incoherence processing, and that trade is why the Hadamard transform (multiplier-free, $O(d\log d)$; \Cref{sec:live}) is the default fixed-rate transform.

Note that the invariant this guarantee turns on is the crest factor of \Cref{lem:subg}, an $\ell_\infty/\ell_2$ (extreme-value) quantity, not the kurtosis, which does not control an AbsMax objective. The sub-Gaussian marginal bound of \Cref{prop:rht} is all an AbsMax scale needs, and it is less than a codebook needs: \citet{benbasat2026quantizing} show a single randomized Hadamard can leave individual coordinates far from Gaussian on adversarial inputs, and prove that \emph{composing} two brings every marginal within $O(d^{-1/2})$ of a standard Gaussian, and that even two may not suffice for vector quantization, a composition of three being what makes the coordinate covariance decay. The lattice codebooks of \Cref{sec:vq} are deployed with a randomized Hadamard applied once per matrix axis, so along the axis a codebook block spans they see a \emph{single} randomized Hadamard: neither the two-fold marginal guarantee nor the three-fold covariance one is what they rest on. (\citet{federici2026dissecting} associate their concentration term with kurtosis, but define it through the group \emph{range}, the same $\ell_\infty$ quantity the shared scale is pinned to, so the extreme-value reading is the operative one here.)

\Cref{fig:crest} has been drawing this move all along: applied to our sixteen-value running group, a per-group Hadamard spreads the lone outlier's energy and pulls the crest factor from $3.7$ down to $1.9$. That is nearly the halving that \Cref{cor:6db} prices at $\approx6$\,dB, and worth a measured $+4.4$\,dB on this group (\Cref{fig:crest}). The flat pole of \Cref{prop:schur} is therefore not merely optimal but essentially free to reach. One catch has been suppressed throughout, however. We have spoken as if a transform's job is to hand a flattened group to an innocent rounder, but the rounding step is not innocent. Error-feedback rounding also reduces the coupled quantization error, and it may already be capturing much of the benefit the rotation was meant to supply. Whether the transform and the rounder \emph{compose}, or merely \emph{substitute} for one another, is the question we turn to next.

\subsection{Why rounding is not independent: an error-feedback primer}
\label{sec:rounding}

Every step so far has treated rounding as an afterthought: flatten the group, then snap each value to its nearest grid point. But nearest-point rounding (round-to-nearest, RTN) rests on an independence assumption the layer objective \eqref{eq:bilateral-proxy} flatly denies. RTN rounds each coordinate on its own; the proxy, through the off-diagonal of $\HX$, couples them. That coupling is not a nuisance but an opportunity. If rounding one weight will, through $\HX$, worsen the error carried by its neighbors, we can round it and then adjust those neighbors to compensate, spending the coordinates not yet quantized to cancel the damage already committed. This is error-feedback rounding (a second, independent lever on the exact objective a transform reshapes), and its optimal single step is a short exercise in constrained least squares.

\begin{proposition}[Optimal single-weight update; Optimal Brain Surgeon \citep{hassibi1993obs}; quantization form \citep{frantar2022obc}]
\label{prop:obs}
Assume $\HX$ invertible; in practice a small diagonal damping is added before inversion, both because $X^{\!\top}X$ is singular where a layer has dead input channels and, in \citeauthor{frantar2022gptq}'s implementation, because the repeatedly-applied inverse accumulates numerical error at scale (they add $1\%$ of the mean diagonal). Minimizing the layer proxy $\Delta w^\top \HX\,\Delta w$ subject to quantizing coordinate $q$ to value $w_q^{\mathrm q}$ (constraint $e_q^\top\Delta w = w_q^{\mathrm q}-w_q$) gives optimal remaining update
\begin{equation}
\Delta w \;=\; -\,\frac{w_q-w_q^{\mathrm q}}{[\HX^{-1}]_{qq}}\;\HX^{-1} e_q,
\qquad
\text{incurring cost}\quad
\frac{(w_q-w_q^{\mathrm q})^2}{[\HX^{-1}]_{qq}} .
\end{equation}
\end{proposition}
\begin{proof}[Proof sketch]
Write $\delta=w_q^{\mathrm q}-w_q$ for the constrained change. Lagrangian $\Delta w^\top \HX \Delta w + 2\lambda(e_q^\top\Delta w - \delta)$; stationarity gives $\Delta w=-\lambda \HX^{-1}e_q$, and the constraint fixes $\lambda=-\delta/[\HX^{-1}]_{qq}$; substitute back for the cost. \end{proof}
Applying \Cref{prop:obs} greedily, one coordinate at a time and with $\HX$ restricted at each step to the coordinates not yet quantized, is Optimal Brain Compression \citep{frantar2022obc}; GPTQ \citep{frantar2022gptq} is its scalable form, which drops the greedy order for a fixed column order with lazy blocked feedback (\Cref{sec:comp-rounding}). A later section lays out the full algorithm and reports its proven equivalence, when that column order is run from the last coordinate to the first, to Babai's nearest-plane solver on the lattice whose basis is the Cholesky factor of the layer Hessian; GPTQ thereby inherits Babai's error bound in the no-clipping regime (\Cref{sec:comp-rounding,sec:comp-substitution}). The key fact for now is the \emph{composability question}: because error feedback already reduces the coupled quantization error, a transform applied \emph{before} rounding may find that much of its benefit has already been captured by the rounding stage. Quantifying this overlap is a central task of our later composition analysis, and a recurring evaluation pitfall.

Note the boundary of what error feedback can do, though. Error feedback re-rounds within a \emph{fixed} grid, so, with scales frozen before rounding (under static groups; GPTQ's default instead refits each group's scale on the already-compensated weights, applying the same AbsMax rule to the compensated peak), it never escapes the shared AbsMax step's dependence on that peak, hence never the \Cref{cor:6db} noise floor that the group's crest factor sets; compensation that pushes a weight past the frozen range is clipped rather than re-scaled. Error feedback therefore cannot by itself remove the dynamic-range wall. What it lowers is the realized coupled error a transform would otherwise have to prevent. This is the precise sense in which, as \Cref{sec:intro} put it, better rounding does not by itself remove the wall. It also names the last of the levers a transform must compete with. One question remains before the machinery is complete: granting that a transform helps, what does it cost to apply at inference, and can that cost be driven to zero? That is where the section ends.

\subsection{Where a transform can live: absorption, invariance, and cost}
\label{sec:live}

 \Cref{def:fpt} already localized the cost: the weight side $WT^{-1}$ is folded in offline and free, so the entire inference price of the input-axis case is its activation side $XT^\top$. The question that closes the section is how small that price can be made, and the answer ranges from a full matmul per token down to exactly zero, according to the structure of $T$ and where it attaches. A transform may act at five sites, each reshaping a different tensor's exposure to its shared scale: (a)~weight-only, absorbed offline; (b)~the residual stream feeding a linear, online unless absorbed by invariance; (c)~a mid-block intermediate feeding $W_o$ or $W_{\mathrm{down}}$, absorbed only when both sides land on static weights, as the head-wise value rotation does, and otherwise online; (d)~the KV cache; (e)~the attention probability path. What each of them costs is what we take up in turn: the weight-only, residual-stream and mid-block sites in this section, the KV-cache and attention-path sites in the later sections devoted to them.

Whether that price is paid at all comes down to absorption. The activation side $XT^\top$ (the whole of an input-axis transform's inference cost, by \Cref{def:fpt}) falls into three cases:
\begin{itemize}
\item \textbf{Foldable (zero cost).} If $T$ can be pushed through the preceding operation into its weights, $XT^\top$ is never materialized. \Cref{prop:gamma} is the diagonal case; the orthogonal case is the computational-invariance theorem below.
\item \textbf{Online, cheap.} If $T$ is a structured orthogonal map, a Hadamard/Walsh transform, then $XT^\top$ costs $O(d\log d)$ additions per token and \emph{no multiplications} (a butterfly network), which is why fixed-rate methods overwhelmingly choose it \citep{tseng2024quipsharp,ashkboos2024quarot}.
\item \textbf{Online, expensive.} A dense, data-dependent $T$ costs a full $O(d^2)$ matmul per token, comparable to the projection itself, unless it is folded offline or given cheap structure (a Kronecker or block-diagonal factorization). That is the deployment tax on the non-orthogonal transforms of \Cref{sec:tax-affine}.
\end{itemize}

\begin{theorem}[Computational invariance for RMSNorm \citep{ashkboos2024slicegpt}]
\label{thm:invariance}
Let $Q$ be orthogonal and let every norm be in its gain-free form $x\mapsto x/\mathrm{rms}(x)$, the learned per-channel gain having first been folded into the consuming weights (\Cref{prop:gamma}). Because that norm has no mean-subtraction and $\mathrm{rms}(xQ)=\mathrm{rms}(x)$ per token (norm-preservation), carrying the residual stream in the rotated frame $X\mapsto XQ$ computes the same function, provided every weight reading the stream is replaced by $W\!Q$, every weight writing it by $Q^\top W$, and every bias written into the stream by $Q^\top b$ (biases on the reading side are unchanged). Hence a single global orthogonal rotation of the residual stream is foldable into the surrounding weights at zero inference cost.
\end{theorem}
\begin{proof}[Proof sketch]
The rotation places $XQ$ in the residual stream. Row-wise, $\mathrm{RMSNorm}(XQ)=\mathrm{RMSNorm}(X)\,Q$: norm-preservation gives $\mathrm{rms}(xQ)=\mathrm{rms}(x)$ for each token row $x$ (as $Q$ is orthogonal), and with $\gamma$ folded (\Cref{prop:gamma}) the normalization commutes with the rotation. A reading weight $W\!Q$ acting on this normalized rotated stream reproduces the original, $\mathrm{RMSNorm}(XQ)\,(W\!Q)^\top=\mathrm{RMSNorm}(X)\,W^\top$, and a writing weight $Q^\top W$ re-rotates the block output back into the stream. Propagating the rotation through the linear reads/writes and the residual adds (which are linear) thus leaves the block's function invariant. Mean-subtraction would break norm-preservation, which is why the theorem is specific to RMSNorm. \end{proof}
\Cref{thm:invariance} is why orthogonal transforms are the ``free'' family \emph{at the residual-stream site}: there the benefit of \Cref{prop:rht} is had at zero runtime. A mid-block site is free only when both its sides land on static weights, as the head-wise value rotation's do; otherwise it stays online, cheaply for a Hadamard, at a measured cost \Cref{sec:systems} prices. Non-orthogonal maps forfeit this and must justify their online cost: the central tension of \Cref{sec:tax-affine}.

Fixed-rate deployment is concretized by the block-scaled floating-point formats now in silicon, which we reference throughout and fix here, with the formats treated in full in \Cref{sec:format}. An element is a low-precision float (E$e$M$m$: $e$ exponent, $m$ mantissa bits) and a \emph{block} of $G$ elements shares one scale of a prescribed type \citep{rouhani2023microscaling,nvidia2025nvfp4}:
\begin{center}\small
\begin{tabularx}{\linewidth}{@{}lllX@{}}
\toprule
\textbf{Format} & \textbf{element} & \textbf{block scale} & \textbf{note} \\
\midrule
INT4 (per-group) & 4-bit int & FP (arbitrary) & the classical fixed-rate grid; scale is unconstrained \\
MXFP4 & E2M1 (4-bit) & E8M0, $G{=}32$ & block scale is a \emph{power of two} (zero mantissa) \\
NVFP4 & E2M1 (4-bit) & E4M3 (FP8), $G{=}16$ & finer blocks, richer (3-mantissa-bit) scale, plus a per-tensor \texttt{FP32} scale \\
\bottomrule
\end{tabularx}
\end{center}
The scale type is not a detail: an E8M0 power-of-two scale cannot represent an arbitrary group AbsMax, and (as we discuss in \Cref{sec:fmt-flip}) this interacts with the element grid to decide whether flattening helps or hurts: the INT-vs-FP flip. For now the reader needs only that all three are fixed-rate with a per-group shared scale, so in each of them a group's peak-to-typical ratio is what the shared scale must clear; how that ratio is priced depends on the element grid, and the crest-factor mechanism of \Cref{sec:bennett,sec:incoherence} is derived for the uniform one (\texttt{INT4}; see \Cref{sec:fmt-flip} for E2M1).

Quantization error is measured at the layer output \eqref{eq:bilateral-proxy}; under a uniform-grid AbsMax scale that error is governed by one scalar, the crest factor \eqref{eq:crest}, whose price is exactly $6$\,dB per doubling, a bit in the high-resolution limit (\Cref{cor:6db}). Concentration and flattening are the two opposed extremes of the within-group majorization order (\Cref{prop:schur}), and the rate regime decides which the deployed objective rewards. A random rotation comes within a $\sqrt{\log}$ factor of the flat extreme essentially for free (\Cref{prop:rht}); error-feedback rounding contests part of that gain (\Cref{prop:obs}); and an orthogonal transform of the residual stream folds into the surrounding weights at zero inference cost (\Cref{thm:invariance}), while orthogonal maps at mid-block sites stay online but cheap, unless both their sides land on static weights.

With these primitives in hand the paradox we opened with is no longer a paradox but a theorem waiting to be stated. \Cref{sec:classical} follows the classical half (the coders that, free to allocate bits, correctly chose concentration and the KLT), and \Cref{sec:inversion} shows that the deployed low-bit kernel, denied that freedom, is optimized by the exact opposite.

\section{The Classical Inheritance: Transform Coding, 1963--2021}\label{sec:classical}

\Cref{sec:foundations} closed by promising that the classical theory chose \emph{concentration} and that deployment would invert it. This section makes the first half rigorous. Building up from the high-resolution quantization laws, it proves the \emph{concentration pole} of the inversion (that for a coder free to allocate bits across coordinates, the optimal transform packs a signal's energy onto as few of them as possible). It then follows that same theory forward, through its standardization for images and video, into the compression of neural-network weights, where it reaches LLM scale and then goes no further. The high-resolution law fixes what one bit is worth (\Cref{sec:classical-highrate}); entropy coding and the \emph{space-filling gap} fix how close a scalar quantizer can come to the information-theoretic floor (\Cref{sec:classical-entropy}). The KLT and water-filling then prove concentration optimal and say exactly how to spend the bits (\Cref{sec:classical-klt}); two caveats bound the claim, and the same subsection meets the charge that the classical literature had already posed the shared-scale question (\Cref{sec:classical-caveats}). The last two subsections carry the theory into neural-weight coding (\Cref{sec:classical-nn,sec:classical-industry}), where a production, entropy-coded, rate--distortion-optimized codec already exists and stops exactly where the fixed-rate inference datapath begins. Everything here answers to the \emph{allocation-flexible} regime of \Cref{def:rate}; \Cref{sec:inversion} then confronts it with the shared-scale kernel.

\subsection{The high-rate law and the variable-rate distortion}
\label{sec:classical-highrate}

The classical theory is \emph{variable-rate}, and its central quantities are cleanest at high resolution, so we start there; \citet{gray1998quantization} survey this material in full. Recall Bennett's integral (\Cref{lem:bennett}). For a source whose overload distortion is negligible in the sense of \Cref{lem:bennett}, and a quantizer with point density $\lambda(x)$ (reproduction points per unit length, so local step $\Delta(x)\approx 1/\lambda(x)$), the distortion is $D=\tfrac{1}{12}\int \lambda(x)^{-2}p(x)\,dx$, the integral taken over that range.

\begin{proposition}[Panter--Dite optimal point density and the high-rate law \citep{panter1951quantization}]
\label{prop:panterdite}
Assume the high-resolution regime of \Cref{lem:bennett}. For a \emph{fixed-length} (uncoded) quantizer of $R$ bits, with $p$ smooth and light-tailed enough that $\int p^{1/3}$ converges and overload negligible at the rates considered, ($L=2^R$ points, $\int\lambda=L$), minimizing Bennett's integral over $\lambda$ under the count constraint gives the optimal density $\lambda^\star(x)\propto p(x)^{1/3}$ and distortion
\begin{equation}
D_{\mathrm{fl}} \;\approx\; \frac{1}{12}\Big(\int p(x)^{1/3}dx\Big)^{3}\,2^{-2R}.
\end{equation}
For an \emph{entropy-constrained} (variable-length) quantizer, minimizing $D$ at fixed \emph{index entropy} $H$ instead makes the optimal density \emph{uniform} ($\lambda^\star$ constant) \citep{gish1968asymptotically}, giving $D_{\mathrm{ec}}\approx \tfrac{1}{12}2^{2h(X)}2^{-2H}$, where $h(X)$ is the differential entropy (\Cref{sec:rate}).
\end{proposition}
\begin{proof}[Proof sketch]
Fixed-length: Lagrange/H\"older on $\int\lambda^{-2}p$ at fixed $\int\lambda=L$. Entropy-constrained: the index entropy is $H\approx h(X) + \log_2 L + \int p\log_2(\lambda/L)$; minimizing $D$ at fixed $H$ makes $\lambda$ constant, and the constant folds into $2^{2h(X)}$. \end{proof}
Entropy coding changes the accounting outright. With variable-length codewords the price paid is the index \emph{entropy}, not the point count, so the quantizer can afford a uniform grid and let the coder spend the long codewords on the rare cells. The entropy-constrained optimum therefore \emph{flattens} $\lambda$ exactly where the fixed-length optimum \emph{concentrates} it: the first appearance, in miniature, of the concentrate-versus-flatten split that organizes the survey.
Both distortions carry the same $2^{-2R}$ factor, hence the same $6$\,dB/bit slope met in \Cref{cor:6db}, though that corollary states it for the uniform AbsMax grid. What both laws drop, relative to that corollary, is its crest-factor term $-20\log_{10}\mathrm{CF}$: neither $D_{\mathrm{fl}}$ nor $D_{\mathrm{ec}}$ depends on a group's peak, because neither is pinned to an AbsMax scale, the fixed-length one having spent its freedom on the point density instead. What separates the two is a constant that entropy coding removes by assigning shorter codewords to more frequent levels. This contrast is the classical fixed- versus variable-\emph{length} axis, code length for a \emph{single} source, which \Cref{def:rate} treats as one realization of, and narrower than, the per-coordinate bit-\emph{allocation} axis (the KLT and water-filling below) that a shared-scale kernel also forecloses. Both are facets of the classical ``variable-rate'' freedom this survey contrasts with the deployed shared-scale grid.

\subsection{Entropy coding and the space-filling gap}
\label{sec:classical-entropy}

\Cref{prop:panterdite} shows a uniform quantizer is optimal once entropy coding is allowed. How close to the rate--distortion bound does that get?

\begin{theorem}[Gish--Pierce near-optimality \citep{gish1968asymptotically}]
\label{thm:gishpierce}
At high resolution, for a memoryless source and under standard regularity conditions on its density, an entropy-coded uniform scalar quantizer operates within $\tfrac12\log_2\!\frac{2\pi e}{12}\approx 0.255$ bits/sample of the Shannon lower bound on $R(D)$, which the high-resolution regime approaches.
\end{theorem}
That $0.255$ is a purely geometric constant, not a statistical accident, and seeing why explains the entire gap. The entropy-coded uniform quantizer achieves $D_{\mathrm{ec}}=\tfrac{1}{12}2^{2h}2^{-2R}$ (writing $R$ for its entropy-coded rate, which equals the index entropy $H$ of \Cref{prop:panterdite}) while the Shannon lower bound is $D_{\mathrm{SLB}}=\tfrac{1}{2\pi e}2^{2h}2^{-2R}$. The ratio $D_{\mathrm{ec}}/D_{\mathrm{SLB}}=2\pi e/12$ is independent of the source density, the universality that \Cref{thm:gishpierce} asserts. It is the ratio of two \emph{normalized second moments} $G(\Lambda)$: the integral of a cell's squared quantization error, per dimension, divided by the cell's volume raised to the power $1+2/n$ so as to be dimensionless and scale-invariant. An interval, the scalar quantizer's cell, has $G(\Lambda)=\tfrac{1}{12}\approx0.083$; the best cell shape as the dimension grows is a ball, with $G(\Lambda)\to\tfrac{1}{2\pi e}\approx0.059$. The $0.255$ bits is thus the price of tiling space with one-dimensional intervals rather than spheres: a packing-geometry loss, not a statistical one.

This is the \emph{space-filling gap}, and it is what vector and lattice quantizers recover in the limit of large dimension, a finite-dimensional cell taking only part of it ($E_8$ recovers $0.65$\,dB of the $1.53$\,dB those $0.255$ bits are worth, about two fifths of it \citep{conway1982voronoi}). The figure of merit for a \emph{source} coder is the normalized second moment $G(\Lambda)$ of a lattice's Voronoi cell, \emph{not} the packing density that governs \emph{channel} coding \citep{conway1982voronoi}. The two criteria coincide in dimensions one and two (the hexagonal $A_2$ lattice is both densest and lowest-$G$) but part company already at $n=3$, and it is $G$, which falls toward $1/(2\pi e)$ as the dimension grows, that a quantizer must minimize. Conway--Sloane's fast closest-point decoders \citep{conway1982fast} are what make such high-dimensional, low-$G$ cells deployable, replacing an $O(\text{codebook})$ nearest-neighbor search with a closed-form rule costing $O(n)$ for $D_n$, $D_n^*$, $E_n$ and $E_n^*$, $O(n\log n)$ for $A_n$, where a sort dominates, and $O(n^2\log n)$ for $A_n^*$. Trellis-coded quantization \citep{marcellin1990trellis} matches or beats the best lattice quantizers up to dimension 24, within $0.21$\,dB of the distortion-rate bound for a uniform source, by a different route, an expanded codebook whose subsets label the branches of a trellis, so a Viterbi search finds the minimum-distortion path at the encoder while the decoder is a bare table lookup. These are the classical ancestors of the modern weight codebooks: the $E_8$-lattice codebook of QuIP\# and the trellis codebook of QTIP (\Cref{sec:vq}) are \Cref{thm:gishpierce}'s space-filling gain applied to LLM weights, enabled by the incoherence processing of \Cref{sec:incoherence}. Entropy coding and lattices, though, only fix how efficiently one codes a scalar or an \emph{already-chosen} basis; they are silent on \emph{which} basis. Choosing it is the transform's job, and the classical answer is the concentration pole.

\subsection{The KLT, coding gain, and water-filling}
\label{sec:classical-klt}

That answer is a pair of results, and together they are the concentration pole of the inversion. For variable-rate coding of a correlated vector, under the hypotheses the two results below make precise, the optimal transform is the KLT and the optimal bit budget is spent by water-filling. We prove both.

\begin{theorem}[KLT optimality \citep{huang1963block,goyal2001transform}]
\label{thm:klt}
Let $x\sim\mathcal N(0,\Sigma)$ with $\Sigma\succ0$ be coded by an orthogonal transform $U$ followed by independent high-rate quantizers under the optimal logarithmic allocation at total rate $R$, every coordinate receiving positive rate. (Positive definiteness is needed for the logarithmic allocation: a singular $\Sigma$ puts a KLT coordinate at zero variance, so the interior allocation no longer applies and \Cref{prop:waterfill} governs instead.) Among orthogonal $U$, the transform coding gain $G_{\mathrm{TC}}(U)$ of \eqref{eq:codinggain} is maximized, and the distortion minimized, by $U=U_{\mathrm{KLT}}$, the eigenbasis that diagonalizes $\Sigma$.
\end{theorem}
\begin{proof}[Proof sketch]
Under that allocation the high-rate distortion is $D(U)\propto\big(\prod_k (U^\top\Sigma U)_{kk}\big)^{1/d}\,2^{-2R/d}$, so minimizing $D$ maximizes the coding gain \eqref{eq:codinggain}, i.e.\ minimizes $\prod_k(U^\top\Sigma U)_{kk}$ at fixed trace. By Hadamard's inequality $\prod_k(U^\top\Sigma U)_{kk}\ge\det(U^\top\Sigma U)=\det\Sigma$, with equality (for $\Sigma\succ0$) iff $U^\top\Sigma U$ is diagonal, i.e.\ $U$ diagonalizes $\Sigma$; we write $U_{\mathrm{KLT}}$ for any such choice. Equivalently (\Cref{prop:schur}), $-\sum_k\log(U^\top\Sigma U)_{kk}$ is Schur-convex, maximized by the most concentrated diagonal, achieved by the diagonalizing rotation. \end{proof}
The two halves of the theorem rest on different footings. The coding-gain maximization is distribution-free, a statement about $(U,\Sigma)$ alone settled by Hadamard's inequality. The distortion-minimization half needs the rotated marginals to keep their shape, of which the Gaussian case is the familiar instance, because the high-rate distortion carries a per-coordinate quantizer shape constant that is common across coordinates, and so independent of $U$, only under that hypothesis. For non-Gaussian sources the KLT can be strictly suboptimal among orthogonal transforms \citep{effros2004suboptimality}.
For $\Sigma\succ0$ the rotations that saturate Hadamard's inequality are exactly those that diagonalize $\Sigma$: the KLT, up to a permutation and sign flips of its columns when the eigenvalues of $\Sigma$ are distinct, and an entire orthogonal sub-family within each repeated eigenspace when they are not (at $\Sigma=\sigma^2 I$ every rotation saturates it, and there is nothing to concentrate). Fixing any one of them, the KLT does two things at once: it decorrelates the coordinates (which, for a Gaussian source, makes them independent) and it spreads their variances as unequally as possible, the maximal energy concentration that the coding gain \eqref{eq:codinggain} rewards. (Decorrelation equals independence only in the Gaussian case, the caveat of \Cref{sec:classical-caveats}.)

The KLT fixes the basis, but the per-coordinate rates \Cref{thm:klt} took as given are themselves the solution of an allocation problem, and \emph{reverse water-filling} is that solution, reducing to the unclipped logarithmic rule \Cref{thm:klt} assumes exactly when the water level sits strictly below $\min_k\sigma_k^2$.
\begin{proposition}[Reverse water-filling \citep{cover2006elements}; high-rate allocation form as in \citet{goyal2001transform}]
\label{prop:waterfill}
Given transformed variances $\sigma_1^2,\dots,\sigma_d^2$, not all zero, and total rate $R>0$, the real rates $R_k\ge0$ minimizing $\sum_k\sigma_k^2 2^{-2R_k}$ subject to $\sum_k R_k=R$ are $R_k=\max\!\big(0,\tfrac12\log_2(\sigma_k^2/\theta)\big)$, where $\theta>0$ is the unique level solving $\sum_k\max\!\big(0,\tfrac12\log_2(\sigma_k^2/\theta)\big)=R$; coordinates with $\sigma_k^2\le\theta$ receive zero bits.
\end{proposition}
\begin{proof}[Proof sketch]
Minimize $\sum_k\sigma_k^2 2^{-2R_k}$ s.t.\ $\sum R_k=R$, $R_k\ge0$ by KKT; the stationary interior solution equalizes $\sigma_k^2 2^{-2R_k}=\theta$, giving the $\log$ rule, and the non-negativity multiplier zeros out the low-variance coordinates. \end{proof}
Together: \emph{concentrate} the energy (KLT), then allocate bits to where the energy went (water-filling). Both steps require per-coordinate rate: the defining privilege of variable-rate coding, and precisely what fixed-rate hardware removes (\Cref{sec:inversion}).

\subsection{Beyond the linear-Gaussian ideal}
\label{sec:classical-caveats}

Two caveats keep the theory honest. First, the KLT is source-dependent, so its eigenbasis must be estimated and transmitted; it also has no fast algorithm. For a stationary first-order Markov (AR-1) source the eigenvectors of the Toeplitz covariance approach cosines as the correlation $\rho\to1$, so the fixed, signal-independent, $O(n\log n)$ discrete cosine transform \citep{ahmed1974dct} is a near-optimal KLT proxy. Image and video coders accordingly use a \emph{fixed} basis rather than the data-optimal KLT, a direct precedent for the fixed Hadamard's appeal in \Cref{sec:live}. (This fixed-and-data-free versus data-aware tension is the classical rehearsal of the survey's own Hadamard-versus-WUSH choice, though WUSH's transform runs fused, at close to the Hadamard's online cost, \Cref{sec:tax-affine}.) Second, optimality of \emph{linear} transform coding is a high-rate, second-order statement. A linear map removes only second-order correlation, which for a Gaussian is all the dependence there is, but real sources live on curved manifolds whose higher-order structure no linear transform can factorize, so learned \emph{nonlinear} transform codes can beat linear transform coding \citep{balle2020nonlinear}. Both caveats matter for LLMs, whose weight and activation distributions are heavy-tailed and non-stationary; the survey's transforms are the linear, tractable core of a larger design space.

The shared-scale format itself is not a modern invention, and saying so sharpens rather than weakens the inversion. \emph{Block floating point}, in which a run of numbers carries one joint exponent at equal mantissa width, was under formal error analysis from at least \citet{oppenheim1970blockfloat} onward, and \citet{kalliojarvi1996blockfloat} give the closest classical counterpart to the deployed quantizer: they analyze the errors of quantizing \emph{to} a block-floating-point format and report that it beats fixed- and floating-point representation ``with same total number of bits per sample.'' That literature asks a different question of the format than this survey does. Its object is the arithmetic: the roundoff behavior of a filter realization, the error growth of a fixed filter structure, or the number format itself, rather than which transform minimizes the coding loss a shared exponent imposes on a data source. The gap the inversion names is therefore not the format but the transform-design question the format poses, and the format removes a second freedom on top of the per-coordinate rate, the per-coordinate \emph{scale}: \citet{goyal2001transform}'s optimum holds for any bit allocation when every coefficient is quantized by the same scale-invariant family, so that each one's distortion scales with its own variance. That per-coordinate scaling is exactly what a shared scale removes.

The flattening mechanism has a precedent too. \citet{hung1998rotations} rotate \emph{i.i.d.} Laplacian-like data with a Walsh--Hadamard transform, which needs ``only additions and subtractions,'' expressly to ``significantly improve the overload characteristics for quantization,'' the rotation being ``motivated by the geometry of the Laplacian probability distribution.'' The i.i.d.\ hypothesis is what makes this a precursor rather than a coincidence: on a white source a rotation buys nothing by decorrelation or energy compaction, so what remains is distributional reshaping, of which the peak-to-typical ratio is the part a shared scale charges for. \citet{popat1992robust} had earlier reshaped a memoryless source's amplitude distribution before quantizing it, for robustness rather than compaction. The mechanism was then proved outside image coding altogether, in distributed mean estimation: \citet{suresh2017dme} show that a structured random rotation applied before a shared-scale quantizer takes the error from $\Theta(d)$ to $O(\log d)$ with \emph{no} probabilistic assumption on the data, which is \Cref{prop:rht} in all but name, six years before \citet{chee2023quip} reintroduced it to language models. This lineage is live on the information-theory side of the field: \citet{ordentlich2026highrate} cite \citet{hung1998rotations} and \citet{popat1992robust} as prior art for rotation-as-Gaussianization and derive the shared-scale penalty as the peak-to-typical ratio the rotation repairs, and \citet{savkin2025nestquant} cite \citet{hung1998rotations} for the same idea. The rotation papers that define the deployed pipelines, QuaRot, QuIP\#, SpinQuant and FlatQuant, cite none of it, and it is the reading of \Cref{sec:inversion} that makes the connection load-bearing rather than incidental.

\subsection{Neural network weights as a rate--distortion source}
\label{sec:classical-nn}

The template transferred directly to neural-network compression, along two threads. The first is pure coding. Deep Compression \citep{han2016deep} pruned, then learned a $k$-means weight-sharing codebook whose centroids are fine-tuned to the loss, then Huffman-coded the indices. This is entropy-coded scalar quantization applied to weights, with \Cref{thm:gishpierce} pricing the gap such a scheme leaves to the vector-quantization bound. The second thread is \emph{sensitivity weighting}. \citet{choi2017towards} observed that the loss degradation from quantization is, to second order, a Hessian-weighted distortion, simplified there to its diagonal $\Delta w^\top\,\mathrm{diag}(\text{Hessian})\,\Delta w$, so the right objective is not raw weight MSE but a curvature-weighted one, which Choi minimizes with a diagonal-Hessian Lloyd--Max $k$-means, a weighted instance of the alternation of \Cref{prop:lloydmax} and, by his own account, a heuristic rather than an exact solution. Choi's scheme is GPTQ's \emph{sibling}, not its parent. Both schemes are second-order, Hessian-weighted quantization, descending from the curvature-saliency criterion the Optimal Brain Damage and Optimal Brain Surgeon line introduced for pruning \citep{lecun1990obd,hassibi1993obs}, but Choi weights by the diagonal of the \emph{training-loss} Hessian, the assumption OBS was written to drop, whereas GPTQ minimizes the full layerwise \emph{output} error $\|X\,\Delta W^\top\|_F^2$ with the non-diagonal Hessian $2\HX$ (\Cref{sec:rounding}). SqueezeLLM \citep{kim2023squeezellm} carries Choi's idea to LLM scale, replacing the Hessian by its Fisher approximation, and \citet{gao2019rate} supply the theoretical bookend, importing the Shannon rate--distortion function itself to bound a trained model's compressibility. The recurring lesson (that the right error metric is the output/loss-weighted one, not raw weight MSE) is exactly the activation-weighted proxy of \eqref{eq:bilateral-proxy} that the concentration transforms of this part are built to serve.

\subsection{Industrial codification and the LLM-scale descendant}
\label{sec:classical-industry}

The variable-rate pipeline was standardized for neural networks in its own right, and by the same body that standardizes video. DeepCABAC \citep{wiedemann2019deepcabac} carried the context-adaptive binary arithmetic coder (CABAC), the entropy engine of the H.264/AVC and H.265/HEVC video standards, over to weights, pricing each weight by an inverse-variance (diagonal-Fisher) importance in a rate--distortion Lagrangian. DeepCABAC became the entropy-coding core of the ISO/IEC MPEG Neural Network Compression and Representation standard \citep{kirchhoffer2022nnr} (ISO/IEC~15938-17), with an open, standard-compliant implementation in NNCodec \citep{becking2023nncodec}. These are production, entropy-coded, rate--distortion-optimized weight coders, the surviving \emph{coding} half of this section's tradition, and they apply no transform of their own. They compress networks to a small fraction of their size without accuracy loss, yet are almost absent from the LLM quantization citation graph, a disconnect this survey aims to repair.

The most direct LLM-scale continuation of that rate--distortion line, though it inherits the Lagrangian and not the entropy coder, is Radio \citep{young2025radio}, which allocates bit depth by rate--distortion across weight groups up to the $66$--$70$B scale (and argues the method scales to hundreds of billions). Its per-group rule (bit depth $\approx\mathrm{clamp}\big(\tfrac12\log_2(\text{sensitivity}\times\text{variance}/\theta),\,0,\,8\big)$, with a dual level $\theta$ raised until the average-rate budget is met) is the reverse water-filling of \Cref{prop:waterfill} operationalized per weight group at LLM scale: its clip at zero is exactly the proposition's non-negativity, while its ceiling of $8$ bits is an extra deployment constraint the proposition does not impose. Radio itself names transform coding as the natural next step. The reason that next step has not simply subsumed LLM quantization is the subject of \Cref{sec:inversion}: the inference datapath is fixed-rate, so the per-coordinate allocation that makes the KLT (\Cref{thm:klt}) optimal cannot survive into the operand tile the matrix instruction consumes.

In sum, classical transform coding, made rigorous here, prescribes: \emph{concentrate} (the KLT, \Cref{thm:klt}) and allocate bits by water-filling (\Cref{prop:waterfill}), realized, at high resolution, within $0.255$ bits of the Shannon lower bound by an entropy-coded scalar quantizer (\Cref{thm:gishpierce}). The first two steps presume per-coordinate bit allocation under a fixed total rate, whether or not the indices are entropy-coded; the third presumes the entropy coding itself. Either way, every step lives in the allocation-flexible regime of \Cref{def:rate}, and \Cref{sec:inversion} confronts that regime with the deployed shared-scale kernel and derives the opposite prescription.

\section{The Great Inversion and the Optimality Theory}\label{sec:inversion}

The foundations (\Cref{sec:foundations}) equipped us with two things: a precise
account of what fixed- and variable-rate quantizers cost (the crest-factor law
of \Cref{cor:6db}, the majorization geometry of \Cref{prop:schur}), and the
classical verdict that, for a variable-rate coder, the optimal transform
\emph{concentrates} energy (\Cref{thm:klt}). This part confronts that
verdict with the constraint the classical optimality theory never faced: the transform
must run inside a modern inference kernel. We show that this constraint does not
merely weaken the classical prescription; it \emph{reverses} it. The transform
a deployed low-bit LLM wants points the opposite way to the KLT, and the
reversal is not a heuristic observation but a statement, made precise as an
opposition of two surrogate objectives (\Cref{thm:inversion}). We
then use the reversal as a lens: it explains a widespread empirical hazard
(\Cref{sec:inv-proxy}), it organizes the scattered optimality results of the
literature onto two non-communicating sides (\Cref{sec:inv-optimality}), and it
locates the field's true open frontier (\Cref{sec:inv-synthesis}).

The argument relates three objectives that are easily conflated, so we name them once here and keep them distinct throughout. \emph{(i)~The bilateral $L^2$ proxy} \eqref{eq:bilateral-proxy}, $\tr(\Delta W\,\HX\,\Delta W^{\!\top})+\tr(\Delta X\,\HW\,\Delta X^{\!\top})$, the smooth quantity almost every calibration procedure actually optimizes. \emph{(ii)~The shared-scale AbsMax surrogate} $\sum_g M_g^2$ (\Cref{prop:flat-opt}), a high-resolution model of the step size the deployed grid charges. \emph{(iii)~The realized deployed error} (the round-to-nearest layer-output error, and through it end-to-end perplexity), which is what actually ships. The ``inversion'' below opposes the coding surrogate of \Cref{sec:classical}, the variable-rate geometric-mean cost the KLT minimizes, against the shared-scale AbsMax surrogate~(ii). The coding surrogate and~(ii) are opposed under within-group majorization, so their optimal transforms sit at opposite poles. Objective~(i) is the smooth stand-in for~(ii) that calibration actually optimizes, and \Cref{sec:inv-proxy} asks how faithfully stand-ins of that kind rank transforms against~(ii). Whether either surrogate tracks the realized error~(iii) is a separate, open question (\Cref{sec:open}). We name these gaps here rather than re-hedging at every step, and return to them only where a specific result turns on one.

\subsection{Why the deployed kernel is fixed-rate}
\label{sec:inv-fixedrate}

The classical optima of \Cref{sec:classical} are inseparable from one assumption: that the
coder may spend a different number of bits on each coordinate. It is worth
pausing on why that assumption, so natural in signal coding, is simply
unavailable to an LLM inference engine. A quantized matrix multiply reaches
hardware peak only when its operands are laid out uniformly, so that every lane
of a tensor-core or SIMD tile does identical work; and the dequantization must
be a cheap, branch-free map. Both requirements exclude that machinery from the operand tile the matrix instruction consumes, though not from the kernel around it (\Cref{sec:sys-vr}).

\begin{assumption}[The datapath fixes the rate]
\label{prop:datapath}
A quantized GEMM kernel that attains hardware peak throughput requires (i) every element of the operand tile its matrix instruction consumes to arrive at the \emph{same} bit-width, and (ii) at most one shared
scale per contiguous block, the block being the finest granularity the datapath
carries for free (\Cref{sec:live,sec:sys-groupsize}). Neither per-coefficient bit allocation nor entropy-coded indices are expressible in the operand tile that kernel's matrix instruction consumes; a decoder may run inside the kernel, but what it hands the instruction is dense, equal-width, and the quantizer's own reconstruction; the only allocation that could still hide there would restrict some coordinates to a sub-alphabet of that same width, which no deployed coder does (\Cref{sec:sys-vr}). Hence
the dense GEMM at the core of every deployed low-bit LLM linear layer is
shared-scale in the sense of \Cref{def:rate}; mixed-precision schemes that peel
off a few outlier channels \citep{zhao2023atom,ashkboos2023quik} still run their bulk
multiply on exactly such a kernel, and where a scheme does restore per-coordinate allocation it does so outside that instruction's operand tile, by moving coordinates into a different
precision lane (\Cref{sec:tax-learned-orthogonal}). These are the requirements today's fast low-bit kernels are
built around, on the element side \citep{lin2024qserve} and on the block-scale side
\citep{rouhani2023microscaling,nvidia2025nvfp4}; the systems evidence, including the entropy-coded weight coders that must return a dense, equal-width tile to that instruction, whether they decompress before the GEMM, decode tile by tile inside the kernel, or stream a constant-rate hardware decoder, is collected in \Cref{sec:sys-vr}. This is a property of today's tensor cores rather than a theorem
about computation (\Cref{sec:open}).
\end{assumption}

The two block-scaled formats now in silicon (MXFP4 and NVFP4) and the
INT4-per-group grid the software kernels still target (\Cref{sec:live}) are all shared-scale and fixed-allocation in exactly this sense:
equal element width, one scale per block. Only INT4-per-group additionally carries the
\emph{uniform} element grid that the AbsMax surrogate below assumes, a distinction
that becomes the format flip of \Cref{sec:fmt-flip}.
There is nowhere in the operand tile such a kernel's matrix instruction consumes for the reverse water-filling of \Cref{prop:waterfill} to survive, and it is that step that gives the KLT its
advantage: at a common per-coordinate rate, each coordinate carrying its own scale and one quantizer
shape constant across coordinates, every orthogonal transform ties (\Cref{sec:rate}); it is the shared scale that breaks that tie. Removing that step, as the rest of this part shows, does not
leave the classical answer approximately intact; it flips which transform wins.

\subsection{The uniform-grid fixed-rate optimum is flatness}
\label{sec:inv-flatness}

To see the flip, we first compute the fixed-rate cost as a function of the
transformed distribution. Partition a transformed weight (or activation) matrix
into groups of size $G$, and let group $g$ have RMS $\sigma_g$ and maximum
magnitude $M_g$. A $b$-bit AbsMax quantizer on a \emph{uniform} (integer) element grid
sets the group step from the single
largest entry, $\Delta_g=M_g/(2^{b-1}-1)$, so Bennett's model (\Cref{lem:bennett})
gives a group distortion $D_g = G\,\Delta_g^2/12 \propto M_g^2$, and therefore
\begin{equation}
\label{eq:fixedrate-D}
D_{\mathrm{fr}} \;\propto\; \sum_g M_g^2 .
\end{equation}
The uniform-grid hypothesis is load-bearing, because on a non-uniform element grid (the E2M1
grid of \texttt{MXFP4}/\texttt{NVFP4}) the element grid's own contribution is governed not by $\sum_g M_g^2$ but by a joint energy-concentration measure. That change of controlling quantity is the format flip of \Cref{sec:fmt-flip}, and the measure is itself derived for an \emph{ideal} FP grid and has not been evaluated against the deployed formats. The two formats do not behave alike there:
\texttt{MXFP4}'s power-of-two block scale re-imposes an AbsMax-like penalty one
level up, which \texttt{NVFP4}'s mantissa-carrying scale weakens.
Read through \Cref{cor:6db}, the same statement is that each group's SQNR is
$\propto(\sigma_g/M_g)^2=\mathrm{CF}_g^{-2}$: the crest factor introduced in
\Cref{sec:zoo} is the exact per-group penalty. What, then, minimizes
\eqref{eq:fixedrate-D}? The answer is not the classical one.

\begin{proposition}[The uniform-grid fixed-rate optimum is within-group flatness]
\label{prop:flat-opt}
Writing $E=\sum_k v_k^2$ for the total energy over coordinate magnitudes $v_k$,
the objective \eqref{eq:fixedrate-D}, which is the group-maximum objective of
\Cref{prop:schur}(ii) rederived from the quantizer's own step rule, satisfies
\[
D_{\mathrm{fr}}=\sum_g M_g^2 \;\ge\; \frac1G\sum_k v_k^2 \;=\; \frac{E}{G},
\]
with equality iff every group is internally \emph{uniform}. Thus, at fixed $E$, $D_{\mathrm{fr}}$
is minimized precisely by driving every within-group crest factor to its floor;
the \emph{across}-group energy profile is irrelevant once each group is internally
balanced. No within-group concentration step lowers $D_{\mathrm{fr}}$, and any such step that
raises a group's peak $M_g$ raises it strictly; $D_{\mathrm{fr}}\le E$, with equality
when every group's mass sits on a single coordinate.
\end{proposition}
\begin{proof}[Proof sketch]
For each group, $M_g^2=\max_{k\in g}v_k^2\ge\frac1G\sum_{k\in g}v_k^2$ (the maximum
dominates the mean), with equality iff the group is uniform; summing over groups
gives $D_{\mathrm{fr}}\ge E/G$; and $M_g^2\le\sum_{k\in g}v_k^2$ summed over groups gives
$D_{\mathrm{fr}}\le E$. The absolute floor is $\mathrm{CF}_g=1$
(a perfectly uniform group); the bound $\mathrm{CF}_g=O(\sqrt{\log G})$ that a
balanced sub-Gaussian group satisfies (\Cref{lem:subg}, in expectation and with
probability $1-\delta$) is the near-flat regime that incoherence processing
(\Cref{prop:rht}) reaches in practice.
\end{proof}

The contrast with the classical theory could not be sharper. Coding gain rewarded a transform
for making the coordinate energies as \emph{unequal} as possible; the fixed-rate
cost rewards the opposite, penalizing any group whose maximum outruns its bulk.
Where the KLT concentrates, the deployed kernel wants a transform that spreads.

\subsection{The Great Inversion}
\label{sec:inv-duality}

\begin{figure}[t]
\centering
\includegraphics[width=\linewidth]{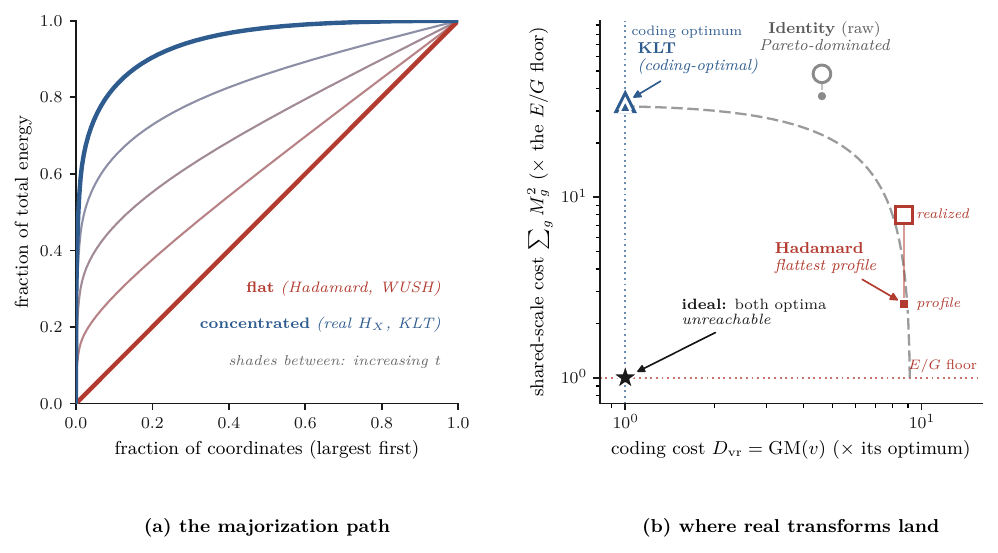}
\caption[The Great Inversion]{\textbf{The Great Inversion (\Cref{thm:inversion}): one majorization axis, two opposed
costs, computed by us on the real $H_X$ of the layer-$17$ $q$-projection of
\texttt{TinyLlama-1.1B} over $3072$ tokens.} \textbf{(a)} A single fixed-energy vector
morphed from internally \emph{flat} to \emph{concentrated} (the measured $H_X$ eigenvalue
spectrum, whose single largest coordinate already holds a quarter of the energy), drawn as
Lorenz curves: height at coordinate-fraction $x$ is the energy share of the $x$ largest
coordinates, so the flat pole is the diagonal and concentration bows the curve toward the
top-left. \textbf{(b)} Where three energy-preserving transforms actually land on the two
costs of \Cref{thm:inversion}, each normalized by its own reference value: the
shared-scale AbsMax cost $\sum_g M_g^2$ at group size $G=128$, as a multiple of the flat floor
$E/G$ (\Cref{prop:flat-opt}), which is a lower bound neither reading attains here,
against the coding cost $D_{\mathrm{vr}}=\mathrm{GM}(v)$, as a multiple of the optimum the KLT
does attain (\Cref{thm:klt}). Every transform appears
twice on the shared-scale axis: the \emph{second-moment} profile cost
$D_{\mathrm{fr}}=\sum_g\max_{k\in g}v_k$ (filled) and the \emph{realized} per-token cost an
AbsMax kernel pays, $\E_t\sum_g\max_{k\in g}(Tx_t)_k^2$ (open), joined by the inflation
between them, which Jensen's inequality keeps positive: a mean of group maxima dominates the
maximum of the per-coordinate means. That inflation is largest for the \textbf{Hadamard}
($3.1\times$, its pair sitting at $2.6\times$ and $8.0\times$ the $E/G$ floor, against
$1.3\times$ for Identity and $1.05\times$ for the KLT), so the
surrogate flatters the flattening pole most; \Cref{lem:subg} bounds inflations of this kind
in amplitude, this ratio being in its square. The \textbf{KLT} sits exactly on the
coding-optimal wall and the randomized \textbf{Hadamard} nearest the flat floor on either
reading, opposite ends of the majorization path (gray dashed, drawn in profile space). The plotted path is a \emph{global} majorization chain, along which the two costs do move oppositely; \Cref{thm:inversion}'s step-by-step guarantee is the within-group one, and \textbf{Identity}, though majorized by the KLT profile, still carries the higher shared-scale cost. The
both-optimal corner ($\star$) is unreachable for this measured spectrum
(\Cref{rem:inv-scope}). Each point is placed
in its transform's own coordinate order, for the KLT the conventional decreasing-eigenvalue
one. The three-transform ordering is the same on both readings, and unchanged at the other group sizes we ran, $G\in\{32,64\}$. This is one layer of one $1.1$B model (the base checkpoint \texttt{TinyLlama-1.1B-intermediate-step-1431k-3T}, $3072$ tokens (six windows of $512$, capture seed fixed) from the \texttt{wikitext-2-raw-v1} train split, activations cached in \texttt{fp16}, figure arithmetic in \texttt{fp64}, the Hadamard sign seed fixed), so it illustrates the two costs rather than measuring their magnitude at deployment scale, and whether the gap widens with model size we do not test.}
\label{fig:inversion}
\end{figure}

We can now state the reversal precisely. The reversal is not a soft ``different
regimes favor different heuristics'' remark, but neither is it a claim that one
transform is globally optimal. The statement is, rather, that the two surrogate objectives
are extremized at opposite ends of the within-group majorization order over the transformed
energy profile $v=(v_k)\ge0$ at fixed total energy $E=\sum_k v_k$. (Here and in
\Cref{prop:schur} $v_k$ denotes coordinate \emph{energy}, the squared magnitude
of \Cref{prop:flat-opt}, so that $\sum_g\max_{k\in g}v_k$ is exactly the
$\sum_g M_g^2$ of \eqref{eq:fixedrate-D}. Read instead as a \emph{second-moment}
profile, which is what \Cref{fig:inversion}'s filled markers evaluate while its open markers
evaluate the realized cost itself, the identification holds only
up to a within-group extreme-value inflation of the kind \Cref{lem:subg} bounds in
amplitude, since the largest per-coordinate second moment in a group understates the mean
of that group's realized maxima.) We are careful
to separate this exact statement about objectives from the (harder, model- and
grouping-dependent) question of which transform attains them.

\begin{theorem}[Inversion of the surrogate objectives]
\label{thm:inversion}
Regard both distortions as functionals of the transformed energy vector $v$ at
fixed $E=\sum_k v_k$, the shared-scale one on a \emph{uniform} grid.
\begin{itemize}
\item The \textbf{allocation-flexible} coding surrogate (\Cref{sec:classical})
$D_{\mathrm{vr}}(v)\propto(\prod_k v_k)^{1/d}$ is the geometric mean, the form the
variable-rate distortion takes while every coordinate draws positive rate
(\Cref{prop:schur}(i); once the water level of \Cref{prop:waterfill} rises above a
coordinate's energy the true distortion is floored by the un-coded coordinates' energy,
so the vanishing limit below is an idealization of the surrogate). It is strictly
Schur-\emph{concave} wherever all coordinates are positive and identically zero elsewhere; it decreases under majorization, so it is driven down by
\emph{concentration}, the direction in which the KLT moves the profile, reaching the
majorization-maximal \emph{reachable} one, the spectrum (\Cref{thm:klt}).
\item The \textbf{shared-scale AbsMax} surrogate (\Cref{sec:inv-flatness})
$D_{\mathrm{fr}}(v)=\sum_g\max_{k\in g}v_k$ obeys the max-vs-mean bound of
\Cref{prop:schur}(ii), so among all $v$ of fixed sum it is minimized at the
\emph{flat} $v$, the direction in which incoherence processing and the Hadamard move the
profile (\Cref{prop:rht}), and maximized under within-group concentration.
\end{itemize}
Over all $v$ of fixed total energy, then, the two surrogates are driven apart. The
geometric mean is strictly Schur-concave on the positive orthant, so every majorization
step toward concentration strictly lowers $D_{\mathrm{vr}}$ while every coordinate remains
positive, and it is maximized at the flat $v$;
while $E/G\le D_{\mathrm{fr}}\le E$, the lower bound attained exactly when every group
is internally uniform and the upper bound when each group's mass sits on a single
coordinate. Concentrating energy \emph{within} a group therefore strictly lowers the first whenever every coordinate remains positive, and never lowers the second, strictly raising it as soon as the group maximum rises; flattening within groups does the reverse. Along no \emph{within-group} step of the order do the two objectives move the same way.

\end{theorem}
\begin{proof}[Proof sketch]
For the coding surrogate, $\prod_k v_k$ at fixed $\sum_k v_k$ is maximal at the
flat vector (AM--GM) and falls strictly under majorization, so $\mathrm{GM}(v)$ is
strictly Schur-concave (\Cref{prop:schur}(i)) and concentration lowers it, to $0$ once
any coordinate vanishes. For the AbsMax surrogate, the max-vs-mean bound
(\Cref{prop:schur}(ii)) gives $D_{\mathrm{fr}}\ge E/G$ with equality iff every group is
internally uniform, while $\max_{k\in g}v_k\le\sum_{k\in g}v_k$ gives
$D_{\mathrm{fr}}\le E$ with equality iff no group carries mass on more than one coordinate.
Within a group the maximum is Schur-convex, so a within-group majorization step weakly
raises $D_{\mathrm{fr}}$ while strictly lowering $\mathrm{GM}$: the two move in opposite
directions. It is not permutation-symmetric, however
($v=(4,1,3,1)$ and $v'=(4,3,1,1)$ give $7$ and $5$ at $G=2$), hence not Schur-convex on
the fixed-energy simplex; and it is not even strictly monotone along majorization, since
any across-group redistribution preserving each group's maximum leaves it unchanged. \end{proof}

One degeneracy is worth naming rather than hiding. Because $D_{\mathrm{vr}}$ vanishes
identically wherever any coordinate is zero, a within-group-flat vector that carries
zeros (for instance $v=(E/2,E/2,0,0)$ at $G=2$) minimizes both surrogates at once.
That is a degeneracy of the geometric mean on the boundary of the simplex, not a regime
in which one transform serves both objectives; requiring every group sum to be strictly positive removes it (\Cref{prop:schur}), and
unlike $v_k>0$ that constraint leaves $D_{\mathrm{fr}}$'s maximum attained; restricting to $v_k>0$ would
also remove it, at the cost of turning $D_{\mathrm{vr}}$'s extreme into an infimum approached
only as a coordinate vanishes. The claim is therefore an opposition of \emph{directions} rather than a separation of minimizers; and because $D_{\mathrm{fr}}$ is grouping-dependent and not permutation-symmetric, it is not monotone along every majorization step. Which profiles an
admissible transform can actually \emph{reach}, and when the two reachable optima can
coincide, is a separate question, taken up in \Cref{rem:inv-scope}.

\begin{remark}[What the theorem does and does not say]
\label{rem:inv-scope}
\Cref{thm:inversion} is a statement about two objectives, not a proof that the KLT
and the Hadamard are the optimal \emph{transforms}. The energy vectors $v$
reachable from a given $W,X,\Sigma$ by an admissible transform form a constrained
set: the diagonal of a fixed-spectrum covariance is majorization-bounded by its
eigenvalues (Schur--Horn) on the coding side, and a single map must act jointly on
all rows, columns, and groups on the AbsMax side, so exact within-group flatness is
generally unattainable for all of them at once. What transfers rigorously is the
pole structure, not a step-by-step monotonicity: concentrating a group's energy
\emph{within} the group helps the coding surrogate and never lowers
$D_{\mathrm{fr}}$, strictly raising it as soon as the group maximum rises (\Cref{prop:flat-opt}), whereas the \emph{across}-group profile is
free, so a reshaping that merely moves energy between groups can change the coding
surrogate while leaving $D_{\mathrm{fr}}$ untouched. Whether the two \emph{reachable}
optima are disjoint therefore turns on how the spectrum sits relative to the
grouping, and whiteness is not the criterion: by Schur--Horn the coding surrogate's
reachable minimizer is the spectrum itself, and were that spectrum constant across
each scale group (plateaus of multiplicity $G$ aligned with the group boundaries),
the sorted KLT would be simultaneously the exact minimizer of $D_{\mathrm{vr}}$ and
sit on the floor $E/G$, at arbitrarily large condition number. For a generic
spectrum, carrying no such aligned degeneracies (the case for the measured $\HX$ of \Cref{fig:inversion}), the two minimizer sets are disjoint; whiteness is only the extreme case, in
which concentration is unavailable at all. The randomized Hadamard realizes the flat pole only
in the high-probability, near-incoherence sense of \Cref{prop:rht}, not as an exact
minimizer of $\sum_g\max_{k\in g}v_k$; and WUSH (\Cref{thm:wush}) does not minimize
surrogate~(ii) either. It optimizes a distinct \emph{stochastic} AbsMax functional
that also charges the inverse map, and attains that model's optimum exactly on the
floating-point grid and only within a $d^{o(1)}$ factor, $d$ the block size, on the integer grid for
Gaussian or Laplacian data
(developed in \Cref{sec:tax-affine}).
One last limit is of \emph{domain} rather than of reachability. Majorization compares vectors
of equal total energy, so this axis prices only what a linear \emph{homogeneous} map does to a
fixed energy budget. It does not price \emph{location}: a per-channel shift changes
$\sum_i v_i$ itself, and so is not a move along the order at all. Nor does it price
\emph{clipping}, which trades overload against a smaller step rather than redistributing
energy. Both matter at deployment, and on the numbers we transcribe they can dominate what a
flattening rotation leaves behind (\Cref{sec:tax-affine}): they are levers beside the axis,
not points on it.
\end{remark}

\begin{remark}[Domain of the inversion, and what is an extension rather than a corollary]
\label{rem:inv-domain}
\Cref{thm:inversion}'s shared-scale surrogate is the AbsMax cost of a \emph{uniform} grid
carrying one scale per group, \eqref{eq:fixedrate-D}, and that is the regime the theorem
covers. It is also the only regime in which flattening is proved optimal against this
shared-scale AbsMax cost itself: row~5's guarantee is a proof too, but against a per-vector
$\ell_2$ MSE on a companded grid, and row~10's against a stochastic \emph{model} of the AbsMax
cost rather than the cost itself; the extensions below rest on a model or a measurement. Four extensions differ in what governs them and in their
evidential status; \Cref{tab:proven} gives the two format extensions
rows of their own, and the KV-cache none at all.
On an \emph{ideal} floating-point grid the governing quantity is the joint concentration
$\Delta_{\mathrm{FP}}$ of the operand \emph{pair} rather than either operand's spread, and the
prescription can reverse (row~8; \Cref{sec:fmt-flip}). On a \emph{deployed} block-scaled
format the global-versus-block-confined \texttt{MXFP4} contrast is measured, and this
survey's surrogate puts both families at one ideal floor without being computed here on the
two deployed maps, and no published account reduces the contrast to a comparable statistic
either (row~9; an open problem of \Cref{sec:open}). For a
\emph{lattice} or vector codebook the guarantee runs through a Gaussianization and a
near-independence that a single Hadamard renders approximate rather than exact
(\Cref{sec:vq,sec:incoherence}); the companded \emph{scalar} case, by contrast, is proved
(row~5). And in
the \emph{KV-cache} the reported wins either bundle the transform with an axis choice, a
nonlinear codebook, attention-sink handling and outlier retention, or, where a transform
\emph{is} isolated (OSCAR against a data-agnostic Hadamard), answer to the attention product's
distortion rather than to the shared-scale AbsMax cost this ledger indexes, so no flattening
theorem accounts for them (\Cref{sec:beyond-kv,sec:emp-kv}). We therefore state the inversion for the uniform grid and
treat the rest as extensions, not corollaries.
\end{remark}

This is the ``Great Inversion'' (\Cref{fig:inversion}), and it is the spine of the survey. The very
quantity that six decades of transform coding taught us to concentrate is the
one a deployed kernel demands we flatten. The transform taxonomy that follows is best read as a map of this primary axis, which fixes the pole a method aims for rather than certifying the method, running from the concentrated pole (the KLT,
serving coding gain) to the flat pole (the Hadamard, refined at the model-optimum by WUSH's data-aware whitening,
serving the AbsMax kernel). Before we tour it, two further points
deserve their own treatment, because each has repeatedly misled the field: one a
direct consequence of the inversion (when the $L^2$ proxy lies), the other a
bridge from single-layer error to end-to-end perplexity.

\subsection{A cautionary consequence: when the L2 proxy lies}
\label{sec:inv-proxy}

The AbsMax objective \eqref{eq:fixedrate-D} is non-smooth in the transform, so
almost every analysis, and many calibration procedures, optimize a
smooth $L^2$ surrogate built from group RMS $\sigma_g$ instead of the group
maximum $M_g$ the hardware actually charges. This substitution is usually made
without comment. The inversion tells us exactly when it is safe, and when it
quietly inverts the conclusion.

\begin{proposition}[A sufficient condition for proxy-faithfulness]
\label{prop:proxy}
Let transforms $A,B$ act on the same operand under a common grouping and produce group-RMS profiles with $\sum_g\sigma_g^2=S_A\ge S_B$ and per-group crest factors $M_g=\mathrm{CF}_g\,\sigma_g$ all lying in a band
$\mathrm{CF}_g\in[\underline c,\bar c]$ (necessarily $1\le\underline c\le\bar c\le\sqrt G$
for a group of size $G$). If that band is tight relative to the proxy gap,
$\bar c^2/\underline c^2 < S_A/S_B$, then the $L^2$ proxy $\sum_g\sigma_g^2$ and the
AbsMax step-size surrogate $\sum_g M_g^2$ rank $A$ and $B$ the \emph{same} way. When
the crest-factor profiles differ enough to break this condition, which a
concentrating transform induces, the ranking can \emph{invert}. Note that the aggregate $\sum_g\sigma_g^2$ is the operand's Frobenius energy divided by $G$, hence unchanged by any orthogonal map: between two rotations $S_A=S_B$ and the sufficient condition is unavailable, so this proposition bites only where the transforms change that energy. What a concentrating rotation still changes is the crest-factor band, and that is the route by which the two surrogates come apart. 
\end{proposition}
\begin{proof}[Proof sketch]
With $\mathrm{CF}_g\in[\underline c,\bar c]$ for both transforms,
$\underline c^2\,S\le\sum_g M_g^2\le\bar c^2\,S$; the gap condition then gives
$\sum_g M_g^2(A)\ge\underline c^2 S_A>\bar c^2 S_B\ge\sum_g M_g^2(B)$, preserving the
order. A concentrating transform violates the premise by creating a few heavy-$\mathrm{CF}$
groups while the rest sit near the floor, spreading the crest factors across the
whole admissible range $[1,\sqrt G]$: bounded, but wide enough to decouple
$\sum_g M_g^2$ from $\sum_g\sigma_g^2$, so the proxy-optimal transform can be the
surrogate-\emph{pessimal} one.
\end{proof}

It is worth being explicit about which statistic controls this, because a natural
shortcut gets it wrong.

\begin{remark}[Crest factor, not kurtosis]
\label{rem:crest}
The governing quantity is an extreme-value ratio, the crest factor $M_g/\sigma_g$
($\ell_\infty$ over $\ell_2$), not the fourth-moment kurtosis. An AbsMax scale is
pinned by the single largest magnitude, so it responds to the maximum, which
kurtosis does not determine: the eight-point groups $(3,4,4,4,4,4,4,4)$ and
$(0,\dots,0,1)$ share the central kurtosis $43/7$ yet sit at the two ends of the
admissible band, $\mathrm{CF}=1.03$ and $\mathrm{CF}=\sqrt8$. This matches the SQNR decomposition of \citet{federici2026dissecting}, derived as their Theorem~2.4 under negligible clipping and uncorrelated quantization noise, whose ``concentration'' term is the squared $\ell_2$-over-range ratio, proportional at fixed group size to the reciprocal of the squared crest factor of \Cref{prop:flat-opt} rather than to a fourth moment. (That work itself notes its concentration is closely related to
kurtosis; what carries the present argument is definitional: the AbsMax scale is
pinned by the range, an extreme-value statistic, not by a fourth moment.)
\end{remark}

\Cref{prop:proxy} bounds when the two agree; the trap is what happens outside that
bound. A transform tuned to an $L^2$ or Frobenius proxy, the right objective for the
compression problem of the classical theory, can be measurably worse on the deployed
benchmark, and it fails for a principled reason. Concentration minimizes the geometric
mean of \Cref{prop:schur}(i) while spreading the crest factors that \Cref{prop:schur}(ii)
charges for, and it is that spread, not the summed energy, that the deployed AbsMax surrogate charges for. We
return to this as an evaluation pitfall in \Cref{sec:pitfalls}; here it is enough to see
that it is a direct corollary of the inversion.

\subsection{From layer error to model error}
\label{sec:inv-linearity}

One gap remains between the theory and practice. Everything above optimizes a
single layer's error, whereas what a practitioner measures is end-to-end
perplexity. A linearity result connects the two. As its
proof makes precise, however, that result connects perplexity to each layer's relative
\emph{weight} error. The weight-side term $\tr(\Delta W\,\HX\,\Delta W^{\!\top})$ of \eqref{eq:bilateral-proxy}
matches that relative weight error only when the activation second moment $\HX$ is isotropic (otherwise the output
proxy carries only a within-layer warrant, the OBS/GPTQ argument of
\Cref{sec:rounding}, which prices a layer's own output error but does not itself
reach end-to-end perplexity).

\begin{theorem}[Linearity \citep{malinovskii2025higgs}]
\label{thm:linearity}
To leading order in the per-layer quantization perturbations (the first-order
term vanishing at the trained optimum), the increase in the
model's perplexity decomposes as a sum of per-layer
contributions, each a layer-specific constant times that layer's relative
squared weight error $\|\Delta W_l\|_F^2/\|W_l\|_F^2$. Minimizing each per-layer
error is therefore aligned, in the small-perturbation regime, with minimizing
end-to-end degradation.
\end{theorem}
\begin{proof}[Proof sketch]
Taylor-expand the loss in the stacked per-layer perturbation; the first-order term
vanishes at a trained optimum. Two approximations then yield the additive form:
(a) the cross-layer blocks of the loss Hessian are dropped (a block-diagonal,
independent-layer approximation: the empirical content of the result), giving a
sum of per-layer quadratics; and (b) within each block the (norm-normalized) \emph{weight-space}
Hessian is treated as isotropic (a scalar multiple of the identity per layer,
the assumption \citet{malinovskii2025higgs} state and validate empirically)
so each per-layer quadratic collapses to the relative Frobenius weight
error $\|\Delta W_l\|_F^2/\|W_l\|_F^2$. This is a weight-only quantity: the input
second moment $\HX$ does not enter, and \citet{malinovskii2025higgs} are
explicit that the theorem ``has no direct bearing on the data-aware layer-wise
MSE'' $\tr(\Delta W\,\HX\,\Delta W^\top)$. That data-aware output proxy of
\eqref{eq:bilateral-proxy} is instead licensed one layer at a time by the
separate OBS/GPTQ argument, for which it \emph{is} the layer output MSE (with
Hessian $2\HX$, \Cref{prop:obs}); the two coincide, up to per-layer constants,
only when $\HX$ is isotropic. The theorem prices weight perturbations only, so the
activation-side term of \eqref{eq:bilateral-proxy} has no counterpart in it.
\end{proof}

The theorem supplies a local justification, under these approximations, for minimizing
per-layer relative weight error one layer at a time. Its hypotheses also mark its limits: the expansion is truncated after the quadratic term, so at
aggressive bit-widths, where the perturbation is no longer small, it can
fray, and a single-layer proxy may then rank transforms in the \emph{opposite}
order to full-model perplexity. That failure mode, together with the proxy trap
of \Cref{prop:proxy}, is why the evaluation practices of \Cref{sec:pitfalls} matter as much
as the theory.

\subsection{The optimality landscape, read through the inversion}
\label{sec:inv-optimality}

With the inversion in hand, the optimality results scattered across the
literature fall into a clean two-column picture: each proves optimality
against a specific objective, on a specific side, and
none transfers across the divide. We state the load-bearing theorems here and
collect the principal results, with the objective each optimizes, in the master
optimality table (\Cref{tab:proven}). The rounding- and format-specific machinery that
some of them invoke is developed in our treatment of rounding and number formats (\Cref{sec:composition,sec:format}).

Take first the side the deployed kernel cannot reach, which holds the concentration optimum of the variable-rate coder and, alongside it, a lattice optimum that is neither concentrating nor variable-rate.
The KLT (\Cref{thm:klt}) is optimal for the allocation-flexible coding objective. Nested-lattice
quantization \citep{ordentlich2024optimal}, for i.i.d.\ Gaussian matrices, asymptotically achieves the rate--distortion bound for the inner-product distortion at a fixed rate per entry, a coding-theoretic optimum realized by nested lattices rather than by an AbsMax kernel. The first relies on per-coordinate rate allocation, the second on a lattice decoder no AbsMax GEMM can run: either way, a privilege the deployed kernel does not have.

On the flattening side, home to the deployed fixed-rate kernel, the guarantees
instead track flatness. Incoherence processing bounds the fixed-rate
error through the coherence $\mu$ \citep{chee2023quip}; a randomized Hadamard followed by a dithered Lloyd/Gaussian-companded scalar quantizer attains the same leading \emph{MSE} constant that was proved for a fully random rotation, up to an $o(1)$ in the bit-width \citep{feng2026provable};
block rotations admit non-asymptotic outlier-suppression bounds
\citep{sanjeet2026mixquant}. The benefit is also \emph{format-dependent}, an
important subtlety we take up in \Cref{sec:format}, being essential for the
INT-AbsMax grid (INT4 elements, AbsMax scale; \Cref{sec:live}) yet far weaker on
the FP-AbsMax grid (E2M1 elements). In the high-rate analysis of
\citet{ordentlich2026highrate}, on the weight--activation pairs they test, the FP-grid benefit is at best neutral and
sometimes harmful, while \Cref{tab:flip} shows the weaker, still-positive half of
the contrast on real activation groups. It is format-dependent \emph{within} FP4 as well: a
block-confined rotation helps \texttt{MXFP4}, whose
power-of-two block scale re-imposes an integer-like penalty, while being roughly neutral (mildly harmful with plain RTN) on \texttt{NVFP4}, whereas the global
integer-style Hadamard craters \texttt{MXFP4} and is roughly neutral on \texttt{NVFP4}
(\Cref{sec:format,tab:res_fp4}) \citep{egiazarian2025mrgptq,chen2025wush}. And the two sides do meet at
one point: water-filling gives the rate--distortion-optimal allocation, which at high rate reduces to
the equal-rate allocation implicit in GPTQ exactly when the Cholesky pivots of $\Sigma_X$
are equal, and otherwise dominates it by $\tfrac12\log$ of their AM/GM ratio
(\Cref{tab:proven}). On the measured
\texttt{Llama-3-8B} covariances of \citet{ordentlich2026highrate2}, GPTQ with a random rotation already lands within about $0.1$ bit of WaterSIC, the scalar-INT realization of that allocation, both sides entropy-coded, so this is not a margin a fixed-length deployed kernel inherits; WaterSIC itself sits, with i.i.d.\ Gaussian weights, its integer output entropy-coded and at high rate, $\approx0.25$ bit off the rate--distortion bound by the space-filling gap of \Cref{thm:gishpierce}, a margin available only to an allocation-flexible realization.

The result the whole inversion builds toward is the shared-scale counterpart of the
KLT: the transform that optimizes the AbsMax model the hardware induces.

\begin{theorem}[WUSH: a model-optimal adaptive block transform \citep{chen2025wush}]
\label{thm:wush}
Under a stochastic model of the per-group AbsMax block quantizer, block losses
approximated as independent, matrix columns treated as i.i.d.\ samples, and
quantization modeled as \emph{unbiased} noise (multiplicative i.i.d.\ for the
floating-point grid, max-scaled i.i.d.\ for the integer grid), with the bilateral
loss split to first order, there is a closed-form, data-aware blockwise transform
that is \emph{optimal} for the FP-AbsMax model and \emph{$d^{o(1)}$-near-optimal}
for the INT-AbsMax model on zero-mean Gaussian or Laplacian data (within a factor
$d$ for arbitrary distributions), among all invertible block-diagonal
transforms, $d$ here being the transform's block dimension, which the
source sets equal to the quantization group size $G$ of \Cref{tab:notation}. It combines a data-dependent whitening/balancing component with a
Hadamard factor applied \emph{after} it, so the Hadamard acts on the
already-whitened representation rather than as the raw-element flattening of
\Cref{prop:rht}. The optimality is of the \emph{model}; the source paper then
applies the resulting transform to deterministic rounding (both round-to-nearest
and GPTQ error-feedback) and reports that the advantage transfers empirically. Its full construction we give with the
rest of the non-orthogonal family in \Cref{sec:tax-affine}.
\end{theorem}

\Cref{thm:wush} closes the arc opened in \Cref{sec:classical}. It is the fixed-rate analogue of
the KLT (\Cref{thm:klt}), and it names a \emph{different} transform for a
principled reason, since it answers to the opposite \emph{side} of the inversion
(\Cref{thm:inversion}), optimizing its own stochastic AbsMax functional, which charges
the inverse map as well, rather than the coding surrogate (\Cref{rem:inv-scope}).
The KLT and WUSH are not two candidates competing on one axis to be benchmarked
against each other; they are the optima of two different axes, and which one is
correct is decided in the first instance by whether the deployment is variable- or fixed-rate, the element grid and the noise model then settling which optimum applies within that side.

That two-axis reading also resolves the apparent tension with the integer--floating-point
flip above. Under WUSH's FP-AbsMax \emph{model}, an orthogonal transform applied to the
raw operands is exactly useless: every orthogonal choice, the identity and the Hadamard
alike, leaves the modeled error at the same untransformed value, so the standalone
flattening of \Cref{prop:rht} buys nothing on an ideal FP grid. What the source shows is
that the same Hadamard is \emph{not} useless once it acts \emph{after} the data-dependent
whitening. \citet{chen2025wush} report the whitening and the Hadamard to be \emph{both}
essential: dropping either returns the modeled error to that same untransformed value, and
only the two together reach the model's floor. The Hadamard earns its place by flattening
the diagonal of the \emph{whitened} second moment, not by flattening raw weights. The measured ablation follows the model on \texttt{INT4} and \texttt{MXFP4}, where removing the Hadamard is costly, but not on \texttt{NVFP4}, where the whitening alone is already as good as the pair, and where a Hadamard applied on its own is mildly harmful because that format preserves the largest element of each block \citep{egiazarian2025mrgptq}. The mantissa-carrying scale behind that effect is what \Cref{sec:format} takes up. That a
data-free rotation of floating-point data is not merely unhelpful but, on the matrices they
test, \emph{harmful} \citep{ordentlich2026highrate} is a separate effect living outside that
model, driven by the joint concentration of the weight--activation pair rather than by the
element grid alone (\Cref{sec:fmt-flip}). WUSH's model settles only where its own Hadamard acts.

\subsection{Synthesis: no optimum transfers}
\label{sec:inv-synthesis}

The two-column picture has a blunt corollary, which we state because it is the
proposition the rest of the survey implicitly relies on.

\begin{corollary}[No transfer]
\label{cor:notransfer}
A transform proven optimal for one regime carries no optimality guarantee for the other. In particular, for a generic spectrum the KLT (\Cref{thm:klt}), optimal for allocation-flexible
coding, leaves the energy profile bounded away from the within-group flatness that the
AbsMax step-size surrogate rewards (\Cref{thm:inversion},
\Cref{prop:schur}(ii)). Because that surrogate is not a functional of the global order at
all, the change relative to an arbitrary starting basis is not even signed: what fails to
transfer is the guarantee, the signed opposition of \Cref{thm:inversion} being a within-group
statement rather than a global one. The one exception is degenerate: if the spectrum is constant on every scale group in the eigenvalue-sorted order (plateaus of multiplicity $G$ aligned with the group boundaries, \Cref{rem:inv-scope}), the sorted KLT sits on the fixed-rate floor $E/G$ and both surrogates are minimized at once.
\end{corollary}
\begin{proof}
Immediate from \Cref{thm:inversion}: the KLT drives $v$ to the majorization-maximal
\emph{reachable} profile, the spectrum, and $D_{\mathrm{fr}}$ attains its floor $E/G$ there
only if that spectrum is within-group uniform in the eigenvalue-sorted order; in that
degenerate case the sorted KLT is simultaneously the exact minimizer of $D_{\mathrm{vr}}$
and sits on the floor $E/G$, but a generic spectrum carries no such aligned plateaus
(\Cref{rem:inv-scope}). The KLT is therefore a minimizer of $D_{\mathrm{vr}}$ that is
bounded away from the minimizer of $D_{\mathrm{fr}}$, so no optimality transfers. Because
$D_{\mathrm{fr}}$ is grouping-dependent and not majorization-monotone, not even the
\emph{sign} of its change against an arbitrary starting basis is determined
(\Cref{fig:inversion}). \end{proof}

The corollary is a statement about the surrogate rather than a measured guarantee on every
deployed instance: it licenses no transform; it only forbids importing an optimality proof
across the divide. To our knowledge,
no result in the ledger of \Cref{sec:t4} establishes any single transform as optimal
for the deployed setting \emph{jointly} across number formats and rounding stages.
That absence is not a defect of the theory but its
current frontier, and it sets the agenda for the open problems of \Cref{sec:open}. The
master table (\Cref{tab:proven}) makes the situation legible at a glance: proofs exist
on both sides, each against its own objective, with no joint optimum yet in view.
Everything after the ledger (the transform taxonomy, the composition and format
analyses, and the extensions beyond weight matrices) is a tour of the transforms the
field has built on the fixed-rate side, and we read each of them through
the same three questions: which Schur objective it targets, what rounding it
composes with, and what it costs to deploy.

\section{What Is Proven, and What Is Not}\label{sec:t4}

\Cref{tab:proven} is the ledger of what the two sides actually prove. It lists the results that bear on \emph{which transform or allocation} is optimal for a regime's own objective, together with the regime-neutral rounding machinery and the perplexity-linkage result; per-method error bounds, general quantizer-design theorems (\Cref{lem:bennett,prop:lloydmax,prop:panterdite,thm:gishpierce}), the classical allocation optimum (\Cref{prop:waterfill}, which returns as a regime-bridging result in row~13), the single-weight rounding step (\Cref{prop:obs}), the SQNR decomposition (\Cref{rem:crest}), and the structural feasibility result the taxonomy rests on (\Cref{thm:invariance}) are treated where they arise. Reading its objective column top to bottom shows why no optimality guarantee transfers across the boundary.

{\footnotesize
\renewcommand{\arraystretch}{1.2}
\begin{longtable}{@{}p{0.3cm} >{\raggedright\arraybackslash}p{4.3cm} >{\raggedright\arraybackslash}p{2.5cm} >{\raggedright\arraybackslash}p{3.45cm} >{\raggedright\arraybackslash}p{3.15cm}@{}}
\caption{\textbf{The master optimality ledger.} Rigorous results exist on both sides of the inversion (\Cref{thm:inversion}), \emph{each under its own stated assumptions and against its own functional}, the point being that none transfers across the boundary. Rows 1--2 prove optimality of \emph{concentration} for allocation-flexible coding (row~3 is the nested-lattice matrix-multiplication optimum); rows 4 and 6 establish achievability and guarantees for \emph{flattening} on a shared-scale uniform grid, and row~5 the same on a shared-scale Lloyd/Gaussian-companded codebook, and row~10 proves model-optimality (under an explicit stochastic-noise model), while rows 7--9 \emph{delimit} flattening's benefit, which is format-dependent and, in row~9, not yet explained by any evaluated surrogate; rows 11--12 and 14 are regime-neutral machinery, and row~13 \emph{bridges} the two regimes.}\label{tab:proven}\\
\toprule
\textbf{\#} & \textbf{Result (what is proven)} & \textbf{Regime} & \textbf{Objective it optimizes} & \textbf{Reference} \\
\midrule
\endfirsthead
\toprule
\textbf{\#} & \textbf{Result (what is proven)} & \textbf{Regime} & \textbf{Objective it optimizes} & \textbf{Reference} \\
\midrule
\endhead
\midrule
\multicolumn{5}{r}{\emph{continued on next page}}\\
\endfoot
\bottomrule
\endlastfoot
1 & For a \emph{Gaussian} source with $\Sigma\succ0$ at high rate, the KLT with optimal (logarithmic) bit allocation, every coordinate receiving positive rate, is MSE-optimal at a fixed total rate \emph{among orthogonal transforms}; the optimal transform \textbf{concentrates} energy & \textbf{Allocation-flexible $\to$ concentrate} & MSE at fixed total rate (per-coefficient allocation) & \cite{huang1963block,goyal2001transform}; \Cref{thm:klt} \\
2 & Coding gain (the arithmetic-over-geometric-mean measure of the transformed variances; the cited source uses the geometric-over-geometric variant) is maximized by maximally \textbf{unequal} variances; when all coefficients share one scale-invariant quantizer family, \emph{a} KLT is optimal \emph{among orthogonal transforms} for \emph{any} bit allocation & Allocation-flexible $\to$ concentrate & Coding gain (distortion reduction at fixed rate) & \cite{goyal2001transform} \\
3 & Nested-lattice quantization is asymptotically optimal for the matrix-multiplication distortion (i.i.d.\ Gaussian matrices) & Lattice / rate--distortion & Inner-product MSE at fixed rate & \cite{ordentlich2024optimal} \\
4 & $\mu$-\textbf{incoherence} processing bounds the proxy quantization error; a random rotation achieves low coherence with high probability & \textbf{Shared-scale $\to$ flatten} & Layer-output MSE under uniform quant & \cite{chee2023quip} \\
5 & A randomized Hadamard with a \textbf{dithered} Lloyd/Gaussian-companded scalar quantizer attains the same leading MSE constant that \citet{zandieh2025turboquant} proved for a \emph{fully random} rotation, up to an $o(1)$ vanishing as the bit-width grows & Shared-scale $\to$ flatten (companded grid) & Per-vector $\ell_2$ MSE (worst-case unit vector) & \cite{feng2026provable} \\
6 & Block-Hadamard rotation has \textbf{non-asymptotic} bounds on the outlier suppression it can achieve & Shared-scale $\to$ flatten & Post-rotation within-group $\ell_\infty$, set by how evenly $\ell_1$ mass is spread across blocks & \cite{sanjeet2026mixquant} \\
7 & Hadamard flattening is \textbf{essential for INT-AbsMax}: an untransformed block can cost a factor as large as the dimension $n$, and a random rotation caps the \emph{expected} penalty at ${\approx}2\ln n$ for $n\ge27$, which the high-rate analysis prices as an effective-rate cost & \textbf{Shared-scale $\to$ flatten} (uniform grid) & AbsMax quant.\ MSE, INT grid & \cite{ordentlich2026highrate} \\
8 & On an \textbf{ideal FP grid the sign reverses}: the effect is set by the joint concentration $\Delta_{\mathrm{FP}}$, which rotation drives toward its isotropic value $\Delta_{\mathrm{FP}}{=}1$ (proved for an i.i.d.\ operand, with the post-rotation \emph{expectation} proved $<3$, and measured ${\approx}1$ after a Hadamard), so rotation helps only when the untransformed $\Delta_{\mathrm{FP}}$ exceeds ${\approx}1$; it is \emph{measured} to be harmful on the matrices tested, a conclusion the source calls specific to them & Shared-scale, \textbf{format-dependent} & AbsMax quant.\ MSE, ideal FP grid (an inner-product, not element-wise, objective) & \cite{ordentlich2026highrate} \\
9 & On \emph{deployed} \texttt{MXFP4}, a block-confined rotation is \emph{measured} to help the coarse power-of-two scale while a \emph{fixed} global one craters. \textbf{Nothing is proved here, and no surrogate comparable across the two families has been evaluated on it}: $\sum_g M_g^2$ puts both families at one \emph{ideal} floor, but the residuals the two fixed maps actually pay were not computed on the published data (\Cref{sec:fmt-flip}) & Shared-scale, \textbf{deployed} format & \emph{Measured} end-task accuracy & \cite{egiazarian2025mrgptq}; \cite{shao2025blockrotation} \\
10 & A closed-form data-aware blockwise transform is \textbf{model-optimal} for FP-AbsMax and $d^{o(1)}$-\textbf{near-optimal} ($d$ the block size) for INT-AbsMax on \emph{zero-mean} tail-bounded (Gaussian/Laplacian) data ($d$ for arbitrary), among all invertible block-diagonal transforms, under a stochastic-noise model & Shared-scale $\to$ flatten (data-aware) & AbsMax block error (stochastic model) & \cite{chen2025wush} \\
11 & GPTQ $\equiv$ LDLQ, and LDLQ is \textbf{optimal} (worst- and average-case, proxy loss $\propto\tr(D)$) among rounders whose linear feedback depends on $H$ but not $W$, when rounding to \emph{integers}, a finite-grid counterexample breaking it otherwise (first source); error-feedback rounding admits quantitative proxy-error bounds on an unbounded grid (second) & \textbf{Regime-neutral (rounding)} & Layer-output MSE given a fixed transform & \cite{chee2023quip}; \cite{zhang2025provable} \\
12 & GPTQ run \textbf{back-to-front} (no basis reduction) $\equiv$ Babai's nearest-plane for the closest-vector problem, for $X$ of full column rank, with the rounding domain matched on both sides; the inherited \emph{Babai} error bound additionally needs an \emph{unbounded} grid (no clipping) & Regime-neutral (rounding) & Lattice CVP distortion & \cite{chen2025geometry,birnick2026lattice} \\
13 & At high rate, water-filling the grid spacings provably dominates GPTQ's implicit \textbf{equal-rate} allocation, by $\tfrac12\log$ of the pivots' AM/GM ratio, which a \emph{random} rotation is measured to shrink to ${\approx}0.1$ bit on \texttt{Llama-3-8B} covariances & \textbf{Bridges regimes} (entropy-coded realization) & Rate--distortion at fixed average bits & \cite{lifar2026watersic,ordentlich2026highrate2} \\
14 & The perplexity \emph{increase} is (locally) a weighted sum of the per-layer relative \textbf{squared weight} (Frobenius) errors, at a trained optimum & Regime-neutral (medium-bitwidth) & Perplexity via per-layer relative weight error & \cite{malinovskii2025higgs} \\
\end{longtable}
}

The objective column is the argument: each result is correct \emph{under its stated assumptions}, and they optimize different functionals, so a transform proven optimal in one column carries \emph{no} guarantee in the other. The clearest instance is that the KLT (row~1) and WUSH (row~10) name genuinely different transforms, for any spectrum without group-aligned plateaus, precisely because they answer to different objectives: \Cref{cor:notransfer} makes the KLT's non-transfer exact against the AbsMax surrogate, while WUSH answers to a third, stochastic functional that \Cref{rem:inv-scope} keeps distinct from it. Rows 7--9 sharpen the boundary from inside the shared-scale regime: even there, the integer-versus-floating-point format changes whether flattening helps at all, while row~10 (WUSH) attains FP-AbsMax model-optimality only by combining that Hadamard with the non-orthogonal whitening it rides on top of, both of which \citet{chen2025wush} report to be essential: dropping either returns the modeled error to the value it takes under \emph{any} orthogonal transform, the identity included. That non-orthogonality is not incidental: \citet{federici2026dissecting} prove that, under negligible clipping and uncorrelated quantization noise, the AbsMax SQNR carries an alignment factor that \emph{no} orthogonal transform can change, the alignment-optimal map being a geometric mean of the weight second moment and the \emph{inverse} activation second moment, which is the balancing the whitening family performs. Rows 11--14 are the regime-neutral machinery, error-feedback rounding and the link from per-layer weight error to perplexity, together with the one result (row~13) that \emph{bridges} the regimes by reintroducing per-coordinate allocation into an otherwise shared-scale pipeline. \Cref{sec:systems} shows the datapath still closing that seam; \Cref{sec:open} poses reopening it as an open problem.

The sharpest gap is row~9: the global-versus-block-confined \texttt{MXFP4} contrast is measured and not yet explained, the AbsMax surrogate this survey works with putting both families at one ideal floor and not being computed here on the two deployed maps (\Cref{sec:fmt-flip}). Evaluating it there, and reducing the published mechanistic accounts (\citealp{shao2025blockrotation}; \citealp{chen2025wush}) to a comparable statistic, are both part of the second open problem of \Cref{sec:open}. More broadly, no result in the ledger proves that any single transform is optimal for deployed LLM inference across all formats and rounding stages. That absence is not an omission; it is the current state of the theory, and it is what defines the open problems of \Cref{sec:open}.

\section{A Taxonomy of Transforms}\label{sec:taxonomy}

The inversion of \Cref{sec:inversion} tells us \emph{what} a good low-bit transform must do (flatten the within-group energy so that no coordinate dominates its shared scale) but not \emph{how}. The literature answers the ``how'' with a design space we organize by \emph{algebraic class}: diagonal rescalings, permutations, orthogonal rotations (fixed or learned), and general invertible affine maps, with sequence-axis and frame transforms off to the side. These classes correlate with, but are not strictly ordered by, the degrees of freedom a transform commands. A diagonal map has one continuous parameter per channel. An orthogonal rotation has $\binom{d}{2}$ angles but preserves every length, while a general invertible map spends the full $d^2$ parameters and gives up length preservation.

\begin{figure}[tbp]
\centering
\definecolor{famDiag}{HTML}{1F7A6D}
\definecolor{famPerm}{HTML}{2C6E9A}
\definecolor{famFixO}{HTML}{34489A}
\definecolor{famLrnO}{HTML}{6A4794}
\definecolor{famAff}{HTML}{97355F}
\definecolor{famOff}{HTML}{464C54}
\footnotesize
\begin{forest}
  forked edges,
  for tree={
    grow'=0,
    fill=white, draw=black!30, line width=0.6pt, rounded corners=2pt,
    inner sep=3pt, font=\footnotesize, align=left,
    anchor=west, child anchor=west, parent anchor=east,
    edge={draw=black!40, line width=0.6pt},
    l sep=5mm, s sep=1.3mm,
  },
  where level=2{text width=70mm, tier=lvs}{},
  [{\textbf{Transform families}}, name=froot, draw=black!70, fill=black!5, line width=0.9pt,
     inner sep=5pt, font=\small\bfseries,
    tikz={\draw[-{Latex[length=2.4mm]}, black!75, line width=1pt]
            ([xshift=-5mm]froot.west |- fdiag.north) -- ([xshift=-5mm]froot.west |- foff.south);
          \node[rotate=90, anchor=south, font=\scriptsize\itshape, text=black!80]
            at ([xshift=-6.6mm]$(froot.west |- fdiag.north)!0.5!(froot.west |- foff.south)$)
            {increasing degrees of freedom};}
    [{Diagonal scaling\\{\normalfont\itshape\scriptsize\color{black!72}rebalance operands}}, name=fdiag,
       draw=famDiag, fill=famDiag!10, line width=1.2pt, text width=33.5mm,
       for children={draw=famDiag!80, edge={draw=famDiag!72, line width=0.7pt}}
      [{\textbf{Migration:} SmoothQuant, AWQ}]
      [{\textbf{Learned scale:} OmniQuant, TEQ}]
      [{\textbf{Centering shift:} OS+}]
      [{\textbf{Two-sided step-size:} Z-Fold}]
      [{\textbf{Data-free root:} CLE}]
    ]
    [{Permutation\\{\normalfont\itshape\scriptsize\color{black!72}regroup channels}},
       draw=famPerm, fill=famPerm!10, line width=1.2pt, text width=33.5mm,
       for children={draw=famPerm!80, edge={draw=famPerm!72, line width=0.7pt}}
      [{\textbf{Range-cluster:} RPTQ}]
      [{\textbf{Outlier-reorder:} Atom}]
      [{\textbf{Processing-order:} GPTQ act-order}]
      [{\textbf{Learned permutation:} PermLLM (N:M sparsity)}]
      [{\textbf{Split--merge:} QLLM}]
    ]
    [{Fixed orthogonal\\{\normalfont\itshape\scriptsize\color{black!72}data-free rotation}},
       draw=famFixO, fill=famFixO!10, line width=1.2pt, text width=33.5mm,
       for children={draw=famFixO!80, edge={draw=famFixO!72, line width=0.7pt}}
      [{\textbf{Global Hadamard:} QuaRot, QuIP\#}]
      [{\textbf{Block-confined:} Block-Rotation$^\dagger$,\\MR-GPTQ$^\dagger$, GSR}]
    ]
    [{Learned orthogonal\\{\normalfont\itshape\scriptsize\color{black!72}learned rotation}},
       draw=famLrnO, fill=famLrnO!10, line width=1.2pt, text width=33.5mm,
       for children={draw=famLrnO!80, edge={draw=famLrnO!72, line width=0.7pt}}
      [{\textbf{Gradient (Cayley):} SpinQuant, OptRot}]
      [{\textbf{Alternating Procrustes:} DFRot, ConQuR}]
      [{\textbf{Distribution-shaping:} KurTail, DartQuant}]
      [{\textbf{Structured / $+$diagonal:}\\ButterflyQuant, HARP, OSTQuant, ParoQuant}]
    ]
    [{Non-orthogonal affine\\{\normalfont\itshape\scriptsize\color{black!72}whitening; full $\mathrm{GL}(d)$}}, name=faff,
       draw=famAff, fill=famAff!10, line width=1.2pt, text width=33.5mm,
       for children={draw=famAff!80, edge={draw=famAff!72, line width=0.7pt}}
      [{\textbf{Whitening (closed-form):} WUSH}]
      [{\textbf{Learned invertible:}\\FlatQuant, AffineQuant, LATMiX, FPTQuant}]
      [{\textbf{$+$ bias / asymmetric:} BASE-Q}]
    ]
    [{Sequence \& frame\\{\normalfont\itshape\scriptsize\color{black!72}compose with any}}, name=foff,
       draw=famOff, fill=famOff!10, line width=1.2pt, text width=33.5mm,
       for children={draw=famOff!82, edge={draw=famOff!76, line width=0.7pt}}
      [{\textbf{Sequence:} PrefixQuant, STaMP$^\dagger$}]
      [{\textbf{Overcomplete frame:} FrameQuant, Kashin}]
    ]
  ]
\end{forest}
\caption[Taxonomy of transform families]{\textbf{A map of the transform families and their characteristic ``moves.''} The five feature-axis families are ordered top to bottom by loosely \emph{increasing degrees of freedom}, and each fans out into its distinct constructions (its ``moves'') and a few representative methods; the sixth group leaves the single-matrix feature-axis setting and composes with any of the five. Randomized fixed orthogonal transforms already reach near-optimal flattening with no calibration (\Cref{prop:rht}, which covers the Haar and randomized-Hadamard members rather than the deterministic block constructions); the affine family attains, in WUSH, the optimum of the floating-point AbsMax model and comes within a $d^{o(1)}$ factor on the integer grid for zero-mean Gaussian or Laplacian data, among block-diagonal transforms and under the theorem's stochastic noise model (\Cref{thm:wush}); its learned members approximate that map without the guarantee. Daggered ($\dagger$) figure entries are format-co-design instantiations treated in \Cref{sec:format}, not among the \(43\) rows of \Cref{tab:taxonomy}.}
\label{fig:taxonomy}
\end{figure}
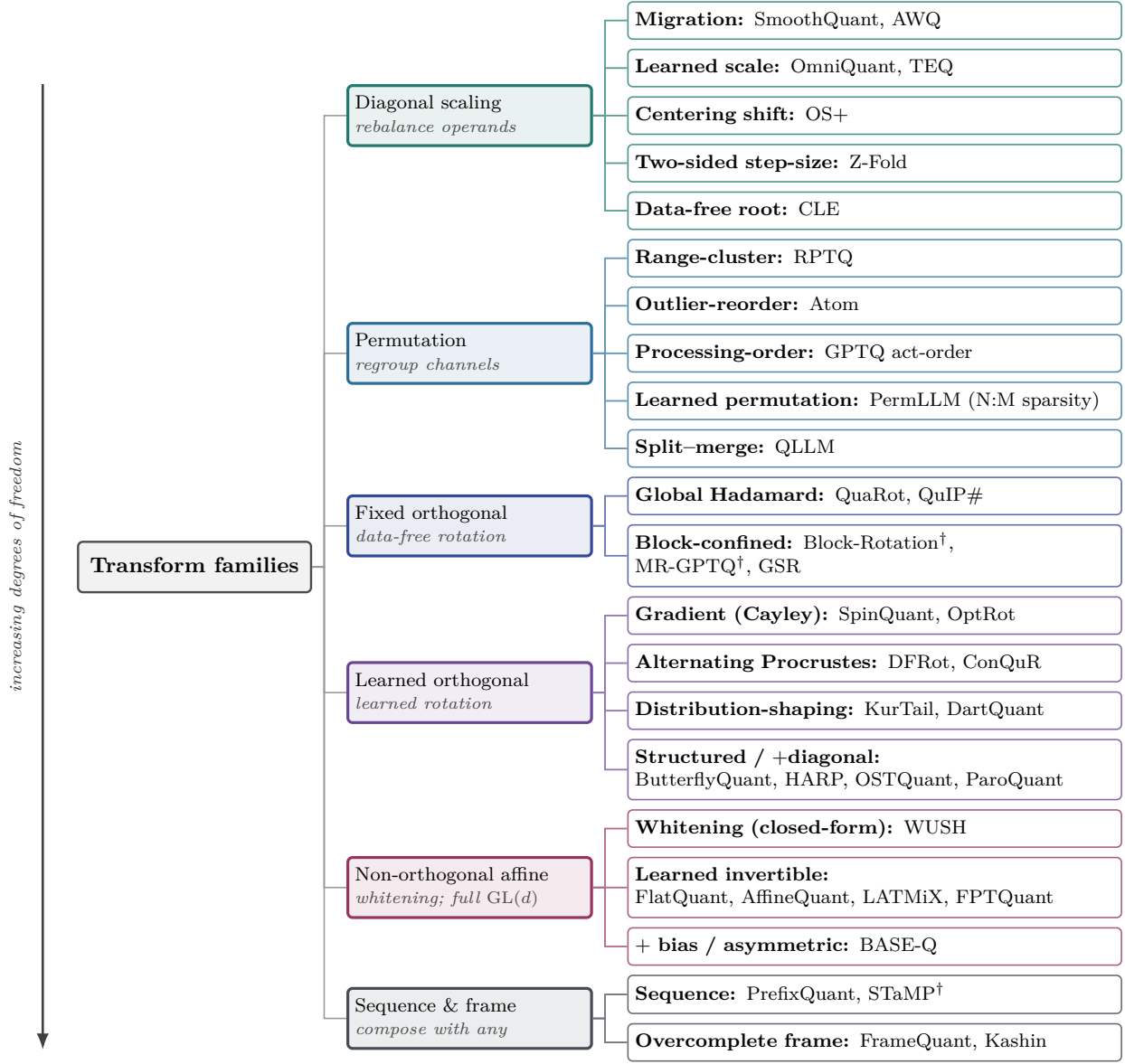

Two caveats keep the degrees-of-freedom ladder from being a clean total order. A permutation is discrete (zero continuous parameters) and so is not simply ``below'' a diagonal map but incomparable to it. The fixed and learned orthogonal families share the \emph{same} feasible set (the orthogonal group), differing only in how the rotation is selected, not in how many degrees of freedom it has. Read with those caveats, the ordering of \Cref{fig:taxonomy} still tracks a real progression in the field. It runs from the 2022--2023 scaling methods that migrated outliers between operands, through the 2024 rotations that made \texttt{W4A4} possible, to the 2025--2026 non-orthogonal maps that narrow the remaining gap toward the (surrogate-model) optimum.

We classify each method along six axes, following the vocabulary fixed in \Cref{sec:foundations}: its \emph{structure} (the algebraic form of $T$); whether it is \emph{data-aware} (uses calibration statistics) or data-free; whether the transform is \emph{learned} by a search over candidate transforms versus \emph{given by a construction}, data-free or from calibration statistics; its \emph{deployment cost} (folded into adjacent weights at zero runtime, versus applied by an online kernel); its \emph{granularity} (per-tensor, per-channel, per-head, per-block); and its composability with error-feedback rounding (\Cref{sec:composition}). \Cref{tab:taxonomy} tabulates the first four of these axes for the 43 methods surveyed here, with granularity carried in the one-line description wherever the source specifies it and composability discussed, where the sources report it, in the narrative and in \Cref{sec:composition}. The subsections that follow narrate each family, and \Cref{sec:t4} has already placed the endpoints of the space against what is actually proven.

\begin{figure}[t]
\centering
\includegraphics[width=\linewidth]{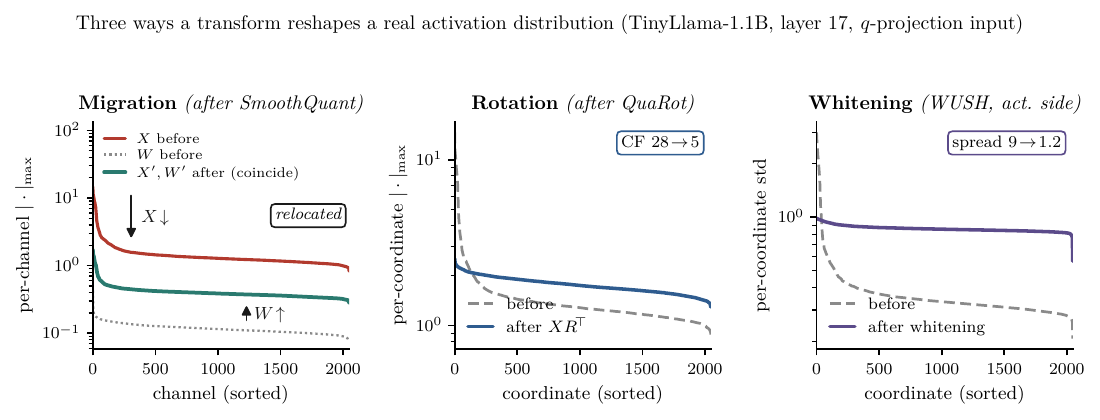}
\caption[The three transform families on a real activation distribution]{\textbf{The three transform families reshaping a real activation distribution}, our own computation on the $q$-projection input of layer~17 of the base \texttt{TinyLlama-1.1B-intermediate-step-1431k-3T} checkpoint over $3072$ \texttt{wikitext-2-raw-v1} train tokens, the activations cached in \texttt{fp16} and the figure arithmetic in \texttt{fp64}: an illustrative single layer, not a benchmark. Each panel sorts a per-coordinate statistic before and after the transform. \emph{Migration} (after SmoothQuant, \citealp{xiao2023smoothquant}) applies SmoothQuant's per-channel scale \eqref{eq:smoothquant} at $\alpha=\tfrac12$, i.e.\ $s_j=\max|X_j|^{1/2}\max|W_j|^{-1/2}$, pulling $X$ down and $W$ up until the two operands coincide at their geometric mean: it relocates crest between the operands rather than reducing the \emph{joint per-channel} burden (\Cref{sec:tax-diagonal}, for that invariance and for what it can still equalize across a group), hence no crest number is quoted for it. \emph{Rotation} (after QuaRot, \citealp{ashkboos2024quarot}; a randomized Hadamard $R=H\diag(\pm1)$) genuinely collapses the crest factor $\mathrm{CF}$ \eqref{eq:crest} of a single operand from $28$ to $5$, stably across Hadamard sign seeds. \emph{Whitening} (an activation-side stand-in for the bilateral WUSH map, \citealp{chen2025wush}, which a single-operand curve cannot fully depict) drives the per-coordinate standard deviations toward a common value (spread, meaning their maximum over their median, $9.0\!\to\!1.2$, the panel using the regularized inverse square root $(\Sigma+\lambda I)^{-1/2}$ with $\lambda=0.02\,\tr(\Sigma)/d$, so the residual is that ridge rather than a limit of whitening: exact whitening drives the spread to $1$ by construction here (those tokens against $d{=}2048$ leave $\Sigma$ full rank), and the ridge is what a deployed data-aware map would need on a finite calibration sample); the panel shows the whitening factor alone, with WUSH's Hadamard omitted, though under WUSH's own model the two are jointly essential (\Cref{thm:wush}); whitening's advantage is spectral and bilateral rather than a further cut to the activation crest.}
\label{fig:reshape}
\end{figure}

Keep the inversion in view throughout. The families near the top of the degrees-of-freedom ladder can, in principle, approach the flattening optimum, while the families near the bottom cannot; under the \emph{floating-point} model of \Cref{thm:wush} that optimum is not reached by flattening the raw operand, since every orthogonal map leaves the modeled error where it started; it is the Hadamard acting on the \emph{whitened} second moment, with the whitening, that attains it. The theory of \Cref{sec:inversion} makes that claim precise, and the closed-form construction of \Cref{sec:tax-affine} attains it for that model. The families also reshape the distribution in qualitatively different ways, as \Cref{fig:reshape} shows on a real layer. A diagonal scaling rebalances per-channel magnitude between the operands, an orthogonal rotation spreads each outlier across the coordinates, and a non-orthogonal whitening isotropizes the covariance.

{\footnotesize
\renewcommand{\arraystretch}{1.15}
\begin{longtable}{@{}>{\raggedright\arraybackslash}p{2.55cm} >{\raggedright\arraybackslash}p{1.35cm} c c >{\raggedright\arraybackslash}p{1.7cm} >{\raggedright\arraybackslash}p{5.5cm}@{}}
\caption{\textbf{The transform methods surveyed, along the classification axes of \Cref{sec:taxonomy}.} The complete cross-role index of all cited works is \Cref{tab:index}. \emph{Fam.}: Diag (diagonal scaling), Perm (permutation), FixO (fixed orthogonal), LrnO (learned orthogonal, read broadly to include data-fitted rotations obtained in closed form), Aff (non-orthogonal affine), and, within the sixth family, Seq (sequence-axis) and Fr (frame); hybrids are listed under their primary lever, and a starred code is glossed in that method's own description. \emph{D}: data-aware. \emph{L}: the transform's parameters are obtained by \emph{searching over candidate transforms} against a quantization objective (gradient descent; an objective-driven search as in AWQ's exponent grid or PQF's search over orderings; an alternating solve as in Z-Fold's ALS or DFRot's and ConQuR's Procrustes iterations). They are graded N where the transform is instead \emph{given by a construction}, for instance: a data-free map (QuaRot's Hadamard, GSR, Kashin, the folded LayerNorm gain of Outlier Suppression), a formula in calibration statistics (SmoothQuant's balanced exponent, WUSH's Cholesky and SVD, ResQ's PCA), or an ordering or selection read off a clustering or ranking of channel or token statistics (RPTQ, Atom, DuQuant, PrefixQuant); a closed-form rule propagated along a chain of layers (CLE) counts as a construction too, even where it iterates internally. \emph{Cost}: deployment cost of the transform, counted on the activation side, the weight-side companion always folding offline (\Cref{def:fpt}): \emph{fold} (the activation side is absorbed too, nothing at inference), \emph{hybrid} (the main map folds but an auxiliary stage stays online), \emph{online} (an inference-time kernel applies it on every token), \emph{none}$^\ast$ (the map is undone before storage; a residual index survives only under group-wise scales). Constructions are taken from the source papers; two of the 43 (GPTQ act-order, QuIP\#) are rows here by structure but belong primarily to the rounding-and-codebook story of \Cref{sec:composition}, where \Cref{tab:index} indexes them by that primary role.}\label{tab:taxonomy}\\
\toprule
\textbf{Method} & \textbf{Fam.} & \textbf{D} & \textbf{L} & \textbf{Cost} & \textbf{Transform, in one line} \\
\midrule
\endfirsthead
\toprule
\textbf{Method} & \textbf{Fam.} & \textbf{D} & \textbf{L} & \textbf{Cost} & \textbf{Transform, in one line} \\
\midrule
\endhead
\midrule
\multicolumn{6}{r}{\emph{continued on next page}}\\
\endfoot
\bottomrule
\endlastfoot
CLE \citep{nagel2019dfq}          & Diag & N & N & fold & Data-free pairwise channel rescale equalizing adjacent-layer weight ranges via ReLU scale-equivariance. \\
Outlier Supp.\ \citep{wei2022outlier}   & Diag & N & N & fold & Fold LayerNorm's per-channel $\gamma$ (the outlier amplifier) into the next weight, with a compensating parameter on the residual branch (data-free); token-wise clipping is a separate calibrated add-on. \\
SmoothQuant \citep{xiao2023smoothquant} & Diag & Y & N & fold & $s_j=\max|X_j|^\alpha/\max|W_j|^{1-\alpha}$ migrates activation outliers into weights. \\
AWQ \citep{lin2023awq}            & Diag & Y & Y & fold & Per-channel scale $s_X^\alpha$ protecting the channels salient by \emph{activation} magnitude; $\alpha$ grid-searched on output MSE. \\
Outlier Supp.+ \citep{wei2023outliersuppplus} & Diag & Y & Y & fold & Per-channel shift (center) then scale (outlier threshold grid-searched); both migrated into the next weight and bias. \\
OmniQuant \citep{shao2023omniquant}& Diag & Y & Y & fold & Gradient-learned per-channel scale and shift (LET) plus learnable weight clipping (LWC). \\
SmoothQuant+ \citep{pan2023smoothquantplus} & Diag & Y & Y & fold & SmoothQuant scale for group-wise \texttt{W4A16}; $\alpha$ grid-searched on whole-model loss. \\
Z-Fold \citep{jeon2023zfold}      & Diag & Y & Y & fold & Two-sided diagonal step-size $S=\zeta\alpha^{\!\top}$ via Hessian ALS; $\zeta$ folds into the previous layer. \\
MergeQuant \citep{wang2025mergequant} & Diag & Y & N & hybrid & Migrate static per-channel activation scales into weights (folds); a lightweight dimensional-reconstruction gather stays online. \\
TEQ \citep{cheng2023teq}          & Diag & Y & Y & fold & Learn a minimal-parameter per-channel equivalent transformation, folded away like SmoothQuant. \\
\midrule
RPTQ \citep{yuan2023rptq}         & Perm & Y & N & fold & $k$-means-cluster channels by $(\min,\max)$ range; reorder so a cluster shares one scale (fused into the LayerNorm write addresses and the linear's indexing, so nothing runs online). \\
GPTQ act-order \citep{frantar2022gptq} & Perm & Y & N & none$^\ast$ & Quantize columns in decreasing $\diag(H)$ order: a processing-order permutation, inverted before storage, not a change of stored basis (zero-cost under a per-row scale; with group-wise scales the inverted order leaves a per-column group index at inference, hence \texttt{-{}-static-groups}). \\
Atom \citep{zhao2023atom}         & Perm & Y & N & hybrid & Reorder top-128 outlier channels to the matrix end; \texttt{INT8} for them, \texttt{INT4} for the rest (weight reorder folds, activation reorder online $<0.5\%$). \\
QLLM \citep{liu2024qllm}          & Perm$^\ast$ & Y & Y & hybrid & Disassemble an outlier channel into $T$ replicated sub-channels, then reassemble similar channels ($^\ast$a channel split/merge, \emph{not} a pure permutation; the low-rank adapter folds, and the split/merge folds into a linear predecessor, running online at $\sim$4\% over \texttt{W4A4}, only after a non-linear op). \\
DuQuant \citep{lin2024duquant}    & Perm & Y & N & online & Smooth, then greedy block-diagonal rotation + zigzag permutation balancing outlier mass across blocks, then a second rotation. \\
PQF \citep{martinez2021pqf}       & Perm & N & Y & fold & Search a function-preserving input-channel permutation, equivalently the predecessor's output order, that eases vector quantization. \\
PermLLM \citep{zou2025permllm}    & Perm & Y & Y & fold & Learned Sinkhorn$\to$Hungarian channel permutation for more accurate N:M pruning (the input-channel reorder is absorbed into the predecessor's output order, as for PQF). \\
\midrule
QuIP \citep{chee2023quip}         & FixO & Y & N & online & Conjugate $W$ and $H$ by Kronecker-factored \emph{random} orthogonals to force $\mu$-incoherence (the rotation is data-free, but it is preceded by a Hessian-derived per-channel rescale $(\operatorname{diag}(X^{\!\top}X)/\operatorname{diag}(W^{\!\top}W))^{1/4}$; the paired LDLQ rounding is Hessian-aware). \\
QuIP\# \citep{tseng2024quipsharp} & FixO & Y & Y & online & Randomized Hadamard (Hadamard $\times$ random sign), $O(n\log n)$; $E_8$-lattice codebook (the RHT is data-free as constructed, but both fine-tuning stages then relax its sign vectors to real values and learn them by gradient, and the BlockLDLQ rounding is Hessian-aware). \\
QuaRot \citep{ashkboos2024quarot} & FixO & N & N & hybrid & Global randomized Hadamard fused into weights by computational invariance; a few online Hadamards. \\
GSR \citep{choi2025grouped}       & FixO & N & N & fold & Block-diagonal sequency-ordered Walsh--Hadamard blocks fused offline; training-free; a drop-in replacement for $R_1$ on its own, and additionally an initialization for learned rotations. \\
RRS \citep{yi2024rotated}          & FixO & N & N & online & Online Hadamard, then a runtime smoothing scale taken as the maximum over each $128$-channel group of magnitude-reordered channels (group $=$ GEMM block, so the scale is constant within each block and rides the per-group dequantization already paid); the raw per-channel maximum is the idealized form the kernel cannot run. \\
OptRot \citep{gadhikar2025optrot} & LrnO & N & Y & hybrid & Data-free rotation \emph{learned} (Cayley--SGD on the Stiefel manifold) against a fourth-power weight-outlier proxy. \\
FrameQuant \citep{adepu2024framequant} & Fr & N & N & online & Quantize in an overcomplete tight-fusion-frame ($r\!\approx\!1.1$), data-free; redundancy averages out noise (only the GPTQ rounding in frame coords is Hessian-aware). \\
Kashin \citep{merkulov2024kashin} & Fr & N & N & online & $x\approx u+Qv$ in a $2\times$ redundant basis so both factors have small $\ell_\infty$. \\
\midrule
SpinQuant \citep{liu2024spinquant}& LrnO & Y & Y & hybrid & Learn residual ($R_1$) and head-wise value ($R_2$) rotations by Cayley SGD; $R_3,R_4$ fixed Hadamard. \\
DFRot \citep{xiang2024dfrot}       & LrnO & Y & Y & hybrid & Alternating refinement of QuaRot's Hadamard, each rotation step a closed-form Procrustes solve, loss up-weighting massive-activation tokens. \\
KurTail \citep{akhondzadeh2025kurtail} & LrnO & Y & Y & hybrid & Learn Stiefel rotations minimizing the gap of rotated-activation kurtosis to the uniform value. \\
ButterflyQuant \citep{xu2025butterflyquant} & LrnO & Y & Y & online & Learnable butterfly of Givens rotations ($\tfrac{n}{2}\log n$ angles), trained on a layerwise quantized-reconstruction loss with a uniformity-KL regularizer on the rotated activations; $O(n\log n)$ online. \\
DartQuant \citep{shao2025dartquant}& LrnO & Y & Y & hybrid & QR-parametrized rotation optimizing a distribution-shaping ``Whip'' loss $\sum e^{-|x|}$. \\
HARP \citep{zagitov2026harp}      & LrnO & Y & Y & online & Sparse butterfly-like block-orthogonal stages, Hadamard-initialized, mixed-radix, backend-aware. \\
ConQuR \citep{thrash2026conqur} & LrnO & Y & Y & hybrid & Alternating Procrustes solve (targets the nearest hypercube corners, re-solved against calibration), no gradient over $O(d)$ parameters. \\
OSTQuant \citep{hu2025ostquant}   & LrnO & Y & Y & hybrid & Learn rotation $+$ diagonal scaling jointly by end-to-end fine-tuning under a KL-Top loss; a quantization-space-utilization (QSUR) metric motivates the form. \\
ReSpinQuant \citep{kim2026respinquant} & LrnO & Y & Y & hybrid & Full-size Cayley-learned rotation per layer, folded into the attention and FFN weights, with a low-rank online correction for the inter-layer basis mismatch. \\
ParoQuant \citep{liang2025paroquant} & LrnO & Y & Y & online & Learned \emph{pairwise} (Givens) rotations + per-channel scaling; PTQ, including a QAT-style weight and step-size fine-tuning stage; $\sim$10\% runtime. \\
ResQ \citep{saxena2025resq} & LrnO & Y & N & hybrid & PCA into the top-variance $\sim\!\tfrac18$ subspace kept at \texttt{INT8}, the rest at \texttt{INT4}, with an independent random orthogonal rotation inside each subspace; a data-aware rotation using KLT concentration for mixed precision. \\
\midrule
WUSH \citep{chen2025wush}         & Aff & Y & N & online & $T\simeq H\,S^{-1/2}U^{\!\top}$ (schematic; full Cholesky-whitened form in \S\ref{sec:tax-affine}): Hadamard backbone $\times$ data-dependent whitened-SVD balance. \\
AffineQuant \citep{ma2024affinequant} & Aff & Y & Y & fold & One invertible $A$ per layer, learned by a gradual diagonal-to-dense mask; diagonal-dominance keeps $A$ invertible. \\
FlatQuant \citep{sun2024flatquant}& Aff & Y & Y & online & Learnable invertible Kronecker $P=P_1\!\otimes\!P_2$; fused online kernel. \\
FPTQuant \citep{vanbreugel2025fptquant} & Aff & Y & Y & hybrid & Function-preserving transforms: Q/K scale-and-rotate, value matrix, MLP scaler, dynamic residual scale. \\
BASE-Q \citep{he2025baseq}        & Aff & Y & Y & hybrid & Residual and value rotations fold into weights; $R_{\mathrm{qk}}$/$R_{\mathrm{down}}$ stay online Hadamards; learnable channel-bias recentering + asymmetric per-quantizer scaling also run online, fused into one quant/dequant kernel. \\
LATMiX \citep{gordon2026latmix}   & Aff & Y & Y & hybrid & Learnable full invertible affine (LU or QR form) mixing mass across channels; MX-format bound (its two learned maps, global and attention-value, fold; a third, unlearned transform stays online in the FFN). \\
\midrule
PrefixQuant \citep{chen2024prefixquant} & Seq & Y & N & hybrid & Prepend a fixed prefix of the token types that most often carry outliers, so the prefix absorbs the token-wise outliers; $+$ fixed Hadamard rotations (the transform is training-free; any fine-tuning trains the quantizer's clipping and the weights, not the transform). \\
\end{longtable}
}

\subsection{Diagonal scalings: migrating the difficulty}
\label{sec:tax-diagonal}

The simplest function-preserving transform is a diagonal one. Writing the diagonal map as $T=\diag(s)^{-1}$ in the invariance $Y=(X\diag(s)^{-1})(W\diag(s))^{\!\top}$ divides the $j$-th input activation by $s_j$ and multiplies the $j$-th weight column by the same factor. The layer therefore computes an identical output, while the quantization difficulty of coordinate $j$ is shifted between the two operands. Because the rescaling can be pushed backwards into a preceding LayerNorm or linear layer, the transform costs nothing at inference. The entire family is foldable when the surrounding nonlinearity permits it. Its lineage predates LLMs. Cross-Layer Equalization \citep{nagel2019dfq} rescales a pair of consecutive convolutional layers so their per-channel ranges match, using the positive scaling-equivariance of ReLU, $f(sx)=sf(x)$. The closed-form data-free factor $s_i=r_i^{(2)-1}\sqrt{r_i^{(1)}r_i^{(2)}}$ equalizes the two layers' channel ranges. This equalization is the conceptual root of everything that follows, and \citet{meller2019same} arrived at it independently and slightly earlier by weight factorization. It is also, with Outlier Suppression's $\gamma$-migration, among the few members of the family needing no calibration data at all, that method's clipping add-on being a separately calibrated stage.

The LLM era began by identifying where the outliers come from. \citet{wei2022outlier} identified the per-channel scaling parameter $\gamma$ of LayerNorm as an amplifier of activation outliers. They removed it by a ``gamma migration'' that replaces the norm with a non-scaling variant and moves $\gamma$ out of the norm, absorbing it into the following weight by $(x\odot\gamma)W^{\!\top}=x(W\diag(\gamma))^{\!\top}$ and carrying a matching factor on the residual branch. The reparameterization is exact and data-free. SmoothQuant \citep{xiao2023smoothquant} generalized this into the canonical migration rule: choose the per-channel factor
\begin{equation}
s_j \;=\; \frac{\max_n |X_{nj}|^{\alpha}}{\max_i |W_{ij}|^{1-\alpha}},
\label{eq:smoothquant}
\end{equation}
with a migration strength $\alpha$ (default $\tfrac12$) that balances how much of each channel's dynamic range is carried by the activation versus the weight, enabling \texttt{W8A8}.

AWQ \citep{lin2023awq} recast the same lever as weight-only protection. It scales up the channels with the largest activation magnitude, the ``salient'' channels whose error matters most for the output. AWQ picks the exponent by a grid search $\alpha^\star=\argmin_\alpha \|X\diag(s)^{-1}\,Q(W\diag(s))^{\!\top} - XW^{\!\top}\|$ that minimizes the reconstruction error directly, rather than fixing it a priori. That it keys on \emph{activation} rather than weight magnitude is not incidental. AWQ's own oracle holds a mere $0.1$--$1\%$ of channels in \texttt{FP16} and recovers most of the quantization loss (\texttt{OPT-6.7B} \texttt{INT3} perplexity $23.5\!\to\!11.4$) when those channels are selected by activation magnitude, but barely improves it when they are selected by weight magnitude ($23.5\!\to\!22.4$). Salience is an activation-side property, and the foldable per-channel scale is a hardware-friendly surrogate for the mixed precision the oracle would otherwise need. Outlier Suppression+ \citep{wei2023outliersuppplus} added the missing degree of freedom, a per-channel \emph{shift} $z_j=\tfrac12(\max_n X_{nj}+\min_n X_{nj})$ that centers each channel before scaling and migrates into the next layer's bias. The shift symmetrizes asymmetric activation distributions that a pure scale cannot address. \Cref{fig:reshape} shows this migration on a real layer in its diagonal-scaling panel. Being rank-preserving, the scaling only rebalances the crest factor between operands and cannot lower it within a single channel (\Cref{prop:flat-opt}). \citet{xiao2023smoothquant} visualize the same migration of difficulty from activations onto weights.

SmoothQuant fixes its exponent a priori, while AWQ and Outlier Suppression+ recover theirs by a one-dimensional search; the next methods put the whole per-channel vector under gradient descent. OmniQuant \citep{shao2023omniquant} makes the scale and shift of its Learnable Equivalent Transformation trainable parameters, optimized block-by-block against the quantized-layer reconstruction error together with a Learnable Weight Clipping that reparametrizes the quantizer's clip range. Both fold away exactly as before, so the added cost is entirely offline. TEQ \citep{cheng2023teq} learns the same kind of per-channel equivalent transformation with a deliberately minimal parameter count, folding it away identically. The migration lever also transfers to weight-only regimes. SmoothQuant+ \citep{pan2023smoothquantplus} reuses \eqref{eq:smoothquant} for group-wise \texttt{W4A16}, grid-searching $\alpha$ against the whole-model loss. Z-Fold \citep{jeon2023zfold} gives the quantizer step size a rank-one structure $S=\zeta\alpha^{\!\top}$ with an out-channel factor $\alpha$ and an in-channel factor $\zeta$, and fits them by alternating least squares, Hessian-weighted on the $\alpha$ step. It then folds $\zeta$ into the preceding norm's affine parameters, or into the preceding linear's own step size where no nonlinearity intervenes, so that only the ordinary per-channel dequant scale $\alpha$ survives at runtime. MergeQuant \citep{wang2025mergequant} pushes the idea to its logical end for static-activation serving, migrating the per-channel activation quantization scale itself into the weights so that the dequantization step overlaps the multiply-add and disappears from the datapath. Only a lightweight dimensional-reconstruction gather stays online.

The whole family shares a hard ceiling, and the theory of \Cref{sec:inversion} says exactly where it sits. A diagonal transform is rank-preserving. It can move the crest factor between the two terms of the bilateral proxy \eqref{eq:bilateral-proxy}, and it can equalize the magnitudes \emph{across} channels. But it cannot destroy dynamic range, only move it between the two operands: whatever it takes off the activation side it puts on the weight side, and it can never spread a single channel's outlier across the coordinates that share its scale, since the product $\max|W_j|\max|X_j|$ of the two channels' maxima is invariant under it. It can lower either operand's crest factor, but only by raising the other's, so for a group of a single channel the \emph{joint} burden the fixed-rate error charges (\Cref{prop:flat-opt}, applied to each operand in turn) is invariant under it; what it can still do is equalize that burden across the channels sharing a group. That is why the migration methods are competitive only up to \texttt{W8A8}/\texttt{W6A6} and weight-only \texttt{W4A16} and cede the aggressive \texttt{W4A4} regime to the rotations and non-orthogonal maps of the subsections that follow.

The ceiling is sharpest at the FFN down-projection, whose inputs carry activation outliers on the order of a thousand times the median magnitude. A diagonal scale large enough to tame them merely manufactures a matching outlier on the weight side. This is why applying SmoothQuant's smoothing there backfires \citep{lin2024duquant}; OmniQuant and AffineQuant also exclude that layer from their learned transform (AffineQuant because the affine matrix in the inflated intermediate dimension is too large to optimize reliably). \citet{lin2024duquant} single the layer out as the case a diagonal scale cannot handle alone, repairing it by following the smoothing with a rotation, a permutation that rebalances the blocks, and a second rotation: a clean instance of the conservation law that a diagonal map moves dynamic range but never destroys it. What a diagonal scaling \emph{can} do (rebalance difficulty between operands, at zero cost) makes it a common first stage, and several hybrid methods carry a scale alongside their rotation for exactly this reason (OSTQuant and ParoQuant both learn one jointly with their rotation), even though the canonical rotations do not.

\subsection{Permutations: relabeling the coordinates}
\label{sec:tax-permutation}

A permutation matrix is an orthogonal transform with no continuous parameters; it only relabels coordinates. It preserves the function of a layer for the same reason a diagonal scale does: permuting the output channels of one layer and the matching input columns of the next leaves the composition unchanged, an invariance made explicit by \citet{martinez2021pqf}. What a permutation buys is control over \emph{grouping}. Under a shared per-group scale, which coordinates land in which group determines how the outlier mass is distributed across the groups. The grouping choice is a genuine lever, even though the global multiset of magnitudes is untouched.

The reorder-based line began with RPTQ \citep{yuan2023rptq}, which clusters activation channels by their $(\min,\max)$ range with $k$-means and reorders channels so that each cluster is contiguous and shares one set of quantization parameters. RPTQ was the first method to reach 3-bit LLM activations. The permutation is not applied online, since it is fused into the LayerNorm write addresses and into the linear layer's row/column indexing. Atom \citep{zhao2023atom} specializes the reorder to mixed precision, moving the 128 highest-energy channels to the end of the matrix so that the \texttt{INT4} bulk and the \texttt{INT8} outlier block are each contiguous and hardware-regular. Atom keeps the same fusion into the preceding operator, and measures its residual cost at less than half a percent of runtime.

QLLM \citep{liu2024qllm} attacks the same activation-channel outliers by \emph{disassembling} an offending channel into $T$ equal replicated sub-channels (each carrying $1/T$ of the magnitude while the replicated weight rows keep the output invariant) and then \emph{reassembling} similar channels by bipartite matching to restore the original width. This is a dimension-preserving reallocation rather than a pure reorder. The disassembly step is the $T$-way form of pre-LLM Outlier Channel Splitting \citep{zhao2019outlier}, which duplicates an outlier channel and halves its values so the network stays functionally identical but \emph{wider}. The reassembly step is QLLM's addition.

Two special cases deserve mention. GPTQ's activation ordering \citep{frantar2022gptq} is a permutation of the \emph{processing} order rather than of the stored layout. Columns are quantized in order of decreasing Hessian diagonal, so the most important columns are rounded while most of the weights remain free to compensate. The permutation is inverted before the weights are written, so with a single scale per row it leaves zero runtime footprint. Group-wise scales are the one case where the ordering reaches inference: the groups are then formed over consecutive columns of the \emph{permuted} matrix, so after inversion each stored column carries its own group index, which is why the authors added a \texttt{-{}-static-groups} flag that fixes the grids on the original order instead. (The \texttt{act-order} ordering is a later addition to the authors' own released code (\texttt{IST-DASLab/gptq}), documented there as one of two extra tricks alongside \texttt{-{}-true-sequential} and re-exposed as \texttt{desc\_act} by AutoGPTQ. The original paper quantizes columns in a single fixed order, arguing it suffices for large layers.) With one scale per row it is thus the cheapest transform in the survey, and it composes with almost everything. At the other extreme, PermLLM \citep{zou2025permllm} \emph{learns} its permutation, relaxing the discrete choice to a doubly-stochastic matrix by Sinkhorn normalization, training it end-to-end with a straight-through estimator, and hardening it to a hard permutation with the Hungarian algorithm. Its target, though, is N:M pruning rather than quantization.

PQF \citep{martinez2021pqf}, from the vision-compression literature, searches for the permutation that makes weights easiest to vector-quantize, connecting reordering directly to the rate--distortion view of \Cref{sec:classical}. Concretely, it minimizes the determinant of the sub-vector covariance, which under a zero-mean Gaussian model of the sub-vectors bounds the codebook distortion, so that the reordered sub-vectors become more compact (lower-volume covariance) and the layer's single shared product-quantization codebook reconstructs them more accurately. This is a between-group \emph{concentration} that is the mirror image of the within-group spreading a scalar shared scale rewards (distinct from the within-vector isotropization that incoherence supplies for lattice codebooks, \Cref{sec:vq}). (SliM-LLM \citep{huang2024slimllm}, sometimes grouped with these reorder methods, is not in fact a permutation; it keeps channels in contiguous 128-element groups and allocates per-group bit-widths by salience, so it belongs to the mixed-precision line outside our scope.)

Permutations reach their most effective form when combined with a rotation. DuQuant \citep{lin2024duquant} smooths first, then interleaves greedy block-diagonal rotations, each built to fold the current worst outlier dimension uniformly across a block, with a ``zigzag'' permutation that deals the ranked outlier channels back and forth across the blocks so that every block receives balanced outlier mass. A proven bound on the per-block \emph{mean} is layered on the rotation's guarantee that the per-block \emph{maximum} stays controlled. This hybrid sits on the boundary between this family and the orthogonal transforms, and it previews the recurring lesson of the rest of the taxonomy. A permutation alone cannot flatten a distribution, but it can equalize the load a subsequent rotation must carry.

\subsection{Fixed orthogonal transforms: incoherence for free}
\label{sec:tax-fixed-orthogonal}

The transforms that unlocked \texttt{W4A4} are orthogonal rotations, and the decisive observation (\Cref{prop:rht}) is that a rotation need not be adapted to the data to work. A single random rotation, applied to any fixed vector, drives its coordinates to a sub-Gaussian profile whose crest factor is $O(\sqrt{\log d})$ in expectation, and with high probability by the tail branch of \Cref{lem:subg}, spreading any outlier across all coordinates. That spreading is incoherence, and realizing it cheaply is the whole game. QuIP \citep{chee2023quip} introduced the principle, first rescaling the channels by a Hessian-derived diagonal $(\operatorname{diag}(X^{\!\top}X)/\operatorname{diag}(W^{\!\top}W))^{1/4}$ and then conjugating both the weight matrix and the calibration Hessian by random orthogonal matrices to make both $\mu$-incoherent (so that no entry, whether of the weights or of the Hessian's eigenvectors, is much larger than its typical $\sim\!1/\sqrt d$ scale) before an adaptive rounding step. To keep the rotation affordable it uses a Kronecker product $U_1\otimes U_2$ of small factors, so the multiply costs $O(d\sqrt d)$ rather than $O(d^2)$. QuIP\# \citep{tseng2024quipsharp} replaced the Kronecker orthogonals with the randomized Hadamard transform, a Hadamard matrix composed with a random sign diagonal, which achieves a strictly better incoherence bound at $O(d\log d)$ with only $\pm1$ arithmetic. It paired that rotation with a codebook on the $E_8$ lattice, chosen for its packing density and, being also a low-second-moment quantizer, recovering part of the space-filling gain of \Cref{thm:gishpierce}.

QuaRot \citep{ashkboos2024quarot} turned incoherence into a deployable recipe by exploiting computational invariance (\Cref{thm:invariance}). With the RMSNorm gains first folded away into the consuming weights (\Cref{prop:gamma}), as \Cref{thm:invariance} requires, a single global orthogonal $Q$ inserted into the residual stream can be absorbed into the adjacent weight matrices ($W_{\mathrm{in}}\!\leftarrow W_{\mathrm{in}}Q$, $W_{\mathrm{out}}\!\leftarrow Q^{\!\top}W_{\mathrm{out}}$, matching the convention of \Cref{thm:invariance}), so that most of the rotation costs nothing at inference. Only a few structured Hadamards remain as cheap online transforms: before the down-projection, on the query/key path (post-RoPE, which is why it cannot be folded), and across heads on the attention output; a further head-wise rotation fused into the value projection is what makes the value cache quantizable. Because a randomized Hadamard is data-free, QuaRot needs no calibration for the rotation itself. It also rotates the KV-cache, on \texttt{Llama-2-70B} taking weights, activations, and KV-cache all to 4 bits (\texttt{W4A4KV4}) at under half a perplexity point. \citet{ashkboos2024quarot} visualize the collapse of the activation-outlier channels that the rotation produces directly, making it the reference architecture that the learned methods of \Cref{sec:tax-learned-orthogonal} refine.

Later data-free constructions probe the residual gap. GSR \citep{choi2025grouped} orders the Walsh functions of a block-diagonal Hadamard by sequency, clustering similar-frequency components to lower error at very low bit-widths, and can stand alone as the $R_1$ of a training-free pipeline; it additionally serves as an improved initialization for SpinQuant and OSTQuant rather than an extra composed factor. It is training-free and data-free. Rotated Runtime Smooth \citep{yi2024rotated} applies a QuaRot-style online Hadamard first and then a runtime smoothing scale to mop up the channel-wise outliers the rotation leaves behind. A raw per-channel scale would need its own rescale inside the reduction, so the deployed form reorders channels by magnitude and takes the maximum over each group of the GEMM's block width, making the scale constant \emph{within} each block so that it rides along with the per-group dequantization the datapath already pays (\Cref{sec:sys-dequant}). It needs no calibration. It is thus a fixed rotation carrying a runtime max-based per-group scale, not SmoothQuant's balanced exponent, both acting on the feature axis. OptRot \citep{gadhikar2025optrot} stays data-free but, rather than a fixed Hadamard, optimizes the rotation over the orthogonal group, for weight-only and \texttt{W4A8} quantization, against a weight-outlier proxy (a fourth-power objective), narrowing the residual gap. It is a data-free yet learned construction, which is why \Cref{tab:taxonomy} classes it \textsc{LrnO} for the learning even though we narrate it here among the data-free rotations.

A distinct branch of this family trades a little redundancy for robustness, and belongs conceptually with the frame methods of \Cref{sec:tax-sequence-frame}. Kashin quantization \citep{merkulov2024kashin} writes each vector as $x\approx u+Qv$ in a $2\times$ overcomplete basis so that both factors have small $\ell_\infty$ norm. The construction is a calibration-free spreading of energy, its guarantee stated for a Haar-random orthogonal map, at the cost of doubling the coefficient count. FrameQuant \citep{adepu2024framequant} quantizes in a tight fusion frame with redundancy $r\approx1.1$, where the overcompleteness averages out quantization noise with an $O(1/r)$ reduction in MSE, and lets GPTQ do the rounding in the frame coordinates. 

Fixed orthogonal transforms occupy a sweet spot the theory explains cleanly. \Cref{prop:rht} guarantees that their randomized members achieve near-optimal flattening in expectation with no calibration in the rotations as constructed (QuIP prefixes a Hessian-derived diagonal rescale, and QuIP\#'s fine-tuning later relaxes its sign vectors by gradient) and, for the global residual-stream rotation that computational invariance (\Cref{thm:invariance}) folds, at near-zero runtime. Several members still keep an online kernel, though, among them QuaRot's few structured Hadamards, QuIP and QuIP\#, the frame constructions of FrameQuant and Kashin, and RRS, whose runtime max-smoothing cannot be folded by construction. What they leave on the table is the gap between a \emph{random} near-flat rotation and the \emph{data-optimal} one, the gap the next two families try to close.

\subsection{Learned orthogonal transforms: adapting the rotation}
\label{sec:tax-learned-orthogonal}

If a random rotation already flattens most distributions, is there anything to gain by fitting the rotation to the data? The learned-orthogonal family answers yes, but modestly, and its interest lies as much in what objective it optimizes as in the rotations themselves. SpinQuant \citep{liu2024spinquant} uses a four-rotation layout concurrent with QuaRot's and makes the two foldable rotations (the residual-stream $R_1$ and the head-wise value rotation $R_2$) learnable, optimizing them on the Stiefel manifold by Cayley SGD to minimize the quantized network's end-task loss. The two non-foldable rotations stay fixed as online Hadamards. Learning the rotation against the true task loss is the most direct objective but also the most expensive. The rest of the family is largely a search for cheaper surrogate objectives that still track the crest factor \Cref{prop:flat-opt} identifies as the quantity fixed-rate error charges.

Those surrogates are revealing. KurTail \citep{akhondzadeh2025kurtail} minimizes the gap between the kurtosis of the rotated activations and the kurtosis of a uniform distribution, a fourth-moment proxy for flatness. Cayley Adam does the optimizing, and KurTail is cheap enough to run on a single GPU at 70B. DartQuant \citep{shao2025dartquant} shapes the whole distribution with a ``Whip'' loss $\sum_i e^{-|x_i|}$ that expands the interval near the origin and compresses the tails. It sidesteps the manifold cost with a QR-based orthogonal parametrization (optimize an unconstrained latent, keep its QR factor), enabling 70B calibration on a single consumer card. ButterflyQuant \citep{xu2025butterflyquant} parametrizes the rotation as a butterfly network of $\tfrac{n}{2}\log n$ Givens rotations (orthogonal by construction, applied online at $O(n\log n)$) trained with an explicit uniformity-KL regularizer.

HARP \citep{zagitov2026harp} uses sparse butterfly-like stages similar to ButterflyQuant's, initialized from the randomized Hadamard and adapted per layer and backend, with mixed-radix support for non-power-of-two dimensions. It also complicates the flattening objective it inherits, reporting that maximizing a generic incoherence score is \emph{not} the true target. Its learned basis barely moves the weight incoherence ($\mu_W\ 5.77\!\to\!5.10$) and even \emph{raises} the classical Hessian incoherence ($\mu_H\ 6.6\!\to\!11.3$). Yet it cuts \texttt{Llama-2-7B} 2-bit \texttt{WikiText} perplexity over the QuIP\#/fixed-RHT baseline from $8.95$ to $7.85$ at context length $2048$ (and $8.22\!\to\!7.23$ at context $4096$); these are HARP's own baseline figures, while \Cref{tab:res_weight} lists the fully fine-tuned QuIP\# configuration at $6.66$ for the same cell. The reason it wins against that baseline is that it optimizes instead a Hessian-weighted, off-block-penalized proxy that aligns the rotated curvature with the quantizer's block structure rather than maximizing incoherence. Part of the margin is bought with bits: HARP spends $2.11$ bits per parameter against the baseline's $2.00$ in both comparisons ($2.05$ against $2.00$ on \texttt{Llama-2-13B}). Raw incoherence and deployed error are thus correlated but not identical, a caveat we return to in \Cref{sec:open}.

DFRot \citep{xiang2024dfrot} takes a corrective view of QuaRot, alternating a closed-form Procrustes solve for the rotation with a re-solve of the quantization scale and zero point. A random Hadamard sharply reduces error for typical tokens but, \citet{xiang2024dfrot} report, only marginally for the few tokens carrying the massive activations characterized by \citet{sun2024massive}, where a random orthogonal rotation would instead \emph{increase} their error, so those massive-activation tokens dominate the residual loss. DFRot therefore refines the rotation by orthogonal Procrustes under a loss that up-weights exactly those tokens, aiming to be simultaneously outlier-free and massive-activation-free. ConQuR \citep{thrash2026conqur} also avoids gradients, alternating a Procrustes solve that aligns normalized activations with the corners of an inscribed hypercube against an online calibration loop that avoids storing activations. OSTQuant \citep{hu2025ostquant} steps toward the next family by learning an orthogonal rotation and a diagonal scaling together against a KL-Top loss, with a quantization-space-utilization metric motivating the construction.

Two further methods depart from the single global rotation, in opposite directions. ReSpinQuant \citep{kim2026respinquant} buys expressivity, learning a full-size Cayley-optimized rotation for \emph{each} layer, and folds those rotations into the attention and FFN weights, paying only a low-rank online correction for the near-identity basis mismatch the residual connection leaves between adjacent layers. ParoQuant \citep{liang2025paroquant} trades generality for cost instead, restricting the learned rotation to a product of pairwise (Givens) rotations together with per-channel scaling, targeted at reasoning models. Unlike the foldable rotations above, its structured rotation is applied online and carries a modest (order-ten-percent) runtime cost, placing it at the boundary with the non-orthogonal family.

A data-aware rotation of a different kind closes the family. ResQ \citep{saxena2025resq} uses PCA, a KLT (\Cref{sec:classical}), to identify the roughly one-eighth of the coordinates carrying the highest activation variance. It keeps those in \texttt{INT8} and quantizes the rest to \texttt{INT4}, with a random orthogonal rotation inside each subspace to suppress residual outliers, and proves that this split minimizes an upper bound on the quantization error under a Gaussian model of the rotated coefficients. It is a telling exception to the inversion: a concentration transform earns its keep in deployment precisely because mixed precision restores a form of per-coordinate bit allocation (\Cref{sec:inversion}), the regime in which concentration, not flattening, is optimal.

The pattern across the family confirms the inversion rather than contradicting it. Because a random Hadamard already achieves near-optimal flattening (\Cref{prop:rht}), a learned rotation can only recover the residual gap between random-flat and data-optimal-flat, and at \texttt{W4A4} on \texttt{Llama-2-7B} the reported gains over QuaRot are correspondingly of order a fraction of a perplexity point, though they grow on the harder \texttt{Llama-3} models, whose activations sit further from flat and so leave a data-aware rotation more of a gap to close. The genuine advances here are in efficiency (QR and butterfly parametrizations that avoid the Cayley manifold cost) and in objective design, where kurtosis, uniformity, and the Whip loss are increasingly explicit stand-ins for the crest factor that \Cref{prop:flat-opt} identifies as the quantity controlling fixed-rate error under an AbsMax scale; where a clipping threshold is fitted instead, part of the error moves into the overload term \Cref{lem:bennett} sets aside, and the crest factor becomes a proxy rather than the exact charge.

\subsection{Non-orthogonal affine transforms: spending the last degrees of freedom}
\label{sec:tax-affine}

The rotations of the last two subsections preserved lengths, and the diagonal maps before them changed only per-coordinate scale. A general $T\in \mathrm{GL}(d)$ gives up both constraints, opening the full $d^2$-parameter design space, and with it the ability to \emph{whiten}\footnote{The same data-aware whitening, applied to \emph{compression} rather than function preservation, underlies an adjacent low-rank line: SVD-LLM \citep{wang2024svdllm} whitens by activation statistics before truncating the weight spectrum, and ASVD \citep{yuan2023asvd} applies a related diagonal activation-aware scaling. Because they compress the layer rather than preserve its function, they sit outside our scope, but SVD-LLM's whitening shares the core that WUSH makes exact and function-preserving.}: to rescale the coordinate axes by the data's own second moments before rotating. Whitening is the one move a pure rotation cannot make. An orthogonal map redistributes a group's energy but cannot reshape its second-moment ellipsoid, so the distribution's anisotropy can only be redistributed across coordinates by a rotation, never removed from the spectrum. A \emph{data-agnostic} rotation does not even redistribute it evenly: on correlated data the rotated diagonal $(H^{\!\top}\Sigma H)_{ll}$ remains uneven and the shared per-group scale is pinned by the largest entry. A data-aware \emph{orthogonal} map can flatten that diagonal (MambaQuant's $KH$ does exactly this, \Cref{sec:beyond-attn}), but it still cannot change the ellipsoid's spectrum. It is that spectrum the bilateral objective of \Cref{thm:wush} charges for, which is why the last gain requires dropping orthogonality. Whitening rescales the coordinate axes by the data's own second moments, driving both operands to one and the same diagonal second moment \emph{before} any rotation acts. That balancing equalizes the two operands against each other, but it leaves the second moment diagonal rather than flat, and \citet{chen2025wush} report that on the outlier-heaviest projections the whitening alone is worse on \texttt{INT4} than no transform at all (a layerwise loss of $213$ against the identity's $170$ on the $q$-projection), because without the Hadamard the non-orthogonal factor can amplify individual coordinates and so raise the AbsMax group scale. The Hadamard that follows is what flattens that diagonal, which is why in their model the two are jointly and not separately essential: setting either factor to the identity leaves the same suboptimal trace term.

A fixed Hadamard, as \citet{chen2025wush} note, ``does not adapt to the statistics of the underlying weights or activations,'' so in what sense is it optimal for quantization? WUSH is the closed-form, data-aware answer in this family. It constructs, per block, the transform
\begin{equation}
T \;=\; H \, S^{-1/2}\, U^{\!\top} {W'}^{\!\top},
\label{eq:wush}
\end{equation}
where $W'$ and $X'$ are the (lower-triangular) Cholesky factors of the regularized weight and activation second moments, ${W'}{W'}^{\!\top}=d_{\mathrm{out}}^{-1}W_{(i)}^{\!\top}W_{(i)}$ (the input-axis weight Gram, matching $\HX$'s dimension) and ${X'}{X'}^{\!\top}=N^{-1}X_{(i)}^{\!\top}X_{(i)}$ ($N$ the token count of \Cref{sec:notation}), the subscript restricting each operand to the block's input coordinates. The factors $U,S,V$ come from the singular value decomposition of the whitened weight--activation cross-correlation ${W'}^{\!\top}X'=USV^{\!\top}$, and $H$ is a fixed \emph{orthonormal} Hadamard ($HH^{\!\top}=I$, which is what makes the companion \eqref{eq:xvsh} exactly $T^{-\!\top}$). Thus $S^{-1/2}$ is a data-dependent balancing that equalizes the two sides of the proxy \eqref{eq:bilateral-proxy}, and the name \textbf{WUSH} is a mnemonic for the four factors $W'\!,U,S,H$.

Despite being built from the weight factor $W'$, \eqref{eq:wush} is the map carried on the \emph{activation} side; the weights carry its companion, built from the activation factor $X'$,
\begin{equation}
T_{\mathrm{xvsh}} \;=\; H\,S^{-1/2}V^{\!\top}{X'}^{\!\top} \;=\; T^{-\!\top},
\label{eq:xvsh}
\end{equation}
the inverse-transpose of \eqref{eq:wush}, so the pair preserves the layer's output exactly by the function-preserving identity of \Cref{def:fpt}, not by the orthogonal-only invariance of \Cref{thm:invariance}. The two sides differ in which SVD factor, $U$ or $V$, and which Cholesky factor, $W'$ or $X'$, they carry. The regularized moments $M+\lambda d^{-1}\tr(M)I$ are Cholesky-factored, so the construction is numerically stable and entirely closed-form, with no gradient training. It is also cheaper than a GPTQ Hessian by a factor of order $d_{\mathrm{in}}/d$ (the input dimension over the block size), ``a negligible overhead on top of GPTQ.'' The $S^{-1/2}$ sitting \emph{between} two orthogonal factors is precisely what an orthogonal transform cannot produce. It whitens, and whitening is what carries WUSH past the rotations of the previous subsection. This is why WUSH attains the optimum of \Cref{thm:wush} (optimal under the floating-point AbsMax \emph{model} and, for the integer grid, near-optimal within a $d^{o(1)}$ factor, $d$ the transform block size, for zero-mean Gaussian or Laplacian data and a factor $d$ for an arbitrary distribution) and why it is the shared-scale mirror image of the KLT (\Cref{thm:klt}). Both diagonalize a second-moment operator, but the KLT concentrates energy for an allocation-flexible coder while WUSH flattens it for a shared-scale one, the two poles that \Cref{cor:notransfer} separates, outside the degenerate spectra it excepts.

Only the activation-side transform is online (the weight side folds into the weights offline), and a fused kernel keeps that online cost within about $1.3\%$ on average of an optimized blockwise-Hadamard \texttt{W4A4} pipeline (QuaRot/MR-GPTQ class) on the shapes and hardware the source reports (RTX~5090, batch~1024, MXFP4), so its deployment cost is essentially that of an online-Hadamard pipeline. \citet{chen2025wush} plot the geometry of the resulting optimal transform directly.

The learned members of this family reach for the same expressiveness by gradient descent rather than closed form. FlatQuant \citep{sun2024flatquant} learns a per-layer invertible transform with a Kronecker structure $P=P_1\otimes P_2$, applied online by a fused kernel that keeps the two small matmuls cheap, and jointly learns clipping thresholds. AffineQuant \citep{ma2024affinequant} learns a single invertible matrix $A$ per layer, dense wherever a linear predecessor can absorb it, keeping it invertible by enforcing diagonal dominance and admitting off-diagonal mass through a mask that starts near-diagonal and densifies over training. After optimization, $A$ folds into the weights at zero runtime, though at the post-LayerNorm positions under weight--activation quantization the source optimizes only its diagonal, so that it can be absorbed into the norm's own scale and bias. LATMiX \citep{gordon2026latmix} parametrizes its full invertible affine map in LU or QR form to guarantee invertibility, targets microscaling block formats, and proves a quantization-error bound in terms of $\|A^{-1}\|$. Its two learned maps, one acting globally on every block's input activations and one on the attention's scaled dot-product activations, fold into the weights, though only approximately function-preserving, since a non-orthogonal map on the residual path does not commute with the RMSNorm and the source restores equivalence by training rather than by algebra; the third transform, inside the FFN across the nonlinearity, is not learned and stays online. BASE-Q \citep{he2025baseq} keeps its residual rotation out of the optimization altogether, constructing it in closed form as $U^{\!\top}\!H$ from the PCA basis of the weight covariance composed with a Hadamard, and learns only the per-block value rotation by Cayley parametrization; the $R_{\mathrm{qk}}$ and $R_{\mathrm{down}}$ rotations stay online Hadamards. To that it adds two learnable quantizer-side corrections, a channel bias that recenters each channel and an asymmetric per-quantizer scaling, applied at the O- and down-projection inputs, that fits the clip range to the post-rotation Gaussian shape, jointly optimized per block. Both corrections run online in a fused quantize/dequantize kernel, their compensating output-bias term folding into the layer bias.

BASE-Q's motivation is a sharp diagnosis of what a rotation leaves behind. The authors report that once the spread is flattened, the residual variation in per-channel means accounts for as much as $85\%$ of the layer's rounding error on Qwen2.5-3B. They also report that MSE-optimal $4$-bit clipping of the rotated, near-Gaussian activations still discards about $18\%$ of the total activation energy, the share a near-Gaussian concentrates in the clipped tails. A mean offset and a clip loss both survive \emph{any} orthogonal flattening (a shift is not a rotation, and neither is a rescaling), which is precisely why the affine family, not the rotations, defines the current frontier. FPTQuant \citep{vanbreugel2025fptquant} assembles lightweight function-preserving transforms tailored to the transformer's structure: mergeable per-head query/key \emph{scale-and-rotate} maps, a per-head value matrix, a diagonal MLP scaler, and a calibration-free per-token residual rescaling.

This family is the most expressive, and it is where the flattening pole's one model-level optimality result sits (\Cref{thm:wush}). The tension it exposes (and it is the tension the whole survey turns on) is between expressiveness and deployability. A general invertible transform that is \emph{folded} into weights (AffineQuant, and the global and attention parts of LATMiX) is free at inference but constrained by what the surrounding architecture allows it to absorb, while one that whitens the activations optimally (WUSH), or that learns a general invertible map against the quantized loss (FlatQuant), must run online. So must BASE-Q, whose learnable channel bias and asymmetric per-quantizer scaling its authors fuse into an online quantize/dequantize Triton kernel, at negligible but non-zero cost. That the online cost has fallen to under a tenth of runtime is what has made this family, rather than the fixed rotations, the current frontier.

\subsection{Sequence-axis and overcomplete transforms}
\label{sec:tax-sequence-frame}

The families above all act within $\R^d$ on the feature (channel) axis of a single matrix. Two smaller lines leave that setting: one moves to the sequence axis, the other keeps the feature axis but leaves $\R^d$ for a redundant superspace. Both matter because they compose with the feature-axis transforms rather than competing with them. The sequence-axis idea is that some structure is a property of \emph{tokens} rather than channels, and is better addressed by acting on the sequence; it admits both poles of the inversion, with \Cref{sec:format}'s STaMP the concentration-side member. PrefixQuant \citep{chen2024prefixquant} prepends a short fixed prefix built from the token types that most often carry outliers, identified offline by the ratio of a token's maximum activation to the median of token maxima, with the prefix's KV entries computed once and held in high precision. With the token-wise outliers thus removed (the token-wise maximum ratio drops from several hundred to a few, whereas a Hadamard rotation alone leaves the token-wise outliers largely intact), the remaining activations can be quantized with cheap static per-tensor scales instead of dynamic per-token ones, atop the usual QuaRot rotations.

The frame line, already met in \Cref{sec:tax-fixed-orthogonal} through Kashin and FrameQuant, is the other departure. Rather than change the basis within $\R^d$, both embed the vector in a larger, redundant space, but for different reasons: FrameQuant's redundancy averages the quantization noise down, while Kashin's spreads the vector so that both coefficient blocks carry a small $\ell_\infty$. Both cost storage overhead and remain online, the redundant basis being non-square and so unfoldable into a weight matrix, but they are the direct low-bit heirs of the redundant-representation and tight-frame constructions of classical coding, and they show that overcompleteness is a lever orthogonal to the choice of transform. Neither the sequence-axis nor the frame methods displace the channel-axis transforms. They are additional stages that the frontier pipelines increasingly stack: a rotation for channel outliers, a prefix for token outliers, a scale for residual imbalance.

Reading the taxonomy as a whole, the six families are neither exclusive nor equally live. The clear direction of travel is up the degrees-of-freedom ladder, from the diagonal scalings that dominated 2022--2023 to the non-orthogonal affine maps that define the 2025--2026 frontier, tempered at every step by the deployability constraint that a transform must either fold into weights or run in a cheap online kernel. The most competitive recent methods are hybrid (DuQuant, OSTQuant, BASE-Q, FPTQuant, and PrefixQuant each combine two or three of these levers) because the levers address different failure modes: a diagonal scale rebalances the operands, a permutation balances the groups, a rotation flattens within a group, whitening balances the two sides of the proxy, and a prefix handles the tokens the channel transforms cannot see. How these stages interact with each other and, crucially, with the rounding algorithm and the number format is the subject of the following section.

\section{Composition: Rounding and Codebooks}\label{sec:composition}

A transform is never deployed alone. It is the first stage of a pipeline whose remaining stages (an adaptive rounding algorithm, and increasingly a codebook) do the actual bit reduction, and the accuracy of the whole depends on how these stages interact. This section makes that interaction precise. We adopt three verbs for it. Two stages \emph{compose} when each does work the other does not, so that stacking them helps. One stage \emph{substitutes} for another when it captures a benefit the other would also have provided, so that stacking them is redundant. And two stages \emph{co-optimize} when the best choice of one depends on the other and they must be designed together. (A limiting case of composition, which we call \emph{enabling}, arises when one stage is not merely helpful to the other but a precondition for it, as the transform's Gaussianization is for the fixed lattice and trellis codebooks of \Cref{sec:vq}, though not for the data-aware codebooks fitted there to the raw weights.) The field has been using these relationships without naming them, and the confusion has real consequences. Reported gains from a new transform can be an artifact of a weak rounder whose work it overlaps with, and a codebook's optimality can silently assume a transform upstream. We first lay out the rounding stage (\Cref{sec:comp-rounding}), then resolve the substitution question the introduction promised (\Cref{sec:comp-substitution}), and finally show how transforms enable the fixed vector and lattice codebooks that reach two bits, and how a parallel data-aware branch reaches low bit-widths without one (\Cref{sec:vq}). \Cref{tab:composition} classifies the rounders, codebooks, allocation rules, and lookup kernels surveyed here by type and by their relationship to the transform, one row per method.

\begin{table}[!htbp]
\centering\footnotesize
\renewcommand{\arraystretch}{1.15}
\caption{\textbf{Rounding, codebook, allocation, and kernel methods (\Cref{sec:composition}), by type and by how each stands to the transform:} \emph{composing} with it, \emph{substituting} for it, or \emph{enabled} by it.}\label{tab:composition}
\begin{tabularx}{\linewidth}{>{\raggedright\arraybackslash}p{3.0cm} >{\raggedright\arraybackslash}p{2.0cm} >{\raggedright\arraybackslash}X >{\raggedright\arraybackslash}X}
\toprule
\textbf{Method} & \textbf{Type} & \textbf{Optimizes} & \textbf{Relation to transform} \\
\midrule
WaterSIC~\citep{lifar2026watersic} & grid/allocation (variable-rate) & geometric-mean (det \ensuremath{\Sigma_X}) distortion; basis-free scalar-INT optimum \emph{under entropy coding} (rates reported as entropy, not log-cardinality) & near-substitute for a rotation at high rate (waterfills grid spacings instead); the worst-case equivalence is left open and the overlap shrinks at $2$--$4$ bits \\
FLUTE~\citep{guo2024flute} & kernel & throughput: LUT dequant keeping pace with tensor cores & none directly; it serves the codebook stage downstream of the transform, and only the scalar lookup-table codebooks, vector and lattice ones being future work in the source \\
GPTQ~\citep{frantar2022gptq} & rounder & quadratic layer loss; fixed-order blocked error feedback & applies no basis-changing transform; rounds on the given basis (composes; at high rate it is a rotation and per-coordinate allocation that are the near-substitutes, not the rounder, \Cref{sec:comp-substitution}) \\
GPFQ~\citep{lybrand2021gpfq} & rounder & layer output error under the mismatched input pair; greedy path-following & composes; corrects the committed previous-layer input mismatch, adjusts no future weights \\
LDLQ (QuIP)~\citep{chee2023quip} & rounder & quadratic proxy loss $\propto\tr(D)$; proved optimal, worst and average case, among rounders whose linear feedback depends on $H$ and not on $W$, when rounding to integers & composes; QuIP pairs it with incoherence processing, which supplies nearly all of the 2-bit gain (\Cref{sec:comp-substitution}) \\
OBC~\citep{frantar2022obc} & rounder & quadratic layer loss; exact greedy per-row order with optimal compensation & rounds on transform's basis (composes) \\
Qronos~\citep{zhang2025qronos} & rounder & true output error including activation-quantization mismatch and error accumulated from previously quantized layers & composes with Hadamard, QuaRot, SpinQuant, SmoothQuant \\
NF4 (NormalFloat)~\citep{dettmers2023qlora} & scalar codebook & equal-probability bins (max index entropy) under Gaussian & relies on weights already Gaussian post-absmax (no transform) \\
SqueezeLLM~\citep{kim2023squeezellm} & scalar codebook & Fisher-weighted k-means; dense-and-sparse outlier split & data-aware fit; substitutes for Gaussianizing rotation \\
AQLM~\citep{egiazarian2024aqlm} & vector/lattice & calibrated output error; additive multi-codebook, beam search & data-aware fit; substitutes for Gaussianizing rotation \\
GPTVQ~\citep{vanbaalen2024gptvq} & vector/lattice & Hessian-weighted k-means over weight sub-vectors & data-aware weighting substitutes for Gaussianizing rotation \\
PCDVQ~\citep{yue2025pcdvq} & vector/lattice & separate direction ($E_8$-sampled) and magnitude ($\chi_8$ Lloyd--Max) codebooks under Gaussianized weights & enabled by incoherence rotation; decouples direction from magnitude \\
QTIP~\citep{tseng2024qtip} & vector/trellis & Gaussian-source distortion; trellis-coded quantization at high dimension & enabled by incoherence rotation (Gaussianized weights) \\
QuIP\#~\citep{tseng2024quipsharp} & vector/lattice & $E_8$-lattice E8P codebook distortion at 2 bits & enabled by incoherence rotation; composes with block LDLQ \\
TurboQuant~\citep{zandieh2025turboquant} & vector (rotation $+$ scalar) & near-optimal distortion rate (within $\sim2.7\times$), MSE and inner-product & data-oblivious; random rotation makes coordinates near-independent, enabling optimal per-coordinate scalar quant \\
VPTQ~\citep{liu2024vptq} & vector/lattice & second-order optimization; channel-independent error compensation & data-aware fit; substitutes for Gaussianizing rotation \\
\bottomrule
\end{tabularx}
\end{table}

\subsection{The error-feedback rounding line}
\label{sec:comp-rounding}

Most of the rounding algorithms that dominate weight quantization descend from the Optimal Brain Surgeon framework introduced in \Cref{prop:obs}, with one independent line beside it: the greedy path-following quantizer of \citet{lybrand2021gpfq}, which corrects the error already committed, including the input mismatch the already-quantized preceding layers leave behind and the OBS line ignores, but does not adjust the not-yet-quantized weights. The two lines converge in Qronos below. Optimal Brain Compression \citep{frantar2022obc} instantiated the OBS framework for quantization. Under the quadratic layer loss with Hessian $H=2X^{\!\top}X$ (the rounding literature's $H$, not the Hadamard $H$ of \Cref{prop:rht}; context distinguishes them) (diagonally damped before inversion, as \Cref{prop:obs} notes) and independence across output rows, it repeatedly quantizes the single weight of least impact and optimally compensates the remainder,
\begin{equation}
p^\star=\argmin_p \frac{(\,Q(w_p)-w_p)^2}{[H^{-1}]_{pp}},
\qquad
\delta = -\,\frac{w_p-Q(w_p)}{[H^{-1}]_{pp}}\,H^{-1}_{:,p},
\label{eq:obc}
\end{equation}
each step being the exact optimum under the OBS model. Its cost is cubic per row, which does not scale to billions of parameters. GPTQ \citep{frantar2022gptq} removed that cost with three changes, only the first of which is a reduction in arithmetic. They observe \emph{empirically} that quantizing every row in one fixed column order costs little against the per-row greedy order, in particular on large, heavily parametrized layers; because $H$ depends only on the shared layer inputs and not on the weights, a common order then lets one $H^{-1}$, and one Cholesky factor of it, serve every row. The feedback is applied lazily in blocks of $128$ columns, which the authors introduce to raise arithmetic intensity rather than to lower the operation count. And the update is taken from a Cholesky factor for numerical reasons: on multi-billion-parameter models the repeatedly-updated $H^{-1}$ otherwise loses positive definiteness and destroys the layer. The result is the workhorse of the field, and the point to hold onto is that GPTQ applies \emph{no} basis-changing transform of its own (its act-order variant permutes the processing order, and \Cref{sec:taxonomy} classifies it on that ground). It rounds on whatever basis it is handed, propagating each column's rounding error $E_{:,j}=(W_{:,j}-Q(W_{:,j}))/[H^{-1}]_{jj}$ forward into the not-yet-quantized columns.

Two refinements matter for the composition story. QuIP's LDLQ rounder \citep{chee2023quip} writes the same forward feedback as $\widehat W = Q\!\big(W+(W-\widehat W)U\big)$, where $U$ is the strictly-upper-triangular factor of the LDL decomposition $H=(U+I)D(U+I)^{\!\top}$. QuIP proves this rounder \emph{optimal} (in both worst and average case, with a proxy loss proportional to $\tr(D)$) among all rounders whose linear feedback depends on $H$ but not on $W$, when rounding to integers. QuIP's theorem is the strongest optimality statement in the rounding line, and its caveat is instructive. It is optimality for the quadratic \emph{proxy} loss of \eqref{eq:bilateral-proxy}, within a restricted class, and that proxy is the layer's own output error only under the calibration set's second moment, never the model's end-to-end loss. The integer-rounding hypothesis is load-bearing too: \citet{chee2023quip} exhibit a finite-grid counterexample in which clamping to a bounded set of levels breaks the guarantee, and give a repaired algorithm for that case.

Qronos \citep{zhang2025qronos} then closed the gap the OBS line shares, and supplied in return the future-weight diffusion the path-following line lacks. GPTQ and LDLQ correct only the error from \emph{weights} rounded so far. They ignore the mismatch between the true activations $X$ and the \emph{quantized} activations $\widetilde X$ that a low-bit deployment actually feeds the layer. Qronos alternates an explicit error-\emph{correction} step that absorbs this activation-quantization error (and errors inherited from previous layers) with a GPTQ-style \emph{diffusion} step, each in closed form, the diffusion step admitting an efficient Cholesky recursion. GPTQ is recovered as the special case in which the activation mismatch vanishes ($\widetilde X = X$), which \citet{zhang2025qronos} record as a new reading of GPTQ itself; on their own statement Qronos subsumes both ancestors, GPTQ and the path-following line. Notably, Qronos is demonstrated running unmodified on top of Hadamard incoherence processing, SmoothQuant scaling, QuaRot, and SpinQuant. That the rounder runs on \emph{any} basis the transform hands it settles that the two are stackable; whether each does work the other does not, which is what composing requires, is the question of \Cref{sec:comp-substitution}. \citet{zhang2025provable} give what they state to be the first quantitative bounds on this proxy error, for OPTQ/GPTQ and Qronos, on an unbounded grid. 

\subsection{Substitute or compose? The transform--rounding relationship}
\label{sec:comp-substitution}

If error-feedback rounding is applied on whatever basis the transform produces, a sharp question follows: how much of a transform's benefit does the rounder already capture on its own? The cleanest answer comes from the high-rate analysis of \citet{ordentlich2026highrate2}. Writing the input covariance as $\Sigma_X=U^{\!\top}U$ (an upper Cholesky factor here, not the strictly-upper LDL factor above), they observe that GPTQ's equal-bit successive-cancellation rounding incurs a high-rate distortion proportional to the \emph{arithmetic} mean $\tfrac1n\sum_i U_{ii}^2$ of the Cholesky pivots. The information-theoretic limit, they note, is proportional to their \emph{geometric} mean $(\prod_i U_{ii}^2)^{1/n}=(\det\Sigma_X)^{1/n}$. The gap between arithmetic and geometric mean is exactly what a transform can close. A random rotation applied before rounding moves energy into the off-diagonal of $U$, which lowers the pivots' arithmetic mean toward their rotation-invariant geometric mean. The pivots themselves do not equalize: the source has the largest of them near the eigenvalues' arithmetic mean and the smallest near their harmonic mean, with the rest in between. On real \texttt{Llama-3-8B} covariances, \citet{ordentlich2026highrate2} report that this brings GPTQ to within about $0.1$ bit, varying with the layer type, of WaterSIC, both sides entropy-coded and their rates counted as entropy rather than log-cardinality, so the margin is not one a fixed-length deployed kernel inherits. WaterSIC is the basis-free scalar-INT optimum (itself within $\approx0.25$ bit of the rate--distortion bound at high rate, as below).

At high rate, a random rotation is a near-substitute for optimal per-coordinate bit allocation, because both attack the same basis-dependence of the rounding error. WaterSIC \citep{lifar2026watersic}, analyzed at high rate by \citet{ordentlich2026highrate2}, makes the duality explicit by taking the other route, keeping the basis fixed and instead waterfilling the per-coordinate grid spacings $\alpha_i \propto |U|^{1/n}/|U_{ii}|=|\Sigma_X|^{1/(2n)}/|U_{ii}|$ on the same GPTQ-style rounder. This is provably basis-free (its distortion depends only on $\det\Sigma_X$, so it is immune to rotation) and, at high rate with i.i.d.\ Gaussian weights and its integer output entropy-coded, within $\approx0.25$ bit of the rate--distortion bound, uniformly over the input covariance, which is the guarantee \citet{lifar2026watersic} themselves prove; \citet{ordentlich2026highrate2} supply the basis-free reading of it and the comparison against rotated GPTQ below. That margin is the $\tfrac12\log_2(2\pi e/12)\approx0.255$-bit space-filling gap of \Cref{thm:gishpierce}, the price of quantizing on the cubic lattice $\mathbb{Z}^n$ rather than an optimal one. Where WaterSIC \emph{provably} attains that geometric-mean ($\det\Sigma_X$) optimum, a random rotation is shown only \emph{empirically} to close most of the arithmetic-to-geometric-mean gap, to within the $\approx0.1$ bit above; the authors leave the worst-case gap between rotated GPTQ and WaterSIC open. So waterfilling is the provable route and rotation the empirically-good one. At high rate they are near-substitutes and layering both buys little, though the equivalence is not proven in the worst case. At 2--4 bits, where clipping and weight structure intrude, the overlap shrinks and is layer-dependent.

QuIP \citep{chee2023quip} marks the low-rate extreme, where the substitution collapses entirely. Its LDLQ rounder \emph{without} the incoherence processing is, on \citeauthor{chee2023quip}'s own statement, a more efficient implementation of OPTQ/GPTQ, so the whole gap between the two at $2$ bits (\texttt{Llama-2-70B} perplexity $123.9$ for the rounder alone against $6.33$ once QuIP's incoherence processing is added) is bought by the pre-processing, not the rounder. That pre-processing is a package rather than a rotation: a Hessian-derived diagonal rescale $(\operatorname{diag}(X^{\!\top}X)/\operatorname{diag}(W^{\!\top}W))^{1/4}$, the Kronecker-factored random-orthogonal conjugation, a random permutation of entries, and a quantization range taken from $\|W\|_F$ rather than the usual $\max_{ij}|W_{ij}|$, which is a clipping choice and no transform at all. \citet{chee2023quip} ablate the sub-steps and report all of them necessary for the full gain, the permutation alone worth $74$ perplexity at $2$ bits on \texttt{OPT-125m} (averaged over WikiText-2, PTB and C4), so the credit belongs to the package and not to the rotation by itself. Where at high rate a rotation and per-coordinate allocation are near-substitutes for one another, either one reaching an optimum the other also reaches, at $2$ bits the rounder does not stand in for that pre-processing on the models measured: the pre-processing does essentially all of the work and the rounder almost none.

The two routes are this survey's two regimes in miniature. GPTQ spends the same number of bits on every coordinate, the equal-allocation half of the fixed-rate side of \Cref{thm:inversion} (the rates here counted as entropies rather than set by a group AbsMax scale, so the parallel is one of allocation and not of objective), and there flattening is what pays: a rotation lowers the pivots' arithmetic mean toward their rotation-invariant geometric mean. WaterSIC instead restores per-coordinate allocation, the side on which the geometric mean is already the operative objective, and it therefore needs no flattening at all, its distortion depending on $\det\Sigma_X$ alone. That two opposed prescriptions should meet at all is a fact about the high-rate limit, not a merging of the regimes: the flat profile that minimizes the shared-scale surrogate is the one that \emph{maximizes} the allocation-flexible surrogate, which is what \Cref{thm:inversion} says and what makes the coincidence worth recording. The high-rate result gives the substitution its quantitative form: after a rotation, GPTQ sits within about $0.1$ bit of the basis-free allocation optimum, itself within $\approx0.25$ bit of the rate--distortion bound, in the high-rate, controlled-clipping regime, for the covariances measured, and under the conditions stated above.

A second and purely empirical overlap runs between the transform and the rounder, and it does not follow from the high-rate result above. Transforms evaluated against a rounder \emph{without} error feedback report larger gains than the same transforms evaluated against GPTQ, because part of what the transform ``fixes'' is what GPTQ's error feedback would have repaired anyway. This is the overlap the section opener warned about, and it is a caution about evaluation protocol rather than a theorem.

Two further results sharpen the relationship rather than reduce it to substitution. \citet{chen2025geometry}, and independently \citet{birnick2026lattice}, show that GPTQ, run from the last coordinate to the first, is \emph{identical} to Babai's nearest-plane algorithm for the closest-vector problem on the lattice whose basis is the Cholesky factor of $H$; \citet{chen2025geometry} carry the equivalence through the per-channel scale stretch, while \citet{birnick2026lattice} proves it for a single scalar step size. GPTQ therefore inherits Babai's error bound (in the no-clipping regime). This recasts the ``transform'' as the choice of lattice basis. A rotation of the input basis changes the lattice, hence the achievable rounding error, and the classical lattice-reduction machinery (LLL) is in principle the transform that would improve it, exact minimization being out of reach since the closest-vector problem is NP-hard and reduction buys only an approximation factor. \citet{chen2025geometry} note, though, that clipping, and the per-output-channel scales that generate a different reduction for each row, disable it in practice, leaving a min-pivot column ordering as the composable lever. \citet{birnick2026lattice} likewise leaves an LLL-style reduction wrapper to future experimental evaluation.

The second result is the linearity theorem of \citet{malinovskii2025higgs}, stated as \Cref{thm:linearity}. It supplies the bridge from a layer's relative weight error to the model's loss. The theorem says the perplexity increase is, to leading order, a weighted sum $\sum_l \alpha_l t_l^2$ of per-layer relative \emph{weight} errors. So a transform that makes the post-transform weights Gaussian renders $t_l^2$ constant and weight-independent, since a fixed-rate grid on a fixed-\emph{shape} source has a relative error that depends only on the bit-width, not on the particular weights. Under the isotropic weight-Hessian assumption \Cref{thm:linearity} requires, this is what makes an MSE-optimal grid end-to-end optimal and turns bit allocation into a separable allocation solved by dynamic programming. The data-aware layer-wise error the theorem omits can, at aggressive bit-widths, reorder that ranking (\Cref{sec:pitfalls}). Here the transform and the quantizer co-optimize, in both directions the definition asks for: the best grid depends on the transform through this Gaussianization, and \Cref{sec:format} supplies the converse, where the element grid decides which transform is optimal. The rotation is not fixing what the rounder would fix; it is making each layer's error a function of its own bit-width alone, so that per-layer choices compose additively rather than fighting one another. The synthesis, then, is that at high rate a rotation and per-coordinate allocation are partial substitutes, each closing most of the same arithmetic-to-geometric-mean gap, of which about $0.1$ bit is left unclosed by the rotation, under the entropy-coded accounting above. But the transform does two things neither the rounder nor the allocation can: it Gaussianizes the weights, which we now show enables a strictly better codebook, and it makes each layer's relative error a function of its bit-width alone, which is what renders the theorem's per-layer sum separably allocatable.

\subsection{Codebooks enabled by incoherence}
\label{sec:vq}

The scalar quantizers above tile the line with uniform intervals, forfeiting the space-filling gain that \Cref{thm:gishpierce} promised to vector and lattice quantizers. Recovering that gain on LLM weights is the second thing a transform enables, and the enabling mechanism is again incoherence. A random Hadamard rotation makes each weight's marginal approximately Gaussian and the block incoherent (the rotated coordinates are sub-Gaussian, though not strictly independent). How close that approximation must be is exactly where a codebook asks more than an AbsMax scale does, and \Cref{sec:incoherence} records the shortfall: along the axis a codebook block spans, the deployed methods see a \emph{single} randomized Hadamard, which carries neither the two-fold marginal guarantee nor the three-fold covariance one of \citet{benbasat2026quantizing}. The Gaussian source in what follows is therefore the model these codebooks are designed against and validated on, not a theorem about the weights they receive. A codebook matched to that approximately isotropic Gaussian ball then applies uniformly to every block. The classical vector-quantization theory of \Cref{sec:classical} thus resurfaces inside weight-only quantization. It is legal there, and not on the activation datapath, because codebook weights are dequantized through a lookup table rather than a shared-scale GEMM.

The scalar entry point is NormalFloat \citep{dettmers2023qlora}. Its 4-bit codebook places reproduction points at the quantiles of the standard normal, $q_i=\tfrac12\big(Q_X(\tfrac{i}{2^k+1})+Q_X(\tfrac{i+1}{2^k+1})\big)$ with $Q_X$ the inverse Gaussian CDF and an asymmetric split that preserves an exact zero, so that each bin carries equal probability mass. Equal mass maximizes the marginal entropy of the fixed-length index (uniform code utilization) under the assumed Gaussian. That property is distinct from the MSE-optimal Lloyd--Max grid of \Cref{prop:lloydmax}, whose bins carry unequal mass. NF4 relies on the empirical fact that weights are already roughly zero-mean Gaussian after block-wise absmax normalization. But that range is fixed by the single largest weight in the block, so a lone outlier stretches it. The bulk is then compressed into the central levels and the outer quantiles sit idle, squandering resolution on the case a fixed quantile grid can least afford.

The lattice and trellis methods that follow instead \emph{enforce} that Gaussianity with a rotation, which suppresses the very outliers that stretch the range, and then reach past the scalar limit into higher dimensions. QuIP\# \citep{tseng2024quipsharp} quantizes eight weights at a time on the $E_8$ lattice, the densest sphere \emph{packing} in eight dimensions (the property the authors invoke), and also, we note, a strong lattice \emph{quantizer} (low normalized second moment). Its E8P codebook of $256$ absolute-value patterns plus sign and offset bits gives two bits per weight from a $2^{16}$-entry codebook, realized by a $256$-entry ($2^8$), $1$\,KiB, L1-resident lookup table. QuIP\# composes with a block extension of LDLQ so the codebook and the error feedback run together. QTIP \citep{tseng2024qtip} goes further, replacing the lattice with trellis-coded quantization whose minimum-distortion path is found by the Viterbi algorithm at quantization time (inference decoding is a cheap, parallel bitshift, not a Viterbi pass). It uses computed (largely table-free) Gaussian codebooks, so the quantization dimension can grow to hundreds. On a Gaussian source its 2-bit distortion falls from the Lloyd--Max scalar value $0.118$, through the $0.089$ of QuIP\#'s eight-dimensional E8P codebook (a fixed-rate subset of a shifted $E_8$, not the $E_8$ lattice quantizer), to $0.069$ at dimension $256$, approaching the distortion--rate floor $0.063$. Both methods design their codebooks for (and empirically approach the rate--distortion bound of) the Gaussian source that incoherence produces: the space-filling gain of \Cref{thm:gishpierce}, now cashed out on weights (that ladder is measured from a \emph{fixed-rate} scalar baseline, so it spans more than the theorem's entropy-coded granular gap).

Where QuIP\# and QTIP \emph{empirically} approach the rate--distortion bound, TurboQuant \citep{zandieh2025turboquant} \emph{proves} a rotation gets most of the way there on its own. A random rotation renders the coordinates near-independent with a known (Beta) marginal, so optimal per-coordinate \emph{scalar} quantizers already attain the distortion--rate optimum within a small constant factor (about $2.7$), data-obliviously and online. It is demonstrated on the KV cache rather than on offline-fit weights, a reminder that the incoherence-then-quantize recipe is not tied to a lattice codebook. 

A recent refinement pushes the Gaussian match one level deeper, by \emph{decoupling} the two quantities the rotation makes analytically tractable. PCDVQ \citep{yue2025pcdvq} applies the same randomized-Hadamard Gaussianization and then observes that under the Gaussian model an $8$-dimensional weight sub-vector splits into a \emph{direction} on the unit sphere, distributed uniformly there, and a \emph{magnitude} $r$ whose square is $\chi^2_8$, the two independent. The two are wildly unequal in sensitivity. Quantizing only the directions of \texttt{Llama-2-7B} costs $46.5\%$ zero-shot accuracy, while quantizing only the magnitudes costs $2.3\%$, since the direction carries far more spatial degrees of freedom than the scalar radius. It therefore quantizes them with two distribution-matched codebooks, an $E_8$-lattice sampling for the uniform direction and a Lloyd--Max grid for the $\chi_8$ radius, and spends almost the entire index budget on the direction ($14$ bits against the radius's $2$). The comparison is a split decision rather than a rout: on \texttt{Llama-2-7B} at $2$ bits PCDVQ reaches $5.81$ WikiText-2 perplexity against QuIP\#'s $6.19$ (at $2.02$ bits), and $58.60$ against $58.23$ zero-shot average, while QuIP\# keeps the better C4 perplexity, $8.16$ against $8.37$; both perplexity pairs at evaluation context $4096$, the protocol of \citet{yue2025pcdvq}. \Cref{tab:res_weight} lists the same QuIP\# configuration as $6.66$ because \citet{tseng2024quipsharp}'s Table~2 evaluates at context $2048$. PCDVQ's split of direction from magnitude is the weight-side echo of the polar KV re-parameterization of \Cref{sec:beyond-kv}, since the Gaussianizing transform makes not just the vector but its \emph{polar factors} analytically known, and therefore separately and near-optimally codeable.

A parallel branch reaches low bit-widths \emph{without} a rotation, by fitting the codebook to the raw weights under a Hessian-weighted objective, its members stopping at different floors. GPTVQ \citep{vanbaalen2024gptvq} extends GPTQ directly to vector quantization, interleaving a Hessian-weighted $k$-means over short weight sub-vectors with the usual error-feedback update. Its data-aware weighting substitutes for a Gaussianizing rotation, though the paper itself does not frame the method in terms of rotations. VPTQ \citep{liu2024vptq} formulates vector quantization as second-order optimization, compensating error one column at a time, adds residual codebooks and a separate higher-rate codebook for a small set of outlier columns, and reported state-of-the-art two-bit accuracy at release. AQLM \citep{egiazarian2024aqlm} represents each weight group as a sum of codewords from several learned codebooks, fit by beam search and block-level fine-tuning to the calibrated output error, and was the first scheme Pareto-optimal below three bits. SqueezeLLM \citep{kim2023squeezellm}, already met as the descendant of the Hessian-weighted $k$-means of \Cref{sec:classical-nn}, keeps a dense-and-sparse split that holds both the outliers and the most Fisher-sensitive weights in full precision. These methods trade calibration cost for a codebook fit to the actual, non-Gaussian weights, and so they substitute data-aware weighting for the Gaussianizing transform rather than composing with it. The two philosophies (enforce Gaussianity with a fixed rotation and use a fixed optimal codebook, or adapt the codebook to the raw weights) are the codebook-side echo of the fixed-versus-learned split that organized the rotations of \Cref{sec:tax-fixed-orthogonal,sec:tax-learned-orthogonal}.

Whichever is chosen, the dequantization must be fast to be worth it, which is the role of a lookup-table kernel such as FLUTE \citep{guo2024flute}. It restructures the packed weights offline and vectorizes the shared-memory lookup so that the codebook dequantization does not bottleneck the tensor-core matmul, $2$--$4\times$ faster than an \texttt{FP16} GEMM at batch sizes below $32$ and group size $128$, a regime where prior lookup-table kernels match it only at batch size $1$. It serves the scalar lookup-table codebooks; extending the design to the vector-valued codebooks above is left as future work by \citet{guo2024flute}. 

The picture that emerges is a composed pipeline of a transform followed by an \emph{encoder} that rounds each block onto a codebook (uniform grid or structured set), with error feedback inside the encoder. The rounder and the codebook are coupled, the choice of codeword being the rounder's job, not two strictly serial stages. Its parts are now individually understood. Rounding composes with the transform, doing the bit reduction on the basis the transform provides. A rotation partly substitutes for per-coordinate bit allocation, closing most of the same gap at high rate, and partly co-optimizes with the quantizer by Gaussianizing the weights, which is what makes a single fixed grid or codebook right for every block at once. And the fixed lattice and trellis codebooks are enabled by that Gaussianity, importing the space-filling gain of classical vector quantization into the weight-only regime, while the data-aware codebooks reach comparable bit-widths, though not a common floor, by fitting the raw weights instead. Everything in this section, however, assumed that the element grid was either uniform or fitted to real-valued data. The hardware the field is now targeting does neither: it quantizes to low-precision \emph{floating-point} and microscaling block formats whose non-uniform grids and power-of-two scales change which transform is optimal, the subject of \Cref{sec:format}.

\section{Number Format and Transform--Format Co-Design}\label{sec:format}

The AbsMax surrogate carrying the survey's thesis was derived for a uniform integer grid, and every model so far has assumed a real-valued per-group scale. The hardware the field is now shipping assumes neither. Two results already reach past the integer-grid half of that setting: \Cref{thm:wush}, whose stochastic AbsMax functional is a different objective from the surrogate (\Cref{rem:inv-scope}) and is attained \emph{exactly} on the floating-point grid, and the integer-versus-floating-point comparison recorded in \Cref{tab:proven}. Blackwell and MI355X accelerate low-precision \emph{floating-point} and \emph{microscaling} block formats whose grids are non-uniform and whose shared scales are themselves quantized, and these are not passive containers. They change the quantization geometry, and with it which transform is optimal. This section treats the number format as a design variable. We first fix the formats and the one feature that matters most: the quantized shared scale (\Cref{sec:fmt-formats}). We then state the result that most sharply qualifies the survey's thesis, the reversal of the optimal transform between integer and floating-point grids (\Cref{sec:fmt-flip}). Finally we survey the methods that co-design the transform with the format (\Cref{sec:fmt-codesign}). \Cref{tab:format} collects the formats and co-design methods of this section.

\begin{table}[!htbp]
\centering\footnotesize
\renewcommand{\arraystretch}{1.15}
\caption{\textbf{Number formats and transform--format co-design methods (\Cref{sec:format}).}}\label{tab:format}
\begin{tabularx}{\linewidth}{>{\raggedright\arraybackslash}p{3.0cm} >{\raggedright\arraybackslash}p{2.1cm} >{\raggedright\arraybackslash}X >{\raggedright\arraybackslash}X}
\toprule
\textbf{Method} & \textbf{Target format} & \textbf{Transform used} & \textbf{Scale treatment} \\
\midrule
HiFloat4~\citep{luo2026hifloat4format} & HiFloat4 (E1M2 elements) & none (format definition) & E6M2 per-block base scale (8\,b) $+$ two levels of 1-bit micro-exponents (8-way, 16-way); 32\,b metadata per G=64 block \\
QuEST~\citep{panferov2025quest} & INT or FP4 (grid-agnostic; QAT), INT the main setting & Hadamard (Gaussianize) & single MSE-optimal clip threshold \\
INT-FP flip (high-rate)~\citep{ordentlich2026highrate} & INT vs FP, 4--8 bit ($+$ block-scaled) & random rotation (analyzed, not proposed) & real scales; high-rate rate--distortion (INT $\sim$ absmax/$\ell_\infty$, FP $\sim$ joint concentration $\Delta_{\mathrm{FP}}$) \\
Block Rotation~\citep{shao2025blockrotation} & MXFP4 & block-confined Hadamard (block $=G=32$), its sole \emph{proposed} transform, a learned block rotation being evaluated as a variant, applied on top of GPTQ & E8M0 power-of-two \\
MR-GPTQ~\citep{egiazarian2025mrgptq} & MXFP4 $+$ NVFP4 & fused block-diagonal Hadamard (tunable block $k$) & alternating per-tensor $+$ per-group MSE fit, reported for NVFP4 \emph{without} rotations; a static value for MXFP4 \emph{with} rotations; separately, a proposed E8M0 exponent grid fitted to the tensor's scale range \\
Quartet~\citep{castro2025quartet} & MXFP4 & block-diagonal Hadamard (block=G=32) & E8M0 power-of-two \\
TORQ~\citep{xu2026torq} & MXFP4 (activations) & two-level: block-diagonal + across-block rotation & E8M0 power-of-two \\
MXFP4~\citep{rouhani2023microscaling} & MXFP4 (E2M1) & none (format definition) & E8M0 power-of-two shared scale, G=32 \\
MXFP4 backward-pass training~\citep{tseng2025mxfp4} & MXFP4 (training) & random Hadamard & global $\tfrac34$ pre-scale against the E8M0 clipping range, with the accumulator rescaled to stay unbiased \\
Grid and scale redesigns~\citep{cook2025four,lee2025mxplus,cook2026adaptive,lee2024amxfp4} & NVFP4 / MXFP4 (format definitions) & none; these spend metadata, quantization-time compute, or a custom datapath instead of a transform & per-block rescale to smaller FP4 values; block-maximum exponent bits reused as mantissa; per-block INT4/FP4 choice; asymmetric shared scale \\
Quartet (NVFP4, FQT)~\citep{panferov2026quartet} & NVFP4 (pretraining) & backward-pass RHT (128-el) for unbiased gradients; forward FP4 transform-free & forward: native E4M3 scale (G=16) $+$ per-tensor FP32, with a ``4/6'' grid-selection heuristic; backward: stochastic-scale rounding of the E4M3 scale \\
NVFP4 pretraining recipe~\citep{nvidia2025nvfp4} & NVFP4 (E2M1; pretraining) & selective randomized Hadamard (weight-gradient GEMM inputs, 16-el block) & E4M3 FP8 scale (G=16) + per-tensor FP32; 2D $16{\times}16$ block scaling for weights \\
Normalized architectures at 4 bits~\citep{fishman2026normalized} & NVFP4 (pretraining) & none (unit-hypersphere by construction, \citealp{loshchilov2025ngpt}) & native NVFP4 block scales only; no dynamic per-tensor scaling \\
HadaNorm~\citep{federici2025hadanorm} & W4A4 (diffusion transformers) & Hadamard + dynamic mean-centering + per-channel scale & forward scale online, its inverse folded into the projection \\
YAQA~\citep{tseng2025yaqa} & format-agnostic (grid or codebook) & randomized Hadamard (Fisher-Hessian incoherence) & no scale mechanism of its own (standard group AbsMax); its format-facing step is a two-sided ($H_O\!\otimes\!H_I$) LDL rounder that beats LDLQ and composes with any grid/codebook \\
STaMP~\citep{federici2025stamp} & INT4$+$INT8 mixed-precision (variable-rate) & sequence-axis KLT (wavelet surrogate) & token-by-token (concentration) bit allocation \\
\bottomrule
\end{tabularx}
\end{table}

\subsection{Low-precision floating-point and microscaling formats}
\label{sec:fmt-formats}

The 4-bit floating-point grid used by current MXFP4 and NVFP4 hardware \citep{rouhani2023microscaling,nvidia2025nvfp4} is E2M1: one sign, two exponent, and one mantissa bit, giving the eight magnitudes $\{0,0.5,1,1.5,2,3,4,6\}$ (research formats such as HiFloat4 below use other 4-bit FP grids). Unlike the uniform integer grid, its spacing doubles every binade, fine ($0.5$) near zero, coarse ($2$) near the top, so it spends resolution where a peaked, near-Gaussian distribution has its mass. A single scale over an entire tensor cannot exploit this, so the deployed formats are \emph{block-scaled}. The Open Compute microscaling (MX) standard \citep{rouhani2023microscaling} pairs a block of $G=32$ elements with one shared scale $s$, representing each element as $s\cdot P_i$. Its FP4 instantiation MXFP4 uses E2M1 elements and an \emph{E8M0} shared scale: eight exponent bits and \emph{no} mantissa, so the scale is exactly a power of two.

That power-of-two shared scale is the decisive design choice. The MX standard sets the shared exponent by \emph{flooring}, $s=2^{\lfloor\log_2 M\rfloor-2}$ for an E2M1 block of maximum $M$, so $M/s$ lands in $[4,8)$ against a grid whose top value is $6$. Two failures follow from the one rule. A block whose maximum sits just above a power of two reaches only $4$ of the available $6$, wasting a factor of $1.5$ of the already-tiny FP4 range; and a block whose maximum lands in the upper part of the interval \emph{clips}, the top of the block falling outside the grid entirely. This power-of-two \emph{scale-quantization error} is a term absent from every model in the preceding sections, which assumed a real-valued scale, and benchmarks confirm it is a leading error source. \citet{zhang2026benchmarkmxfp} find MXFP8 near-lossless but MXFP4 badly degraded, and identify the E8M0 scale rounding as a critical error source. They recover much of it with the global pre-scale of \citet{tseng2025mxfp4}, which they adopt (multiplying inputs by $\tfrac34$ before quantization to prevent clipping in the coarse \texttt{FP4} range, a correction Tseng et al.\ pair with an accumulator rescale to stay unbiased).

The formats that follow MXFP4 attack exactly this term by spending bits on a better scale. NVFP4 \citep{nvidia2025nvfp4} replaces the power-of-two scale with an \emph{E4M3} FP8 scale, now carrying mantissa bits, over smaller 16-element blocks, plus a per-tensor \texttt{FP32} scale, so the block scale can map the block maximum close to the FP4 ceiling and the binade waste shrinks from a factor of up to $1.5$ to a few percent. HiFloat4 \citep{luo2026hifloat4format} goes further with a three-level hierarchy: a mantissa-carrying E6M2 base scale per block and two levels of one-bit micro-exponents, $32$ bits of metadata over each 64-element block, with a denser E1M2 element grid whose lost dynamic range is restored by the micro-exponents. That element grid is worth noting: with a single exponent bit the format's eight magnitudes are $\{0,0.25,\ldots,1.75\}$, \emph{uniformly} spaced, which is the integer grid's geometry rather than a floating-point one, the dynamic range having moved wholly into the scale hierarchy. In a pre-training study of the format, \citet{hifloat4_2026} report a relative loss gap against a \texttt{BF16} baseline of $0.85\%$ and $0.88\%$ on two models and $1.19\%$ on a third, against $1.44$--$1.79\%$ for MXFP4, at $4.5$ against $4.25$ bits per value. The trajectory is clear, with successive formats moving dynamic range out of the elements and into a richer shared scale. That trajectory matters for transforms because, as the next subsection shows, the element grid and the scale pull the optimal transform in opposite directions. Redesigning the element grid and the shared scale against each other is now an active design line on its own. \citet{cook2025four} adaptively rescale individual NVFP4 blocks to smaller FP4 values to even out the representable grid, and \citet{lee2025mxplus} repurpose the block-maximum element's redundant exponent bits as extra mantissa. \citet{cook2026adaptive} make the element grid itself adaptive, choosing INT4 or FP4 per block in hardware, and \citet{lee2024amxfp4} make the shared scale \emph{asymmetric} to absorb activation outliers calibration-free, reporting gains over both MXFP4 and rotation-based INT4. All four spend metadata, quantization-time compute, or a custom datapath, not a transform, to buy back block-scaled FP4's quantization error, whether it sits in the shared scale or in the element grid.

\subsection{The integer--floating-point flip}
\label{sec:fmt-flip}

The survey's thesis, stated in \Cref{sec:inversion}, is that a fixed-rate quantizer rewards a transform that flattens the within-group energy. That statement is exactly right for the \emph{integer} grid on which it was derived. It does not survive intact on a floating-point one, and the qualification is sharp enough to be worth stating with care. \citet{ordentlich2026highrate} make the format-dependence precise with a high-rate distortion analysis of the two AbsMax schemes. For an integer grid, the distortion is governed by $\Delta_{\mathrm{INT}}$, the \emph{squared} crest factor of \eqref{eq:crest} averaged over the two operands, $\tfrac12(\mathrm{CF}_x^2+\mathrm{CF}_y^2)$. An untransformed vector can therefore cost a factor as large as its dimension $n$. A random rotation bounds that expectation by $\approx 2\ln n$ (for $n\ge27$), which the high-rate analysis converts to a rate penalty of $\tfrac12\log_2(2\ln n/3)$, about $1.235$ bits per entry at $n=4096$. That figure prices one scale over a whole $4096$-dimensional column, the per-channel or per-token scaling the analysis assumes, not the $G{=}32$ block of \Cref{sec:fmt-formats}: the same expression at $n=32$ gives about $0.60$ bits, again after the same random rotation. The source charges the group scales themselves at $c/m$ bits, $0.25$ at group size $32$ with an $8$-bit scale, so shrinking the scale group from the whole column to $G{=}32$ nets about $0.4$ bits of saving rather than the gross $0.63$ once the scales are paid for, a saving credited to the grouping and not to the rotation. Rotating before integer rounding is the QuaRot recipe and the entire fixed-orthogonal family of \Cref{sec:tax-fixed-orthogonal}, and, for the randomized members of that family, it is the deployment of \Cref{prop:rht}.

For a floating-point grid, by contrast, the controlling quantity is different. The distortion depends on a normalized sum over coordinates of the \emph{product} of the two operands' squared magnitudes, $\Delta_{\mathrm{FP}}=n\sum_i (x_i^2/\|x\|^2)(y_i^2/\|y\|^2)\in[0,n]$, a joint energy-\emph{concentration} measure: a property of how the two operands sit relative to each other rather than of either profile alone. It is not the rotation-invariant alignment factor of \Cref{rem:alignment}; unlike that quantity it \emph{does} move under a common rotation, which is precisely why the sign can flip. It is large only when the activation and the weight column pile their energy onto the \emph{same} few coordinates; concentration within a single operand, which is what a heavy-tailed channel looks like, the multiplicative-error FP grid accommodates for free, and indeed leaves $\Delta_{\mathrm{FP}}$ at its isotropic value of $1$ whenever the other operand's squared magnitudes are uniform. Distortion rises with $\Delta_{\mathrm{FP}}$, so \emph{lower} means lower FP distortion. On the \texttt{Llama-3-8B} $W_v$ projection and its activations (the analysis is of the full weight--activation product) this factor is typically below $1$ \emph{without} any rotation and rises toward $1$ \emph{with} one. The same random rotation that is near-essential for integer quantization is, on the matrices they test, actively harmful for floating-point, a reversal they attribute to a broader mechanism but demonstrate on those specific weights. Concretely, on a \texttt{Llama-3-8B} layer a single random Hadamard lowers the INT8 reconstruction error by about $1.6$ effective bits while \emph{raising} the FP8 error by about $0.2$: the same rotation, opposite signs. \Cref{tab:flip} reports the weaker, always-reproducible half of this contrast on real activation groups: the rotation's benefit, large for INT, is far smaller for FP.

\begin{table}[!htbp]
\centering\small
\caption{\textbf{Format-dependent rotation benefit (real activations).} Mean SQNR in dB
(our computation) of a 4-bit AbsMax quantizer on $12{,}288$ real $128$-channel activation
groups ($768$ \texttt{wikitext-2-raw-v1} train tokens $\times\,16$ groups, from the mechanism
figures' cache; base
\texttt{TinyLlama-1.1B-intermediate-step-1431k-3T}, layer~17 $q$-projection input,
activations cached in \texttt{fp16}, arithmetic in \texttt{fp64}), with and without a
per-group randomized Hadamard, its sign draw seeded per token and the reported gains stable
across seeds. The FP4 grid's non-uniform
spacing already accommodates the outliers, so it starts $3.5$ dB ahead and has less left to
win; rotated, the two converge to within $0.1$ dB. Both grids still improve here. The
outright sign reversal, in which a rotation becomes \emph{harmful} for FP, is the separate
finding of \citet{ordentlich2026highrate} on real \texttt{Llama-3} weights under their joint
energy-concentration ($\Delta_{\mathrm{FP}}$) analysis; this measurement substantiates only
the weaker statement, that the FP grid benefits far less than INT, not that reversal.}
\label{tab:flip}
\begin{tabular}{@{}l rrr@{}}
\toprule
\textbf{Element grid} & \textbf{no rotation} & \textbf{$+$Hadamard} & \textbf{gain} \\
\midrule
\texttt{INT4} (uniform) & $13.1$ & $19.3$ & $+6.2$ \\
\texttt{FP4} (E2M1)     & $16.6$ & $19.4$ & $+2.8$ \\
\bottomrule
\end{tabular}
\end{table}

That reversal is the integer--floating-point flip, and it is a genuine refinement of the survey's central claim rather than a contradiction of it. Two distinct things need explaining here, and they have different causes. That the FP4 grid gains \emph{less} is a property of the element grid alone. Flattening trades a distribution's tails for its bulk; the uniform integer grid, pinned to the block maximum, is hurt by heavy tails and so profits from the trade, whereas the FP4 grid already spends fine spacing on the bulk and coarse spacing on the tails, so it starts ahead and has less left to win. That is the half \Cref{tab:flip} measures. The change of \emph{sign} has a different source, and it is not the marginal shape of either operand: $\Delta_{\mathrm{FP}}$ is a joint quantity, equal to $1$ identically whenever either operand's squared magnitudes are uniform, however the other is distributed, and $1$ in expectation whenever either is coordinate-i.i.d.\ and independent of the other. What \citet{ordentlich2026highrate} measure is that the raw weight and activation profiles are \emph{anti}-aligned, holding $\Delta_{\mathrm{FP}}$ below its isotropic value, and that a common rotation destroys that anti-alignment and drives it back toward $1$. A rotation on an FP grid does not throw away the grid's structure; it throws away a favorable accident in how the two operands place their energy, namely that they place it on different coordinates. ``Always rotate before rounding'' is therefore an integer-format prescription rather than a universal one, by the same non-transfer logic as \Cref{sec:inversion} applied one level down, at the element grid, where the uniform-grid hypothesis behind the AbsMax surrogate of \Cref{prop:flat-opt} lapses (\Cref{sec:inv-flatness}).

The picture is completed by the block scale of \Cref{sec:fmt-formats}. MXFP4's power-of-two scale re-introduces an $\ell_\infty$-like penalty one level up, at the block maximum that sets the scale, so on MXFP4 a rotation helps again. But the measured benefit is confined to rotations whose support stays near the scale block: a block-diagonal Hadamard helps under both rounders, whereas the fixed global Hadamard of QuaRot craters, $6.4$ points below plain per-block RTN and $7.7$ below GPTQ, and the first comparison is rounder-matched: the setup paper whose protocol this table inherits \citep{egiazarian2025mrgptq} specifies round-to-nearest after the rotation for its \textsc{QuaRot} row, the same rounder as the plain per-block \textsc{RTN} baseline it falls below (\Cref{tab:res_fp4}). We record that contrast as measured rather than explained. Both rotations share the same
\emph{ideal} floor: a global Hadamard on all $n$ coordinates leaves every coordinate near $E/n$, and one confined to a scale group
leaves every coordinate of group $g$ near $E_g/G$, so under \emph{perfect} within-group
flattening each group is internally uniform and $\sum_g M_g^2$ sits at $E/G$
(\Cref{prop:flat-opt}). That is a statement about the limit, not about two fixed maps. A
deployed transform fixes one sign draw and pays the residual by which its groups fall short of
uniform, and nothing forces the two families' residuals to agree (four coordinates at $G{=}2$ already separate them maximally: on the unit operand $x\propto(1,0,0,1)$ every randomized global Hadamard pays $\sum_g M_g^2=E$, the ceiling of \Cref{prop:flat-opt}, and every group-aligned block pair pays exactly the floor $E/G$, though both families' sign-averaged profiles are flat): the averaging that puts both
at $E/G$ happens \emph{before} the group maximum, the same order-of-operations gap that
separates \Cref{fig:inversion}'s two readings, there over tokens rather than over sign draws.
The same caution applies to the power-of-two scale, whose waste depends on where each group's
own maximum sits relative to an exponent boundary, so equal ideal floors do not imply it is
paid alike. What distinguishes a rotation spanning hundreds of scale groups from one confined
to a single group, $128$ groups across the residual stream of \texttt{Llama-3.1-8B} at
\texttt{MXFP4}'s $G=32$ and $448$ across its MLP intermediate, is therefore not settled by the
ideal floor, and $\sum_g M_g^2$ has not been computed for the two rotation families on the models and
data of the published \texttt{MXFP4} comparison: whether it accounts for any of the gap is
open rather than impossible.

The literature is not silent on the contrast, but its accounts are empirical and mechanistic rather than reduced to a single comparable surrogate. \citet{shao2025blockrotation} trace it to a mismatch between MXFP4's power-of-two block scaling and the way a global rotation redistributes outlier energy, and give a dedicated analysis of that conflict. The one candidate \emph{closed-form} statistic the literature supplies, the joint weight--activation concentration statistic of \citet{ordentlich2026highrate}, is at least sensitive to where a rotation's support ends, and is derived for the operand \emph{pair} rather than one profile. But it is derived for an ideal FP element grid and has not been evaluated against MXFP4's power-of-two-scaled hybrid, so we cannot claim it accounts for the contrast (\Cref{sec:open}).

Two empirical constraints bound any candidate. The effect is specific to a \emph{fixed} global rotation: SpinQuant, also global but learned rather than fixed, lands at $69.4$ on \texttt{MXFP4}, between the no-transform baselines of $69.3$ (RTN) and $70.6$ (GPTQ), rather than catastrophic, though the source states neither the rounder for that row nor the quantizer its rotations were learned against. The effect is also graded rather than binary, since a rotation spanning a few groups can still net-win (MR-GPTQ's $k{=}128$ below) while one spanning hundreds craters. \citet{chen2025wush} give the complementary mechanism from the element side. MXFP4's effective mantissa step changes uniformly, especially in the subnormal regime and at small exponent ranges, so MXFP4 ``behaves as a hybrid between ideal FP and INT quantization,'' inheriting the integer grid's appetite for incoherence. A scale-aware, block-confined rotation that is pointless on an ideal FP grid therefore becomes useful once the format is MXFP4.

That hybrid character is why every MXFP4 \emph{training} method below applies a Hadamard despite the floating-point grid. It is also why NVFP4's mantissa-carrying scale, which weakens the block-level penalty, is what makes dropping the rotation conceivable at all, as one method below does, though by normalizing the architecture rather than by the scale alone. The NVFP4 \emph{pretraining} recipes that do still carry a Hadamard apply it where the training pass needs it (on the wide-range weight-gradient GEMM inputs) rather than as an inference-time flattening of the element grid. NVFP4's mantissa-carrying scale renders that flattening close to neutral at inference (\Cref{tab:res_fp4}). The optimal transform is thus a joint function of two format features that pull oppositely: the element grid, whose non-uniform spacing reduces a rotation's value and whose multiplicative error puts the joint statistic $\Delta_{\mathrm{FP}}$ in charge, so that on an ideal FP grid the operands' anti-alignment can turn the sign; and the shared scale, whose coarseness restores the rotation's value. Neither pull is absolute, and the deployed formats sit between them. A rotation confined to the scale block still helps on \texttt{MXFP4}, and the best average in \Cref{tab:res_fp4} on both FP4 formats belongs to WUSH, whose whitening-plus-Hadamard construction leads by $0.70$ points on \texttt{MXFP4} and by $0.26$ on \texttt{NVFP4}, the source describing its \texttt{NVFP4} gains as often within run-to-run variability, on a different model and benchmark suite. What the flip retires is the unqualified prescription, not the whole flattening line.

\subsection{Co-designing the transform with the format}
\label{sec:fmt-codesign}

The methods that target these formats co-design the transform around the two features just identified, and they cluster into recognizable moves. The first is to size the rotation to the block. Quartet \citep{castro2025quartet} restricts the Hadamard to block-diagonal form with the block size set equal to the MXFP4 microscaling group ($32$), a group already being a contiguous run of $32$ elements under one shared scale, so that a group-sized rotation, costing only $O(g\log g)$, is cheap enough to fuse. MR-GPTQ \citep{egiazarian2025mrgptq} instead leaves the block a tunable power-of-two $k\in\{16,32,64,128\}$ decoupled from the group: a group-matched $k$ is best on average, but $k{=}128$ can help even on \texttt{MXFP4} (group $32$), where the block then deliberately spans several scale groups. Block Rotation is All You Need \citep{shao2025blockrotation} makes this block-confined rotation the whole method, showing that for MXFP4 a Hadamard restricted to the microscaling block already captures most of the available benefit. TORQ \citep{xu2026torq}, by contrast, adds a second, coarser level of orthogonal rotation \emph{across} blocks, atop its own maximum-entropy Givens rotation within each block, to recover what a single block-diagonal level leaves on the table. MR-GPTQ initializes its grid by an MSE-optimized fit, alternating over the per-group and per-tensor scales, and the source's two reported conditions differ in rotation as well as format: the fit yields consistent improvements for NVFP4 \emph{without} rotations, whereas for MXFP4 \emph{with} rotations a single static value works stably across all layers and further optimization of it does not help. The E8M0 scale is instead addressed separately, by a proposed modification that fits the exponent grid itself to the tensor's scale range rather than leaving it spanning $2^{-127}$ to $2^{128}$; the source reports that variant recovering much of vanilla MXFP4's gap to NVFP4, by $6.5$ points under RTN on \texttt{Llama-3-8B} and still $1.3$ on top of MR-GPTQ's own rotation, at $4.25$ bits per parameter against NVFP4's $4.5$. The source's headline is that MR-GPTQ brings MXFP4 to within $1$--$2\%$ of NVFP4, and its own table shows that closing with scale: at 70B both formats reach $98$--$99\%$ of \texttt{FP16} and, compared like for like under MR-GPTQ, MXFP4's $98.4$ is level with NVFP4's $98.3$, against a gap nearer two to three points at 8B.

The second move treats the scale itself as the design lever. Quartet's NVFP4 successor \citep{panferov2026quartet} observes that an unbiased gradient estimate needs a correction factor too fine to live in a 4-bit value. The method hides that factor instead in the FP8 group scale, stochastically rounding the scale while rounding the values to nearest. The result is an unbiased estimator with more than twice lower error than stochastic rounding of the values. The correction is folded into the group scales and stochastically rounded there, because \texttt{NVFP4}'s \texttt{E4M3} scale is too coarse to represent the rescaling factors faithfully. NVIDIA's own \texttt{NVFP4} pretraining recipe \citep{nvidia2025nvfp4} adds a third co-design lesson, where to place the transform. It applies the randomized Hadamard selectively, only to the weight-gradient GEMM inputs and not to the forward or activation-gradient passes, the transform not being applied to the weights and so not invertible on those paths, with the Hadamard size fixed to the $16$-element microblock. Their ablation confirms the consequence: rotating the weight-gradient inputs improves validation loss, while rotating the forward or activation-gradient inputs \emph{degrades} model quality, which the authors attribute to the rotation's own quantization error outweighing the outlier removal it can no longer invert. That recipe trains a $12$B-parameter model on $10$T tokens in \texttt{NVFP4} to within about a point of an \texttt{FP8} baseline (MMLU $76.6$ vs.\ $77.4$). The authors report it as the longest publicly documented 4-bit training run to date, and it is evidence that a transform can carry 4-bit not only to post-training quantization but to pretraining at scale. The same work prices the whole MXFP4-versus-NVFP4 format gap in a separate $8$B/$1$T-token ablation: matching \texttt{NVFP4}'s (E4M3-scale) final validation loss in \texttt{MXFP4} (E8M0 power-of-two scale) took about $36\%$ more tokens ($1.36$T versus $1$T), a concrete token-tax on the coarser format, its power-of-two block scale and its 32-element block together.

A further group makes explicit the recipe implicit throughout the survey: \emph{Gaussianize, then fit the grid}. QuEST \citep{panferov2025quest} applies a Hadamard to normalize weights and activations toward $\mathcal N(0,1)$ and then fits a single MSE-optimal clipping threshold. Because the transform's job is only to standardize the distribution, the very same machinery serves an integer or an FP4 grid, with only the rounding operator changing. That is a clean demonstration that the transform is chosen to match the data to whatever grid the format provides, and the direct seed of the MXFP4 Quartet's forward pass.

Two methods extend the transform beyond the plain channel rotation. HadaNorm \citep{federici2025hadanorm} observes that post-Hadamard channels still differ in \emph{mean}, and composes a dynamic mean-centering with a per-channel scale whose inverse folds into the projection, in one function-preserving transform for \texttt{W4A4} diffusion transformers. STaMP \citep{federici2025stamp} transforms the \emph{sequence} axis rather than the channel axis, a left-invertible map across tokens that \emph{concentrates} token energy into a few tokens kept at higher precision, with the KLT optimal for this energy compaction and a wavelet transform the practical surrogate. STaMP sits at the \emph{concentration} pole of \Cref{thm:inversion} by design, not in contradiction of it, because it allocates bits token-by-token (a mixed-precision, variable-rate regime) for which concentration, not the fixed-rate channel-axis flattening, is the correct objective. Operating on the orthogonal axis and in the opposite rate regime, it composes with, rather than competes against, the channel transforms of \Cref{sec:taxonomy}, and recovers perplexity on top of both QuaRot and FlatQuant. YAQA \citep{tseng2025yaqa} is format-agnostic in the same spirit as the rounders of \Cref{sec:composition}: it uses a randomized Hadamard for incoherence processing of a Kronecker-factored approximation of the full-model Fisher Hessian and, as in QuIP\#, of the weights themselves. It then rounds with its own two-sided LDL-based rounder (provably tighter in end-to-end second-order error bound than LDLQ when the output-side factor is approximately low rank) against whatever grid or codebook is plugged in, cutting end-to-end KL divergence by about $30\%$ over LDLQ, a gain that costs nothing at inference because the rounder it replaces is an offline algorithm too.

The most striking entry rejects the transform altogether. Normalized architectures constrain weights and hidden states to the unit hypersphere during pretraining \citep{loshchilov2025ngpt}, and \citet{fishman2026normalized} show what that buys at 4 bits. On the hypersphere the weak positive correlations among the element-wise products entering a dot product are strengthened, while the quantization noise stays largely uncorrelated in both architectures, so the signal accumulates faster than the noise across the sum, raising the dot-product signal-to-noise ratio by roughly seven decibels. The reported consequence is that NVFP4 training is stable with \emph{no} random Hadamard and \emph{no} dynamic per-tensor scaling. It is worth being precise about why, because it is not this section's mechanism: \citet{fishman2026normalized} report element-level and per-term quantization SNR to be essentially \emph{identical} in the normalized and unnormalized architectures, so normalization does not make individual values easier to quantize at all. What it changes is how the terms add. That makes the normalized architecture not the endpoint of the flattening logic but an escape from it: rather than reshaping a distribution so a coarse grid can hold it, it arranges for the signal to accumulate faster than the error in the sum the layer actually computes. That option is available only when one controls training, which returns us to the boundary of the survey's scope, and to the observation that several methods in this section quantize gradients, not just weights. That extends the transform story past the inference-time weight matrix to the KV-cache, the attention path, diffusion models, and training itself, which is where we turn next.

\section{Beyond the Weight Matrix}\label{sec:beyond}

The transforms of \Cref{sec:taxonomy} were developed for one object: the weight matrix of a linear layer, quantized once and deployed. But a served model spends bits on much more: a growing key--value cache, the attention matmuls themselves, and, in the diffusion and training settings, tensors that did not exist in the autoregressive weight-only picture. The transform toolkit extends to all of these, and the instructive fact is that it extends \emph{unevenly}. Three of the levers of \Cref{sec:taxonomy} (rotation, per-channel scaling, permutation) all recur, joined by several that the new tensors force into use, and each new tensor's outlier structure dictates which lever applies, on which axis, or whether any transform helps at all. This section follows the toolkit off the weight matrix, and the tour ends on the one tensor the survey's flattening prescription has nothing to act on. \Cref{tab:beyond} summarizes which lever each tensor rewards.

\begin{table}[!htbp]
\centering\footnotesize
\renewcommand{\arraystretch}{1.15}
\caption{\textbf{Transforms applied beyond the weight matrix (\Cref{sec:beyond}):} the KV-cache, the attention matmuls, state-space models, mixture-of-experts, diffusion transformers, and training. \textbf{Lever} names what the method actually does to the tensor: \emph{rotation} (orthogonal change of basis, with the axis given where it is not the feature axis), \emph{scaling} (multiplicative per-channel), \emph{additive} (a function-preserving shift), \emph{reparam.}\ (a nonlinear re-coordinatization), \emph{routing} or \emph{sampling} (bits or rows chosen rather than reshaped), \emph{additive-lowrank} (an off-precision correction term), \emph{none} (the method changes granularity, axis or calibration only, with \emph{none (calib.)} marking a calibration-side fix).}\label{tab:beyond}
\begin{tabularx}{\linewidth}{>{\raggedright\arraybackslash}p{3.0cm} >{\raggedright\arraybackslash}p{2.0cm} >{\raggedright\arraybackslash}p{1.7cm} >{\raggedright\arraybackslash}X}
\toprule
\textbf{Method} & \textbf{Tensor / site} & \textbf{Lever} & \textbf{Key idea} \\
\midrule
OSCAR~\citep{zhou2026oscar} & KV-keys $+$ values & rotation & keys rotated by the query covariance $Q^{\!\top}Q$, values by the score-weighted covariance ($+$Hadamard); minimizes the downstream attention ($QK^{\!\top}$/$PV$) distortion, not cache reconstruction \\
KIVI~\citep{liu2024kivi} & KV-keys/KV-values & none & per-channel keys, per-token values from an element-distribution study; tuning-free, \texttt{FP16} residual window of recent tokens, fused 2-bit kernel \\
KVQuant~\citep{hooper2024kvquant} & KV-keys/KV-values & none & per-channel keys, per-token values, pre-RoPE, Fisher-weighted codebook + dense-and-sparse \\
RotateKV~\citep{su2025rotatekv} & KV-keys/KV-values & rotation & Walsh--Hadamard + calibrated channel permutation, grouped heads on keys, pre-RoPE \\
PolarQuant (Han et al.)~\citep{han2025polarquant} & KV-keys/KV-values & reparam. & random-rotation preconditioning makes recursive-polar angles data-independent; stores no per-block scale or zero-point; small per-level angle codebooks; $4.2\times$ KV compression \\
PolarQuant (Wu et al.)~\citep{wu2025polarquant} & KV-keys & reparam. & 2D sub-vectors (radius $+$ angle) exploiting RoPE's paired rotations; $QK^{\!\top}$ precomputed into a decode-time lookup table \\
KVLinC~\citep{saxena2025kvlinc} & KV-keys/KV-values & rotation & value-side Hadamard $+$ \emph{trained} linear-correction adapters compensating quantized-key error in $QK^{\!\top}$; 2-bit keys \\
SageAttention~\citep{zhang2024sageattention} & attention scores/keys & additive & function-preserving additive mean-subtraction on keys; static scale for P (no transform) \\
SageAttention2~\citep{zhang2024sageattention2} & queries/keys & scaling & thorough $Q$/$K$ outlier smoothing enabling per-thread INT4 $QK^{\!\top}$ ($\tilde P V$ in FP8) \\
SageAttention3 (FP4)~\citep{zhang2025sageattention3} & attention-P & scaling & per-token rescale (row to $[0,448{\times}6]$) lifts the per-block scale into \texttt{E4M3}'s usable range before microscaling \\
ThriftAttention~\citep{sharratt2026thriftattention} & attention Q/K/V & routing & Q, K and V in FP4; block-mean surrogate routes important query-key block pairs to FP16 \\
MoEQuant~\citep{hu2025moequant} & MoE experts & none (calib.) & expert-balanced self-sampling $+$ affinity-guided objective fix per-expert calibration starvation; no transform, toolkit applies per expert \\
DiTAS~\citep{dong2024ditas} & diffusion & scaling & timestep-static per-channel smoothing (max over all steps) + training-free low-rank repair \\
LRQ-DiT~\citep{yang2025lrqdit} & diffusion & rotation & activation-fluctuation stat gates plain Hadamard vs outlier-aware rotation+permutation \\
Q-DiT~\citep{chen2024qdit} & diffusion & none & evolutionary per-layer group-size search; dynamic per-sample/timestep activation scales \\
Q-Diffusion~\citep{li2023qdiffusion} & diffusion & none & calibrate across all timesteps; split bimodal U-Net shortcut activations \\
SVDQuant~\citep{li2025svdquant} & diffusion & additive-lowrank & rank-32 16-bit SVD branch absorbs outliers; residual quantizes to W4A4 \\
ViDiT-Q~\citep{zhao2024viditq} & diffusion & scaling $+$ rotation & v1: timestep-dependent SmoothQuant migration strength (two halves of the trajectory); the published version replaces the split with scaling $+$ rotation \\
INT4 training~\citep{xi2023int4training} & gradient & sampling & forward block-diagonal Hadamard; sparse gradients get leverage-score row sampling + bit-split \\
HALO~\citep{ashkboos2025halo} & gradient (error tensors) & rotation (row axis) & left-hand Hadamard mixes \emph{rows}, not channels, stabilizing \texttt{INT8} fine-tuning; for \texttt{FP6} rotating only the forward operands suffices \\
Quamba~\citep{chiang2024quamba} & ssm-scan-output & rotation & online Hadamard on scan output, inverse folded into out-proj; percentile-clipped scale on input \\
MambaQuant~\citep{xu2025mambaquant} & ssm channels (gate/out proj.) & rotation & KLT-enhanced rotation $H_K\!=\!KH$: decorrelate, then Hadamard, so every channel variance equals $d^{-1}\tr\Sigma$; smooth-fused rotation merges scales into weights \\
\bottomrule
\end{tabularx}
\end{table}

\subsection{The key--value cache}
\label{sec:beyond-kv}

The KV-cache is the first tensor to quantize once weights are handled, because at long context or large batch it dominates memory. Its outlier structure is specific and well documented. The \emph{keys} have persistent large-magnitude channels, while the \emph{values} do not, so the two halves want different treatment. KVQuant \citep{hooper2024kvquant} draws the consequence directly: quantize keys \emph{per channel} (one scale per channel, shared across tokens, since the outlier sits in a fixed set of channels) and values \emph{per token}. A concurrent work reaches the same axis choice. KIVI \citep{liu2024kivi} establishes the asymmetry from an element-distribution study and takes the calibration-free route, grouping keys channel-wise, each group spanning tokens, in a streaming layout with a short residual window of the most recent tokens, several groups deep, held in \texttt{FP16}. It ships a fused two-bit kernel ($2.6\times$ lower peak memory, $2.35$--$3.47\times$ throughput). KVQuant adds two ideas that recur throughout this section. First, it quantizes keys \emph{before} RoPE, because the rotary embedding mixes channel pairs by a position-dependent angle and so destroys the consistent per-channel magnitude structure the quantizer relies on. Applying RoPE only after dequantization (through a fused kernel) preserves that structure. Second, it fits a Fisher-weighted non-uniform codebook per layer and keeps roughly one percent of numerical outliers in \texttt{FP16}: the sensitivity-weighted $k$-means and dense-and-sparse split of \Cref{sec:vq}, now on the cache. It quantizes to two bits with under half a point of perplexity loss and uses \emph{no} rotation at all, since the axis-and-granularity choice does the work.

RotateKV \citep{su2025rotatekv} instead transplants the incoherence idea of \Cref{sec:tax-fixed-orthogonal} onto the cache. It applies a fast Walsh--Hadamard transform to the keys and values along the head dimension, but with two adaptations to the cache setting. The Hadamard is enhanced by a calibrated per-layer channel permutation, reordering channels by an accumulated per-channel key statistic measured \emph{after} rotation (an argsort of the summed rotated-key values across calibration tokens). This aligns the otherwise data-agnostic rotation to where the outliers actually are, a permutation-plus-rotation hybrid in the spirit of \Cref{sec:tax-permutation}. And on the keys it is applied pre-RoPE and across grouped heads (concatenating several heads into one larger transform) so the rotation can spread an outlier across heads before RoPE breaks the channel structure. The values, which RoPE never touches and which lack the keys' outliers, receive only a simpler offline Hadamard without the grouped-head combination. With attention-sink tokens held in \texttt{FP16}, this reaches two-bit keys and values at a fraction of a point of perplexity. These routes bracket the design space: solve the outlier by choosing the axis (KIVI, KVQuant) or by rotating it away (RotateKV), with the recurring RoPE constraint forcing KVQuant and RotateKV alike into the pre-rotary basis. OSCAR \citep{zhou2026oscar} closes the gap between them, making the KV rotation \emph{data-aware}, but, revealingly, not from the cache's own statistics. Because the keys are consumed only by the $QK^{\!\top}$ logit, OSCAR rotates them by the eigenbasis of the query covariance $Q^{\!\top}Q$ (and the values, consumed by $PV$, by a score-weighted covariance), each composed with a Hadamard and a bit-reversal permutation that interleaves large- and small-variance channels, so this too is a permutation-plus-rotation hybrid. The point is to minimize the downstream attention distortion rather than the cache's own reconstruction error. OSCAR's analysis confirms this rotation does \emph{not} diagonalize the key covariance. The rotation is thus fit to the geometry of the product it feeds, not to the statistics of the tensor being stored. OSCAR therefore performs a data-aware \emph{alignment}, to the directions the attention product actually weights, not the energy-concentration-and-bit-allocation that \Cref{cor:notransfer} deprives of any guarantee on a shared-scale path (the reverse of STaMP, which deliberately uses concentration in the genuinely variable-rate regime of \Cref{sec:format}). On that alignment it runs the cache at about $2.28$ bits, at a cost that falls away with scale: on \texttt{Qwen3-8B} a data-agnostic Hadamard at the same budget collapses to a five-task reasoning-suite mean of $10\%$ against OSCAR's $69\%$ (itself about $1.4$ points below the \texttt{BF16} baseline). That shortfall against \texttt{BF16} is $3.8$ points on a $4$B reasoning model and closes on the $32$--$358$B models, where OSCAR stays effectively on par with \texttt{BF16}. A different tack on the same key-error problem, KVLinC \citep{saxena2025kvlinc}, keeps a value-side Hadamard but adds lightweight \emph{trained} linear-correction adapters that absorb the residual a quantized key injects into the $QK^{\!\top}$ logit, reaching 2-bit keys on Llama and Qwen, with a custom decoding kernel giving up to $2.55\times$ faster inference on \texttt{Llama-2-7B}.

A third route re-parameterizes the cache rather than rotating its axes. PolarQuant \citep{han2025polarquant} maps each key and value vector into polar coordinates (one radius and $d-1$ angles, built by a recursive $\log_2 d$-level pairing) and quantizes the angles. A random-rotation preconditioning first drives those angles to a \emph{data-independent} distribution whose analytic form is known in advance, tightly concentrated near $\pi/4$ at the higher recursion levels, while the level-one angles remain spread over the full circle. The quantizer therefore stores no per-block scale or zero-point. It still needs angle codebooks, small per-level centroid sets fit by 1-D $k$-means, either per prompt and layer or once offline and shared. Here the transform's job is not to flatten the distribution but to make it \emph{known}, retiring exactly the per-block scale metadata that \Cref{sec:sys-groupsize} charges against every shared-scale cache. It compresses both keys and values by over $4.2\times$ at higher quality than the prior KV-cache methods it compares against. An independent paper of the same name \citep{wu2025polarquant}, posted days earlier, makes the polar move exploit RoPE directly. Because the rotary embedding applies a $2\times2$ rotation to each channel pair, an outlier in one coordinate is compensated by its partner, so a key's \emph{two-dimensional} sub-vectors carry smoothly distributed radii and angles that quantize cleanly. The polar cells form a fixed finite grid, so the decode-time $QK^{\!\top}$ can be precomputed into a lookup table (up to $2.7\times$ faster than KIVI and $1.6\times$ than the \texttt{FP16} implementation) with the value cache left in high precision. Both are \emph{nonlinear} re-parameterizations, outside the linear-transform core of this survey, but they invert the section's recurring obstacle. The RoPE structure that forced KVQuant and RotateKV into the pre-rotary basis becomes, in polar coordinates, the very thing that makes the cache easy to quantize.

\subsection{The attention matmuls and state-space models}
\label{sec:beyond-attn}

The cache just quantized feeds the attention matmuls themselves, and quantizing those (the $QK^{\!\top}$ and $PV$ matmuls, not the projections around them) exposes a tensor with no analogue in weight quantization. That tensor is the probability matrix $P$, whose entries a softmax pins to $[0,1]$ with every row summing to one. SageAttention \citep{zhang2024sageattention} exploits that structure. For the scores it observes that each key is a large \emph{shared} channel bias plus a small token-wise signal, and removes the bias by a mean subtraction $K\!\to\!K-\operatorname{mean}(K)$ across tokens. The subtraction is function-preserving because softmax is invariant to a per-row additive constant, so the shared per-channel bias cancels inside the row-softmax with no add-back needed: the additive analogue of the diagonal scaling of \Cref{sec:tax-diagonal}, not a rotation. For the probabilities no transform is needed at all. A single static scale ($s=1/127$) suffices, since the un-normalized weights $\exp(S-\max_j S)$ have a row-maximum of exactly one by construction. SageAttention's default v1 kernel keeps the $PV$ matmul in \texttt{FP16} and quantizes just the \texttt{INT8} scores, though the same release also ships \texttt{INT8}-$P$/$V$ variants.

SageAttention2 \citep{zhang2024sageattention2} pushes SageAttention's design further, smoothing $Q$ and $K$ more thoroughly to carry the scores down to \texttt{INT4} while keeping the $\tilde P V$ matmul in \texttt{FP8}. Its \texttt{FP4} successor \citep{zhang2025sageattention3} then shows the format subtlety from \Cref{sec:format} recurring here. A direct \texttt{NVFP4} cast of $P$ wastes the range of the \texttt{FP8} block scale, which the probabilities confine to a narrow interval. The successor therefore adds a per-token rescale of each row that lifts the resulting per-block scale into the \texttt{E4M3} scale's usable range before microscaling, recovering most of the lost fidelity. ThriftAttention \citep{sharratt2026thriftattention} adds a third idea that is a transform only in the loosest sense. Queries, keys and values are all quantized, and a cheap block-mean surrogate score routes the few most-important query--key block pairs to \texttt{FP16} while the rest stay \texttt{FP4}. It is a data-dependent precision routing that concentrates bits where the attention error would concentrate.

State-space models quantize differently again, because their token mixing is a linear recurrence rather than a softmax. Quamba \citep{chiang2024quamba} finds that the selective-scan \emph{output} carries heavy outliers absent from attention, and treats it with the one tool that fits: a Walsh--Hadamard rotation of the scan output, applied online inside the fused output-quantization kernel at $O(d\log d)$ additions per token, with its inverse absorbed offline into the output projection (\Cref{def:fpt}). Its \emph{input}, by contrast, has only a sparse fringe of extreme values, so it is handled not by a rotation but by a percentile-clipped scale. The same paper thus uses a rotation on one tensor and a robust scale on another, chosen by their differing outlier shapes: the section's theme in miniature.

MambaQuant \citep{xu2025mambaquant} attacks the same Mamba architecture as Quamba, but from the channel side, and supplies the sharpest test of this survey's framing. It reports that on Mamba a Hadamard \emph{alone} fails to equalize channel variance: on the correlated channels of the gate and output projections and the SSM matmul the rotated diagonal $(H^{\!\top}\Sigma H)_{ll}=d^{-1}\sum_{jk}\pm\Sigma_{jk}$ stays uneven. Its fix is an offline transform that pre-multiplies the Hadamard by the KLT eigenbasis, $H_K=KH$, applied to the LoRA module and the inter-block connection, where the output, gate and state projections are transformed; decorrelating first makes the rotated diagonal exactly $(H^{\!\top}\!\diag(\lambda)H)_{ll}=d^{-1}\tr\Sigma$ for every channel, so one shared scale fits them all equally. The KLT appears here as a \emph{decorrelating pre-conditioner in the service of flattening}, the same role BASE-Q's closed-form $U^{\!\top}\!H$ plays on the residual stream (\Cref{sec:tax-affine}), and it is evidence for the inversion rather than against it. It is not the rate-allocation concentration that \Cref{cor:notransfer} deprives of any guarantee, nor ResQ's mixed-precision subspace split (\Cref{sec:tax-learned-orthogonal}). $H_K$ is the orthogonal-only relative of WUSH's whitening-then-Hadamard construction (\Cref{sec:tax-affine}) with the whitening factor dropped, since it equalizes the second moment's \emph{diagonal} while, being orthogonal, leaving the second moment's spectrum intact. A smooth-fused rotation that merges per-channel smoothing scales into the weights completes the recipe, giving under one point of accuracy loss at \texttt{W8A8} on Vim and Mamba language models.

\subsection{Mixture-of-experts}
\label{sec:beyond-moe}

The sparsely-activated mixture-of-experts (MoE) models at the current frontier change nothing about the transform itself. Each expert is an ordinary linear layer, so its weight matrix takes the full toolkit of \Cref{sec:taxonomy}, and a residual-stream rotation folds into the expert and router projections exactly as it does into a dense layer's, the block's RMSNorm gain folding away into them first as \Cref{thm:invariance} requires (\Cref{prop:gamma}). What MoE changes is the \emph{calibration} a data-aware transform depends on. Because the router sends each token to only a few experts, a fixed calibration set is split unevenly, and the rarely-routed experts receive too few, and biased, samples to estimate the second moments that whitening (\Cref{sec:tax-affine}) and error-feedback rounding (\Cref{sec:comp-rounding}) rely on: below $d_{\mathrm{in}}$ routed tokens an expert's $X^{\!\top}X$ is not merely noisy but singular, so the rounder runs only on the damped Hessian of \Cref{prop:obs}. MoEQuant \citep{hu2025moequant} isolates exactly this failure, generating a data-free, expert-balanced calibration set by self-sampling from the model and weighting the objective by sample--expert affinity to recover the accuracy that per-expert statistical starvation otherwise costs. The expert dimension additionally opens a bit-allocation axis (more bits for heavily-routed experts) that is orthogonal to, and composes with, the transforms surveyed here. Per-coordinate mixed precision is the opposite case: by restoring a measure of allocation flexibility it changes which transform is optimal (\Cref{sec:inversion}). The transform question in an MoE is thus unchanged; only the statistics feeding a data-aware transform must be gathered per expert.

\subsection{Diffusion transformers and the time axis}
\label{sec:beyond-diffusion}

Diffusion models add an axis absent from language models: the denoising timestep, across which activation distributions drift substantially. This makes the central design question not \emph{which} transform but how it should vary with time, and the early methods answer without any reparameterizing transform at all. Q-Diffusion \citep{li2023qdiffusion} handles the drift purely in the calibration data, sampling calibration inputs across all timesteps, and otherwise only splits the bimodal shortcut activations of the U-Net into their two concatenation branches, quantized separately. Q-DiT \citep{chen2024qdit} likewise uses no transform, allocating per-layer group sizes by evolutionary search and recomputing activation scales dynamically per sample and timestep. The transform proper enters as the SmoothQuant migration of \Cref{sec:tax-diagonal} transplanted onto the DiT, and the interesting variation is temporal. ViDiT-Q \citep{zhao2024viditq}, in its original form, makes the per-channel smoothing scale \emph{timestep-dependent}, using different migration strengths for the two halves of the trajectory; its later version drops that split in favor of combining scaling with a rotation for the time-varying component, tracking the field's scaling-to-rotation drift. DiTAS \citep{dong2024ditas} takes the opposite view, aggregating the activation maximum over \emph{all} timesteps into a single static per-channel factor and repairing the residual weight error with a training-free low-rank branch. Whether the scale should track the timestep or be robust to it is the diffusion-specific form of the granularity axis.

The aggressive low-bit DiT methods bring in the rotations and low-rank branches of the weight-quantization literature. LRQ-DiT \citep{yang2025lrqdit} gates the transform on the data: a per-layer activation-fluctuation statistic decides whether a plain Hadamard suffices or whether a stronger outlier-aware rotation-plus-permutation is warranted, a within-model instance of the fixed-versus-adaptive choice. SVDQuant \citep{li2025svdquant} is the most distinctive. After the usual smoothing, it decomposes the weight by SVD and carries the dominant rank-32 subspace in a 16-bit \emph{low-rank branch}, leaving a well-conditioned residual that quantizes cleanly to \texttt{W4A4}, reaching near-\texttt{BF16} FID on \texttt{FLUX.1-dev} at a $3\times$ speedup over a 4-bit weight-only (\texttt{W4A16}) baseline, the comparison that isolates the \texttt{W4A4} datapath gain.

What a DiT gives up is not the rotation but its \emph{free} version. Adaptive normalization generates its per-channel scale and shift at runtime, so, unlike the static RMSNorm gain that QuaRot folds away before rotating the residual stream (\Cref{sec:tax-fixed-orthogonal}), there is no static diagonal for the residual-stream rotation of \Cref{thm:invariance} to commute past. Rotations therefore have to enter at the projection inputs, at the online-Hadamard cost priced in \Cref{sec:sys-kernels}, which is exactly what LRQ-DiT above and HadaNorm (\Cref{sec:fmt-codesign}) do, alongside a wider rotation-for-DiT line \citep{shao2025trdq,huang2025convrot,sharify2026dirotq}. Unlike the activation-by-activation matmuls of attention, where neither operand is static, and unlike RoPE, which denies the key rotation its free fold outright, so that the rotation is paid online while the value-path one still folds, the obstruction here is a runtime \emph{diagonal}, not a missing weight. What SVDQuant uses instead is neither a rotation nor a scaling but an \emph{additive off-precision} correction: the outliers the four-bit grid cannot hold are absorbed by a high-precision low-rank term, and a fused kernel cuts the low-rank branch's otherwise-substantial latency to near-negligible. It is the clearest example in the survey of an \emph{adjacent} mechanism. It changes the \emph{arithmetic} rather than the basis, so it sits beside the function-preserving transforms of \Cref{sec:taxonomy} rather than among them, and it composes with, rather than replaces, the smoothing that precedes it.

\subsection{Training-time gradients, and where flattening fails}
\label{sec:beyond-training}

The transforms so far all serve inference, but several of the format methods of \Cref{sec:format} quantize the training pass, where a new tensor appears: the gradient. Full \texttt{INT4} training \citep{xi2023int4training} is the sharpest illustration of this section's thesis, because it applies opposite treatments to the forward and backward tensors on principle. The forward activations carry concentrated per-coordinate outliers, so they and the weights alongside them get a block-diagonal Hadamard (the incoherence rotation of \Cref{sec:tax-fixed-orthogonal}) which spreads each outlier across its block. In a training pass there is no static weight to absorb the inverse into, so the transform is applied to both operands and the two copies cancel inside the product, paid online rather than folded in the sense of \Cref{def:fpt}. The gradients carry no per-\emph{channel} outlier for such a rotation to spread. Their structure is a between-row sparsity, a few token-rows carrying almost all the norm and most rows near zero, a consequence of over-parameterization, and under a shared scale those rows are themselves the range-setting extremes. A flattening rotation is the wrong instinct here, though not through the row-mixing image it first suggests. The forward Hadamard acts on the channel axis, and a feature-side rotation preserves each token-row's norm, so it can neither erase nor exploit the between-row sparsity. We read the paper's design as treating that sparsity as a structure to \emph{select from} rather than a range to flatten. The gradients get no transform at all, but instead a leverage-score importance sampling, scoring each row by its gradient norm together with that of the activation row it pairs with, that retains the high-scoring rows with high probability and subsamples the rest, rescaled to stay unbiased, together with a bit-split into two integer planes. \citet{ashkboos2025halo} see the same row-dominant error structure and show it is not beyond a rotation's reach, only beyond a \emph{feature-side} one: they apply a left-hand Hadamard to the error tensors, mixing rows rather than channels, and report that this is what makes \texttt{INT8} fine-tuning stable, while for the wider dynamic range of \texttt{FP6} it suffices to rotate only the forward-pass operands, the inputs and the weights, leaving the error tensor untransformed. The axis, not the tensor, is what decides whether a rotation can help here. The gradient exception marks the boundary of the survey's central prescription. Flattening is the surrogate-optimal response when the error is dominated by a heavy tail under a shared scale (the fixed-rate, AbsMax regime of \Cref{sec:inversion}) and it is simply inert, on the feature axis, against between-row sparsity. The gradient case sits with the \emph{ideal} floating-point grid of \Cref{sec:format} and the concentration optimum of \Cref{sec:classical} among the settings where flattening \emph{along the feature axis} is not the answer. On an ideal FP grid a common rotation can destroy a favorable anti-alignment between the weight and activation profiles, as measured on the matrices \citet{ordentlich2026highrate} test, the element grid's spacing accounting only for the milder fact that FP gains less; variable-rate coders reward concentration; and sparse gradients reward selection, or a rotation on the \emph{row} axis, but not one on the feature axis. The deployed FP4 formats sit between the poles, as \Cref{sec:fmt-flip} records. The transform toolkit is powerful and general, but it is not universal, and knowing which tensor rewards which lever (rotation, and on which axis; multiplicative scaling or an additive shift; re-parameterization; low-rank absorption; selection or routing; or nothing) is the practical skill this section has tried to convey. \Cref{sec:composition,sec:format} and this section together map where each lever pays off; the next section turns from the algorithms to the systems that run them and to the one coding regime the matrix instruction still cannot host.

\section{Systems and the Variable-Length Lane}\label{sec:systems}

An optimal transform that cannot run at speed is of no use, and the systems layer is where the survey's abstractions meet the tensor cores. Two questions matter here. The first is how the transforms of \Cref{sec:taxonomy,sec:beyond} are made cheap enough to deploy: the reason a non-foldable, data-dependent map can now sit on the inference path at all. The second is sharper, and closes a loop opened in \Cref{sec:classical}. The shared-scale operand tile a matrix instruction consumes admits no per-coordinate allocation in the sense \Cref{def:rate} names, for the datapath reasons of \Cref{prop:datapath}, and \Cref{cor:notransfer} is why no classical optimum transfers into that regime, so what has become of the entropy-coded, variable-length lineage that classical transform coding drew on? We take the kernels first (\Cref{sec:sys-kernels}), then the dequantization bottleneck and the GPU datapath that is reshaping it (\Cref{sec:sys-dequant}). After that comes how the block size couples to the hardware scale multiplier that removes it (\Cref{sec:sys-groupsize}), then the variable-length lane and the precise sense in which it survives (\Cref{sec:sys-vr}), and finally what the shipped 4-bit releases actually do (\Cref{sec:sys-deployed}). \Cref{tab:systems} lists the kernels, serving stacks and weight coders of this section; the deployed releases of \Cref{sec:sys-deployed} are named in the text rather than tabulated.

\begin{table}[!htbp]
\centering\footnotesize
\renewcommand{\arraystretch}{1.15}
\caption{\textbf{Systems: fused kernels, serving stacks, and lossless/variable-length weight coders (\Cref{sec:systems}).}}\label{tab:systems}
\begin{tabularx}{\linewidth}{>{\raggedright\arraybackslash}p{3.0cm} >{\raggedright\arraybackslash}p{1.9cm} >{\raggedright\arraybackslash}X >{\raggedright\arraybackslash}p{1.5cm}}
\toprule
\textbf{Method} & \textbf{Role} & \textbf{What it provides} & \textbf{Rate} \\
\midrule
CCQ~\citep{zhou2025ccq} & fixed-rate coder & $\sim$2-bit (2.06 bpw) \emph{lookup-free} convolutional code, bit-shift decode (no codebook lookup) & fixed-rate \\
HadaCore~\citep{agarwal2024hadacore} & kernel & online Walsh--Hadamard rotation for incoherence processing & fixed-rate \\
QUIK~\citep{ashkboos2023quik} & kernel & mixed-precision \texttt{W4A4} GEMM; $\sim$256 outlier columns kept in FP16, summed into the INT4 GEMM epilogue (no extra round-trip) & fixed-rate \\
DFloat11~\citep{zhang2025dfloat11} & lossless-coder & BF16 exponent field, GPU decode at transformer-block granularity, \ensuremath{\sim}30\% weight-memory cut & variable-length \\
NeuZip~\citep{hao2024neuzip} & lossless-coder & BF16 exponent field, GPU pre-GEMM decode & variable-length \\
ZipNN~\citep{hershcovitch2024zipnn} & lossless-coder & BF16 exponent field for storage/network transfer & variable-length \\
ZipServ~\citep{fan2026zipserv} & lossless-coder & tensor-core-aware bitmap encoding, decode fused into the GEMM mainloop & \emph{fixed}-length \\
Tile-level ANS~\citep{tan2026shannon} & lossless-coder & rANS on low-bit formats, decoded per tile inside the GEMM kernel & variable-length \\
Huff-LLM~\citep{yubeaton2025huffllm} & lossless-coder (hardware) & FP16/BF16 fields (exponent coded; FP16 also the mantissa; sign raw), one weight/clock streaming decode & variable-length \\
QServe~\citep{lin2024qserve} & serving-system & W4A8KV4 quantized GEMM on INT8 tensor cores & fixed-rate \\
LLM Compressor~\citep{vllm_llmcompressor} & toolkit & ships GPTQ, AWQ, SmoothQuant, AutoRound, RTN $+$ SpinQuant/QuIP rotations (online Hadamard runs at inference in vLLM via HadaCore kernels~\citep{agarwal2024hadacore}) & fixed-rate \\
TensorRT Model Optimizer~\citep{nvidia_modelopt} & toolkit & ships SmoothQuant, AWQ, GPTQ + fast Hadamard, NVFP4, MXFP4 PTQ export & fixed-rate \\
\bottomrule
\end{tabularx}
\end{table}

\subsection{Kernels that make transforms cheap}
\label{sec:sys-kernels}

The single most important systems fact for this survey is that the online transform has become cheap enough to sit on the inference path once it is \emph{fused} with the quantizer. \citet{sun2024flatquant} measure a $0.07\times$ end-to-end slowdown for their five fused affine maps, against $0.26\times$ for QuaRot's three online Hadamards, which run as a separate kernel stage ahead of the quantize step (\Cref{sec:emp-cost}). Fusion, not the transform's mere existence, is what buys the low figure. Not free, and for an unfused online rotation not yet negligible, but no longer prohibitive. A rotation that folds into weights by computational invariance (\Cref{thm:invariance}) costs nothing. But the data-dependent maps of \Cref{sec:tax-affine} and the online Hadamards that QuaRot and its successors insert before the down-projection do not fold, and their viability rests on fast kernels. HadaCore \citep{agarwal2024hadacore} is the representative, recasting the fast Walsh--Hadamard transform with a size-16 base case computed directly on the tensor cores. The online rotation used for incoherence processing then runs $1.1$--$1.4\times$ faster than the previous best kernel on an A100 (peak $3.5\times$), and $1.0$--$1.3\times$ on an H100 (peak $3.6\times$), at comparable quantization error. This is what makes \Cref{prop:rht} deployable rather than merely provable. Together with the fused transform-and-quantize kernels of FlatQuant and WUSH (\Cref{sec:tax-affine}) and the lookup-table engine FLUTE for the scalar codebooks (\Cref{sec:vq}), HadaCore is what keeps the fixed-rotation rung cheap, the migration up the degrees-of-freedom ladder being carried by the fused transform-and-quantize kernels alongside it. 

Around the transform sits the rest of the serving stack, and two systems illustrate how the quantized GEMM is kept on the fast path. QServe \citep{lin2024qserve} targets \texttt{W4A8KV4}, keeping every GEMM on the \texttt{INT8} tensor cores, its attention staying in \texttt{FP16} on the CUDA cores, so that no partial-sum rescale stalls the integer datapath (that mechanism, and the dequantization overhead it sidesteps, are the subject of \Cref{sec:sys-dequant}). For the transform stage it carries a SmoothAttention scaling that migrates the fixed per-channel key outliers of \Cref{sec:beyond} onto the queries. QUIK \citep{ashkboos2023quik} keeps a small fraction of outlier channels (256 columns by default, ${\approx}3\%$ of the hidden dimension at \texttt{OPT-66B} scale the fraction being set by the hidden width rather than the parameter count) in \texttt{FP16} and the rest in \texttt{INT4}. It fuses the \texttt{INT4} dequantization into the GEMM epilogue and accumulates the result directly into the output of the separate \texttt{FP16} outlier matmul, so the two precisions combine without an extra write-and-read pass over the \texttt{INT32} partial sums (its end-to-end speed-up is collected in \Cref{tab:res_systems_eff}). The common lesson is that the transform, the rounding, and the mixed-precision handling all have to be \emph{fused} into the GEMM to be worth their accuracy. A transform reported without its kernel is an incomplete result, as \Cref{sec:pitfalls} warns.

These techniques are no longer confined to research code: the transforms this survey catalogs ship in production post-training-quantization toolkits. NVIDIA's TensorRT Model Optimizer implements SmoothQuant, AWQ, and GPTQ, and exposes a fast Hadamard transform in its quantizer for QuaRot-style rotation, alongside \texttt{NVFP4} and \texttt{MXFP4} post-training quantization with direct export to TensorRT-LLM, vLLM, and SGLang \citep{nvidia_modelopt}. The vLLM \texttt{llm-compressor} toolkit ships GPTQ, AWQ, and SmoothQuant together with the rotation methods SpinQuant and QuIP as first-class calibration modifiers. There the Hadamard is applied at inference time through tensor-core HadaCore kernels \citep{agarwal2024hadacore} rather than only at calibration time \citep{vllm_llmcompressor}. The incoherence rotations of \Cref{sec:tax-fixed-orthogonal}, the migration scalings of \Cref{sec:tax-diagonal}, and the error-feedback rounders of \Cref{sec:comp-rounding} appear as named, composable stages in both toolkits, with the same fused-kernel discipline the research prototypes established. That is the clearest sign that the transform stage has become standard deployment machinery rather than a research curiosity.

\subsection{The dequantization bottleneck and the GPU datapath}
\label{sec:sys-dequant}

\begin{table}[!htbp]
\centering\small
\renewcommand{\arraystretch}{1.25}
\caption{\textbf{Low-precision matmul support across GPU generations, and its bearing on
the dequantization cost.} The datapath the transforms must survive is a moving target:
integer \texttt{INT4} was a first-class tensor-core type on Ampere, silently lost its tensor-core path on Hopper, though Ada kept it, and has been superseded on Blackwell and CDNA4 by
\emph{hardware-native} 4-bit floating-point with microscaling, where the tensor core
applies the per-block scale itself. Sources: \citet{nvidia_ampere,nvidia_hopper,nvidia_blackwell,nvidia_ptx_isa,amd_cdna3,amd_cdna4,luo2024hopper}, the NVIDIA block-scale instruction semantics being documented in the PTX ISA reference.}
\label{tab:hardware}
\begin{tabularx}{\linewidth}{@{}l l c >{\raggedright\arraybackslash}X@{}}
\toprule
\textbf{Accelerator (arch.)} & \textbf{Low-precision matmul} & \textbf{MX/NVFP4} & \textbf{Bearing on the dequantization cost} \\
\midrule
A100 (Ampere, '20)      & \texttt{INT4}, \texttt{INT8}, \texttt{FP16}/\texttt{BF16} & -- & \texttt{INT4} GEMM is tensor-core-native ($4\times$ \texttt{FP16}), but W4A4's per-group scales still rescale INT32 partial sums on the CUDA cores. \\
H100/H800 (Hopper, '22)      & \texttt{FP8}, \texttt{INT8}, \texttt{FP16}/\texttt{BF16} & -- & \texttt{INT4} dropped from the tensor core: it compiles to CUDA-core \texttt{IMAD}; low-bit compute shifts onto the \texttt{FP8}/\texttt{INT8} path. \\
B200/GB200 (Blackwell, '24) & \texttt{FP4}, \texttt{FP6}, \texttt{FP8}, \texttt{INT8}, \texttt{FP16}/\texttt{BF16} & \texttt{NVFP4}, \texttt{MXFP4/6/8} & The tensor core applies the block scale in hardware, so the \texttt{FP4} block-scale ``dequantization'' is not a separate software step. \\
MI300X (CDNA3, '23)     & \texttt{FP8}, \texttt{INT8}, \texttt{FP16}/\texttt{BF16} & -- & \texttt{FP8}/\texttt{INT8} matrix-core path; no native \texttt{FP4}/\texttt{FP6}. \\
MI355X (CDNA4, '25)     & \texttt{FP4}, \texttt{FP6}, \texttt{FP8}, \texttt{INT8}, \texttt{FP16}/\texttt{BF16} & \texttt{MXFP4/6/8} & First AMD data-center \texttt{FP4}; the MX block scale is handled in the matrix core. \\
\bottomrule
\end{tabularx}
\end{table}

A transform that is cheap to apply still has to feed a matrix multiply whose operands are quantized, and here the \texttt{W4A4} regime pays a cost the accuracy analysis never sees. An \texttt{INT4}$\times$\texttt{INT4} tensor-core multiply accumulates in \texttt{INT32} along the contraction ($K$) dimension. The fate of the quantization scale turns entirely on \emph{how finely} it varies along $K$. A per-tensor scale, a per-token activation scale, or a per-output-channel weight scale is constant along $K$, so it factors out of the whole reduction and is applied \emph{once}, to the finished \texttt{INT32} accumulator, in the GEMM epilogue, essentially free. QuaRot takes that path \citep{ashkboos2024quarot}, running a pure integer matmul with a single epilogue rescale, made accurate enough by a Hadamard that flattens the tensor so a coarse scale suffices. QUIK \citep{ashkboos2023quik} also reaches that path by other means, pulling the few outlier channels out to \texttt{FP16} rather than rotating (\Cref{sec:sys-kernels}).

A \emph{per-group} scale, by contrast, changes several times \emph{within} the reduction, so it cannot be folded into one output scale. The partial sums must be converted and rescaled group by group, mid-reduction, and that work lands not on the tensor cores but on the general-purpose CUDA cores. In the per-group \texttt{W4A4} regime this toll is paid twice. Both operands are $4$-bit with a scale that varies along $K$ (the weight's per-input-group scale and the activation's per-group scale), so two independent rescales, not one, must be unfolded inside the reduction. The mid-reduction \emph{partial-sum} rescale is what \texttt{W4A16} (only the weight is quantized) and \texttt{W4A8} (the activation stays at $8$ bits, its scale absorbed into the \texttt{INT8} accumulation) each avoid. That is ruinous in the only currency that matters here. On an A100-class GPU a single CUDA-core operation costs about as much as fifty \texttt{INT4} tensor-core operations, so a rescale interleaved into the inner loop dominates the multiply it serves. \citet{lin2024qserve} put the resulting penalty, across existing \texttt{INT4} methods and for dequantizing weights or partial sums, at $20$--$90\%$ of runtime, the partial-sum case sitting at the upper end, and \citet{zhao2023atom} fuse the dequantize-and-accumulate into the matmul pipeline to run it in place. The systems lesson sharpens the accuracy one: the transform is not only an accuracy device but a \emph{cost} device. By flattening the tensor it lets a coarse, epilogue-cheap scale stand in for the per-group scale the datapath charges for (\Cref{prop:flat-opt}).

\begin{table}[!htbp]
\centering\small
\renewcommand{\arraystretch}{1.2}
\caption{\textbf{The four-bit lanes, compared by where the per-scale-group multiply
(the ``dequantization'') lands.} The cost of low-bit inference is set less by the
nominal bit-width than by \emph{how many} operands are quantized and at \emph{what
granularity} the shared scale varies, because a scale that varies along the
matmul's contraction ($K$) axis cannot be folded into a single output scale and must
be applied mid-reduction. Coarse (per-tensor/channel) scales rescale once in the
epilogue; per-group scales rescale in the main loop on the CUDA cores; and native
\texttt{FP4} moves the per-block multiply into the tensor core itself.}
\label{tab:w4lanes}
\begin{tabularx}{\linewidth}{@{}l l X@{}}
\toprule
\textbf{Lane} & \textbf{W\,/\,A} & \textbf{Where the per-scale-group multiply lands, and representative kernels} \\
\midrule
\texttt{W4A16} & \texttt{INT4}\,/\,\texttt{FP16} & Weight-only: an \texttt{INT4}$\to$\texttt{FP16} upconversion that a well-pipelined kernel \emph{hides} behind the weight loads it saves at the batch sizes decode runs at (Marlin holds near-$4\times$ to batch $16$--$32$; Machete), or eliminates outright by tabulating partial products (LUT-GEMM); at larger batch the GEMM turns compute-bound and the memory saving has nothing left to hide behind. \\
\texttt{W4A8} & \texttt{INT4}\,/\,\texttt{INT8} & \texttt{INT4} weights decoded to \texttt{INT8} (register-level, or the scale folded offline), then \texttt{INT8}$\times$\texttt{INT8} on the tensor cores: no partial-sum rescale (QServe/QoQ, QQQ, LiquidGEMM). \\
\texttt{W4A4} \texttt{INT}, coarse & \texttt{INT4}\,/\,\texttt{INT4} & Any scale constant along $K$ (per-tensor, per-token activation, or per-output-channel weight) $\Rightarrow$ one rescale of the \texttt{INT32} accumulator in the \emph{epilogue} (cheap); viable once a strong transform flattens the tensor (QuaRot) or the outliers are pulled to \texttt{FP16} (QUIK), with MergeQuant folding the scale away and PrefixQuant making it static. \\
\texttt{W4A4} \texttt{INT}, per-group & \texttt{INT4}\,/\,\texttt{INT4} & Per-group scale varies along $K$ $\Rightarrow$ partial sums rescaled in the \emph{main loop} on the CUDA cores, the upper end of the $20$--$90\%$ dequantization tax QServe measures across \texttt{INT4} methods (Atom fuses it into the pipeline; COMET, APEX4 attack it directly). \\
\texttt{W4A4} \texttt{FP4}, native & \texttt{FP4}\,/\,\texttt{FP4} & Per-block ($16$ or $32$) scale applied \emph{inside} the tensor core by a block-scaled MMA: no software step (Blackwell \texttt{tcgen05.mma}; CDNA4 documents the capability without naming the instruction). \\
\bottomrule
\end{tabularx}
\end{table}

\Cref{tab:w4lanes} lays this out across the four-bit lanes. \texttt{W4A16} is the mild case. Only the weights are $4$-bit, so the scale multiply is a one-time \texttt{INT4}$\to$\texttt{FP16} weight expansion, and because weight-only decode is memory-bound it hides behind the traffic it saves. Marlin \citep{frantar2024marlin} pipelines the upconversion with the tensor-core math (streaming the weights through asynchronous loads) to hold near the ideal $4\times$ up to batch $16$--$32$ (up to $2.8\times$ end to end in vLLM). Its Hopper successor Machete \citep{machete2024} does the same through \texttt{TMA} and \texttt{wgmma}, and LUT-GEMM \citep{park2022lutgemm} removes the arithmetic dequantization entirely by tabulating partial products.

\texttt{W4A8} is the case a serving stack engineers around. QServe's \texttt{QoQ} \citep{lin2024qserve}, QQQ \citep{zhang2024qqq}, and LiquidGEMM \citep{hu2025liquidgemm} all decode the \texttt{INT4} weights up to \texttt{INT8} (in registers, or with the dequant scale folded offline into the quantization scales) and run every GEMM on the \texttt{INT8} tensor cores, so no \texttt{INT32} partial-sum rescale survives \emph{inside} the reduction, the per-channel scale being fused into the epilogue instead. The activations stay at $8$ bits, trading some memory saving for a datapath that keeps the tensor cores fed (QQQ reports $3.7\times$ and $3.3\times$ per-channel and per-group \texttt{W4A8} GEMM speed-ups over \texttt{FP16}). \texttt{W4A4} then splits along the granularity line above. The coarse-scale systems (QuaRot and QUIK, and MergeQuant \citep{wang2025mergequant}, which folds the scale into the adjacent weights so no runtime rescale survives, and PrefixQuant \citep{chen2024prefixquant}, which freezes it so a static per-tensor scale suffices) keep the epilogue-cheap path. The per-group systems (Atom; the \texttt{W4A4KV4} server COMET \citep{liu2024comet}, mixed \texttt{W4A4}/\texttt{W4A8} in practice, with about $16\%$ of activations left at $8$ bits; and the pure-\texttt{W4A4} kernel co-design APEX4 \citep{guo2026apex4}), by contrast, attack the main-loop tax head-on. The main-loop tax is why an integer \texttt{W4A4} GEMM's per-operation gain trails its arithmetic $4\times$ once the rescale is paid for. It is also why QServe, reporting $2.5$--$2.9\times$ higher throughput on A100 than the \texttt{W4A4} systems it compares against (Atom and QuaRot), argues the \texttt{W4A8} compromise is often the better systems bet on today's integer datapath (the measured systems figures are collected in \Cref{tab:res_systems_eff}).

Underneath all of this the hardware is moving, and it changes the problem rather than solving it in place (\Cref{tab:hardware}). Integer \texttt{INT4} was a first-class tensor-core type on Ampere, at $4\times$ the \texttt{FP16} rate \citep{nvidia_ampere}. On Hopper it quietly lost its tensor-core path, and an \texttt{INT4} matrix instruction now compiles to \texttt{IMAD} operations on the CUDA cores \citep{luo2024hopper,nvidia_hopper}, so the very datapath the \texttt{W4A4} integer methods target has narrowed on the generation most deployments run. What replaced it is floating-point: Hopper added \texttt{FP8}, and Blackwell's fifth-generation tensor cores add \texttt{FP4}/\texttt{FP6} with \emph{hardware-native} microscaling, applying the per-block \texttt{NVFP4} (and MX) scale in silicon during the multiply \citep{nvidia_blackwell}. AMD's CDNA4 (MI355X) likewise adds native \texttt{FP4}/\texttt{FP6} with MX support \citep{amd_cdna4}, where CDNA3 had stopped at \texttt{FP8} \citep{amd_cdna3}. On this hardware the block-scale ``dequantization'' that costs software so dearly is simply not a separate step. That is the systems reason the frontier has swung from \texttt{INT4} to the \texttt{FP4}/microscaling formats of \Cref{sec:format}, and the reason the transform question there is posed on \texttt{MXFP4}/\texttt{NVFP4} rather than \texttt{INT4}.

Two caveats keep the problem alive. First, the installed base is Ampere and Hopper, so the software dequant tax is the reality for most current serving. The routes around it are either to sidestep the partial-sum rescale (QServe's \texttt{W4A8}, or MergeQuant \citep{wang2025mergequant}) or to avoid arithmetic decode entirely (the lookup-table GEMM of \Cref{sec:vq}). MergeQuant merges the per-channel scale into the adjacent linear operators so that no runtime rescale survives, though a lightweight dimensional-reconstruction stage stays online (\emph{hybrid} in \Cref{tab:taxonomy}), a static \texttt{W4A4} path reporting up to $2.06\times$ end to end. Second, hardware that removes the block-scale cost does so only for the coarse (power-of-two or \texttt{FP8}) block scale the format defines, so the accuracy penalty of that scale (\Cref{sec:fmt-formats}) is the price of the free dequantization. The cost has moved off the datapath and into the number format, which is exactly where \Cref{sec:format}'s transforms go to work. And that free hardware scale comes at exactly \emph{one} granularity, the format's block size, which we take up next.

\subsection{Group size and the hardware scale multiplier}
\label{sec:sys-groupsize}

The hardware that removes the dequantization cost does not remove it for free at any granularity: it removes it at one granularity, the format's, and that turns block size into the decisive design variable. Blackwell's fifth-generation tensor core applies each block's shared scale \emph{inside} a single block-scaled matrix instruction (\texttt{tcgen05.mma}, \citealp{nvidia_ptx_isa}), multiplying every $K$-block's partial product by its scale during the accumulate rather than after it \citep{nvidia_blackwell}. AMD's CDNA4 adds instruction and hardware support for the same microscaling formats, the cited architecture reference stating the capability without naming the instruction \citep{amd_cdna4}. The block-scaled instruction supplies the ``multiplier per scale group'' the software \texttt{W4A4} path lacks. The mid-reduction rescale, part of the $20$--$90\%$ dequantization tax QServe measures, is now a wire in the tensor core. But the instruction exposes the scale only at the block sizes the format defines: $32$ elements for the MX formats and $16$ for NVFP4. The scale-vector length is an enumerated hardware choice rather than a free parameter. The hardware exposes its scale only for the format's own element types and only at the block sizes it defines, so an \texttt{INT4} scheme gets no hardware scale path at any group size, and a weight-only one at group $128$ falls back to the software rescale on both counts. Group size, then, is no longer a continuous accuracy knob traded against a continuous cost. On the block-scaled datapath it is quantized to what the silicon multiplies for free.

Within that constraint the block size is a real accuracy--overhead lever, and the two shipping formats sit at different points on it. A smaller block gives the shared scale fewer, more homogeneous values to cover, so less of each block's range is wasted. Finer blocks are more accurate, one reason NVFP4 halves the MX block to $16$, though \Cref{sec:fmt-formats} makes the scale's \emph{representation} the larger term in the error \citep{nvidia2025nvfp4}. The cost is metadata. An $8$-bit scale shared over $k$ elements adds $8/k$ bits per element, so MXFP4's $32$-element block costs $4.25$ effective bits while NVFP4's $16$-element block costs $4.5$ (plus a small per-tensor \texttt{FP32} scale) \citep{rouhani2023microscaling,nvidia2025nvfp4}. NVFP4 spends that extra quarter-bit, and a mantissa-carrying \texttt{E4M3} block scale in place of MX's power-of-two \texttt{E8M0}, precisely to buy back the block-scale rounding error \Cref{sec:fmt-formats} identified. An \texttt{NVFP4}-pretrained 12B model reports MMLU within about a point of an \texttt{FP8} baseline at the same token budget ($76.6$ vs.\ $77.4$; \Cref{sec:fmt-codesign}). No matched \texttt{MXFP4}-versus-\texttt{FP8} downstream comparison at that scale has been published; the post-training evidence for the coarser scale's cost is \Cref{tab:res_fp4}. The block size also couples to the format flip of \Cref{sec:fmt-flip}, because the finer the block, the lower its within-block crest factor, and the more a flattened integer grid competes with the floating-point one at the same granularity. \citet{chen2025intvsfp} confirm that crossover empirically (fine-grained \texttt{NVINT4} with a Hadamard rotation matches \texttt{NVFP4}). So the transform, the format, and the block size are one joint choice, not three.

On the block-scaled datapath, in short, native microscaling is a first, deliberately restricted return of fine-grained adaptive \emph{scaling} (gain adaptation), not yet of per-coordinate bit allocation, to the inference datapath (\Cref{sec:open}). The hardware will carry one adaptive scale per $16$ or $32$ elements at full speed, and the block size decides how fine that free adaptation is and what it costs in effective bits. That finer, multi-level direction is not new. VS-Quant \citep{dai2021vsquant} already gave each $16$--$64$-element vector its own integer scale under a coarser floating-point one, and the shared-microexponents Block Data Representation of \citet{rouhani2023shared} (the ISCA-2023 origin of the MX formats) mapped a two-level design space down to sub-block scaling with mantissas as narrow as two bits. What remains open is how much finer the free per-block scale can profitably go on the tensor core, against the area each halving spends, and which transform co-designs with it (\Cref{sec:open}).

\subsection{The surviving variable-length lane}
\label{sec:sys-vr}

The variable-length, entropy-coded machinery of \Cref{sec:classical} did not disappear from serving, but what survives is its \emph{coding} half, lossless field-entropy compression, not its \emph{concentration} half. Its industrial descendants, the DeepCABAC and neural-network compression coders (\Cref{sec:classical-industry}), live on in a family of lossless weight coders, and looking at exactly where they run is instructive. Those lossless coders compress the \emph{stored} representation without applying a transform or concentrating energy, and they must hand a dense, equal-width tile back to the matrix instruction. It is \Cref{prop:datapath} that forbids per-coordinate allocation and entropy-coded indices in the operand tile the matrix instruction consumes, in the sense \Cref{def:rate} names, and \Cref{cor:notransfer} says only that no classical optimum transfers into it; neither makes lossless storage compression impossible, and it is the latter that these coders exploit.

The family starts from one redundancy. A \texttt{BFloat16} weight's exponent carries only about two to three bits of real information in its eight-bit field, so entropy-coding the exponent (and leaving the near-random mantissa alone) losslessly removes roughly a third of the footprint. That particular redundancy is a property of the \emph{unquantized} float: once weights are cast to a fixed-rate low-bit format there is no eight-bit exponent field left for a byte-oriented coder to work on. Field entropy itself survives the cast, and a tile-level ANS coder still reaches within about $0.01$--$0.1$ bits of the Shannon limit on the low-bit formats themselves \citep{tan2026shannon}. That is a storage-and-bandwidth gain too, but taken differently from the others here: the decode is fused into the GEMM kernel rather than run before it, staged through shared memory a tile at a time; the MMA instruction it feeds still receives a dense, equal-width tile. ZipNN \citep{hershcovitch2024zipnn} reports exactly this asymmetry: already-quantized checkpoints compress far less (\texttt{GGUF} ones not at all, others only to about $85$--$91\%$ of their size), whereas \texttt{BF16} checkpoints still shrink by roughly a third. That is the sharpest sign that this lossless coding is a storage-and-bandwidth layer, disjoint from the fixed-rate arithmetic the matrix instruction performs (\Cref{prop:datapath}).

What differs is where the variable-length decode happens relative to the GEMM. ZipNN keeps it entirely \emph{off} the datapath. It Huffman-codes the exponent for storage and network transfer and decompresses the whole model once on the CPU at load, after which standard fixed-rate kernels see only dense \texttt{BF16}. NeuZip \citep{hao2024neuzip} and DFloat11 \citep{zhang2025dfloat11} move the decode onto the GPU but confine it to a \emph{pre-GEMM} step. They entropy-code the exponent (ANS and Huffman respectively) and decompress into a scratch tile in a dedicated kernel just before use (NeuZip per weight matrix, DFloat11 at transformer-block granularity) then free it. The variable-length decode therefore never reaches the matrix engine, and the GEMM itself stays fixed-rate. DFloat11's two-phase kernel, with hierarchical lookup tables sized to fit in SRAM, achieves this at a decode cost that is constant per block and therefore amortizes over large token batches, the source reporting it constant and independent of batch size and warning that it still bites latency-sensitive small-batch serving, while cutting weight memory by thirty percent. The one coder that reaches the GEMM mainloop does so by giving up variable length altogether. ZipServ \citep{fan2026zipserv} replaces the variable-length bitstream with a fixed-length, tensor-core-aware bitmap encoding whose decode is constant-time and parallel, fused into the mainloop so that no dense scratch tensor is materialized. It belongs with CCQ below rather than with the coders above: it bought its place on the datapath at exactly the price the inversion predicts.

Huff-LLM \citep{yubeaton2025huffllm} is the most instructive, because it answers the constraint in hardware. It splits the \texttt{FP16} weight into small fields and Huffman-codes the exponent and mantissa (leaving the sign raw) with a tiny content-addressable memory. Between the weight buffer and the systolic array it inserts a row of streaming decoders that emits exactly one decompressed weight per clock cycle. Because the decoder's constant output rate is clamped to the array's fixed consumption rate, the variable-length storage is hidden entirely behind a constant-rate decode stage, and the matrix engine, which is fixed-rate by construction, needs no redesign. The precise systems statement of the inversion is this. Variable-rate coding cannot execute \emph{inside} the matrix instruction, but it can feed one three ways: by decompressing before it (in software, per layer); by decoding tile by tile inside the GEMM kernel itself, warp-specialized so that a decoded tile lands in shared memory exactly as the microkernel consumes it \citep{tan2026shannon}; or by pipelining a hardware decoder that emits one symbol per clock without bubbles, which Huff-LLM buys by splitting the word into short fields, an amortized rate being precisely what would stall the array. What none of them changes is the instruction: the tensor-core MMA still consumes a dense tile of equal-width values, which is where the fixed-rate property actually binds. Every coder here is bit-exact on the values it stores, so none restricts any coordinate to its own smaller value set; that is the premise \Cref{prop:datapath} leans on. The contrast that proves the point is CCQ \citep{zhou2025ccq}, a lossy quantizer rather than a lossless coder, but one whose decode stage is the instructive comparison because it is \emph{fixed}-rate. Its convolutional code maps indices to values by a lookup-free linear bit-shift (no codebook to index), so decode is synchronous arithmetic with no entropy stage. CCQ composes directly with a grouped GEMM at two bits per weight, exactly the fixed-rate property that lets its decode sit in the mainloop arithmetic itself, where an entropy stage cannot.

Variable-length coding, then, survives in precisely the form the inversion predicts: as a lossless storage and bandwidth technology that must be converted back to a fixed-rate representation before or as it enters the matmul, never as coding inside the MMA instruction itself. That boundary is a property of today's tensor cores, not a theorem about computation. Whether it will persist, and with it the separation of the two poles of the inversion, is one of the open problems we take up in \Cref{sec:open}.

\subsection{What the shipped 4-bit models actually do}
\label{sec:sys-deployed}
The frontier open-weight releases of 2025--2026 are a natural test of this survey's
premise, and they answer it in a way that sharpens the boundary drawn in
\Cref{sec:live} rather than blurring it.

On the 4-bit \emph{weight} path the frontier mixture-of-experts releases agree on a
transform-free recipe, though the shipped record is not unanimous. \texttt{gpt-oss-120b} converts its mixture-of-experts weights to \texttt{MXFP4}
while excluding attention, the router, the embeddings and the output head
\citep{openai2025gptoss}; Kimi~K2~Thinking ships its routed experts in \texttt{INT4}
obtained by quantization-aware training at group size~32, keeping those same four families in \texttt{BF16}, along with the shared experts and the dense layers \citep{moonshot2026kimik2thinking}; and DeepSeek's 4-bit release puts the routed experts in \texttt{FP4} and, in the card's own words, most other parameters in \texttt{FP8} \citep{deepseek2026v4}. In each
case the accuracy is bought by controlling training and by scoping the quantization to the
memory-dominant, outlier-poor tensors, which on the weight path substitutes for the transform of \Cref{def:fpt} rather than instancing it: the rounding noise is absorbed during training instead of the values being made easier to round. That recipe is not the whole shipped record, and the exceptions are both entries in our own taxonomy: first-party \texttt{AWQ} releases at the same bit-width, \texttt{Qwen3-32B-AWQ} among them, carry a 4-bit weight path whose per-channel diagonal scale is grid-searched on output error and folds offline \citep{lin2023awq,qwen2025qwen3awq}, a map classified as the \texttt{AWQ} row of \Cref{tab:taxonomy} and function-preserving by \Cref{def:fpt}; Meta's rotated release below is the other. Where the frontier cards substitute training control for the transform, those releases instance it.

On the activation path the transform appears where \Cref{sec:live} says there is nothing to fold into, and the sharpest evidence is inside one of those same checkpoints. DeepSeek reports \texttt{FP4} quantization-aware training for one activation path, the query--key path of its sparse-attention indexer, and that is the path it rotates. Its reference implementation, as of the June~2026 revision, makes the pairing explicit: \texttt{q = rotate\_activation(q)} precedes a call that simulates the \texttt{FP4} grid at block size~32 under a power-of-two scale, the rotation being applied for real and the grid modeled. The same file
instantiates the sibling compressor on the ordinary attention path with rotation left off
\citep{deepseek2026v4}. Within a single released model, then, the rotation is switched on
for the 4-bit activation path and off elsewhere. An earlier release had already put the
same rotation in front of an \texttt{FP8} quantizer on that path
\citep{deepseek2025v32code}. The activation side divides by format rather than agreeing on
one recipe: alongside that rotated path, \texttt{NVFP4} releases quantize weights \emph{and}
activations at group size~$16$ with no rotation named, the vendor toolkit shipping a Hadamard
among its options (\Cref{tab:systems}), \texttt{Llama-3.3-70B-Instruct-FP4} being the instance
we examine \citep{nvidia2025llama33fp4}. A 4-bit activation path therefore ships both
with a transform and without one, and across the releases we examine, which of the two it takes tracks the block scale rather than the bit-width. The pattern also shows up as a controlled pair from Meta,
which shipped one model at one target twice, 4-bit groupwise weights in the transformer-block linears with 8-bit per-token dynamic activations and the embedding and classifier left at 8 bits: once with fine-tuned SpinQuant rotations followed by GPTQ, and
once through QLoRA with no transform at all \citep{meta2024llama32quantized}. In that controlled pair, the transform is what the post-training route needs and the training-in-the-loop route does not. The two are not exclusive in general: DeepSeek applies FP4 quantization-aware training to the very indexer path it also rotates, which sharpens rather than weakens the pattern, since the rotation is kept even where training is controlled, precisely because nothing folds on an activation path.

Two further data points bound the claim. In pretraining, NVIDIA's \texttt{NVFP4} recipe
applies a random Hadamard only to the weight-gradient GEMM inputs, reporting that
transforming the forward and data-gradient inputs instead \emph{degrades} quality at the
scales tested, with the structural reason that the transform is not applied to the weights
and so cannot be inverted on those paths \citep{nvidia2025nvfp4}. And in serving,
\texttt{llama.cpp} applies a Walsh--Hadamard to the \texttt{KV} cache by default whenever
that cache is quantized and the head dimension is a multiple of~64
\citep{llamacpp2026kvrot}, paid online because a general-purpose runtime cannot fold a transform into an already-quantized checkpoint.

The reading we take from this is not that transforms are unnecessary at 4 bits. It is that the frontier mixture-of-experts weight paths ship without one, because training is
controlled and the tensors that carry the outliers are held out of the format entirely, which is
the regime in which \Cref{sec:emp-weightonly} already found the transform's margin smallest,
while the first-party \texttt{AWQ} and SpinQuant-rotated releases at the same bit-width do carry one. And the 4-bit activation path that does rotate rotates at precisely the site where
\Cref{thm:invariance} and \Cref{prop:gamma} offer nothing to fold into.

\section{The Empirical Landscape}\label{sec:empirical}

The theory tells us which transform is optimal (under its surrogate objective) for which regime. The literature tells
us how much it buys in practice. This section reads representative numbers off the
published record and assembles them into the picture the preceding sections predict.
Two disciplines make the exercise honest, and both follow from the pitfalls of
\Cref{sec:pitfalls}. First, we run \emph{no} new benchmark and estimate \emph{no}
number: every reported measurement below is transcribed verbatim from a cited table or figure, and the few differences and margins we quote are arithmetic on those transcribed values. Second, each comparison fixes its
protocol, the discipline that makes a cross-method comparison meaningful at all. In most
figures and tables that means drawing the numbers from a \emph{single} source, so
that model, bit-width, group size, rounder, and number format are held constant by
construction. The displays that combine sources are flagged where they appear. The
six-family overview of \Cref{tab:res_families} stitches two \texttt{W4A4} sources shown
to agree to within $1.5\%$ on the methods they share, and the systems-efficiency
landscape of \Cref{tab:res_systems_eff} states each row's own metric and baseline rather
than pretending to a common one. Numbers may therefore be compared \emph{within} a
block but not \emph{across} blocks.
The \texttt{W4A4} integer perplexities of \Cref{tab:res_act} and the \texttt{FP4}
accuracies of \Cref{tab:res_fp4} are different models under different protocols and are
not commensurable.

One consequence of fixing the protocol per comparison is worth stating plainly. Each
single-source table here is transcribed from the paper that proposes its own best-performing
row: \Cref{tab:res_act} from \citet{sun2024flatquant}, \Cref{tab:res_weight} from
\citet{tseng2024quipsharp}, \Cref{tab:res_fp4} from \citet{chen2025wush}, and \Cref{tab:res_kv} from \citet{hooper2024kvquant}; \Cref{tab:res_families} stitches two such papers, and its bolded winner is a row of one of them. Authors tune their own method under their
own protocol, so in each table the margin between the top row and the baselines beneath it is
the quantity most likely to be generous. The orderings we read off are the orderings those
papers report, and we flag the places below where a within-table margin carries weight rather
than treating any of them as protocol-neutral.

What survives this discipline is not a leaderboard but a set of
\emph{shapes}, and those shapes are consistent with the theory, which is as much as a transcribed record can establish. Two regimes the preceding sections
cover, diffusion transformers and quantized training, appear there through inline published
numbers (SVDQuant's near-\texttt{BF16} FID on \texttt{FLUX.1-dev}, and the \texttt{NVFP4} $12$B
pretraining run's $76.57$ MMLU against an \texttt{FP8} baseline's $77.36$) rather than in a
dedicated table here. Each of the two rests on a single headline result without a matched multi-method
comparison to tabulate.

\subsection{Activation quantization: climbing the transform ladder}
\label{sec:emp-w4a4}

\begin{figure}[t]
\centering
\includegraphics[width=\linewidth]{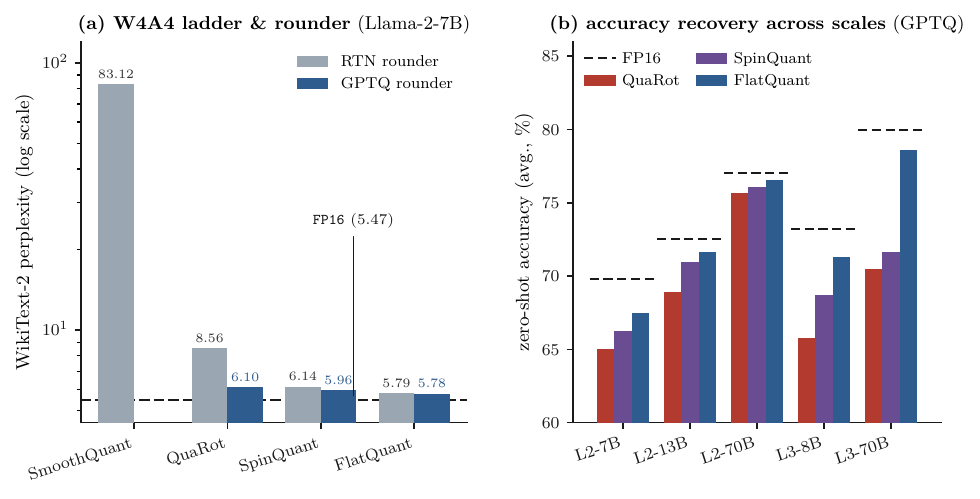}
\caption[The W4A4 transform ladder]{\textbf{Activation quantization (\texttt{W4A4},
\texttt{INT}): the transform ladder.} All numbers from \citet{sun2024flatquant}
(Tables~1--2). \textbf{(a)} Llama-2-7B WikiText-2 perplexity (log scale) climbing the degrees-of-freedom ladder, diagonal (SmoothQuant) $\to$ fixed rotation (QuaRot) $\to$ learned rotation (SpinQuant) $\to$ learned affine (FlatQuant),
each with the RTN and GPTQ rounder where reported. \textbf{(b)} The same ordering among the three
transform rungs (the diagonal rung is not reported here) in zero-shot accuracy across
five models (GPTQ rounder), against each model's \texttt{FP16} reference (dashed, as in (a)).}
\label{fig:res_w4a4}
\end{figure}

The taxonomy sorts transforms into six families (\Cref{fig:taxonomy}); the natural
first question is how they compare head to head. \Cref{tab:res_families} places a representative of each at \texttt{W4A4}
(Llama-2-7B WikiText-2 perplexity, across the two stitched protocols noted above), and the picture is stark. A pure diagonal scaling
(SmoothQuant) leaves perplexity at $83.12$, off the chart. By contrast, a permutation with
outlier handling (Atom, \citealp{zhao2023atom}), a fixed rotation (QuaRot), a learned
rotation (SpinQuant), a non-orthogonal affine map (FlatQuant), and an off-axis prefix
(PrefixQuant, \citealp{chen2024prefixquant}) all land between $5.78$ and $6.19$,
within about $0.7$ of the $5.47$ \texttt{FP16} floor. (A rotation-plus-permutation method,
DuQuant, \citealp{lin2024duquant}, as reproduced by \citet{chen2024prefixquant}, sits just above them at $6.20$; DuQuant's own paper reports $6.28$; the two come from different setups, \citet{chen2024prefixquant} re-running the baselines under its own \texttt{W4A4KV4} protocol, so they are not a like-for-like pair.) Two lessons follow.
First, \emph{some} sufficient transform is essential and diagonal scaling alone is not
it. Second, once past that bar several families are viable and close, so the choice
among them turns on the secondary axes (rounder, format, granularity, and cost)
that the rest of this section isolates. (The two families that compose rather than
stand alone, permutation and the off-axis prefix, are shown by their strongest members;
\Cref{tab:res_families} states the caveat.) The four families that one controlled protocol spans are compared there as whole published pipelines rather than as transforms in isolation, so their order is not a degrees-of-freedom ablation; that reading rests on the single-source ladder of \Cref{fig:res_w4a4}, read with the component-ablation caveat below.

\begin{table}[!htbp]
\centering\small
\renewcommand{\arraystretch}{1.15}
\caption{\textbf{The six transform families across two \texttt{W4A4} protocols.} A representative
method from each family of the taxonomy (\Cref{fig:taxonomy}) on Llama-2-7B WikiText-2
perplexity ($\downarrow$). Rows marked $\dagger$ are from \citet{sun2024flatquant}
(Table~1); the rest from \citet{chen2024prefixquant} (Table~2; \texttt{W4A4KV4}, context 2048, per-token-dynamic activations except Atom, which is group-wise at $g{=}128$ in both operands and adds mixed precision). The two protocols agree to within
$1.5\%$ on the methods they share, QuaRot ($6.10^{\dagger}$ vs $6.19$) and SpinQuant
($5.96^{\dagger}$ vs $5.95$), so the rows can be read together for the family-level contrast they are used for here, though the two protocols are not identical and differ in calibration corpus, and SpinQuant is the one row the source grays as trained on the evaluation corpus, so the $0.01$ inversion there is not independent evidence for the stitch. Every value
verbatim. The $\dagger$ rows come from that source's own rounder blocks (SmoothQuant is
reported with RTN only, FlatQuant with GPTQ); the \citet{chen2024prefixquant} rows carry that paper's Atom and QuaRot entries from \citet{lin2024qserve} and its own reproductions for the rest. The diagonal-versus-rotation verdict is not a rounder
artifact: under matched RTN in the same source it is SmoothQuant $83.12$ against QuaRot
$8.56$ (\Cref{sec:emp-w4a4}). Pure diagonal scaling alone fails; every other family, at its best, lands
within $\sim\!0.7$ perplexity of \texttt{FP16}. (Permutation and sequence/frame maps usually
\emph{compose} with the others rather than run alone; Atom, e.g., composes its permutation with mixed precision.) Best low-bit entry in \textbf{bold}.}
\label{tab:res_families}
\begin{tabular}{@{}ll r@{}}
\toprule
\textbf{Transform family} & \textbf{Representative} & \textbf{W4A4 PPL} \\
\midrule
Diagonal (rebalance)          & SmoothQuant$^{\dagger}$ & 83.12 \\
Permutation (regroup)         & Atom                    & 6.12 \\
Fixed orthogonal (Hadamard)   & QuaRot                  & 6.19 \\
Learned orthogonal            & SpinQuant               & 5.95 \\
Non-orthogonal affine         & FlatQuant$^{\dagger}$   & \textbf{5.78} \\
Sequence \& frame             & PrefixQuant             & 5.93 \\
\midrule
\texttt{FP16} reference       & --                      & 5.47 \\
\bottomrule
\end{tabular}
\end{table}

\begin{table}[!htbp]
\centering\small
\renewcommand{\arraystretch}{1.18}
\caption{\textbf{Activation quantization (\texttt{W4A4}, \texttt{INT}): the transform
ladder, quantified.} All values transcribed verbatim from \citet{sun2024flatquant}
(Tables~1--2): one consistent protocol (4-bit weights, activations \emph{and} KV cache,
integer grid, GPTQ rounder for the rotation/affine methods; SmoothQuant uses RTN).
WikiText-2 perplexity ($\downarrow$) and zero-shot average accuracy ($\uparrow$, \%).
Reading down each block climbs the degrees-of-freedom ladder of \Cref{sec:taxonomy}
(diagonal $\to$ fixed rotation $\to$ learned rotation $\to$ learned affine); the best transform recovers to within $2.4$ accuracy points of the \texttt{FP16} reference, and $0.22$--$0.91$ perplexity of it. Best
low-bit entry per column in \textbf{bold}.}
\label{tab:res_act}
\begin{tabular}{@{}l rrrrr@{}}
\toprule
\textbf{Method (family)} & \textbf{L2-7B} & \textbf{L2-13B} & \textbf{L2-70B} & \textbf{L3-8B} & \textbf{L3-70B} \\
\midrule
\multicolumn{6}{@{}l}{\emph{WikiText-2 perplexity} $\downarrow$}\\
\texttt{FP16} (reference)        & 5.47  & 4.88  & 3.32  & 6.14   & 2.86 \\
SmoothQuant (diagonal)           & 83.12 & 35.88 & 26.01 & 210.19 & 9.60 \\
QuaRot (fixed rotation)          & 6.10  & 5.40  & 3.79  & 8.16   & 6.60 \\
SpinQuant (learned rotation)     & 5.96  & 5.24  & 3.70  & 7.39   & 6.21 \\
FlatQuant (learned affine)       & \textbf{5.78} & \textbf{5.11} & \textbf{3.54} & \textbf{6.90} & \textbf{3.77} \\
\midrule
\multicolumn{6}{@{}l}{\emph{Zero-shot average accuracy} $\uparrow$}\\
\texttt{FP16} (reference)        & 69.79 & 72.55 & 77.05 & 73.23 & 79.95 \\
QuaRot                           & 65.01 & 68.91 & 75.68 & 65.79 & 70.45 \\
SpinQuant                        & 66.23 & 70.93 & 76.06 & 68.70 & 71.66 \\
FlatQuant                        & \textbf{67.47} & \textbf{71.64} & \textbf{76.53} & \textbf{71.33} & \textbf{78.58} \\
\bottomrule
\end{tabular}
\end{table}

The regime the transforms were built for is \texttt{W4A4} on the integer grid, where
the activation crest factor is the wall (\Cref{sec:inversion}). \Cref{fig:res_w4a4}
and \Cref{tab:res_act} read the transform ladder of \Cref{sec:taxonomy} straight off
\citet{sun2024flatquant}. On Llama-2-7B the ladder starts from the same diagonal-scaling collapse \Cref{tab:res_families} showed:
SmoothQuant at $83.12$ against the $5.47$ \texttt{FP16} floor, because the shared-scale integer grid is destroyed by the outliers a diagonal map only relocates (\Cref{prop:flat-opt}). Each added degree of freedom is then accompanied by a smaller gap: a fixed
Hadamard rotation (QuaRot, \citealp{ashkboos2024quarot}) reaches $6.10$, a learned
rotation (SpinQuant, \citealp{liu2024spinquant}) $5.96$, and a learned affine map
(FlatQuant) $5.78$, within half a perplexity point of \texttt{FP16}.
That ordering is a ranking of published methods rather than a measurement of the mechanism, and the source's own component ablation marks the difference. On \texttt{Llama-3-8B}, and at the \texttt{RTN} rounder rather than the \texttt{GPTQ} one the ladder's rotation and affine rungs quote, \citet{sun2024flatquant} decompose their result into the learned transform alone at $8.50$ WikiText-2 perplexity, $7.95$ once a per-channel scaling is added, and $6.98$ once learnable clipping thresholds are added on top, from $1266.60$ with none of the three. That is a different model and a different rounder from the ladder, so we read the ablation for its internal decomposition and not against the ladder's absolute numbers. Part of the affine family's margin at the top of the ladder therefore comes from fitting a clip range rather than from reshaping the distribution, and a fitted clip threshold moves error into the overload term \Cref{lem:bennett} sets aside, where the crest factor becomes a proxy rather than the exact charge. We read the ladder as the ordering the published numbers support, and do not attribute the whole of the top rung's advantage to the transform's extra degrees of freedom.
The learned diagonal and affine baselines that \citet{sun2024flatquant} also report
(OmniQuant~\citep{shao2023omniquant} at $14.74$ and AffineQuant~\citep{ma2024affinequant}
at $12.69$ on \texttt{Llama-2-7B}) land between the diagonal collapse and the rotations. AffineQuant, though, trails the
rotations rather than topping them as its family's ceiling FlatQuant does, a gap in realized degrees of freedom, since it trails QuaRot $12.69$ to $8.56$ at matched \texttt{RTN}, rather than in the family's nominal expressiveness, the diagonal-dominance constraint itself limiting how far the map can depart from a diagonal one. The ordering is
not a Llama-2-7B artifact. It holds across five models and two model families (Llama-2 and Llama-3)
(\Cref{tab:res_act}), and the learned affine map's advantage \emph{widens} at the hard
end. On Llama-3-70B it recovers a zero-shot accuracy of $78.6\%$ against QuaRot's
$70.4\%$ and an \texttt{FP16} reference of $80.0\%$.

The same figure exposes the rounding confound of \Cref{sec:pitfalls} as a measurable
quantity. Each rotation and affine method appears twice, once with round-to-nearest and once with GPTQ
error feedback (SmoothQuant is reported with RTN only). The gap between the two rounders shrinks monotonically as the transform
improves. QuaRot moves from $8.56$ (RTN) to $6.10$ (GPTQ), a swing of $2.5$ perplexity,
while FlatQuant moves only from $5.79$ to $5.78$. That monotone shrinkage makes visible the transform--rounder overlap of \Cref{sec:composition}, in which a strong transform flattens the
distribution the rounder would otherwise have to correct, so the better the transform,
the less error feedback is left to buy. The shrinking gap is also a warning. A rotation evaluated
against RTN and compared to a GPTQ baseline would misattribute the rounder's work to
the transform.

\subsection{Weight-only quantization: the low-bit regime}
\label{sec:emp-weightonly}

\begin{table}[!tbp]
\centering\small
\renewcommand{\arraystretch}{1.18}
\caption{\textbf{Weight-only quantization: the low-bit regime is where the codebook and fine-tuning pay off, on top of an RHT-plus-scalar-LDLQ baseline doing most of the work.} WikiText-2 perplexity ($\downarrow$), transcribed verbatim from
\citet{tseng2024quipsharp} (Table~2; Llama-2, context length 2048, activations
\texttt{FP16}). At 4 bits every method is near \texttt{FP16}; as the budget drops the
scalar methods (AWQ, OmniQuant) diverge, while QuIP\#'s incoherence rotation plus
lattice codebook (\Cref{sec:vq}), together with the method's inter-layer fine-tuning,
stays close; at 2 bits it stays within $1.2$ perplexity of \texttt{FP16}, close to 3-bit OmniQuant on one fewer bit, where the
(per-channel) scalar baselines diverge, a raw gap a finer group size would narrow but the
codebook's space-filling advantage sustains. ``--'' = not reported. Best per column within each bit-width block, excluding the \texttt{FP16} reference, in \textbf{bold}.}
\label{tab:res_weight}
\begin{tabular}{@{}ll rrr@{}}
\toprule
\textbf{Bits} & \textbf{Method} & \textbf{L2-7B} & \textbf{L2-13B} & \textbf{L2-70B} \\
\midrule
16 & \texttt{FP16} (reference) & 5.47 & 4.88 & 3.32 \\
\midrule
\multirow{3}{*}{4}
   & AWQ        & 6.15 & 5.12 & --  \\
   & OmniQuant  & 5.74 & 5.02 & 3.47 \\
   & QuIP\#     & \textbf{5.56} & \textbf{4.95} & \textbf{3.38} \\
\midrule
\multirow{3}{*}{3}
   & AWQ        & 24.0 & 10.5 & --  \\
   & OmniQuant  & 6.58 & 5.58 & 3.92 \\
   & QuIP\#     & \textbf{5.79} & \textbf{5.10} & \textbf{3.56} \\
\midrule
\multirow{2}{*}{2}
   & OmniQuant  & 37.4 & 17.2 & 7.81 \\
   & QuIP\#     & \textbf{6.66} & \textbf{5.74} & \textbf{4.16} \\
\bottomrule
\end{tabular}
\end{table}

With activations kept in high precision, the weight distribution alone matters, and the picture inverts with the bit budget (\Cref{tab:res_weight},
from \citet{tseng2024quipsharp}). At 4-bit weight-only every method is within a
fraction of a point of \texttt{FP16} (Llama-2-7B sits at $5.56$--$6.15$ against $5.47$). That near-tie
is why a rotation buys little here and per-channel scaling with error feedback is
usually enough (\Cref{tab:guide}). The transform earns its place only as the budget
tightens. At 3 bits AWQ's~\citep{lin2023awq} scalar scaling jumps to $24.0$ and at 2 bits OmniQuant reaches
$37.4$. QuIP\#, which composes an incoherence rotation with a lattice codebook
(\Cref{sec:vq}) and inter-layer fine-tuning, holds at $6.66$, close to 3-bit OmniQuant's $6.58$ on one fewer bit and within $1.2$ perplexity of \texttt{FP16}, where the per-channel scalar baselines diverge (those baselines are reported groupless, so a finer group size would
narrow the raw gap). The source's own ablation without that inter-layer fine-tuning and the $E_8$ codebook reads $12.3$, still far below the scalar baselines. That row isolates nothing on its own, since it drops the codebook and the fine-tuning together, so most of the margin over the scalar baselines is already bought by the incoherence rotation and scalar LDLQ error feedback, the ablation rounding to a one-dimensional half-integer grid. The vector codebook and the fine-tuning are jointly what the extreme-compression regime rewards, on top of an \textsc{RHT}-plus-scalar-\textsc{LDLQ} baseline that is doing most of the work. The comparison is read at nominal
bit-width. A fully fair accounting would add each method's metadata, QuIP\#'s codebook and its \texttt{FP16} sign vectors among them (the source prices the sign vectors at under $0.01$ bits per weight), to the effective rate, as \Cref{sec:pitfalls}
insists. At these gaps, though, the ranking is robust to that correction.

\subsection{The format flip}
\label{sec:emp-flip}

\begin{table}[!htbp]
\centering\small
\renewcommand{\arraystretch}{1.18}
\caption{\textbf{The format flip: on \texttt{FP4} grids the near-essential-for-\texttt{INT4}
rotation reverses sign.} \texttt{W4A4} accuracy (\%, $\uparrow$) on
Llama-3.1-8B-Instruct: seven of the nine method rows of \citet{chen2025wush}'s
Table~2, transcribed verbatim (\textbf{Avg.}\ as printed by the source, which can differ from the mean of its four task columns by up to $0.05$). The two omitted rows read SmoothQuant $75.70$\,/\,$70.30$
 and RTN-WUSH $75.28$\,/\,$73.21$ (\texttt{NVFP4}\,/\,\texttt{MXFP4} averages); the second is
 that source's own headline for pairing its transform with plain RTN, and the first is the source's
 SmoothQuant baseline, notable for how much better a pure diagonal scaling fares here than
 the same family fares in the \texttt{INT4} literature (\Cref{tab:res_families}), though on a
 different model, metric and scale granularity, so that contrast is suggestive rather than
 a matched instance of the flip.
Row labels are shortened from the source (RTN $=$ RTN-I, RTN\,$+$\,block-H $=$ RTN-H,
GPTQ $=$ GPTQ-I, MR-GPTQ $=$ GPTQ-H, WUSH $=$ GPTQ-WUSH). Two
floating-point formats: \texttt{NVFP4} (block 16, \texttt{FP8} scale) and
\texttt{MXFP4} (block 32, power-of-two \texttt{E8M0} scale). Rows are grouped by \emph{rounder}, QuaRot's taken from the setup paper below and the one row whose rounder no source states placed last, so each transform reads against the no-transform baseline at matched rounding (\Cref{sec:pitfalls}): a Hadamard \emph{confined to the scale block}
helps \texttt{MXFP4} under both rounders ($69.32\!\to\!70.45$ with RTN,
$70.62\!\to\!73.65$ with GPTQ) and is neutral-to-mildly-harmful on \texttt{NVFP4}
($74.73\!\to\!74.05$, $75.72\!\to\!75.84$). The \emph{global} Hadamard
(QuaRot) that is near-essential on the integer grid (\Cref{tab:res_act}) instead
\emph{craters} \texttt{MXFP4}, $6.4$ points below plain per-block RTN and $7.7$ below GPTQ. The AbsMax
surrogate does not account for that contrast between a block-confined and a global rotation:
it puts both families at the same \emph{ideal} floor, and neither map's actual residual was
computed on the published models; \Cref{sec:fmt-flip} states what is and is not
explained.
The source states that for its block-transform rows ``the transform block size is
always the same as the quantization group size,'' a condition it notes is ``not
applicable to SmoothQuant, QuaRot, and SpinQuant'' (hence global); it does not state
the rounder used for the SpinQuant row; for QuaRot, the setup paper whose protocol it adopts
specifies round-to-nearest after the rotation \citep{egiazarian2025mrgptq}. The adaptive whitening transform (WUSH, \Cref{thm:wush}) reports the best average in both formats, though \citet{chen2025wush} describe their \texttt{NVFP4} improvements as often within run-to-run variability, on a different model and benchmark suite, so the format-level contrast rather than the within-format ranking is what these averages settle. Best per column \emph{within each format block}, excluding the \texttt{BF16} reference
and the two source rows omitted above, in \textbf{bold}.}
\label{tab:res_fp4}
\begin{tabular}{@{}ll rrrr r@{}}
\toprule
\textbf{Format} & \textbf{Method} & \textbf{MMLU-CoT} & \textbf{GSM8K} & \textbf{HellaSw.} & \textbf{WinoG.} & \textbf{Avg.} \\
\midrule
\texttt{BF16} & (reference) & 72.76 & 85.06 & 80.01 & 77.90 & 78.93 \\
\midrule
\multirow{7}{*}{\texttt{NVFP4}}
 & RTN                  & 68.26 & 78.39 & 78.15 & 74.11 & 74.73 \\
 & RTN $+$ block-H      & 67.41 & 78.01 & 77.31 & 73.48 & 74.05 \\
 & QuaRot (global)      & 66.50 & 77.40 & 77.25 & 75.14 & 74.10 \\
 & GPTQ                 & 68.85 & \textbf{81.25} & 78.26 & 74.51 & 75.72 \\
 & MR-GPTQ              & 69.12 & 80.80 & 78.17 & 75.24 & 75.84 \\
 & WUSH                 & \textbf{69.69} & 80.11 & \textbf{78.52} & \textbf{76.09} & \textbf{76.10} \\
 & SpinQuant (global)   & 66.50 & 76.10 & 76.96 & 75.32 & 73.70 \\
\midrule
\multirow{7}{*}{\texttt{MXFP4}}
 & RTN                  & 62.21 & 67.85 & 73.99 & 73.24 & 69.32 \\
 & RTN $+$ block-H      & 62.38 & 72.48 & 75.29 & 71.67 & 70.45 \\
 & QuaRot (global)      & 49.86 & 56.94 & 73.50 & 71.43 & 62.90 \\
 & GPTQ                 & 63.49 & 68.46 & 76.01 & 74.51 & 70.62 \\
 & MR-GPTQ              & 67.19 & 75.70 & 76.91 & \textbf{74.80} & 73.65 \\
 & WUSH                 & \textbf{67.79} & \textbf{77.41} & \textbf{77.44} & 74.78 & \textbf{74.35} \\
 & SpinQuant (global)   & 61.80 & 68.16 & 74.87 & 72.93 & 69.40 \\
\bottomrule
\end{tabular}
\end{table}

The most consequential shape is the one that reverses. \Cref{tab:res_fp4} reads \citet{chen2025wush}'s \texttt{W4A4} accuracies on
Llama-3.1-8B-Instruct for two floating-point formats. On \texttt{MXFP4}, whose block
scale is a power-of-two \texttt{E8M0} value, the global Hadamard that is
\emph{near-essential} for the integer grid becomes actively harmful: QuaRot falls to
$62.9\%$ average accuracy, \emph{below} plain per-block RTN at $69.3\%$. The contrast is measured, not explained (\Cref{sec:fmt-flip,sec:open}). The same table makes it visible that the fault lies with the
\emph{global, fixed} rotation and not with rotation as such. A Hadamard confined to the
scale block \emph{helps} \texttt{MXFP4} under either
rounder ($69.3\!\to\!70.5$ with RTN, $70.6\!\to\!73.7$ with GPTQ).

Move to \texttt{NVFP4}, whose \texttt{FP8} block scale
carries a mantissa, and the penalty all but vanishes. QuaRot recovers to $74.1\%$,
just below RTN's $74.7\%$, while the block-confined Hadamard is roughly neutral there, mildly harmful under RTN ($74.73\!\to\!74.05$) and mildly helpful under GPTQ ($75.72\!\to\!75.84$). Across both formats the adaptive whitening of WUSH
(\Cref{thm:wush}) reports the best average, at $74.35\%$ (\texttt{MXFP4}) and $76.1\%$
(\texttt{NVFP4}), with the block-Hadamard MR-GPTQ~\citep{egiazarian2025mrgptq} at
$73.7\%$ and $75.8\%$. Every row in \Cref{tab:res_fp4} is transcribed from \citet{chen2025wush}'s own table. The WUSH rows are that paper's; the baseline rows match \citet{egiazarian2025mrgptq}'s table cell for cell, one \texttt{GSM8K} entry excepted, and the source states it uses that paper's setup, so the QuaRot and SpinQuant rows are not those papers' self-reported numbers; the MR-GPTQ row is its own authors', carried over under the same setup. The transforms that respect the format
(block-confined, scale-aware) win where the global rotation designed for integers does
not. One and the same
map is the right answer and the wrong answer depending on the grid beneath it, and that
single reversal is why the taxonomy's format axis is not a detail. An
independent single-protocol evaluation is consistent with the weaker half of this
picture, finding that rotation strategies effective for \texttt{INT4} offer only limited
gains on \texttt{MXFP4} and \texttt{NVFP4} \citep{liu2025compeval}. The sign reversal
itself rests on \citet{chen2025wush}'s table alone.

\subsection{Beyond the weight matrix: the KV cache}
\label{sec:emp-kv}

\begin{table}[!htbp]
\centering\small
\renewcommand{\arraystretch}{1.18}
\caption{\textbf{KV-cache quantization: axis choice and outlier retention.} WikiText-2
perplexity ($\downarrow$) on the first-generation LLaMA-7B with \emph{only} the KV cache quantized (weights
and activations \texttt{FP16}), transcribed verbatim from \citet{hooper2024kvquant}
(arXiv v6, Table~1). The sub-4-bit Atom and FlexGen entries are
\citet{hooper2024kvquant}'s re-implementations of those 4-bit methods extended to 3 and
2 bits; the Atom and FlexGen rows quantize per-token post-RoPE with grouped uniform scales, so their edge over the naive row is finer granularity, not axis choice.
At 4-bit every scheme is near \texttt{FP16}; at 2-bit a naive per-token
integer scale explodes, because the keys carry per-channel outliers that RoPE mixes
across positions (\Cref{sec:beyond}), whereas KVQuant's pre-RoPE per-channel key layout,
non-uniform codebook, attention-sink-aware quantization (the first token kept in
\texttt{FP16}), and $1\%$ full-precision outlier retention hold within about a third of a perplexity point. The source's groupless NormalFloat rows (nf4 $5.87$, nf3 $7.33$, nf2 $3210$) are omitted for space. Best low-bit entry per column in \textbf{bold}.}
\label{tab:res_kv}
\begin{tabular}{@{}l rrr@{}}
\toprule
\textbf{Method} & \textbf{KV4} & \textbf{KV3} & \textbf{KV2} \\
\midrule
\texttt{FP16} (reference) & \multicolumn{3}{c}{5.68} \\
\midrule
INT (naive, per-token)  & 5.98 & 10.87 & 11779\phantom{.00} \\
Atom                    & 5.77 & 6.17  & 37.37 \\
FlexGen                 & 5.73 & 5.93  & 11.09 \\
KVQuant                 & 5.72 & 5.87  & 7.23 \\
KVQuant $+1\%$ outliers & \textbf{5.69} & \textbf{5.75} & \textbf{6.01} \\
\bottomrule
\end{tabular}
\end{table}

Neither attention matmul has a static operand (\Cref{sec:beyond}), so the cache is quantized directly rather than through a weight. The levers the published evidence isolates here are the choice of axis and the treatment of outliers, with a rotation available, folded on the value path but paid online on the keys, where RoPE blocks the fold. \Cref{tab:res_kv} reads \citet{hooper2024kvquant}. At 4-bit KV the grouped and per-channel schemes sit within about a tenth of the
\texttt{FP16} $5.68$ (the naive per-token integer scale is already $0.30$ higher). The regime separates decisively at 2 bits, where a naive per-token
integer scale reaches a perplexity of $11{,}779$. The reason is that the keys carry fixed per-channel
outliers that a per-token scale cannot see, and RoPE smears them across positions. The
finer-grouped baselines recover partially (FlexGen to $11.09$, Atom~\citep{zhao2023atom} to $37.37$,
both as re-implemented and extended below 4 bits by \citet{hooper2024kvquant}). By contrast, the combination KVQuant deploys (per-channel keys quantized pre-RoPE, a non-uniform,
nonlinear codebook, attention-sink-aware quantization of the first token, and $1\%$ of entries kept in full precision) lands at $6.01$, within about a third of a point of \texttt{FP16}. The table separates the last of those ingredients from the rest: without the retained outliers the same scheme sits at $7.23$, so the axis, the codebook and the attention-sink handling carry the fall from $11{,}779$, and outlier retention supplies the final $1.22$. The shape echoes the weight-only story of
\Cref{sec:emp-weightonly}: at moderate bit-widths the details barely matter, and the non-uniform codebook with outlier retention pays for its lookup only at the aggressive end.

\subsection{Is the transform worth it? Speed and memory}
\label{sec:emp-cost}

Accuracy is only half the case for a transform. The reason to quantize at all is speed
and memory, and a transform that is not free must earn its overhead. The published
efficiency numbers make the trade concrete. On the payoff side,
QuaRot's \texttt{INT4} kernels reach, over an \texttt{FP16} baseline, prefill
speed-ups of up to $2.16\times$ on Llama-2-7B (from $1.97\times$ at batch~1) rising to
$3.33\times$ on the 70B model, with decoding-stage peak-memory reductions the authors summarize as at least $3.63\times$, the best factors being $3.75\times$ at 7B and $3.89\times$ in almost all 70B cases, from the low-bit weights and KV cache
(\citealp{ashkboos2024quarot}, RTX 3090; prefill at sequence length $2048$ over batch sizes $1$--$64$, memory over prefill lengths $256$--$4096$ at batch~16; both figures are measured on a \emph{single} transformer block, the whole model not fitting on the authors' cluster at large batch sizes. They expect the \emph{memory} factors to grow with depth, since the effect of constant-size objects becomes less significant as layers are added, and state no whole-model expectation either way for the speed-ups). On the cost side, the
foldable transforms are free by construction: an orthogonal map of the residual stream absorbs by
computational invariance (\Cref{thm:invariance}, the norm gains folded out first), a diagonal scaling by folding
(\Cref{prop:gamma}). So only the \emph{online}, non-absorbable maps carry a runtime
charge, and that charge is small. \citet{sun2024flatquant} measure the end-to-end slowdown of
their five fused affine transforms at $0.07\times$, against $0.26\times$ for QuaRot's
three online Hadamards, both measured by those authors on Llama-2-7B (RTX~3090, batch~64, prefill). Both systems still reach prefill speed-ups near or above $2\times$ (FlatQuant reports up to $2.30\times$ prefill and $1.76\times$ decoding at batch~64). This is the quantitative content of the claim that opened \Cref{sec:systems}: the online cost of a fused transform has fallen to under a tenth of end-to-end runtime ($0.07\times$), while three separate online Hadamards still cost nearly four times that ($0.26\times$). It is what makes the accuracy gains of the previous four subsections worth collecting.

Those two systems are representative; \Cref{tab:res_systems_eff} collects
the reported gains across the wider systems literature. The metrics and baselines differ
enough that the factors cannot be lined up directly (a latency speed-up, a
throughput-at-fixed-latency ratio, and a kernel micro-benchmark are three different
measurements). The direction, though, is uniform. On the integer \texttt{W4A4} path, QUIK
reports up to $3.4\times$ end-to-end over \texttt{FP16} on Llama-2-70B
\citep{ashkboos2023quik}, and Atom, measuring serving throughput at a fixed latency
target, up to $7.7\times$ \citep{zhao2023atom}. On the \texttt{W4A8KV4} path, QServe
reports $1.2$--$3.5\times$ throughput, though over a production TensorRT-LLM server
rather than raw \texttt{FP16} \citep{lin2024qserve}. The kernels that carry the transforms
are themselves fast: HadaCore's tensor-core Hadamard runs $1.1$--$1.4\times$ on an A100 (peak $3.5\times$, and $1.0$--$1.3\times$ on an H100) over the previous best implementation \citep{agarwal2024hadacore}.
FLUTE's lookup-table GEMM, which serves the scalar lookup-table codebooks of \Cref{sec:vq},
runs $2$--$4\times$ over an \texttt{FP16} GEMM at batch sizes below $32$ and group size $128$ \citep{guo2024flute}. The table's lesson matches those two systems' at larger scale: the transform stage is no longer the bottleneck, and low-bit
inference is a multiplicative rather than marginal win.

\begin{table}[!htbp]
\centering\small
\renewcommand{\arraystretch}{1.15}
\caption{\textbf{Reported efficiency of low-bit inference systems.} Each factor is
transcribed verbatim from the cited paper; the metrics and baselines \emph{differ}
(throughput vs.\ latency; \texttt{FP16} vs.\ a production TensorRT-LLM server vs.\ a
prior kernel), so the numbers state each system's own claim and are not a controlled
cross-comparison. $^{\P}$HadaCore's peaks are $3.5\times$ (A100) and $3.6\times$ (H100). $^{\S}$QUIK's $3.4\times$ keeps the down-projections at 8 bits. $^{\ddagger}$FlatQuant's figures are at batch~64. $^{\|}$FLUTE's $2$--$4\times$ is at batch sizes below $32$ and group size $128$. $^{\dagger}$QuaRot's factors are kernel-level: the prefill figures are
upper ends of a batch sweep and both are measured on a \emph{single} transformer block
rather than the full model. The direction is nonetheless uniform: low-bit inference buys a
\emph{multiplicative} speed and memory gain, and the transforms and codebooks that
enable it are themselves fast. Throughput measured at a fixed latency target (Atom)
yields larger factors than a latency speed-up.}
\label{tab:res_systems_eff}
\setlength{\tabcolsep}{3pt}
\begin{tabular}{@{}l l l l@{}}
\toprule
\textbf{System} & \textbf{Regime} & \textbf{Reported gain (metric)} & \textbf{Baseline} \\
\midrule
QuaRot \citep{ashkboos2024quarot}    & \texttt{W4A4}     & $2.16$--$3.33\times$ prefill; ${\ge}3.6\times$ memory$^{\dagger}$   & \texttt{FP16} \\
FlatQuant \citep{sun2024flatquant}   & \texttt{W4A4}     & up to $2.30\times$ prefill; $1.76\times$ decode$^{\ddagger}$        & \texttt{FP16} \\
QUIK \citep{ashkboos2023quik}        & \texttt{W4A4}$^{\S}$ & up to $3.4\times$ end-to-end                     & \texttt{FP16} \\
Atom \citep{zhao2023atom}            & \texttt{W4A4}     & up to $7.7\times$ throughput (fixed latency)     & \texttt{FP16} \\
QServe \citep{lin2024qserve}         & \texttt{W4A8KV4}  & $1.2$--$3.5\times$ throughput                    & TensorRT-LLM \\
HadaCore \citep{agarwal2024hadacore} & Hadamard kernel   & $1.1$--$1.4\times$ A100; $1.0$--$1.3\times$ H100$^{\P}$ & prior FWHT kernel \\
FLUTE \citep{guo2024flute}           & LUT (weight-only) & $2$--$4\times$ GEMM$^{\|}$; $1.5$--$2\times$ end-to-end  & \texttt{FP16} GEMM \\
\bottomrule
\end{tabular}
\end{table}

\medskip
\noindent Read together, these blocks are the survey's thesis in numbers, in the weak sense of a
record consistent with it rather than a controlled test of it: on the integer grid,
the attainable published pipelines order themselves by the transform's degrees of freedom, an
ordering the single-source ladder supports, with that source's own component ablation showing
that part of the top rung's margin comes from clipping rather than from reshaping
(\Cref{sec:emp-w4a4}). We stop short of a proportionality to flatness, since no
per-method flatness statistic accompanies the published ladder. The scalar baselines diverge from three bits down, where a rotation with error feedback, and a lattice codebook and fine-tuning on top, keeps the loss bounded (\Cref{sec:emp-weightonly}); the sign of the rotation's
benefit flips with the number format (\Cref{sec:emp-flip}). On the KV cache, where
the key-path rotation cannot fold, axis choice and outlier retention play the role the transform
plays elsewhere in the results we transcribe (\Cref{sec:emp-kv}), a rotation regaining that role
once it is made data-aware (\Cref{sec:beyond-kv}). The whole enterprise pays off at deployment:
prefill speed-ups near or above $2\times$, and, in the one system reporting a decode-stage figure, $1.76\times$ at decode (\Cref{tab:res_systems_eff}), once the online transform is fused down to under a tenth of runtime (\Cref{sec:emp-cost}). None of
this is a normalized leaderboard, and it is not offered as one. The blocks above are the published
record, read under the one discipline (one protocol per comparison) that keeps
such a reading honest.

\section{Evaluation Pitfalls and Practical Guidance}\label{sec:practice}

The numbers of \Cref{sec:empirical} were read under a one-protocol-per-comparison discipline. This section states that discipline in general form and then turns to using the transforms as well as evaluating them. We first collect the evaluation pitfalls that recur in the literature and repeatedly produce misleading comparisons (\Cref{sec:pitfalls}), then distill the survey into a which-transform-when guide indexed by deployment scenario (\Cref{sec:guide}).

\subsection{Evaluation pitfalls}
\label{sec:pitfalls}

The single most consequential mistake is to optimize a surrogate the deployment does not, a failure that follows directly from the inversion (\Cref{thm:inversion}). Two surrogate gaps recur. First, a transform is often tuned against a \emph{pre-quantization} statistic (a group's RMS, or the $\ell_2$ expression $\|X\widehat W^{\!\top} - X W^{\!\top}\|$ evaluated as if the scale were free) rather than the realized error of the deployed AbsMax quantizer, whose group step is pinned by the group maximum (an $\ell_\infty$ statistic) and whose per-group penalty is the crest factor $M_g/\sigma_g$. The two can be \emph{anti-correlated}, because a concentrating transform lowers the variable-rate distortion of \Cref{prop:schur}(i), the geometric mean the classical proxy rewards, while raising the group maxima the shared scale must clear (\Cref{prop:schur}(ii)). Second, even the \emph{exact} layer-output error is only a first-order predictor of end-to-end loss, and can rank transforms in the opposite order at aggressive bit-widths; \Cref{thm:linearity} gives the corresponding statement for the per-layer \emph{weight} error, and its authors are explicit that it has no direct bearing on the data-aware layer-wise objective. Any claim about a transform must therefore be validated on the deployed metric (perplexity or downstream accuracy at the actual bit-width, group size, and scale rule), not on a surrogate the hardware does not optimize. (To be precise, $\|X\widehat W^{\!\top} - X W^{\!\top}\|$ computed with the actually quantized weights is a legitimate layer-output error; the traps are the pre-quantization or RMS-based stand-ins for it, and the leap from layer error to task loss.) The remaining pitfalls are, in effect, ways of accidentally changing the deployed setting between the method and its baseline.

The first is a confound of \emph{granularity}. A transform's apparent win can be an artifact of the quantization granularity it is silently compared under. Per-token activation scales, per-group weight scales, and per-tensor scales induce entirely different crest factors, and a rotation evaluated with per-token scales against a baseline with per-tensor scales measures the granularity, not the rotation. The group size directly sets the achievable crest factor, whose ceiling is $\sqrt G$ by \eqref{eq:crest}. Comparisons must fix the group size and the scale rule across all arms.

A second confound is the \emph{rounder}. As \Cref{sec:composition} showed, a transform empirically overlaps with the rounder, so a transform evaluated against round-to-nearest will look far stronger than the same transform evaluated against GPTQ, because part of what it fixes is what GPTQ's error feedback would have repaired anyway (quantified in \Cref{fig:res_w4a4}, where the RTN$\to$GPTQ gap shrinks sharply once a strong transform is in place). The rounder must be held fixed, and ideally be the strong one, across arms, or the comparison conflates transform quality with rounder weakness.

The \emph{number format} is a third. The integer--floating-point flip of \Cref{sec:format} means a result established for \texttt{INT4} does not transfer to \texttt{FP4} or microscaling. Concretely, the random Hadamard that is near-essential for integer AbsMax can be counterproductive on an \emph{ideal} floating-point element grid, and is helpful again (in block-confined, scale-aware form) on \texttt{MXFP4} because of its power-of-two block scale. Reporting an INT4 transform result and implying it carries to \texttt{MXFP4} is unsound; the format must be stated and matched. A fair comparison must also count \emph{effective} bits (the element width plus the amortized shared scale, any zero-point, outlier channels held in high precision, a codebook, and the transform's own stored parameters) since two methods nominally at ``4-bit'' can differ materially once this metadata is included.

Three smaller but common errors close the list, around calibration hygiene, cost honesty, and evaluation at short context alone. Genuine \emph{leakage} (calibrating or tuning on the very examples used for evaluation, or repeatedly selecting hyper-parameters against the test split) inflates reported accuracy and must be avoided; calibrating on a WikiText \emph{training} split and reporting perplexity on the disjoint \emph{test} split is not itself leakage. Because a transform fitted to calibration statistics can nonetheless carry that corpus's idiosyncrasies, a cross-corpus evaluation is the more informative robustness check even when no split is reused. And a transform's \emph{runtime} must be reported alongside its accuracy. ``Zero-cost'' is warranted only for a transform absorbed into adjacent weights: an orthogonal map of the residual stream by computational invariance (\Cref{thm:invariance}, which first requires the learned norm gains folded out), a diagonal scaling by folding (\Cref{prop:gamma}), or a full invertible map the surrounding architecture lets it absorb offline. A data-dependent, non-absorbable map, by contrast, incurs an online kernel whose cost, even at the $7\%$ that fused kernels now achieve (\Cref{sec:emp-cost}), belongs in the comparison. Finally, because the KV-cache, long-context, and multi-step reasoning behavior (\Cref{sec:beyond}) may degrade separately from, and more than, short-context perplexity, a method that quantizes the cache should be evaluated at length and on generation, not only on a short perplexity proxy.

\subsection{A which-transform-when guide}
\label{sec:guide}

\Cref{tab:guide} distills the survey into a first recommendation indexed by deployment scenario. The table is deliberately a starting point, not a verdict, and its evidentiary status warrants a plain statement. The entries synthesize results reported across different models, tasks, group sizes, calibration sets, and kernels rather than a single normalized head-to-head benchmark, which no published source provides at this coverage. \Cref{sec:empirical} gives the per-source head-to-head numbers where a single protocol makes the comparison fair; the entries here should be read as hypotheses to test first under one's own accuracy budget and available kernels. The rationale column names the governing result where a theorem exists (e.g.\ \Cref{prop:rht,thm:wush}) and otherwise the representative empirical finding it rests on. The through-line is the inversion: rotate to flatten when the grid is integer and the scale is shared; prefer a rotation confined to the scale block where that block scale is coarse (\texttt{MXFP4}'s power-of-two E8M0), and little or no rotation once the block scale carries a mantissa (\texttt{NVFP4}), the element grid pulling against a rotation and the coarse scale pulling for one; avoid rotating a bounded tensor at all, and avoid rotating \emph{along the feature axis} one whose row sparsity the quantizer should exploit instead, or whose sparsity only a row-axis map can reach.

{\small
\renewcommand{\arraystretch}{1.25}
\begin{longtable}{@{}>{\raggedright\arraybackslash}p{3.0cm} >{\raggedright\arraybackslash}p{3.4cm} >{\raggedright\arraybackslash}p{\dimexpr\linewidth-6.9cm-10pt\relax}@{}}
\caption{\textbf{A first-recommendation guide indexed by deployment scenario.} Entries name the transform family and representative methods to try first, with the governing result. The unifying rule is the inversion of \Cref{sec:inversion}, refined by the format flip of \Cref{sec:format}.}\label{tab:guide}\\
\toprule \textbf{Scenario} & \textbf{First choice} & \textbf{Rationale / governing result} \\ \midrule
\endfirsthead
\toprule \textbf{Scenario} & \textbf{First choice} & \textbf{Rationale / governing result} \\ \midrule
\endhead
\midrule \multicolumn{3}{r}{\emph{continued on next page}}\\ \endfoot
\bottomrule\endlastfoot
Weight-only \texttt{W4A16}, \texttt{INT} &
Per-channel scaling + error-feedback rounding; codebook (NF4) if a LUT kernel is available &
Activations stay high-precision, so the weight distribution alone matters; AWQ's scaling is reported orthogonal to GPTQ and composes with it \citep{lin2023awq}; NF4 matches the codebook to the Gaussianized weight distribution, while QuIP\#'s lattice codebook adds the space-filling gain of \Cref{thm:gishpierce} (\Cref{sec:vq}). A rotation typically helps little at 4-bit weight-only, though it becomes valuable at 2--3 bits (\Cref{tab:res_weight}). \\
\texttt{W4A4}, \texttt{INT} &
Randomized Hadamard rotation (fixed, i.e.\ data-free) $\to$ learned rotation $\to$ learned affine (FlatQuant); + GPTQ &
The activation crest factor is the wall; incoherence is near-optimal at zero calibration (\Cref{prop:rht}), while a data-aware whitening is near-optimal for the integer grid within a $d^{o(1)}$ factor, $d$ the transform block size, on Gaussian or Laplacian data (a factor $d$ for arbitrary distributions; WUSH, \Cref{thm:wush}), worth adding where a fused kernel exists. This is the regime the rotations were built for (\Cref{fig:res_w4a4}; six families, \Cref{tab:res_families}). \\
\texttt{W4A4}, \texttt{MXFP4} (power-of-two E8M0 block scale) &
Block-sized (group-matched) Hadamard for the \emph{scale}; \emph{avoid} a global rotation; add a WUSH-style data-aware block whitening where a fused kernel exists &
The FP element grid already handles per-coordinate range, so a rotation buys less, and on an ideal grid it can cost by destroying a favorable weight--activation anti-alignment (\Cref{sec:fmt-flip}); the coarse power-of-two E8M0 block scale then re-imposes an $\ell_\infty$-like penalty at the block maximum, so a rotation helps again provided its support stays near that block (MR-GPTQ, whose block is a tunable power of two, \Cref{tab:res_fp4}; Quartet fixes its block to the microscaling group at training time). The source reports a single static grid value sufficing once a block rotation is in place, so per-layer scale re-optimization is not the lever here. \\
\texttt{W4A4}, \texttt{NVFP4} (mantissa-carrying \texttt{FP8} block scale) &
Little or no plain rotation; a data-aware block whitening (WUSH) where a fused kernel exists; per-block scale fitting; consider a normalized architecture if training &
The mantissa-carrying \texttt{FP8} scale removes most of the block-level penalty (\Cref{tab:res_fp4}); normalized architectures show the rotation can sometimes be dropped entirely (\Cref{sec:format}). A data-aware whitening still leads: WUSH, exactly optimal for the floating-point AbsMax model (\Cref{thm:wush}), holds the best average on both FP4 formats in \Cref{tab:res_fp4}, by $0.70$ points on \texttt{MXFP4} and $0.26$ here, the source describing its \texttt{NVFP4} gains as often within run-to-run variability. \\
KV-cache, low-bit &
Per-channel keys / per-token values, pre-RoPE; at 2-bit either add a codebook with outlier retention or switch to a per-token rotation &
Keys carry per-channel outliers and RoPE destroys the structure (\Cref{sec:beyond}); axis choice suffices at 3--4 bit (KVQuant, \Cref{tab:res_kv}). At 2-bit the two routes diverge: keep the per-channel axis and add a non-uniform codebook with outlier retention (KVQuant) or a residual window (KIVI), or go per-token under a pre-RoPE grouped-head Hadamard (RotateKV), whose rotation replaces the per-channel axis rather than composing with it. \\
Attention matmuls &
Mean-subtraction smoothing of K; exploit softmax boundedness for P; two-level scale for P in FP4 &
The probability matrix is bounded and needs a scale trick, not a transform; K has a shared-bias outlier (SageAttention, \Cref{sec:beyond}). \\
Diffusion / DiT &
Per-channel smoothing (timestep-aware or aggregated); low-rank branch (SVDQuant) for \texttt{W4A4} &
The time axis is the new variable: choose per-timestep vs robust scaling; SVDQuant's 16-bit low-rank branch absorbs outliers the 4-bit grid cannot (\Cref{sec:beyond}). \\
State-space / Mamba (\texttt{W8A8}) &
Hadamard on the scan output (online, inverse folded); KLT-preconditioned Hadamard $H_K{=}KH$ on correlated gate and output projections; percentile-clipped scale on the scan input &
The scan output carries heavy outliers absent from attention while its input has only a sparse fringe, so the two want different levers (Quamba, \Cref{sec:beyond-attn}). On correlated channels a Hadamard \emph{alone} leaves the rotated diagonal uneven; decorrelating first makes it exactly $\tr\Sigma/d$ for every channel (MambaQuant), which is the orthogonal-only relative of WUSH's whitening-then-Hadamard. \\
Training (forward + gradients) &
Hadamard on forward activations/weights (feature axis); on gradients, leverage-score sampling or a \emph{row}-axis (left-hand) Hadamard, not a feature-axis rotation &
Forward tensors have concentrated outliers (rotate); gradients are row-sparse, and a feature-axis rotation preserves each row's norm, with no per-channel outlier to spread, so the sparsity is a structure to exploit by leverage-score selection or by mixing \emph{rows}, not to flatten along the feature axis (HALO, \Cref{sec:beyond}); the one at-scale \texttt{NVFP4} pretraining recipe rotates only the weight-gradient GEMM inputs, its rotation not being applied to the weights and so not invertible on the other paths (\Cref{sec:fmt-codesign}). \\
\end{longtable}
}
\FloatBarrier

\section{Open Problems and Conclusion}\label{sec:open}

The inversion that organizes this survey also exposes what the field has not yet solved. We close with four open problems it makes visible, the limits that bound what precedes, and a brief synthesis.

The first open problem is the joint optimality of transform and rounding. Every optimality result in the survey holds one stage fixed. The KLT and WUSH optima (\Cref{thm:klt,thm:wush}) assume a given quantizer, and LDLQ's optimality (\Cref{sec:composition}) is among rounders for a \emph{fixed} basis. The linearity theorem (\Cref{thm:linearity}) decouples layers but not the transform from the grid within a layer. Yet \Cref{sec:composition} showed the transform standing in all three of this survey's relations at once: it \emph{substitutes} at high rate for per-coordinate allocation, closing most of the gap waterfilling closes; it \emph{composes} with the rounder, while empirically overlapping with what that rounder's error feedback would have repaired anyway; and it \emph{co-optimizes} with the grid, Gaussianizing the weights being what makes an MSE-optimal grid end-to-end optimal.

What is missing is a theory of the \emph{jointly} optimal $(T, Q)$ pair: the transform and the rounding (or codebook) chosen together to minimize the deployed error, rather than a good transform followed by a good rounder. The lattice view of \citet{chen2025geometry}, independently established by \citet{birnick2026lattice}, casts GPTQ as Babai's algorithm on the Hessian lattice and suggests the right language. A transform is a change of lattice basis, and rounding is nearest-plane decoding. Learned-transform methods already choose $T$ against the quantized loss \emph{empirically} (SpinQuant, FlatQuant, OSTQuant). What stays open is the provable optimality of the pair, and the lattice basis reduction both proofs leave for future work (scale-aware in \citeauthor{chen2025geometry}'s formulation, LLL-style in \citeauthor{birnick2026lattice}'s), not the fact of joint optimization.

A second and subtler gap, the absence of an optimality guarantee for the deployed objective, runs underneath the first. Almost every optimality theorem is proved for a \emph{surrogate}: the $\ell_2$ proxy of \eqref{eq:bilateral-proxy}, an i.i.d.-Gaussian source model, or a distribution-free upper bound on the crest factor. The deployed error, by contrast, is the realized round-to-nearest error of the shared-scale AbsMax kernel. That kernel's group step is governed (to high resolution) by the crest factor, an $\ell_\infty$ statistic. \Cref{sec:pitfalls} noted the two can be anti-correlated. The sharpest live instance of the gap is that $\sum_g M_g^2$ places a scale-block-confined rotation and a \emph{fixed} global one at the same \emph{ideal} floor on \texttt{MXFP4}, and we do not compute it on the two deployed maps, although their accuracies differ by more than seven points (\Cref{tab:res_fp4}). Reducing the published mechanistic accounts of that contrast (\citealp{shao2025blockrotation}; \citealp{chen2025wush}) to a statistic comparable with that surrogate, and evaluating the surrogate on the two deployed maps, are both part of this gap. WUSH is the closest thing to an exception, being \emph{model-}optimal for the floating-point AbsMax quantizer (\Cref{thm:wush}) under a stochastic unbiased-noise model. But even that is near-optimal only within a $d^{o(1)}$ factor, $d$ the transform block size, for the integer case, and only on Gaussian or Laplacian data, the factor degrading to $d$ for arbitrary distributions; it also assumes a second-moment model of the data.

Distribution-free control of the extreme-value statistic itself is not what is missing. The incoherence-processing idea of \citet{chee2023quip}, realized as the randomized Hadamard transform \citep{tseng2024quipsharp}, drives the crest factor to $O(\sqrt{\log d})$ in expectation (\Cref{prop:rht}), and to the same order with high probability by the tail branch of \Cref{lem:subg}, and Kashin quantization gives an $O(1)$ crest guarantee, again for a Haar-random orthogonal map, at the price of a $2\times$ overcomplete representation \citep{merkulov2024kashin}. The missing piece is a \emph{matching} result: a data-dependent transform that provably \emph{minimizes} the realized deployed error on heavy-tailed weights and activations, rather than bounding it in the worst case. That matching result, not any extreme-value bound, is the theoretical question the inversion points at.

A third problem is whether allocation-flexible coding might return to the inference path. The inversion rests on the operand tile being fixed-rate, and that (\Cref{prop:datapath}) is an assumption about today's tensor cores closed empirically, not a law of nature. \Cref{sec:systems} already showed the constraint softening. Lossless, entropy-coded weight coders reach the inference path three ways: by decompressing to a dense format before the GEMM, by decoding tile by tile inside the GEMM kernel, or by pipelining a hardware decoder that emits one symbol per clock, never by coding inside the MMA instruction itself. That recovers part of the \emph{lossless-coding} (bandwidth) advantage of that variable-length lineage, though not the KLT-and-water-filling \emph{concentration} gain. That gain, \Cref{sec:systems} shows, does not reappear on a fixed-rate datapath. A second softening is already in silicon: the block-scaled tensor cores of Blackwell and CDNA4 (\Cref{sec:sys-groupsize}) carry one adaptive scale per $16$ or $32$ elements at full speed. That is a deliberately restricted return of fine-grained adaptive scaling (not of per-coordinate bit allocation), though still shared-scale \emph{within} the block. Whether that constraint is permanent or an artifact of the current tensor-core design is the open systems question. Were a future datapath to serve an allocation-flexible representation as cheaply as a shared-scale one, the concentration optimum would return to relevance and the field's recent turn to flattening would read as a transitional episode rather than a permanent regime. Which of the two the hardware ultimately rewards is unsettled, and it decides whether the two poles of the inversion stay separate.

A fourth problem is the co-design of transform and format. \Cref{sec:format} treated the number format as a design variable, but only downstream of a format the hardware had already frozen. When the element grid, the shared-scale format, and the block size are themselves in play, as they increasingly are when accelerators are co-designed with their formats, the optimal transform is a joint function of all of them. The power-of-two scale calls for a block-level rotation, a mantissa-carrying scale weakens the need, and a normalized architecture removes it entirely. Designing the transform and the format together (choosing how many bits to spend on the scale versus the element grid versus a rotation, under a fixed hardware budget) is an open co-design problem. The pure-algorithm and pure-format literatures have so far approached it from opposite sides. The format side has begun to move on its own (adaptive per-block scales, exponent-to-mantissa reallocation, and adaptive INT4/FP4 element grids in hardware; \citealp{cook2025four,lee2025mxplus,cook2026adaptive}). What stays genuinely open is the \emph{joint} choice, the transform and the format designed against each other rather than one frozen while the other moves.

The sharp, already-live instance of that joint choice is the block size itself, which today's block-scaled hardware fixes at $16$ or $32$ (\Cref{sec:sys-groupsize}). The two-level, sub-block direction is not virgin ground. VS-Quant \citep{dai2021vsquant} explored per-vector integer scaling under a coarse floating-point scale and the shared-microexponents Block Data Representation \citep{rouhani2023shared} explored sub-block scaling with mantissas as narrow as two bits, and \citet{luo2026hifloat4format} carries two levels of micro-exponent inside a $64$-element block, reporting post-training inference accuracy above \texttt{NVFP4}. The open question is how much finer the \emph{free} per-block scale can profitably go on a tensor core, weighed against the area each halving costs, and which transform is optimal at that granularity.

Beyond the open problems, three limits bound what precedes. First, we run no benchmarks of our own. Every reported result in \Cref{sec:empirical} is transcribed from a cited table or figure, the differences and margins we quote being arithmetic on those transcribed values, so the picture inherits its sources' models, calibration sets, and any errors. The exceptions are our own: the mechanism figures of \Cref{sec:foundations,sec:inversion,sec:taxonomy} and the measurement in \Cref{tab:flip}, computed on a single layer of one $1.1$B model (\Cref{fig:crest} on a synthetic sixteen-value group) and offered as illustrations of the theory rather than as evaluations. The three scripts behind them, one capturing the activation cache, one drawing every real-data panel from it and one the single synthetic figure, will be released with the paper. Each head-to-head table is also transcribed from a paper proposing one of its own rows, so within-table margins favor that row. We mitigate this by re-checking each figure against its primary source and fixing one protocol per comparison, but the numbers are not commensurable across blocks and we offer no home-grown leaderboard.

Second, and most consequential, the organizing inversion and nearly all the optimality results we assemble are proved against surrogate objectives (the $\ell_2$ bilateral proxy, the AbsMax crest factor, or an i.i.d.\ unbiased-noise model). They are not proved against the realized extreme-value error of the deployed shared-scale kernel on heavy-tailed data. This gap we can only flag (it is the second open problem above) and not close. Third, the corpus carries a June~2026 cutoff in a field that moves in months, so particular method rankings, especially in the still-consolidating \texttt{FP4}/microscaling and learned-affine lines, will date. What we claim outlasts those rankings is the pair of structural axes, not the leaderboard. We classify the 43 transform methods proper (\Cref{tab:taxonomy}) but carry no table covering all 200 cited works, so the corpus can be read for its families and its protocol but not audited work by work from the paper alone. Our scope is likewise deliberate (\Cref{sec:intro}). We survey function-preserving linear transforms, treating quantization-aware retraining of the base weights, non-transform mixed-precision allocation, and architectural change only where they touch the transform stage.

The transform stage that opens most competitive low-bit post-training pipelines, and the 4-bit activation path that ships behind a rotation (\Cref{sec:sys-deployed}), is not a new invention but the sixty-year-old machinery of transform coding, re-run in a regime its optimality theory never treated (\Cref{sec:classical} records the block-floating-point line that analyzed the shared-scale format itself, the robust-quantization line that already rotated for overload, and a 2017 proof of the rotation mechanism in distributed mean estimation, none of them deriving an optimal transform for the regime). Classical transform coding is allocation-flexible, and for it energy \emph{concentration} (the KLT) is optimal. The deployed LLM kernel is shared-scale, and for the objective it induces, on a uniform grid, energy \emph{flattening} is optimal (the Hadamard near-optimally, and WUSH at the model-optimum via a data-aware whitening beyond any rotation). Per-coordinate allocation, together with the per-coordinate scale the operand tile removes alongside it, is the field's \emph{primary} organizing axis, ordering the poles a transform can aim for rather than the transforms themselves, with the number format a genuine second axis, since on a floating-point grid the flattening prescription weakens and can reverse.

Read through those two axes, the allocation-and-scale constraint and the number format, with operand alignment (\Cref{rem:alignment}) as a further factor that neither axis prices and no rotation can move, and remembering that the optimality results on each side hold against different surrogate objectives (none transferring across the divide, \Cref{cor:notransfer}), an otherwise sprawling literature becomes legible. Two further levers sit beside the axes rather than on them: centering, because a per-channel shift changes the energy budget majorization holds fixed, and clipping, because it trades overload against step size instead of redistributing energy. That both survive \emph{any} orthogonal flattening is why the affine family, and not the rotations, defines the current frontier (\Cref{sec:tax-affine}). That reading explains why flattening transforms win at \texttt{W4A4} where the KLT would lose, and why a rotation substitutes at high rate for bit allocation but not for a codebook. It explains why the integer prescription changes under floating-point, and why a rotation that flattens a weight matrix is inert on a sparse gradient unless it is moved to the token axis. The transforms, the rounders, the codebooks, the formats, and the tensors beyond the weight matrix are, read this way, a conditional design map rather than a single verdict. Drawing that map (naming the axes, assembling the objective-specific optimality results, indexing the corpus and classifying the 43 transforms, naming the composition patterns, pricing the transform as a cost device as well as an accuracy device, distilling the which-transform-when guide of \Cref{sec:practice}, and recording which transform each regime rewards) has been the work of this survey. The open problems above are what remains once it is drawn.

\subsubsection*{Broader Impact Statement}

The transforms surveyed here reduce the memory, bandwidth and energy cost of serving large language models. That lowers the barrier to running capable models on modest hardware and on device, with the attendant gains in access, cost and privacy, and it reduces the energy drawn per token served. The same reduction lowers the cost of deploying models whose behavior is harmful, and it does so without discriminating between uses.

One concern is specific to this survey's subject. Quantization changes a model's outputs, so safety evaluations, red-teaming results and alignment properties established on a full-precision checkpoint do not automatically transfer to a low-bit deployment of it. The literature assembled here optimizes for perplexity and task accuracy, which are not sensitive to the behaviors such evaluations target, and the transform stage in particular is selected against a proxy for weight-reconstruction error (\Cref{sec:inversion}). We are aware of no work that measures how the choice of transform affects safety-relevant behavior, and we record that as a gap rather than a settled matter.

\bibliographystyle{tmlr}
\bibliography{refs}

\newif\ifappglossary  \appglossarytrue
\newif\ifapprepro     \appreprofalse
\newif\ifappindex     \appindextrue

\appendix
\ifappglossary 

\section{Notation}
\label{app:notation}

\Cref{tab:notation} collects the symbols used throughout, fixed in \Cref{sec:foundations}.

\begin{table}[!htbp]
\centering
\small
\caption{Notation used throughout the survey.}
\label{tab:notation}
\begin{tabular}{@{}l@{\quad}>{\raggedright\arraybackslash}p{0.62\textwidth}@{}}
\toprule
\textbf{Symbol} & \textbf{Meaning} \\
\midrule
$d$ & ambient dimension: the number of coordinates a transform acts on ($d_{\mathrm{in}},d_{\mathrm{out}}$ for a layer's input and output). Wherever the WUSH guarantee is stated ($d^{o(1)}$ or factor-$d$ near-optimality, \Cref{thm:wush}), $d$ is instead the transform's \emph{block} dimension, which \citet{chen2025wush} set equal to $G$; in the \texttt{W}$b$\texttt{A}$c$\texttt{KV}$d$ shorthand it is a bit-width \\
$W\in\R^{d_{\mathrm{out}}\times d_{\mathrm{in}}}$ & a layer's weight matrix \\
$X\in\R^{N\times d_{\mathrm{in}}}$ & the layer's input activations ($N$ tokens as rows) \\
$\bar X$ & the normalized stream: the output of whichever RMSNorm opens the sub-block under discussion, that norm taking the layer input in attention and the post-attention residual in the MLP (\Cref{sec:anatomy}) \\
$Y=XW^\top$ & the layer output the quantized layer must preserve \\
$T\in\mathrm{GL}(d)$ & a function-preserving transform; $Y=(XT^\top)(WT^{-1})^\top$ \\
$Q(\cdot)$ & the (scalar, vector, or codebook) quantizer, normally written with its argument; a bare $Q$ is an orthogonal matrix in the transform sections (\Cref{thm:invariance}, and the $\mu$ row below), the attention query matrix wherever attention is the subject, and the quantizer itself wherever its argument is suppressed, as in the signature $Q:\R\to\mathcal C$ or a transform--rounder pair $(T,Q)$; context distinguishes them \\
$P$ & the row-stochastic attention probability matrix, the softmax output feeding the context product $PV$ (\Cref{sec:anatomy}); elsewhere $P$ is FlatQuant's learned Kronecker factor (\Cref{sec:tax-affine}) and, subscripted as $P_i$, a microscaling element value (\Cref{sec:format}); context distinguishes them \\
$\HX=X^{\!\top}X$ & input second-moment / activation Gram matrix (batch-first; equals the $XX^{\!\top}$ of GPTQ's feature-first convention). The OBC/GPTQ (layerwise) Hessian is $2X^{\!\top}X=2\HX$ (\Cref{prop:obs}). \\
$\HW$ & input-axis weight Gram $W^{\!\top}W\in\R^{d_{\mathrm{in}}\times d_{\mathrm{in}}}$ (same dimension as $\HX$; pairs with the activation-side proxy). \\
$\Delta W,\ \Delta X$ & quantization errors $Q(W)-W$, $Q(X)-X$ \\
$\Delta$ & quantizer step size, giving per-element noise power $\Delta^2/12$ (\Cref{lem:bennett}); $\Delta=M/(2^{b-1}-1)$ under a symmetric AbsMax scale \\
$\Sigma$ & source covariance (classical transform coding) \\
$U_{\mathrm{KLT}}$ & Karhunen--Lo\`eve transform: an eigenbasis diagonalizing $\Sigma$ (unique up to column permutation and sign, and up to an orthogonal rotation within each repeated eigenspace) \\
$\mathrm{CF}=M/\sigma$ & crest factor: block maximum over RMS (an $\ell_\infty/\ell_2$ ratio) \\
$G_{\mathrm{TC}}(U)$ & transform coding gain of an orthogonal transform $U$ \\
$\lambda(x)$ & quantizer point density (Bennett's integral) \\
$R,\ R_k$ & total rate; per-coordinate rate (water-filling); in the incoherence sections $R$ is instead the randomized rotation $R=HD$ (\Cref{prop:rht}); context distinguishes them \\
$\mu$ & incoherence parameter. For an orthogonal $Q$, $\mu$-incoherent means $|Q_{ij}|\le\mu/\sqrt n$; the main text (\Cref{sec:incoherence}) states the corresponding coherence for a general (non-orthogonal) matrix, scaled by its Frobenius norm. \\
$H$ (as a matrix) & the (normalized) Hadamard matrix, the incoherence backbone \\
$G$ & quantization group size (e.g.\ $128$ for INT, $32$ for MXFP4) \\
$\alpha$ & migration/smoothing strength (diagonal family) or clip threshold \\
\bottomrule
\end{tabular}
\end{table}
\FloatBarrier

\section{Glossary of concepts and formats}
\label{app:glossary}

\paragraph{Rate and coding.}
\emph{Shared-scale (``fixed-rate'') quantizer}: equal bits per coordinate under a single shared per-group scale, with no per-coordinate bit allocation (the dense-GEMM regime of every deployed low-bit kernel; \Cref{def:rate}, grounded in the datapath by \Cref{prop:datapath}). The element grid is a second, independent axis: \texttt{INT4}-per-group adds a uniform grid, whereas \texttt{MXFP4}/\texttt{NVFP4} are shared-scale on a \emph{non-uniform} E2M1 grid (\Cref{sec:format}). \emph{Allocation-flexible (``variable-rate'') quantizer}: a different number of bits spent on different coordinates under a fixed total rate, by unequal deterministic allocation or by entropy coding (the regime of classical transform coding). The operative distinction is the availability of per-coordinate allocation, not code length. \emph{Transform coding}: the decorrelate--allocate--quantize pipeline (\Cref{sec:classical}). \emph{Coding gain}: the arithmetic over geometric mean of the transformed coordinate variances, \eqref{eq:codinggain}; equivalently, in a fixed basis $U$, the ratio of the distortion at a common per-coordinate rate to the distortion under optimal allocation. \emph{Reverse water-filling}: the rate allocation $R_k=\max(0,\tfrac12\log_2(\sigma_k^2/\theta))$, which spends bits on the coordinates whose transformed variance exceeds the water level $\theta$ and none on those at or below it (\Cref{prop:waterfill}). \emph{Space-filling gain}: the asymptotic $0.255$-bit gap between \emph{entropy-coded} scalar and high-dimensional lattice quantization, under the stated high-resolution/regularity assumptions (\Cref{thm:gishpierce}); it is the limiting advantage, not the finite-dimension gain of any particular codebook.

\paragraph{The inversion.}
\emph{Concentration}: a transform that packs energy onto few coordinates, optimal for variable-rate coding (the KLT). \emph{Flattening} (incoherence): a transform that spreads energy evenly, optimal for fixed-rate quantization \emph{on a uniform element grid} (\Cref{sec:format} gives the floating-point exception; the Hadamard near-optimally; WUSH attains its own model's optimum, exactly for the FP-AbsMax grid and within a $d^{o(1)}$ factor for the integer grid on Gaussian or Laplacian data (\Cref{thm:wush}), by a data-aware whitening applied before that same Hadamard; under that model both are needed and an orthogonal map alone buys nothing, the flattening acting on the whitened second moment rather than the raw one, though the measured ablation departs from the model on \texttt{NVFP4}). \emph{Crest factor}: the extreme-value ratio that controls AbsMax error; flattening minimizes it. \emph{The Great Inversion}: concentration and flattening are the opposed optima of the two surrogate objectives, the shared-scale one taken on a uniform grid, each proven on its own side, with no optimality transferring between them except where the spectrum is group-constant (\Cref{thm:inversion,cor:notransfer}).

\paragraph{Transforms and composition.}
\emph{Function-preserving transform}: an invertible $T$ whose action on one operand is exactly undone on the other, leaving the layer's output unchanged (\Cref{def:fpt}); whether it is then \emph{absorbed} offline is a separate, cost question. \emph{Computational invariance}: that a gain-free RMSNorm commutes with an orthogonal $Q$, so once the learned gain is folded into the consuming weights (\Cref{prop:gamma}) a global residual-stream rotation folds into the surrounding linear weights at zero runtime (\Cref{thm:invariance}). \emph{Absorption / folding}: folding $T^{-1}$ offline into the consuming weight ($W\!\leftarrow\!WT^{-1}$) while the activation-side factor $T^{\!\top}$ is pushed through the preceding RMSNorm or linear layer, so $XT^{\!\top}$ is never materialized (the orthogonal, zero-cost fold of \Cref{thm:invariance} is RMSNorm-specific; SliceGPT first converts LayerNorm to RMSNorm). \emph{Composes / enables / substitutes / co-optimizes}: the four relationships between a transform and the rounding or codebook stage (\emph{enabling} a limiting case of composing; \Cref{sec:composition}). \emph{Error-feedback rounding}: the OBS/GPTQ/LDLQ family that propagates each rounding error into the not-yet-quantized weights.

\paragraph{Precision lanes and formats.}
\emph{W$b$A$c$ / W$b$A$c$KV$d$}: weights in $b$ bits, activations in $c$, KV-cache in $d$ (e.g.\ \texttt{W4A16} weight-only, \texttt{W4A4} both operands, \texttt{W4A4KV4} and \texttt{W4A8KV4} the cache as well). \emph{AbsMax scale}: a group's shared scale pinned to its absolute maximum $M$ (no clipping), i.e.\ step $\Delta=M/(2^{b-1}-1)$ on a symmetric $b$-bit grid. \emph{INT4}: uniform 4-bit integer grid. \emph{FP4 / E2M1}: 4-bit floating point ($1$ sign, $2$ exponent, $1$ mantissa), the non-uniform grid $\{0,0.5,1,1.5,2,3,4,6\}$. \emph{Microscaling (MX)}: a block of (typically) $32$ elements sharing one scale. \emph{MXFP4}: E2M1 elements with an \emph{E8M0} power-of-two shared scale. \emph{NVFP4}: E2M1 elements with an E4M3 (FP8) shared scale over $16$-element blocks plus a per-tensor scale. \emph{E8M0 / E4M3}: an $8$-bit exponent-only (power-of-two) scale versus a $4$-exponent, $3$-mantissa FP8 scale. \emph{NF4}: the NormalFloat quantile codebook for Gaussian weights.
 \fi

\ifapprepro 

\section{Reproducibility of the mechanism figures}
\label{app:repro}

Every \emph{result} number in this survey is transcribed from a cited source, a limit
\Cref{sec:open} states plainly. The \emph{mechanism} figures are the exception: they are
our own computations, and this appendix states their provenance so that they are
reproducible and cannot be mistaken for transcribed results. They are illustrations of the
theory on one layer of one small model, not evaluations.

\textbf{Data.}\enspace All distribution figures use one layer of the base checkpoint
\texttt{TinyLlama/TinyLlama-1.1B-intermediate-step-1431k-3T}. We capture the inputs to
the layer-$17$ $q$-projection over six windows of $512$ tokens drawn from WikiText-2
(\texttt{wikitext-2-raw-v1}, \emph{train} split), i.e.\ $3072$ tokens against a width of
$d=2048$, so the activation second-moment matrix $\HX$ is full rank. Capture uses a
fixed seed; activations are cached in \texttt{fp16} and all figure arithmetic is
\texttt{fp64}.

\textbf{Per figure.}\enspace \Cref{fig:reshape} and \Cref{fig:inversion} use all $3072$ captured tokens;
the element-grid diagrams, when enabled, use the same layer's weight matrix;
\Cref{tab:flip} averages $12{,}288$ groups of $128$ channels. \Cref{fig:inversion} evaluates both its
shared-scale costs, the second-moment profile $D_{\mathrm{fr}}=\sum_g\max_{k\in g}v_k$ and
the realized per-token maxima, at group size $G=128$; the ordering of the three transforms is unchanged at the other deployed group
sizes $G\in\{32,64\}$. Randomized-Hadamard sign seeds are fixed (one seed for the global
$d=2048$ rotation, a per-token seed in \Cref{tab:flip}); the reported crest factors are
stable across seeds, so the choice is immaterial. Both readings plotted in \Cref{fig:inversion}(b) come from the same cached activations: the
filled marker is the second-moment profile $v=\mathrm{diag}(T\Sigma T^{\!\top})$, the open one
the realized per-token per-group maxima of $XT^{\!\top}$, with the sign draw
\texttt{make\_real\_figs.py} uses for that panel. At $G{=}128$ the randomized Hadamard's pair
sits at $2.6\times$ and $8.0\times$ the $E/G$ floor, the $3.1\times$ inflation the caption
names. \Cref{fig:crest} is \emph{not} a
measurement: it is the synthetic sixteen-value running group of \Cref{sec:zoo}, fifteen
standard-normal draws plus one injected outlier.

\textbf{The whitening curve.}\enspace The whitening panel of \Cref{fig:reshape} applies a \emph{regularized} inverse square
root, $(\Sigma+\lambda I)^{-1/2}$ with $\lambda=0.02\,\tr(\Sigma)/d$. Exact whitening
would drive the per-coordinate variance spread to $1$ by construction; the residual
$1.2$ reported in that caption is the ridge, not a limitation of the transform. We keep
the ridge because it is what a deployed data-aware map would need on a finite
calibration sample.

\textbf{Availability.}\enspace The parameters above fully specify each computation: a
few hundred lines of \textsc{NumPy} over a cached activation tensor, reproducible from
the checkpoint and corpus named. The three scripts that produce them (\texttt{make\_real\_figdata.py}, which captures the
activation cache, \texttt{make\_real\_figs.py}, which draws every real-data panel from it,
and \texttt{make\_figures.py}, which draws the one synthetic figure) will be released with
the paper.

\section{Corpus construction}
\label{app:corpus}

The corpus was seeded from the transform-stage methods named in \Cref{sec:intro}, extended
by backward and forward citation tracing from those seeds, and checked against the arXiv
\texttt{cs.LG} and \texttt{cs.CL} listings and the main machine-learning and systems venues,
with a June~2026 freeze. The classical lineage was traced back from the modern works'
own citations to the 1948--2021 sources that ground it, the quantization theory dating from 1948 and transform coding proper from 1963. Every work cited anywhere in the survey appears in \Cref{tab:index}. As \Cref{sec:intro} states, the aim is complete
coverage of the transform-stage \emph{families} rather than a census of every instance, so
a method that instantiates a family already represented may be cited in passing or not at
all without affecting the classification.
 \fi

\ifappindex
\section{Complete index of cited works}\label{sec:index}
For auditability, \Cref{tab:index} classifies every one of the works cited in this survey into the role it plays (classical foundation, weight/activation transform, rounding or codebook, number format, beyond-weights tensor, system, or cross-cutting background), with a one-line statement of its contribution and the section that treats it. Methods also appear in the per-domain tables of the corresponding section; this index is the single place every cited work is accounted for.
{\footnotesize
\renewcommand{\arraystretch}{1.12}
\begin{longtable}{@{}>{\raggedright\arraybackslash}p{3.3cm}@{\hspace{8pt}}>{\centering\arraybackslash}p{1.2cm}>{\raggedright\arraybackslash}p{\dimexpr\linewidth-5.25cm\relax}@{}}
\caption{Complete index of the 200 cited works, grouped by role. Each is classified once; methods also appear in the per-domain tables of the corresponding section.}\label{tab:index}\\
\toprule \textbf{Work} & \textbf{Section} & \textbf{Role} \\ \midrule
\endfirsthead
\toprule \textbf{Work} & \textbf{Section} & \textbf{Role} \\ \midrule
\endhead
\midrule \multicolumn{3}{r}{\emph{continued on next page}}\\ \endfoot
\bottomrule\endlastfoot
\multicolumn{3}{@{}l}{\emph{Classical transform coding and quantization (\Cref{sec:classical})}}\\[1pt]
Ahmed et al.\ (DCT)~\citep{ahmed1974dct} & 3 & The discrete cosine transform; the fixed, data-independent transform that approximates the KLT for correlated sources. \\
Ben-Basat et al.\ (RHT proven)~\citep{benbasat2026quantizing} & 2 & Composing two randomized Hadamards puts every marginal within $O(d^{-1/2})$ of a Gaussian, and three makes the coordinate covariance decay, so Haar-designed codebooks transfer to Hadamards. \\
Bennett (1948)~\citep{bennett1948spectra} & 2 & Bennett's integral and the additive-white-noise high-resolution model behind the 6 dB/bit law. \\
Choi Hessian-weighted~\citep{choi2017towards} & 3 & Diagonal-Hessian-weighted Lloyd--Max quantization; the CNN-era \emph{sibling} of GPTQ's second-order objective, not its parent. \\
Conway--Sloane lattices~\citep{conway1982fast} & 3 & Fast closest-point decoding algorithms for the classical lattices; this is what makes a high-dimensional codebook usable at all. \\
Cover \& Thomas (information theory)~\citep{cover2006elements} & 3 & The standard text; source of the reverse water-filling rate allocation (their Thm.~10.3.3) that \Cref{prop:waterfill} states in its high-rate form. \\
Conway--Sloane (Voronoi regions)~\citep{conway1982voronoi} & 3 & Second-moment analysis of lattice Voronoi cells; quantifies the space-filling (granular) gain a vector codebook captures over a scalar grid. \\
DeepCABAC~\citep{wiedemann2019deepcabac} & 3 & Context-adaptive arithmetic weight coder; became the core of the ISO/IEC MPEG Neural Network Compression and Representation standard. \\
Deep Compression~\citep{han2016deep} & 3 & Pioneered the prune, $k$-means weight-sharing, Huffman-coded-index pipeline for network compression. \\
Hung \& Meng (robust rotation)~\citep{hung1998rotations} & 3 & Walsh--Hadamard rotation of \emph{i.i.d.} Laplacian-like data chosen to improve \emph{overload} characteristics, not to compact energy; the classical precursor of the flattening mechanism. \\
Kalliojärvi \& Astola (block floating point)~\citep{kalliojarvi1996blockfloat} & 3 & Roundoff analysis of quantizing \emph{to} a block-floating-point format; the classical shared-scale quantizer, on a matched bits-per-sample basis, with no transform designed for it. \\
NestQuant~\citep{savkin2025nestquant} & 3 & Nested-lattice quantization for matrix products; cites \citet{hung1998rotations} as prior art for rotation-as-Gaussianization. \\
Max (minimum-distortion quantizer)~\citep{max1960quantizing} & 2 & The independent derivation of the optimal fixed-rate scalar quantizer conditions, three years after Lloyd's 1957 Bell Labs memorandum and twenty-two years before it appeared in print; the other half of the Lloyd--Max name. \\
Oppenheim (block floating point)~\citep{oppenheim1970blockfloat} & 3 & Block-floating-point filter realization and its roundoff-noise analysis; an early formal treatment of shared-exponent arithmetic. \\
Popat \& Zeger (robust quantization)~\citep{popat1992robust} & 3 & Reshapes a memoryless source's amplitude distribution before quantizing, for robustness rather than compaction; predates the overload-motivated rotation. \\
Suresh et al.\ (distributed mean estimation)~\citep{suresh2017dme} & 3 & Proves the rotate-then-shared-scale mechanism: a structured random rotation before quantization takes the error from $\Theta(d)$ to $O(\log d)$ with no distributional assumption, exactly \Cref{prop:rht} six years before its rediscovery. \\
Effros et al. (KLT suboptimality)~\citep{effros2004suboptimality} & 3 & The KLT can be strictly suboptimal among orthogonal transforms for non-Gaussian sources; its distortion optimality needs a same-shape-marginals hypothesis, of which the Gaussian case is the familiar instance. \\
Federici et al. (dissecting quant error)~\citep{federici2026dissecting} & 4 & Analytic SQNR decomposition (their Theorem~2.4) splitting quantization error into a concentration (range/extreme-value) term and an alignment term; corroborates the crest-factor view of AbsMax error. \\
Feng et al. (provable RHT)~\citep{feng2026provable} & 4 & A randomized Hadamard with a dithered Lloyd/Gaussian-companded scalar codebook attains the same leading MSE constant proved for a fully random rotation (up to an o(1) that vanishes with bit-width). \\
Gray \& Neuhoff (quantization survey)~\citep{gray1998quantization} & 3 & The standard survey of quantization theory, covering the high-rate laws, entropy-constrained quantization and the lattice/space-filling results this part assembles. \\
Gish--Pierce~\citep{gish1968asymptotically} & 3 & At high resolution, an entropy-coded uniform scalar quantizer lies within 0.255 bits of the Shannon lower bound for a memoryless source (the space-filling gap). \\
Goyal transform coding~\citep{goyal2001transform} & 3 & Transform-coding review: fixed-rate = fixed-length indices; a KLT is optimal among orthogonal transforms for any bit allocation when every coefficient is quantized by a member of one scale-invariant family, each scaled to its own variance; the coding-gain formula, with entropy coding optional. \\
LeCun et al.\ (OBD)~\citep{lecun1990obd} & 3 & Optimal Brain Damage: the diagonal-Hessian saliency criterion for pruning, the assumption OBS was written to drop and the one Choi's quantizer inherits. \\
Hassibi \& Stork (OBS)~\citep{hassibi1993obs} & 2 & Optimal Brain Surgeon: the second-order constrained-least-squares single-weight update that GPTQ-class error-feedback rounding descends from. \\
Huang \& Schultheiss (KLT)~\citep{huang1963block} & 3 & KLT with optimal log bit allocation is MSE-optimal at high resolution for correlated-Gaussian block quantization: the classical pole the quantization inversion overturns. \\
Lloyd--Max~\citep{lloyd1982lsq} & 2 & Lloyd--Max nearest-neighbor and centroid conditions for the MSE-optimal fixed-rate scalar quantizer; ancestor of data-driven weight codebooks. \\
Malinovskii et al. (linearity theorem)~\citep{malinovskii2025higgs} & 4 & Perplexity increase is, to leading order, a weighted sum of per-layer relative WEIGHT (Frobenius) errors; justifies minimizing per-layer relative weight error one layer at a time, under its approximations, and a separable (dynamic-programming) bit allocation. \\
MPEG NNR standard~\citep{kirchhoffer2022nnr} & 3 & Standardized neural-network weight compression: quantization plus context-adaptive arithmetic coding (DeepCABAC), with sparsification and pruning as pre-processing (ISO/IEC MPEG NNR). \\
NNCodec~\citep{becking2023nncodec} & 3 & Open-source tooling implementing the MPEG Neural Network Compression and Representation entropy-coded weight standard. \\
Nonlinear transform coding~\citep{balle2020nonlinear} & 3 & Learned nonlinear transform codes surpass linear KLT on non-Gaussian sources. \\
Ordentlich \& Polyanskiy (nested lattice)~\citep{ordentlich2024optimal} & 4 & Nested-lattice quantization is asymptotically optimal for the matrix-multiplication (inner-product) distortion. \\
Panter--Dite~\citep{panter1951quantization} & 3 & High-resolution companded-scalar-quantizer distortion formula; an early ancestor of modern high-resolution quantization theory. \\
Radio~\citep{young2025radio} & 3 & Rate-distortion bit-depth allocation demonstrated up to the 66--70B scale (argued to extend to hundreds of billions); anticipates transform coding as future work. \\
Rate-distortion model compression~\citep{gao2019rate} & 3 & Made rate-distortion theory for neural-network model compression explicit, from theory to practice. \\
Sanjeet et al. (PeRQ)~\citep{sanjeet2026mixquant} & 4 & Non-asymptotic bounds on the post-rotation $\ell_\infty$ of block-Hadamard rotation, controlled by how evenly $\ell_1$ mass is spread across blocks. \\
SliceGPT (invariance theorem)~\citep{ashkboos2024slicegpt} & 2 & Computational-invariance theorem: with the RMSNorm gain folded away first, a global orthogonal rotation of the residual stream folds into the surrounding linear weights for free. \\
Trellis-coded quantization~\citep{marcellin1990trellis} & 3 & Viterbi-searched trellis quantizer reaching within $0.21$\,dB of the distortion-rate bound for a uniform source, matching or beating lattices up to dimension 24. \\
Young et al. (transform CNN compression)~\citep{young2021transform} & 2 & CNN-era transform-coding treatment of data-aware quantization, weighting by the covariance of the end-to-end output gradients $\partial y/\partial\theta$ rather than by the layer's input second moment. \\
Zhang et al. (provable EF rounding)~\citep{zhang2025provable} & 7 & What they state to be the first quantitative proxy-error bounds for error-feedback (OPTQ/GPTQ and Qronos) rounding, on an unbounded grid. \\
\addlinespace
\multicolumn{3}{@{}l}{\emph{Weight/activation transforms (\Cref{sec:taxonomy})}}\\[1pt]
AffineQuant~\citep{ma2024affinequant} & 6 & One invertible affine A per layer, learned by a gradual diagonal-to-dense mask preserving invertibility. \\
ASVD~\citep{yuan2023asvd} & 6 & Adjacent (not function-preserving): activation-aware SVD with a diagonal per-input-channel rescale from mean $|X|$ before low-rank truncation. \\
Atom~\citep{zhao2023atom} & 6 & Reorder top-128 outlier channels to the matrix end; INT8 for them, INT4 for the rest. \\
AWQ~\citep{lin2023awq} & 6 & Per-channel scale protecting the weight channels salient by \emph{activation} magnitude; exponent grid-searched on output MSE. \\
BASE-Q~\citep{he2025baseq} & 6 & Closed-form residual rotation and learned value rotation fold in, $R_{\mathrm{qk}}$/$R_{\mathrm{down}}$ stay online; learnable channel-bias recentering plus asymmetric per-quantizer scaling run online. \\
ButterflyQuant~\citep{xu2025butterflyquant} & 6 & Learnable butterfly of (n/2)log n Givens rotations, trained on a layerwise quantized-reconstruction loss with a uniformity-KL regularizer, O(n log n). \\
CLE~\citep{nagel2019dfq} & 6 & Data-free pairwise channel rescale equalizing adjacent-layer weight ranges via ReLU scale-equivariance. \\
ConQuR~\citep{thrash2026conqur} & 6 & Alternating Procrustes rotation aligning activations to hypercube corners, re-solved against calibration rather than learned by gradient. \\
DartQuant~\citep{shao2025dartquant} & 6 & Learned rotation optimizing a distribution-shaping ``Whip'' loss via a QR parametrization. \\
DFRot~\citep{xiang2024dfrot} & 6 & Alternating refinement of QuaRot's Hadamard, each rotation step a closed-form Procrustes solve, under a loss up-weighting massive-activation tokens. \\
DuQuant~\citep{lin2024duquant} & 6 & Greedy block-diagonal rotation plus zigzag permutation balancing outlier mass across blocks (perm/orthogonal hybrid). \\
FlatQuant~\citep{sun2024flatquant} & 6 & Learnable invertible Kronecker P = P1 (x) P2, fused into an online kernel, with joint clipping. \\
FPTQuant~\citep{vanbreugel2025fptquant} & 6 & Function-preserving transforms: Q/K scale-and-rotate, value matrix, MLP scaler, dynamic residual scale. \\
FrameQuant~\citep{adepu2024framequant} & 6 & Quantize in an overcomplete tight fusion frame (r\ensuremath{\sim}1.1); redundancy averages out noise. \\
GSR~\citep{choi2025grouped} & 6 & Block-diagonal sequency-ordered Walsh--Hadamard blocks; training-free; a drop-in $R_1$ replacement that also initializes learned rotations. \\
HARP~\citep{zagitov2026harp} & 6 & Sparse butterfly-like block-orthogonal stages, Hadamard-initialized, mixed-radix, backend-aware. \\
Kashin~\citep{merkulov2024kashin} & 6 & Represent x = u + Qv in a 2x-redundant frame basis so both factors have small L-infinity. \\
KurTail~\citep{akhondzadeh2025kurtail} & 6 & Learn Stiefel rotations minimizing the gap of rotated-activation kurtosis to the uniform value. \\
LATMiX~\citep{gordon2026latmix} & 6 & Learnable full invertible affine (LU or QR form) mixing channel mass; MX-format bound. \\
Meller et al.\ (same, same)~\citep{meller2019same} & 6 & Weight-factorization equalization of consecutive layers, concurrent with and slightly ahead of Cross-Layer Equalization. \\
MergeQuant~\citep{wang2025mergequant} & 6 & Migrate static per-channel activation scales into weights; dimensional reconstruction plus clipping. \\
OmniQuant~\citep{shao2023omniquant} & 6 & Gradient-learned per-channel scale and shift (LET) plus learnable weight clipping (LWC). \\
OptRot~\citep{gadhikar2025optrot} & 6 & Data-free optimization of a rotation over the orthogonal group against a fourth-power weight-outlier proxy. \\
OSTQuant~\citep{hu2025ostquant} & 6 & Learn a rotation plus diagonal scaling jointly under a KL-Top loss; a quantization-space-utilization metric motivates the form. \\
Outlier Channel Splitting~\citep{zhao2019outlier} & 6 & Pre-LLM channel duplication-and-halving that keeps the network functionally identical; ancestor of QLLM's decomposition. \\
Outlier Suppression~\citep{wei2022outlier} & 6 & Fold LayerNorm per-channel gamma (the outlier amplifier) into the next weight; token-wise clipping. \\
Outlier Suppression+~\citep{wei2023outliersuppplus} & 6 & Per-channel shift (center) then scale, both migrated into the next weight and bias. \\
ParoQuant~\citep{liang2025paroquant} & 6 & Product of learned pairwise Givens rotations plus per-channel scaling, applied online (\ensuremath{\sim}10\% runtime). \\
PermLLM~\citep{zou2025permllm} & 6 & Learned Sinkhorn-to-Hungarian channel permutation for more accurate N:M pruning. \\
PQF~\citep{martinez2021pqf} & 6 & Search a function-preserving input-channel permutation, equivalently the predecessor's output order, that eases vector quantization. \\
PrefixQuant~\citep{chen2024prefixquant} & 6 & Prepend a fixed prefix of the token types that most often carry outliers, so the prefix absorbs the token-wise outliers, plus Hadamard. \\
QLLM~\citep{liu2024qllm} & 6 & Disassemble an outlier channel into replicated sub-channels, then reassemble similar channels. \\
QuaRot~\citep{ashkboos2024quarot} & 6 & Global randomized Hadamard fused into weights by computational invariance, plus a few online Hadamards. \\
QuIP~\citep{chee2023quip} & 6 & Introduced incoherence processing: a Hessian-derived per-channel rescale, then conjugation of W and H by Kronecker-factored random orthogonals forcing mu-incoherence (the paired LDLQ rounding is Hessian-aware). \\
ReSpinQuant~\citep{kim2026respinquant} & 6 & Full-size Cayley-learned rotation per layer, folded into the attention and FFN weights, with a low-rank online correction for the inter-layer basis mismatch. \\
ResQ~\citep{saxena2025resq} & 6 & PCA-rotation into the top-variance subspace kept at 8-bit, the rest at 4-bit, with an independent random orthogonal rotation inside each of the two subspaces; a data-aware rotation using KLT concentration for mixed precision. \\
RPTQ~\citep{yuan2023rptq} & 6 & k-means-cluster channels by (min,max) range and reorder so a cluster shares one scale. \\
RRS (Rotated Runtime Smooth)~\citep{yi2024rotated} & 6 & Online Hadamard plus a runtime smoothing scale taken as the group maximum over magnitude-reordered channels, the group matching the GEMM block so the scale is constant within each block. \\
SliM-LLM~\citep{huang2024slimllm} & 6 & Salience-driven mixed-precision (per-group bit-widths); cited as a non-permutation, out-of-scope contrast in the reorder family. \\
SmoothQuant+~\citep{pan2023smoothquantplus} & 6 & SmoothQuant scale for group-wise W4A16; alpha grid-searched on whole-model loss. \\
SmoothQuant~\citep{xiao2023smoothquant} & 6 & Per-channel scaling that migrates activation outliers into the more-quantizable weights. \\
SpinQuant~\citep{liu2024spinquant} & 6 & Learn residual (R1) and head-wise value (R2) rotations by Cayley SGD; R3,R4 fixed Hadamard. \\
SVD-LLM~\citep{wang2024svdllm} & 6 & Adjacent (not function-preserving): truncation-aware data whitening for low-rank compression, sharing WUSH's activation-whitening core. \\
TEQ~\citep{cheng2023teq} & 6 & Learn a minimal-parameter per-channel equivalent transformation, folded away like SmoothQuant. \\
WUSH~\citep{chen2025wush} & 6 & Cholesky whitening and a whitened-SVD balancing factor between two orthogonal factors, on a Hadamard backbone. \\
Z-Fold~\citep{jeon2023zfold} & 6 & Rank-one two-sided diagonal step-size $S=\zeta\alpha^{\!\top}$ fit by ALS (Hessian-weighted on the $\alpha$ step); the in-channel factor $\zeta$ folds into the previous layer, leaving $\alpha$ as the runtime per-channel scale. \\
\addlinespace
\multicolumn{3}{@{}l}{\emph{Rounding and codebooks (\Cref{sec:composition})}}\\[1pt]
AQLM~\citep{egiazarian2024aqlm} & 7 & Additive quantization: sum of learned codewords via beam search plus block fine-tuning; Pareto-optimal below 3-bit. \\
Chen et al. (GPTQ geometry)~\citep{chen2025geometry} & 7 & GPTQ run back-to-front equals Babai's nearest-plane for the CVP on the Hessian lattice (up to per-channel scale); recasts the transform as choice of lattice basis. \\
FLUTE~\citep{guo2024flute} & 7 & Lookup-table kernel restructuring packed weights offline; codebook dequantization 2-4x faster than an FP16 GEMM at batch sizes below 32 and group size 128. \\
GPFQ (greedy path-following)~\citep{lybrand2021gpfq} & 7 & Error-feedback rounding line independent of OBS: corrects the error already committed, including the input mismatch the already-quantized preceding layers leave behind, but does not adjust the not-yet-quantized weights. \\
GPTQ$=$Babai (lattice)~\citep{birnick2026lattice} & 7 & Independent short proof that GPTQ equals Babai's nearest-plane algorithm on the Hessian lattice, for a single scalar step size; leaves an LLL-style basis reduction to future work. \\
GPTQ~\citep{frantar2022gptq} & 7 & Scalable OBC: fixed column order, lazy blocked feedback, single Cholesky; the field workhorse, applies no basis-changing transform of its own (its act-order option is the processing-order permutation classified in \Cref{tab:taxonomy}). \\
GPTVQ~\citep{vanbaalen2024gptvq} & 7 & Extends GPTQ to VQ; interleaves Hessian-weighted k-means over weight sub-vectors with the error-feedback update. \\
NF4 (NormalFloat)~\citep{dettmers2023qlora} & 7 & 4-bit scalar codebook at standard-normal quantiles; equal-mass bins maximize fixed-length index entropy, relying on post-absmax Gaussian weights. \\
OBC~\citep{frantar2022obc} & 7 & OBS instantiated for quantization: per-row greedy least-impact quantize plus optimal compensation, cubic cost. \\
PCDVQ~\citep{yue2025pcdvq} & 7 & Randomized Hadamard, then decouples each 8-weight vector into an $E_8$-sampled direction and a $\chi_8$ magnitude, quantized by separate distribution-matched codebooks with most bits on the direction. \\
Qronos~\citep{zhang2025qronos} & 7 & Alternates activation-quantization-error correction with GPTQ diffusion; subsumes GPTQ when activation mismatch vanishes; composes with Hadamard/QuaRot/SpinQuant/SmoothQuant. \\
QTIP~\citep{tseng2024qtip} & 7 & Trellis-coded quantization via Viterbi over long blocks, approaching the Gaussian distortion-rate floor; enabled by incoherence rotation. \\
QuIP\#~\citep{tseng2024quipsharp} & 7 & Randomized Hadamard incoherence plus an 8-weight E8-lattice codebook (E8P, 2 bits); composes with a block extension of LDLQ. \\
SqueezeLLM~\citep{kim2023squeezellm} & 7 & Fisher-weighted k-means scalar codebook; keeps outliers full-precision via a dense-and-sparse split. \\
TurboQuant~\citep{zandieh2025turboquant} & 7 & Data-oblivious online VQ: a random rotation makes coordinates near-independent (Beta marginal), then per-coordinate scalar quantizers reach distortion-rate within $\sim2.7\times$; shown on KV cache. \\
VPTQ~\citep{liu2024vptq} & 7 & VQ as second-order optimization with residual and outlier codebooks; state-of-the-art two-bit. \\
WaterSIC (water-filling)~\citep{lifar2026watersic,ordentlich2026highrate2} & 7 & Waterfills per-coordinate grid spacings on the GPTQ rounder (Lifar et al.); high-rate basis-freeness and rotation-immunity established by Ordentlich \& Polyanskiy; basis-free, rotation-immune scalar-INT optimum that provably beats equal-rate rounding. \\
\addlinespace
\multicolumn{3}{@{}l}{\emph{Number format and co-design (\Cref{sec:format})}}\\[1pt]
Adaptive block-scaled (IF4)~\citep{cook2026adaptive} & 8 & Selects INT4 vs FP4 per 16-value block (an adaptive element grid) with a dedicated MAC unit, encoding the choice in the spare sign bit. \\
AMXFP4~\citep{lee2024amxfp4} & 8 & Asymmetric microscaling FP4: an asymmetric shared scale absorbs activation outliers calibration-free with a custom MAC engine; reports gains over both MXFP4 and rotation-based INT4 with no transform. \\
Block Rotation~\citep{shao2025blockrotation} & 8 & A Hadamard confined to the microscaling block alone captures most of the available MXFP4 benefit. \\
Four Over Six (4/6)~\citep{cook2025four} & 8 & Adaptively rescales individual NVFP4 blocks to smaller FP4 values for a more uniform representable grid; a format/scale co-design with no transform. \\
HadaNorm~\citep{federici2025hadanorm} & 8 & Composes a dynamic mean-centering with a per-channel scale whose inverse folds, and a Hadamard for W4A4 diffusion transformers. \\
HiFloat4 (format)~\citep{luo2026hifloat4format} & 8 & Three-level FP4: denser E1M2 grid, E6M2 global scale, two levels of 1-bit micro-exponents (8-way and 16-way) over G=64, averaging 4.5 bits per value. \\
HiFloat4 (pre-training study)~\citep{hifloat4_2026} & 8 & Pre-trains at HiF4 and reports a 0.85--1.19\% relative loss gap vs \texttt{BF16}, against 1.44--1.79\% for MXFP4. \\
INT-FP flip (high-rate)~\citep{ordentlich2026highrate} & 8 & High-rate analysis of the both-operands matmul: a rotation is needed for the INT grid but not for an ideal FP grid with a real-valued scale, so the number format decides whether flattening helps. \\
INT vs FP (fine-grained)~\citep{chen2025intvsfp} & 10 & Empirical INT-vs-FP study across block granularities: FP wins at 4-bit, but fine-grained NVINT4 with a Hadamard rotation matches NVFP4, and MXINT8 beats FP at 8-bit. \\
Microscaling (MX standard)~\citep{rouhani2023microscaling} & 8 & OCP microscaling (MX) standard: a block of G=32 elements shares one scale; defines MXFP4 as E2M1 elements with a power-of-two E8M0 shared scale. \\
MR-GPTQ~\citep{egiazarian2025mrgptq} & 8 & MR-GPTQ: block-diagonal Hadamard with a tunable power-of-two block $k$ decoupled from the microscaling group, MSE grid search, and E8M0 scale fitting; recovers up to 98-99\% of FP16 on large models. \\
MX+~\citep{lee2025mxplus} & 8 & Repurposes the block-maximum element's redundant exponent bits as extended mantissa via a small auxiliary unit beside the dot-product engine (core MAC unchanged); higher accuracy than MXFP4 at negligible overhead. \\
MXFP benchmark~\citep{zhang2026benchmarkmxfp} & 8 & MXFP8 near-lossless but MXFP4 badly degraded; isolates E8M0 power-of-two scale rounding as the critical error, much recovered by adopting the global 3/4 pre-scale of the MXFP4-training row below. \\
MXFP4 training~\citep{tseng2025mxfp4} & 8 & Backward-only MXFP4 with stochastic rounding under a random Hadamard; introduces the $3/4$ pre-scale that prevents clipping, with the accumulator rescaled by $16/9$ to stay unbiased. \\
nGPT (normalized architecture)~\citep{loshchilov2025ngpt} & 8 & Constrains embeddings, weights and hidden states to the unit hypersphere; the architecture the next row's 4-bit result rests on. \\
Normalized architectures at 4 bits~\citep{fishman2026normalized} & 8 & Unit-hypersphere pretraining makes NVFP4 stable with no Hadamard and no dynamic per-tensor scaling. \\
NVFP4 (pretraining)~\citep{nvidia2025nvfp4} & 8 & NVFP4 format (E2M1 elements, mantissa-carrying E4M3 FP8 block scale over 16-element blocks); pretraining recipe applies selective randomized Hadamard only to weight-gradient GEMM inputs, training 12B/10T tokens within a point of FP8. \\
Quartet~\citep{castro2025quartet} & 8 & MXFP4 training with a block-diagonal Hadamard sized to the microscaling group (G=32). \\
Quartet (NVFP4)~\citep{panferov2026quartet} & 8 & Hides the unbiased-gradient correction in stochastically-rounded FP8 scale, >2x lower error than value stochastic-rounding, using NVFP4's mantissa-carrying scale. \\
QuEST~\citep{panferov2025quest} & 8 & QAT that Hadamard-normalizes weights and activations toward Gaussian, then fits one MSE-optimal clip threshold; grid-agnostic (INT or FP4). \\
Shared microexponents (BDR)~\citep{rouhani2023shared} & 10 & ISCA-2023 Block Data Representation: the two-level design space (coarse block scale plus sub-block shared microexponents, mantissas as low as 2 bits) the MX standard descends from. \\
STaMP~\citep{federici2025stamp} & 8 & Sequence-axis KLT concentrates token energy into a few high-precision tokens; composes with channel transforms. \\
TORQ~\citep{xu2026torq} & 8 & Two-level MXFP4 rotation: a coarser across-block orthogonal rotation atop its own block-diagonal maximum-entropy (Givens) intra-block rotation. \\
VS-Quant~\citep{dai2021vsquant} & 10 & Per-vector scaled quantization: an integer scale per 16--64-element vector under a coarse floating-point scale; an early sub-block scaling scheme prefiguring microscaling. \\
YAQA~\citep{tseng2025yaqa} & 8 & Randomized Hadamard makes a Kronecker-factored full-model Fisher incoherent; its own two-sided LDL rounder cuts end-to-end KL by about $30\%$ over LDLQ at no inference cost. \\
\addlinespace
\multicolumn{3}{@{}l}{\emph{Beyond the weight matrix (\Cref{sec:beyond})}}\\[1pt]
ConvRot~\citep{huang2025convrot} & 9 & Group-wise Hadamard rotation suppressing row- and column-wise outliers for plug-and-play \texttt{W4A4} diffusion-transformer inference. \\
DiRotQ~\citep{sharify2026dirotq} & 9 & Rotation-aware 4-bit quantization for diffusion transformers. \\
DiTAS~\citep{dong2024ditas} & 9 & Timestep-static per-channel smoothing (max over steps) plus training-free low-rank weight-error repair. \\
HALO~\citep{ashkboos2025halo} & 9 & Right-hand Hadamard on the forward operands suffices for \texttt{FP6}; a further left-hand Hadamard on the row-outlying output-gradient tensors is what \texttt{INT8} fine-tuning needs. \\
INT4 training~\citep{xi2023int4training} & 9 & Forward block-diagonal Hadamard; sparse gradients use leverage-score row sampling and bit-split, not flattening. \\
KIVI~\citep{liu2024kivi} & 9 & Tuning-free asymmetric 2-bit KV cache: per-channel keys, per-token values, \texttt{FP16} residual window, fused kernel. \\
KVLinC~\citep{saxena2025kvlinc} & 9 & Value-side Hadamard plus \emph{trained} linear-correction adapters compensating quantized-key error in the $QK^{\top}$ logit; 2-bit keys, up to $2.55\times$ faster decoding on \texttt{Llama-2-7B}. \\
KVQuant~\citep{hooper2024kvquant} & 9 & Per-channel keys, per-token values, pre-RoPE, Fisher-weighted codebook with dense-and-sparse; no rotation, 2-bit KV. \\
LRQ-DiT~\citep{yang2025lrqdit} & 9 & Activation-fluctuation stat gates plain Hadamard vs outlier-aware rotation-plus-permutation per layer. \\
MambaQuant~\citep{xu2025mambaquant} & 9 & KLT-enhanced rotation $H_K=KH$ equalizes Mamba channel variances; a decorrelating pre-conditioner in service of flattening. \\
MoEQuant~\citep{hu2025moequant} & 9 & MoE PTQ: expert-balanced self-sampling and an affinity-guided objective fix per-expert calibration starvation; no transform, composes with the toolkit per expert. \\
OSCAR~\citep{zhou2026oscar} & 9 & Data-aware KLT-flavored offline rotation fitted to the covariance attention consumes (the query covariance for keys) enables 2-bit KV-cache keys. \\
PolarQuant (Han et al.)~\citep{han2025polarquant} & 9 & Random-rotation preconditioning makes recursive-polar angles data-independent, so no per-block scale or zero-point is stored, only small per-level angle codebooks; $4.2\times$ KV compression. \\
PolarQuant (Wu et al.)~\citep{wu2025polarquant} & 9 & 2D sub-vector radius-and-angle exploiting RoPE's paired rotations; precomputes the query-key product into a decode-time lookup table. \\
Q-Diffusion~\citep{li2023qdiffusion} & 9 & No transform; calibrate across all timesteps and split bimodal U-Net shortcut activations. \\
Q-DiT~\citep{chen2024qdit} & 9 & No transform; evolutionary per-layer group-size search plus dynamic per-sample/timestep activation scales. \\
Quamba~\citep{chiang2024quamba} & 9 & Online Hadamard on the SSM scan output with its inverse folded into the output projection; percentile-clipped scale on the input. \\
RotateKV~\citep{su2025rotatekv} & 9 & Walsh--Hadamard plus calibrated channel permutation on grouped heads, pre-RoPE, 2-bit KV. \\
SageAttention2~\citep{zhang2024sageattention2} & 9 & Thorough $Q$/$K$ outlier smoothing enabling per-thread INT4 $QK^{\top}$ (with $\tilde P V$ in FP8); the INT4 step between SageAttention v1 and the FP4 v3. \\
SageAttention3 (FP4)~\citep{zhang2025sageattention3} & 9 & FP4 attention: a per-token rescale lifts the per-block scale into \texttt{E4M3}'s usable range before microscaling. \\
SageAttention~\citep{zhang2024sageattention} & 9 & Mean-subtracts keys (shared-bias removal); a static scale would suffice for the bounded probabilities, though the default v1 kernel keeps PV in FP16 and the FP4/FP8 successors quantize P. \\
SVDQuant~\citep{li2025svdquant} & 9 & Rank-32 FP16 low-rank branch absorbs outliers; the residual quantizes to W4A4. \\
ThriftAttention~\citep{sharratt2026thriftattention} & 9 & Q, K and V in FP4; a block-mean surrogate routes the important query-key block pairs to FP16. \\
TR-DQ~\citep{shao2025trdq} & 9 & Timestep-varying rotation for diffusion transformers. \\
ViDiT-Q~\citep{zhao2024viditq} & 9 & Originally a timestep-dependent SmoothQuant migration, different per-channel smoothing for the two halves of the trajectory; the published version drops the per-timestep split and combines scaling with a rotation for the dynamic component. \\
\addlinespace
\multicolumn{3}{@{}l}{\emph{Systems, kernels, and coders (\Cref{sec:systems})}}\\[1pt]
AMD CDNA3 (MI300X)~\citep{amd_cdna3} & 10 & AMD matrix-core generation with \texttt{FP8}/\texttt{INT8} but no native \texttt{FP4}/\texttt{FP6}. \\
AMD CDNA4 (MI355X)~\citep{amd_cdna4} & 10 & First AMD data-center architecture with hardware-native \texttt{FP4}/\texttt{FP6} and OCP microscaling. \\
APEX4~\citep{guo2026apex4} & 10 & Pure-\texttt{INT4} W4A4 GEMM with $\rho$-aware granularity adaptation; names group dequantization on the CUDA cores as the W4A4 bottleneck and makes the tensor-core-to-CUDA-core throughput ratio the viability indicator. \\
CCQ~\citep{zhou2025ccq} & 10 & Convolutional-code codebook with lookup-free bit-shift decode, no entropy stage; composes with grouped GEMM. \\
CUDA PTX ISA~\citep{nvidia_ptx_isa} & 10 & Defines \texttt{tcgen05.mma} and its block-scaled operands: a scale vector every 32 elements for the MX formats, every 16 for \texttt{NVFP4}. \\
COMET~\citep{liu2024comet} & 10 & Practical W4A4KV4 serving: block-wise mixed precision and channel permutation confine 8-bit activations to a small fraction of blocks, keeping most tiles on the 4-bit path. \\
DFloat11~\citep{zhang2025dfloat11} & 10 & Huffman-codes the exponent with a two-phase SRAM-LUT GPU decode pre-GEMM; cuts weight memory \ensuremath{\sim}30\%. \\
HadaCore~\citep{agarwal2024hadacore} & 10 & Tensor-core fast Walsh--Hadamard kernel (size-16 base case) running the online incoherence rotation 1.1-1.4x faster than the prior kernel on an A100. \\
Huff-LLM~\citep{yubeaton2025huffllm} & 10 & Hardware Huffman decoders between weight buffer and systolic array emit one FP16 weight per clock. \\
LiquidGEMM~\citep{hu2025liquidgemm} & 10 & Hardware-efficient W4A8 GEMM; a fast, overflow-safe \texttt{INT4}$\to$\texttt{INT8} dequant that keeps pace with the tensor cores. \\
LLM Compressor (vLLM)~\citep{vllm_llmcompressor} & 10 & vLLM llm-compressor toolkit: ships GPTQ/AWQ/SmoothQuant plus SpinQuant/QuIP rotations as composable calibration modifiers, Hadamard via fused HadaCore inference kernels. \\
Luo et al.\ (Hopper dissection)~\citep{luo2024hopper} & 10 & Microbenchmark study showing Hopper's \texttt{INT4} \texttt{mma} compiles to CUDA-core \texttt{IMAD}: no tensor-core \texttt{INT4}. \\
LUT-GEMM~\citep{park2022lutgemm} & 10 & Dequantization-\emph{free} weight-only GEMV over a binary-coding (BCQ) representation, into which uniform \texttt{INT4} is rewritten: tabulates partial products so the matmul needs no \texttt{INT4}$\to$\texttt{FP16} upconversion. \\
Machete~\citep{machete2024} & 10 & Hopper-optimized Marlin successor (TMA/wgmma) that overlaps the \texttt{INT4}$\to$\texttt{FP16} upconversion for compute-bound W4A16 serving. \\
Marlin~\citep{frantar2024marlin} & 10 & W4A16 \texttt{FP16}$\times$\texttt{INT4} GEMM that hides the weight dequant behind memory traffic, near-$4\times$ up to batch $16$--$32$. \\
NeuZip~\citep{hao2024neuzip} & 10 & ANS-codes the exponent; GPU decompresses each weight into a scratch tile pre-GEMM, then frees it. \\
NVIDIA Ampere (A100)~\citep{nvidia_ampere} & 10 & Third-gen tensor cores with native \texttt{INT4} ($4\times$ \texttt{FP16}) and \texttt{INT8}; the integer datapath the W4A4 methods target. \\
NVIDIA Blackwell~\citep{nvidia_blackwell} & 10 & Fifth-gen tensor cores with hardware-native \texttt{FP4}/microscaling (\texttt{NVFP4}, MX): the block scale is applied in silicon. \\
NVIDIA Hopper (H100)~\citep{nvidia_hopper} & 10 & Adds \texttt{FP8} via the Transformer Engine but drops \texttt{INT4} from the tensor core. \\
QQQ~\citep{zhang2024qqq} & 10 & W4A8 with a \texttt{FastINT4toINT8} decode and offline scale folding; per-channel/per-group W4A8 GEMM $3.7\times$/$3.3\times$ over \texttt{FP16}. \\
QServe~\citep{lin2024qserve} & 10 & W4A8KV4 serving: INT4 dequant lands in INT8 registers so all linear-layer GEMMs run on INT8 tensor cores (attention stays FP16). \\
QUIK~\citep{ashkboos2023quik} & 10 & Keeps 256 outlier channels in FP16 (${\approx}3\%$ of \texttt{OPT-66B}'s hidden size), the rest INT4; fuses the INT4 dequantization into the GEMM epilogue, accumulating into the separate FP16 outlier matmul. \\
Shannon-bound coding~\citep{tan2026shannon} & 10 & Tile-level ANS lossless decompression aligned with the GEMM tiling, within 0.01--0.1 bits of the entropy limit; effective weight entropy 2--10$\times$ below stored width. \\
TensorRT Model Optimizer~\citep{nvidia_modelopt} & 10 & NVIDIA production PTQ toolkit shipping SmoothQuant/AWQ/GPTQ and a fast Hadamard for QuaRot rotation, with NVFP4/MXFP4 export to TensorRT-LLM, vLLM, SGLang. \\
ZipNN~\citep{hershcovitch2024zipnn} & 10 & Huffman-codes the BF16 exponent; CPU decompresses the whole model once at load, off the datapath. \\
ZipServ~\citep{fan2026zipserv} & 10 & Hardware-aware lossless weight compression: a fixed-length tensor-core-aware bitmap encoding and a fused kernel decompressing into tensor-core registers; up to 30\% smaller, GEMM stays fixed-rate. \\
\addlinespace
Kimi K2 Thinking~\citep{moonshot2026kimik2thinking} & 10 & Ships routed experts in \texttt{INT4} by quantization-aware training (group 32); no transform, attention and router kept in \texttt{BF16}. \\
gpt-oss~\citep{openai2025gptoss} & 10 & \texttt{MXFP4} mixture-of-experts weights with attention, router, embeddings and head excluded; no transform. \\
DeepSeek-V4-Pro~\citep{deepseek2026v4} & 10 & \texttt{FP4} routed experts, most other parameters \texttt{FP8}; no transform on the weight path, but an online Hadamard before the \texttt{FP4} indexer activations. \\
Qwen3-32B-AWQ~\citep{qwen2025qwen3awq} & 10 & First-party 4-bit \texttt{AWQ} weight release; the per-channel diagonal scale folds offline, so a shipped weight path instances \Cref{def:fpt}. \\
DeepSeek-V3.2 indexer~\citep{deepseek2025v32code} & 10 & Earlier shipped online Hadamard on the sparse-attention indexer's queries and keys, ahead of \texttt{FP8} activation quantization. \\
Llama-3.3-70B-Instruct-FP4~\citep{nvidia2025llama33fp4} & 10 & \texttt{NVFP4} weights \emph{and} activations at group 16 with no rotation named; the shipped 4-bit activation path with no transform named. \\
Llama 3.2 quantized pair~\citep{meta2024llama32quantized} & 10 & Same model at \texttt{W4A8} twice: fine-tuned SpinQuant rotations plus GPTQ, and a QLoRA variant with no transform. \\
\texttt{llama.cpp} \texttt{KV} rotation~\citep{llamacpp2026kvrot} & 10 & Walsh--Hadamard on the \texttt{KV} cache, enabled by default whenever the cache is quantized and the head dimension is a multiple of 64. \\
\addlinespace
\multicolumn{3}{@{}l}{\emph{Background, surveys, and cross-cutting theory}}\\[1pt]
Ainslie et al.\ (GQA)~\citep{ainslie2023gqa} & 2 & Grouped-query attention; the key/value-head sharing that sets the granularity of KV-cache quantization. \\
Bhadane et al.\ (PBA)~\citep{bhadane2021pba} & 2 & Principal Bit Analysis; the classical result that a Schur-concave rate objective is minimized by the energy-concentrating KLT, the concentration pole of the inversion's Schur-order framing. \\
Czako et al. review~\citep{czako2025review} & 1 & Adjacent 2025 review bucketing equivalent transforms into activation-outlier mitigations, frozen early-2025, without unifying theory. \\
Gong et al. low-bit survey~\citep{gong2024lowbitsurvey} & 1 & Prior broad low-bit-LLM survey; treats the transform as one step among many rather than the organizing axis. \\
Liu et al.\ (evaluation)~\citep{liu2025compeval} & 1, 11 & Single-protocol evaluation decoupling PTQ into pre-quantization transformation and error mitigation across \texttt{INT4}/\texttt{MXFP4}/\texttt{NVFP4}. \\
LLM.int8()~\citep{dettmers2022llmint8} & 1 & Isolated emergent large-magnitude outlier features in transformer activations at scale: the wall low-bit transforms attack. \\
Marshall \& Olkin (majorization)~\citep{marshall2011majorization} & 2 & Standard majorization reference; supplies the separable-sum half (Schur-concavity for a concave summand) and the Schur-convexity of the maximum. Pairing them along the within-group order is ours. \\
Massive activations~\citep{sun2024massive} & 6 & Names and characterizes the few tokens carrying massive activations that the off-axis prefix and outlier-weighted rotation methods target. \\
Shazeer (GLU)~\citep{shazeer2020glu} & 2 & Gated linear units; the gated-MLP (SwiGLU) form whose gate/up/down projections the survey quantizes. \\
Su et al.\ (RoPE)~\citep{su2021roformer} & 2 & Rotary position embedding; its position-dependent key rotation is why KVQuant and RotateKV quantize KV-cache keys pre-RoPE. \\
Vaswani et al.\ (Transformer)~\citep{vaswani2017attention} & 2 & The Transformer architecture; defines the attention and MLP linear layers on which the transforms act. \\
Xiong et al.\ (pre-LN)~\citep{xiong2020layernorm} & 2 & Analyzes normalization placement and shows the pre-norm form trains without learning-rate warm-up; the placement now standard in the models this survey quantizes. \\
Zhang \& Sennrich (RMSNorm)~\citep{zhang2019rmsnorm} & 2 & RMSNorm, the normalizer that opens each sub-block and subtracts no mean; its learned gain folds into the following projections at zero cost. \\
\addlinespace
\end{longtable}}

\fi

\end{document}